%% file: neurips_2026.tex
\documentclass{article}

\usepackage[preprint]{neurips_2026}

\usepackage[utf8]{inputenc} 
\usepackage[T1]{fontenc}    
\usepackage[colorlinks=true, citecolor=teal, urlcolor=magenta, linkcolor=magenta]{hyperref}  
\usepackage{url}            
\usepackage{booktabs}       
\usepackage{amsfonts}       
\usepackage{nicefrac}       
\usepackage{microtype}      
\usepackage{xcolor}         
\usepackage{graphicx}
\usepackage{dsfont}
\usepackage{float}
\usepackage{caption}
\usepackage{multirow}
\usepackage{fancyhdr}
\usepackage{titletoc}
\usepackage{enumitem}
\usepackage{amsmath}
\usepackage{amssymb}
\usepackage{mathtools}
\usepackage{amsthm}
\usepackage{wrapfig}
\usepackage{verbatim}
\usepackage{fancyvrb}
\usepackage{fvextra}
\usepackage{lipsum}
\usepackage{longtable}
\usepackage[most]{tcolorbox} 
\usepackage[table]{xcolor}
\usepackage[textsize=tiny]{todonotes}
\usepackage[capitalize,noabbrev]{cleveref}
\usepackage{array}
\usepackage{tabularx}
\usepackage{algorithm}
\usepackage{algpseudocode}
\usepackage{makecell}
\usepackage{float}
\usepackage{subcaption}
\usepackage{bm}
\usepackage{svg}
\usepackage{graphicx}
\usepackage{array}
\usepackage{ragged2e}
\usepackage{soul}

\sethlcolor{yellow} 

\newcommand{\name}{GauntletBench}
\definecolor{crimsonorange}{RGB}{220,60,30}

\title{
{\includegraphics[height=0.95em]{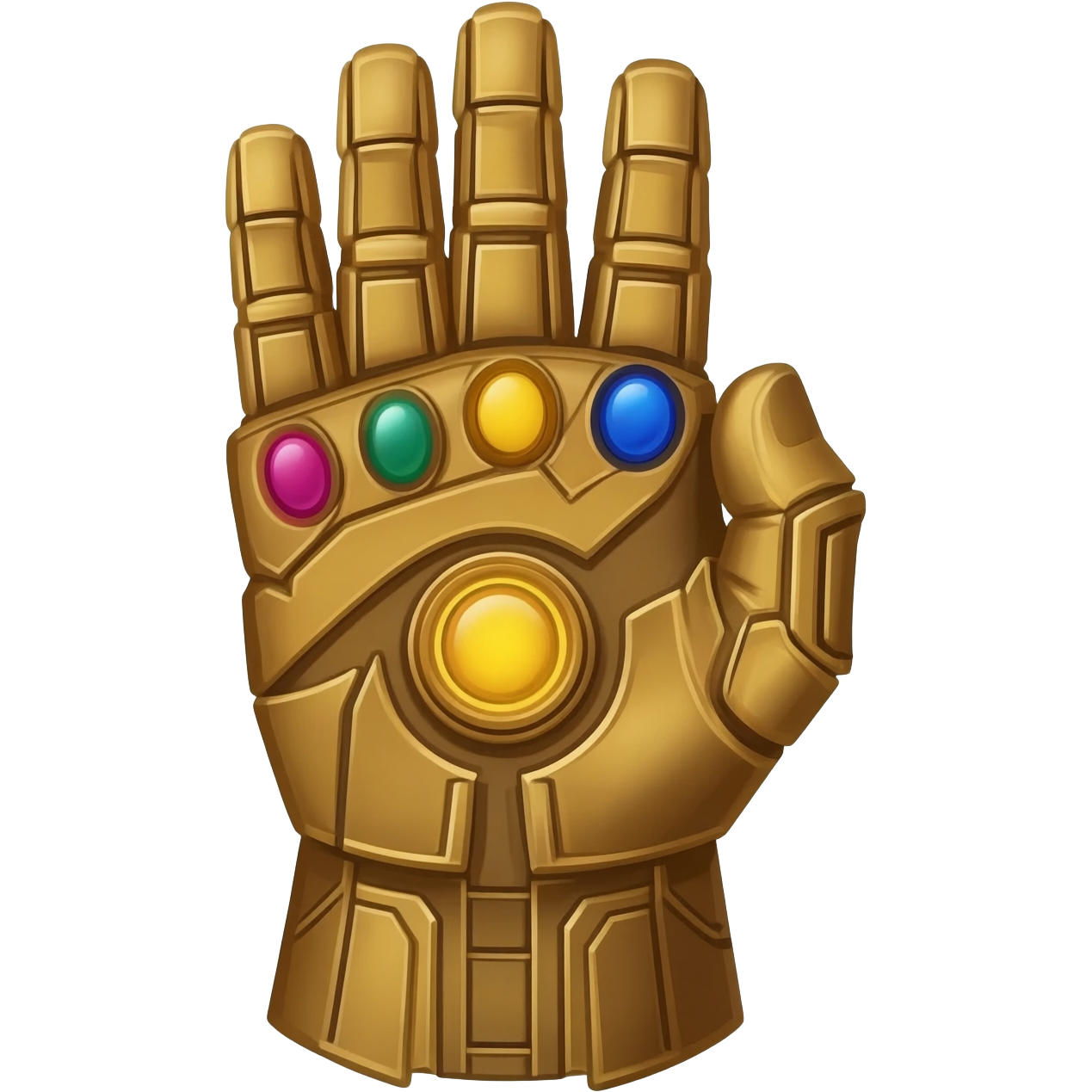}} Running the Gauntlet: Hard Agentic Tasks
}
\author{
\textbf{Mykola Vysotskyi}$^{2,7}$\thanks{Equal contribution}\quad\,
\textbf{Runqi Lin}$^{1}$\footnotemark[1]\quad\,
\textbf{Grzegorz Biziel}$^{2}$\quad\,
\textbf{Michal Zakrzewski}$^{2}$\\
\textbf{Sebastian Montagna}$^{2}$\quad\,
\textbf{Damian Rynczak}$^{2}$\quad\,
\textbf{Shreyansh Padarha}$^{1}$\\
\textbf{Kumail Alhamoud}$^{3}$\quad\,
\textbf{Zihao Fu}$^{4}$\quad\,
\textbf{William Lugoloobi}$^{1}$\quad\,
\textbf{Kai Rawal}$^{1}$\\
\textbf{Hanna Yershova}$^{2}$\quad\,
\textbf{Taras Rumezhak}$^{2}$\quad\,
\textbf{Guohao Li}$^{5}$\quad\,
\textbf{Fazl Barez}$^{1}$\\
\textbf{Baoyuan Wu}$^{6}$\quad\,
\textbf{Arkadiusz Drohomirecki}$^{2}$\quad\,
\textbf{Chris Russell}$^{1}$\\
\textbf{Christopher Summerfield}$^{1}$\quad\,
\textbf{Adam Mahdi}$^{1}$\quad\,
\textbf{Volodymyr Karpiv}$^{2}$\\
\textbf{Philip Torr}$^{1}$\quad\,
\textbf{Adel Bibi}$^{1,2}$\thanks{Corresponding author: adel.bibi@eng.ox.ac.uk}\vspace{-0.4em}\\
\\
$^{1}$University of Oxford \quad
$^{2}$SoftServe \quad 
$^{3}$Massachusetts Institute of Technology \\
$^{4}$CUHK \quad
$^{5}$Eigent.AI \quad
$^{6}$CUHK, Shenzhen \quad
$^{7}$Ukrainian Catholic University
}

\begin{document}

\maketitle
\setcounter{footnote}{0}
\input{0_Abstract}







\begin{flushright}

\itshape
``The greatest obstacle to discovery is not ignorance --- it is the illusion of knowledge.''\\

--- Daniel J. Boorstin
\end{flushright}







\input{1_Introduction}

\input{2_Methodology}

\input{3_Experiments}
\input{4_Related_Works}
\input{5_Conclusion}

\begin{ack}
The authors express their gratitude to Igor Ivanyshyn for his participation in the human annotation.
The authors also thank Svyatoslav Marutyak for his valuable comments on the task design.
Runqi Lin was funded by the Horizon Europe grant 101213369 (DVPS).
Adel Bibi and Philip Torr acknowledge the UKRI Turing AI Fellowship (EP/W002981/1).
Adel Bibi also acknowledges support from Coefficient Giving and is affiliated with the Institute for Decentralized AI.

\end{ack}

{\small
\bibliographystyle{plain}
\bibliography{neurips_2026}
}

\clearpage
\newpage
\appendix

\section*{Appendix: Table of Contents}
\startcontents[appendix]
\printcontents[appendix]{}{1}{\setcounter{tocdepth}{3}}

\newpage

\input{A_Appendix}
\input{B_Appendix}
\input{C_Appendix}
\input{D_Appendix}
\input{E_Appendix}
\input{F_Appendix}
\input{G_Appendix}


\end{document}

%% file: 0_Abstract.tex
\begin{abstract}
As agentic systems continue to evolve and are widely deployed in real-world scenarios, there is a growing demand to faithfully evaluate their capabilities.
However, current benchmarks are typically built on popular applications with relatively simple tasks and focus on a narrow set of capabilities while overlooking broader dimensions, resulting in saturated performance on modern agents and failing to probe their limitations.
To this end, we introduce \textit{GauntletBench}, a web-based benchmark for evaluating agent generalisation in challenging scenarios, focusing on three underexplored capabilities (temporal perception, graphical understanding, and 3D reasoning), across five less-covered professional applications (\texttt{Video Editor}, \texttt{Workflow Builder}, \texttt{3D Modeller}, \texttt{Flight Analyser}, and \texttt{Circuit Designer}), each with 27 vision-intensive tasks (135 in total).
Our benchmark provides a modular pipeline that comprises an environment compatible with both open- and closed-source agent frameworks, a controlled web-based application, a well-structured task suite, and an automated evaluation engine with diverse metrics.
Contrary to widespread expectations, our empirical results reveal that frontier agentic systems remain far from achieving human-level performance.
Even the state-of-the-art agent achieves only a 28.2\% success rate on our GauntletBench, highlighting the limitations in these overlooked capabilities and generalisation\footnote{This paper was originally released on June 12, 2026. Following the release of GPT-6 Astra on September 3, 2026, we re-evaluated GauntletBench and observed a substantial capability shift within only three months: GPT-6 Astra achieves 83.0\%, becoming the first evaluated agent to surpass the human baseline of 80.1\%. GauntletBench was designed around the capability frontier at the time, with task formulations (e.g., step-by-step instructions) that made challenging capabilities meaningfully assessable for then-weaker agents. We are working on a future version revisiting such scaffolding to remain challenging for frontier agents while preserving human solvability.}.
By comparison, non-expert human annotators achieve over 80\% success on our challenging yet feasible tasks, revealing the substantial gap between current agent capabilities and those required for complex real-world scenarios\footnote{The \href{https://gauntlet-landing-page.vercel.app/}{homepage}, \href{https://github.com/gauntlet-benchmark/evaluation-harness}{code}, and \href{https://github.com/gauntlet-benchmark/tasks}{tasks} of GauntletBench are available publicly.}.

\end{abstract}

%% file: 1_Introduction.tex
\section{Introduction}
\vspace{-0.25cm}

The rapid advancement of multimodal large language models (MLLMs) has significantly improved their cross-modal perception and reasoning capabilities~\cite{anthropic2025system, pichai2025new, singh2025openai}, establishing a strong foundation for the transition from passive assistants to intelligent agents~\cite{yao2022react, park2023generative, shen2023hugginggpt}.
Agentic systems built on these models are now widely deployed for autonomous interactions and execution across diverse real-world domains, spanning from consumer-facing~\cite{liuagentbench, yao2022webshop, zhou2024webarenarealisticwebenvironment} to enterprise applications~\cite{saxena2025continuous, mo2026entworld}.
As modern agents evolve, they are expected to handle increasingly complex scenarios, necessitating more challenging evaluations to reflect their capabilities in such settings~\cite{bommasani2021opportunities, park2023generative}.

While existing benchmarks provide valuable assessments, they often operate within highly homogeneous settings with relatively simplistic tasks, leading to saturated performance on modern agents and failing to reflect real-world complexity~\cite{bommasani2021opportunities,liuagentbench}.
Specifically, current benchmarks face two fundamental limitations: (i) they are built on popular applications or near-identical replicas (e.g., Amazon, Booking, or CRM systems~\cite{garg2025realbenchmarkingautonomousagents, mo2026entworld}), with the tasks built upon them exhibiting high similarity and simplicity, making them insufficient to probe the limits of current agentic systems; and (ii) the task design typically targets a narrow set of capabilities, such as UI understanding, tool use, and long-term reasoning~\cite{zhou2024webarenarealisticwebenvironment, wang2025odysseybench}, while failing to capture broader yet critical aspects of agentic performance.
Moreover, these benchmarks often struggle to continuously support new agentic frameworks due to intricate pipelines, limited compatibility, and rigid configurations~\cite{liuagentbench}.
These limitations lead to a mismatch between the evaluations provided by current benchmarks and the challenges of complex real-world scenarios, resulting in an inaccurate assessment of agentic capabilities.

To bridge this gap, we propose \name{}, a web-based benchmark targeting underrepresented capabilities in less-covered applications via visually grounded tasks to evaluate agents’ generalisation in challenging scenarios.
Specifically, our benchmark focuses on three previously overlooked capabilities, namely temporal perception, graphical understanding, and 3D reasoning, which are increasingly important as agentic systems operate in more complex settings and expand across broader modalities.
In our benchmark, we build five professional applications, including \texttt{Video Editor}, \texttt{Workflow Builder}, \texttt{3D Modeller}, \texttt{Flight Analyser}, and \texttt{Circuit Designer}, which are well-suited for evaluating our target capabilities and provide less familiar interfaces for assessing generalisation.
Each application includes 20 challenging tasks across well-calibrated difficulty levels, covering a spectrum of capabilities from vision-based UI understanding to the targeted aspects of our benchmark.


To support compatibility, controllability, and reproducibility, our platform adopts a modular design consisting of four components: environments, applications, tasks, and the evaluation engine.
The environments provide high-level observation and action interfaces, supporting both open-source and API-based MLLM agents, as well as closed-source agent frameworks.
Built on these environments, we develop our applications using modern web technologies to support natural interaction patterns and flexible configuration.
We provide tasks across three difficulty levels: easy, medium, and hard, following a standardised structure comprising application background, task description, and result format.
Finally, we provide an automated evaluation engine that integrates tailored objective evaluators to assess success rate, MLLM-judged progress rate, and efficiency measures, supporting automated evaluation even for closed-source agent frameworks.

Our evaluations indicate that most open-source MLLM agents, such as Qwen-3-VL~\cite{bai2025qwen3vltechnicalreport} and Llama-4-Maverick~\cite{meta_llama4_2026}, achieve near-zero performance, and no API-based MLLM agents achieve success rates above 13.2\% on GauntletBench.
Even the state-of-the-art (SOTA) closed-source agentic frameworks like GPT-5.4 Computer Use~\cite{openai_computer_use_2026}, Gemini Enterprise~\cite{openai2025chatgptatlas}, and Claude Opus Computer Use~\cite{anthropic2024computeruse} achieve success rates of only 7.4\%, 25.7\%, and 28.2\%, respectively.
In contrast, human annotators without specific domain knowledge of these applications achieve a success rate of over 80\% on our benchmark while requiring {30\%} fewer steps than the frontier agentic system.
This substantial gap in our challenging yet feasible benchmark indicates that current agentic systems still have critical limitations in these previously overlooked capabilities and generalisation, which are not faithfully reflected by existing benchmarks.
We believe our benchmark provides a new perspective on evaluating agent generalisation in complex real-world scenarios, enabling a more accurate assessment to meet the growing demands of such settings.
Our major contributions are summarised as follows:

\begin{itemize}[leftmargin=20pt, topsep=1pt, itemsep=1pt]

\item We introduce the \name{}, a challenging benchmark designed to evaluate agents’ generalisation by targeting three underexplored capabilities: temporal perception, graphical understanding, and 3D reasoning.

\item Our benchmark provides a web-based environment, comprising five less-covered professional applications, 135 challenging yet human-feasible tasks, and an automated evaluation engine.

\item Empirical results show that SOTA agents achieve only a 28.2\% success rate on \name{}, indicating that they remain far from human-level performance in complex real-world scenarios.

\end{itemize}

%% file: 2_Methodology.tex
\begin{figure}[t!]
    \centering
    \includegraphics[width=0.95\linewidth]{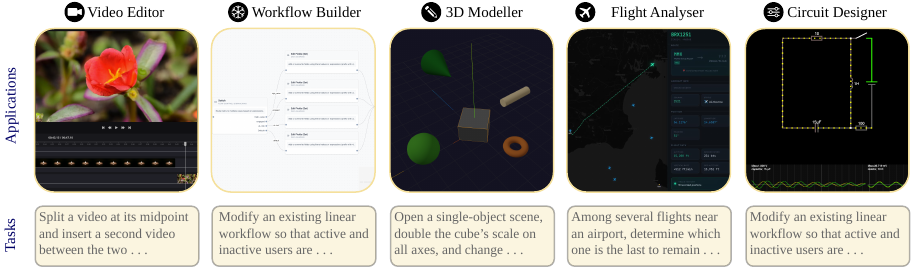}
    \caption{\textbf{Overview of \name{}.} Our benchmark contains five controlled web-based applications with 135 vision-intensive tasks. Detailed application descriptions are provided in Table~\ref{tab:benchmark_apps}, and full application screenshots are shown in Appendix~\ref{app:human-evaluation}.}
    \label{fig:benchmark_apps}
\vspace{-1.em}
\end{figure}

\section{\name{}}
\vspace{-0.25cm}
In this section, we introduce \name{}, a challenging web-based benchmark targeting agent generalisation in complex real-world scenarios.
Our platform follows a modular pipeline comprising a unified web-based environment (Section~\ref{sec:3.1}), where agents interact with controlled applications (Section~\ref{sec:3.2}) to execute well-structured tasks (Section~\ref{sec:3.3}), and outcomes are evaluated by an automated engine (Section~\ref{sec:3.4}).

\subsection{Environments}
\label{sec:3.1}
\vspace{-0.25cm}
Following standard formulations~\cite{mo2026entworld, garg2025realbenchmarkingautonomousagents}, we define a unified environment as $\mathcal{E} = (\mathcal{S}, \mathcal{A}, \mathcal{O}, \mathcal{T}, \mathcal{R})$, where $\mathcal{S}$ denotes the state space, $\mathcal{A}$ and $\mathcal{O}$ indicates the action and observation space, $\mathcal{T}$ is the transition function, and $\mathcal{R}$ refers to the reward function.
At each time step $t$, the agent receives an observation $o_t \in \mathcal{O}$, executes an action $a_t \in \mathcal{A}$, transitions to the next state according to $s_{t+1} = \mathcal{T}(s_t, a_t)$, and use a binary reward $r_t \in \{0,1\}$ to evaluate task success.

\textbf{Observation Space}\hspace*{2mm}GauntletBench exposes agents to all information naturally available through standard web interaction, without artificially restricting the observation space.
In \texttt{3D Modeller}, \texttt{Flight Analyser}, and \texttt{Circuit Designer}, advanced functionality is implemented through canvas-based rendering, whose underlying implementations inherently do not expose structured representations through browser interfaces.
As a result, in these three applications, agents can only use \textit{Screenshots} to perceive and reason about task-relevant content. 
On the other hand, for the \texttt{Video Editor} and \texttt{Workflow Builder}, we additionally provide the \textit{Accessibility Tree} to expose available semantic information about UI elements.
We would like to emphasise that all tasks in GauntletBench can be completed using visual observations alone, and that this visual dependence reflects a common characteristic of real-world professional web applications, whose core interactive workspaces are often not directly exposed through structured browser interfaces.

\textbf{Action Space}\hspace*{2mm}Agents act by selecting $a_t \in \mathcal{A}$ from a set of high-level commands that emulate standard user interactions, such as \texttt{click}, \texttt{type}, \texttt{select}, and \texttt{scroll}, as well as navigation operations.
These actions are executed via a browser automation layer (e.g., Playwright) and, when available, are optionally guided by elements identified from the accessibility tree.

\textbf{Agent Interface}\hspace*{2mm}We provide a unified interface that supports open-source and API-based MLLM agents, as well as closed-source agent frameworks. 
For MLLM agents, we leverage components of the REAL framework~\cite{garg2025realbenchmarkingautonomousagents} to instantiate agents via MLLM-webpage interactions. 
For closed-source agent frameworks, the standardised interface we provide enables their execution without requiring access to the internal invocation process.
This design decouples agent implementation from environment execution, enabling consistent evaluation across different systems.

\subsection{Applications}
\label{sec:3.2}
\vspace{-0.25cm}
\begin{table*}[t]
\footnotesize
\centering
\caption{\textbf{Applications in GauntletBench.} We provide anonymous links for these applications, along with their corresponding descriptions and targeted capability categories.}
\label{tab:benchmark_apps}
\renewcommand{\arraystretch}{1.25} 
\setlength{\tabcolsep}{5pt}        

\begin{tabular}{p{3.1cm} p{6.8cm} p{2.9cm}}
    \toprule
    \textbf{Application} & \textbf{Description} & \textbf{Target Capabilities} \\
    \midrule

    \href{https://gauntlet-video-editor-app.vercel.app/}{Video Editor} &
    A timeline-based media editing interface for previewing clips and performing common editing operations, including trimming, audio overlay, colour adjustment of video segments, transition effects, and text insertion. &
    \parbox[t]{2.9cm}{\raggedright Temporal perception;\\Graphical understanding} \\[0.3em]

    \href{https://gauntlet-workflow-builder-app.vercel.app/workflows}{Workflow Builder} &
    A node-based workflow builder in which users build automation pipelines by adding, connecting, and configuring nodes on a visual canvas, including trigger, branching, data, API, system, and AI nodes. &
    Graphical Understanding \\[0.3em]

    \href{https://gauntlet-3d-modeller-app.vercel.app/editor}{3D Modeller} &
    An interactive 3D scene workspace for navigation, object inspection, and spatial manipulation, including translation, rotation, scaling, and foreground property adjustment. &
    Graphical understanding; 3D reasoning \\[0.3em]
    
    \href{https://gauntletbench-flight-analyser-app.hf.space}{Flight Analyser} &
    A flight radar map for searching aircraft, replaying historical traffic, inspecting trajectories, and extracting telemetry from detailed flight panels. &
    \parbox[t]{2.9cm}{\raggedright Temporal perception;\\Graphical understanding} \\[0.3em]

    \href{https://gauntlet-circuit-app.up.railway.app/}{Circuit Designer} &
    A browser-based circuit environment, adapted from CircuitJS1~\cite{sharpcircuitjs1}, for composing and editing circuits consisting of logical and electrical components. &
    Graphical understanding \\
    \bottomrule
\end{tabular}
\vspace{-1.5em}


\end{table*}
To faithfully evaluate agent generalisation in complex real-world scenarios, our benchmark is designed to assess previously underexplored capabilities while introducing less familiar interfaces.
To achieve these objectives, we identify five less-covered professional applications spanning diverse domains: \texttt{Video Editor}, \texttt{Workflow Builder}, \texttt{3D Modeller}, \texttt{Flight Analyser}, and \texttt{Circuit Design}.

As shown in Table~\ref{tab:benchmark_apps}, these applications not only support the evaluation of basic capabilities, such as UI understanding, tool use, and long-term reasoning, but are also well-suited for assessing our targeted capabilities: temporal perception, graphical understanding, and 3D reasoning.
Moreover, the interfaces of our applications differ substantially from those of popular applications, further supporting the evaluation of generalisation, as illustrated in Figure~\ref{fig:benchmark_apps}.
These characteristics make our applications an effective testbed for assessing the limitations of modern agentic systems.

\textbf{Website Technology Stack}\hspace*{2mm}All applications in our benchmark are implemented using modern web technologies to ensure realistic interaction patterns. 
Specifically, the websites are implemented using JavaScript frameworks (e.g., React or Next.js), while their visual interfaces are rendered using canvas-based techniques.
Due to the lack of structured information exposed by canvas-based interfaces, structured semantic observations are available only in two of the five applications. 
This implementation not only ensures consistent interaction logic, but also enables a genuine evaluation of the targeted visual perception capabilities, as the structured observations heavily relied upon by previous LLM-based agents~\cite{garg2025realbenchmarkingautonomousagents} are not directly available.

\subsection{Tasks}
\label{sec:3.3}
\vspace{-0.25cm}
Our benchmark provides 27 challenging yet human-feasible tasks for each application, totalling 135 tasks. 
Each task is designed to evaluate one or more agent capabilities, either basic ones or those targeted by GauntletBench.
We further support customised task design by enabling users to define tasks and upload ground truth, allowing seamless integration with our automated evaluation engine.


\textbf{Task Complexity}\hspace*{2mm}We categorise tasks into three levels: \textit{easy}, \textit{medium}, and \textit{hard}.
For each application, we construct 9 easy, 9 medium, and 9 hard tasks.
Easy tasks require agents to follow explicit and straightforward instructions, whereas medium tasks involve composing multiple features or functionalities within the environment, requiring agents to coordinate several operations to achieve the goal. In contrast, hard tasks demand more complex capabilities, including multi-step reasoning, fine-grained understanding, and the combination of multiple capabilities.
We provide detailed descriptions of all 135 tasks in {Appendix~\ref{appendix:tasks}}.

\textbf{Task Structure}\hspace*{2mm}Each task follows a standardised three-level structure: \textit{application background}, \textit{task description}, and \textit{result format}.
The application background consists of three components: an application overview (describing the application’s purpose), features \& capabilities (describing the available functionalities), and environment architecture \& interaction paradigms (defining agent interaction with the interface).
The task description specifies the objective through a clearly defined goal (describing the high-level objective) and steps (enumerating the conditions required for successful task completion).
The result format defines the expected output format and evaluation target. 
We consider two types of output formats: information retrieval and action-based. 
For information retrieval tasks, it specifies the exact output format and required content. 
For action-based tasks, agents are required to export the application state as a JSON file or return a completion signal (e.g., a \texttt{done} JSON object) upon task completion.
The detailed three-level structure and background information for each application are provided in {Appendix~\ref{app:task_design_format}}.

\subsection{Evaluation Engine and Metrics}
\vspace{-0.25cm}
\label{sec:3.4}

\begin{figure}[t!]
    \centering
    \includegraphics[width=0.90\linewidth]{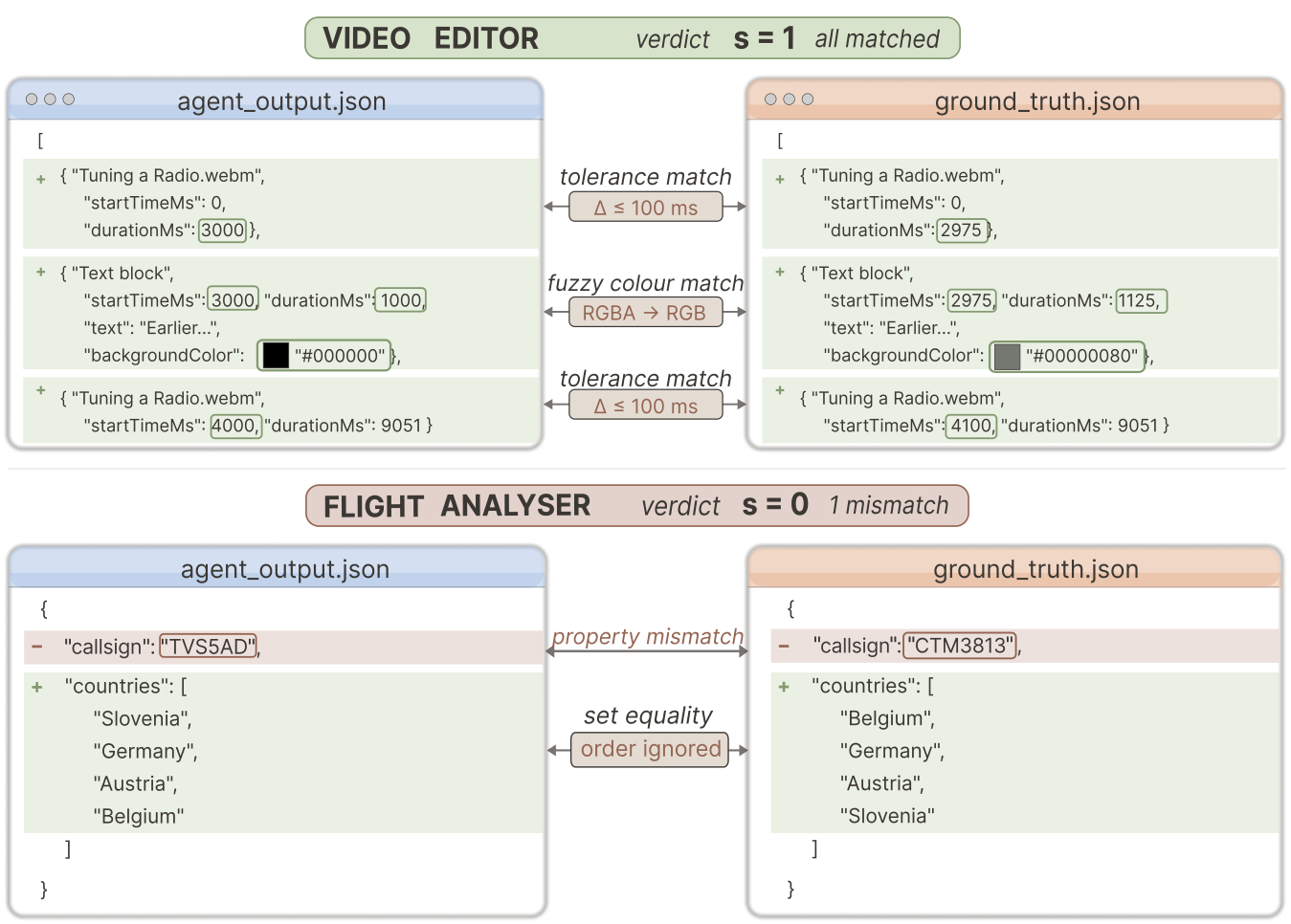}
    \caption{\textbf{Tailored Evaluator.} Each application uses a dedicated comparator that aligns agent output with ground truth via task-aware rules rather than strict string equality. \texttt{Flight Analyser} (top) combines exact property matching with order-invariant set equality on list fields, while \texttt{Video Editor} (bottom) applies $\Delta \leq 100$\,ms timing tolerance and format-normalised fuzzy colour matching, enabling human-aligned yet rigorous scoring.}
    \label{fig:tailored_evaluator}
\vspace{-1.5em}
\end{figure}

Our evaluation engine supports fully automated evaluation for both open- and closed-source agent frameworks with diverse evaluation metrics. 
To achieve this, we embed a tailored evaluator for each application, enabling the system to automatically execute the corresponding evaluation once an agent signals task completion.
This design not only improves compatibility by enabling evaluation even for closed-source agent frameworks, but also enables more flexible and sophisticated evaluation metrics.
As shown in Figure~\ref{fig:tailored_evaluator}, the evaluator for \texttt{Video Editor} allows a 100 ms tolerance when matching start times and video durations between agent outputs and ground truth, as even human annotators cannot perform such edits with frame-level precision. 
Our evaluator also supports fuzzy matching for order-invariant tasks and colour differences.
More importantly, this tailored evaluator provides more fine-grained localisation and detection of actual errors and their variants, such as property mismatches in \texttt{Flight Analyser}.
This design makes success rate assessment more reasonable and human-aligned without compromising rigour, and similarly extends to progress rate and efficiency metrics.
More case analyses of tailored evaluators on different applications can be found in {Appendix~\ref{app:app-specific-evaluation}}.

\textbf{Success Rate}\hspace*{2mm}We define the success rate (SR) as the fraction of tasks in which the agent successfully completes the task according to the tailored objective evaluator.
Formally, let $\mathcal{D} = \{d_1, \dots, d_N\}$ denote the set of $N$ tasks and let $\mathds{1}[\cdot]$ be the indicator function. 
For each task $d_i$, the objective evaluator returns a binary label $s_i \in \{0, 1\}$, where $s_i = 1$ if and only if the agent output matches the ground truth across all checkpoints.
The SR is then given by $    \mathrm{SR} = \frac{1}{N} \sum_{i=1}^{N} \mathds{1}[s_i = 1]$. 
This metric is a strict all-or-nothing criterion that reflects the agent’s ability to fully satisfy all the requirements of the given tasks.

\textbf{Progress Rate}\hspace*{2mm}
To capture meaningful intermediate progress on tasks that the agent partially completes, we additionally report the progress rate (PR).
For each task $d_i$, an MLLM-based judge assigns a score $p_i \in \{1, 2, 3, 4, 5\}$ based on visible final state and trajectory evidence, where $1$ denotes no meaningful progress and $5$ denotes fully successful completion.
The detailed scoring criteria are described in {Appendix~\ref{app:llm-as-judge}}.
The PR is computed as the mean judge score $\mathrm{PR} = \frac{1}{N} \sum_{i=1}^{N} p_i$. 
Unlike the SR, the PR rewards agents that make substantial progress toward the goal, providing a more fine-grained view of agent capability.

\textbf{Efficiency Metrics}\hspace*{2mm}In addition to effectiveness, efficiency is another important aspect of agent performance.
We report the following efficiency metrics: consumed tokens and consumed steps.
Let $\mathcal{D} = \{d_1, \dots, d_N\}$ denote the set of $N$ tasks.
For each task $d_i$, let $c_i$ be the total number of tokens consumed by the agent, $a_i$ the number of interaction steps, and $s_i \in \{0,1\}$ the binary success indicator returned by the objective evaluator. 
consumed tokens (CT) is defined as $\mathrm{CT} = \frac{1}{N} \sum_{i=1}^{N} c_i$. 
consumed steps (CS) is defined as $\mathrm{CS} = \frac{1}{N} \sum_{i=1}^{N} a_i$.

%% file: 3_Experiments.tex
\section{Experiment}
\vspace{-0.25cm}

\subsection{Evaluation Setup}
\vspace{-0.25cm}
\label{section:4_1}
We evaluate \name{} using 14 frontier agents, spanning open-source and API-based MLLM agents as well as closed-source agent frameworks, assessing both task performance and efficiency. The details of human annotation are provided in {Appendix~\ref{app:human-evaluation}}.

\textbf{Open-Source MLLM Agents}\hspace*{2mm}We evaluate a diverse set of widely used open-source MLLMs, including Gemma-3~\cite{gemmateam2025gemma3technicalreport}, Mistral-Large-3~\cite{mistral_mistral3_2025}, Qwen3-VL~\cite{bai2025qwen3vltechnicalreport}, and Llama-4-Maverick~\cite{meta_llama4_2026}.
To further verify the vision-intensive nature of our tasks, we additionally evaluate text-only LLM agents, including Qwen3~\cite{yang2025qwen3technicalreport} and DeepSeek-V3.2~\cite{deepseekai2025deepseekv32pushingfrontieropen}, on the \texttt{Video Editor} and \texttt{Workflow Builder}, where minimal structured information is exposed.

\textbf{API-Based MLLM Agents}\hspace*{2mm}We also provide the result from proprietary MLLMs, including Qwen-Max~\cite{qwen3max}, OpenAI o3-pro~\cite{openai_o3pro_2025}, GPT-5.4~\cite{openai_gpt54thinking_systemcard_2026}, Gemini-3.1-Pro~\cite{googledeepmind_gemini31pro_modelcard_2026}, and Claude-Opus-4.6~\cite{anthropic_claudeopus46_systemcard_2026}.

\textbf{Closed-Source Agent Frameworks}\hspace*{2mm}Finally, we evaluate advanced closed-source agent frameworks to further demonstrate the challenge and compatibility of our benchmark, including GPT-5.4 Computer Use (GPT-5.4 CU)~\cite{openai_computer_use_2026}, Gemini Enterprise~\cite{google_gemini_computer_use_2026}, and Claude-Opus-4.6 Computer Use (Claude-Opus-4.6 CU)~\cite{anthropic2024computeruse}.

\begin{table*}[t]
\setlength{\tabcolsep}{3pt} 
\fontsize{6.58}{7.0}\selectfont
\caption{\textbf{Performance of Agents on \name{}.} The success rate (\%) and progress rate (1-5) are reported in the left and right panels, respectively. ``-'' indicates applications that provide screenshot-only access and are therefore unsuitable for LLMs or vision-disabled MLLM agents. Results for each application are computed over 27 tasks and averaged over 3 independent runs. \textbf{Bold} and \underline{underlined} entries denote the best and second-best results, respectively.}

\label{table:main}
\centering
\begin{tabular}{l c c c c c c | c c c c c c}
    \toprule
    \multirow{3}{*}{\vspace{-0.6em}\textbf{Agent}} & \multicolumn{6}{c}{\textbf{Success Rate ($\uparrow$)} }  & \multicolumn{6}{|c}{\textbf{Progress Rate ($\uparrow$)} }\\
    \cmidrule(l{1pt}r{1pt}){2-7}
    \cmidrule(l{1pt}r{1pt}){8-13}
     & {Video} & {Workflow} & {3D} & {Flight} & {Circuit} & \multirow{2}{*}{{Avg.}} & {Video} & {Workflow} & {3D} & {Flight} & {Circuit} & \multirow{2}{*}{{Avg.}}\\
     & {Editor} & {Builder} & {Modeller} & {Analyser} & {Designer} & & {Editor} & {Builder} & {Modeller} & {Analyser} & {Designer} & \\
    \midrule
    \multicolumn{7}{l}{\textcolor{blue}{\textit{Open-source MLLM Agent}}} \\
    \midrule
    Gemma-3-27B &
    $0.0$ &
    $0.0$ &
    $0.0$ &
    $2.5$ &
    $0.0$ &
    $0.0$ &
    $1.9$ &
    $1.8$ &
    $2.2$ &
    $1.9$ &
    $1.0$ &
    $1.8$ \\

    Mistral-Large-3 &
    $0.0$ &
    $1.2$ &
    $1.2$ &
    $2.5$ &
    $0.0$ &
    $1.0$ &
    $1.9$ &
    $1.6$ &
    $1.6$ &
    $1.8$ &
    $1.2$ &
    $1.6$ \\

    Qwen3-32B (LLM)&
    $0.0$ &
    $0.0$ &
    - &
    - &
    - &
    $0.0$ &
    $1.7$ &
    $1.4$ &
    - &
    - &
    - &
    $1.6$ \\

    Qwen3-VL-32B &
    $0.0$ &
    $0.0$ &
    $1.2$ &
    $1.2$ &
    $0.0$ &
    $0.5$ &
    $2.0$ &
    $1.9$ &
    $1.7$ &
    $1.9$ &
    $1.0$ &
    $1.7$ \\

    DeepSeek-V3.2 (LLM)&
    $0.0$ &
    $2.5$ &
    - &
    - &
    - &
    $0.9$ &
    $2.0$ &
    $2.0$ &
    - &
    - &
    - &
    $2.0$ \\

    Llama-4-Maverick &
    $0.0$ &
    $0.0$ &
    $0.0$ &
    $3.7$ &
    $0.0$ &
    $0.7$ &
    $2.0$ &
    $1.8$ &
    $1.8$ &
    $2.0$ &
    $1.0$ &
    $1.7$ \\

    \midrule
    \multicolumn{7}{l}{\textcolor{blue}{\textit{API-based MLLM Agent}}} \\
    \midrule

    Qwen-Max &
    $0.0$ &
    $4.9$ &
    $7.4$ &
    $6.2$ &
    $0.0$ &
    $3.7$ &
    $2.0$ &
    $2.6$ &
    $2.6$ &
    $2.4$ &
    $1.0$ &
    $2.1$ \\

    GPT-5.4 (w/o vision)&
    $1.2$ &
    $4.9$ &
    - &
    - &
    - &
    $1.7$ &
    $2.1$ &
    $2.3$ &
    - &
    - &
    - &
    $2.2$ \\

    GPT-5.4 &
    $8.6$ &
    $2.5$ &
    $1.2$ &
    $6.2$ &
    $0.0$ &
    $3.7$ &
    $2.8$ &
    $2.3$ &
    $2.0$ &
    $2.1$ &
    $1.0$ &
    $2.0$ \\

    Claude-Opus-4.6 &
    $24.7$ &
    $\textbf{24.7}$ &
    $14.8$ &
    $\underline{22.2}$ &
    $0.0$ &
    $17.3$ &
    $3.1$ &
    $3.2$ &
    $2.5$ &
    $2.9$ &
    $1.0$ &
    $2.6$ \\

    Gemini-3.1-Pro &
    $4.9$ &
    $9.9$ &
    $\underline{30.9}$ &
    $21.0$ &
    $\textbf{22.2}$ &
    $17.8$ &
    $2.5$ &
    $3.0$ &
    $3.3$ &
    $3.0$ &
    $3.0$ &
    $3.0$ \\

    \midrule
    \multicolumn{7}{l}{\textcolor{blue}{\textit{Closed-source Agent Framework}}} \\
    \midrule

    GPT-5.4 CU &
    $12.3$ &
    $0.0$ &
    $1.2$ &
    $16.0$ &
    $7.4$ &
    $7.4$ &
    $2.9$ &
    $2.3$ &
    $2.5$ &
    $2.6$ &
    $1.2$ &
    $2.3$ \\

    Gemini Enterprise &
    $\underline{37.0}$ &
    $12.3$ &
    $\textbf{59.3}$ &
    $13.6$ &
    $6.2$ &
    $\underline{25.7}$ &
    $3.6$ &
    $3.4$ &
    $3.9$ &
    $2.5$ &
    $2.5$ &
    $3.2$ \\

    Claude-Opus-4.6 CU &
    $\textbf{54.3}$ &
    $\underline{19.8}$ &
    $21.0$ &
    $\textbf{30.9}$ &
    $\underline{14.8}$ &
    $\textbf{28.2}$ &
    $3.9$ &
    $3.2$ &
    $3.3$ &
    $3.3$ &
    $2.7$ &
    $3.3$ \\


    \midrule
    \multicolumn{7}{l}{\textcolor{blue}{\textit{Human Annotation}}} \\
    \midrule
    Human &
    $83.9$ &
    $82.7$ &
    $77.8$ &
    $76.5$ &
    $79.6$ &
    $80.1$ &
    $4.3$ &
    $4.0$ &
    $3.8$ &
    $4.3$ &
    $4.2$ &
    $4.1$ \\
    \bottomrule
\end{tabular}
\vspace{-0.8em}
\end{table*}

\begin{figure*}[t]
    \centering
    \includegraphics[width=\linewidth]{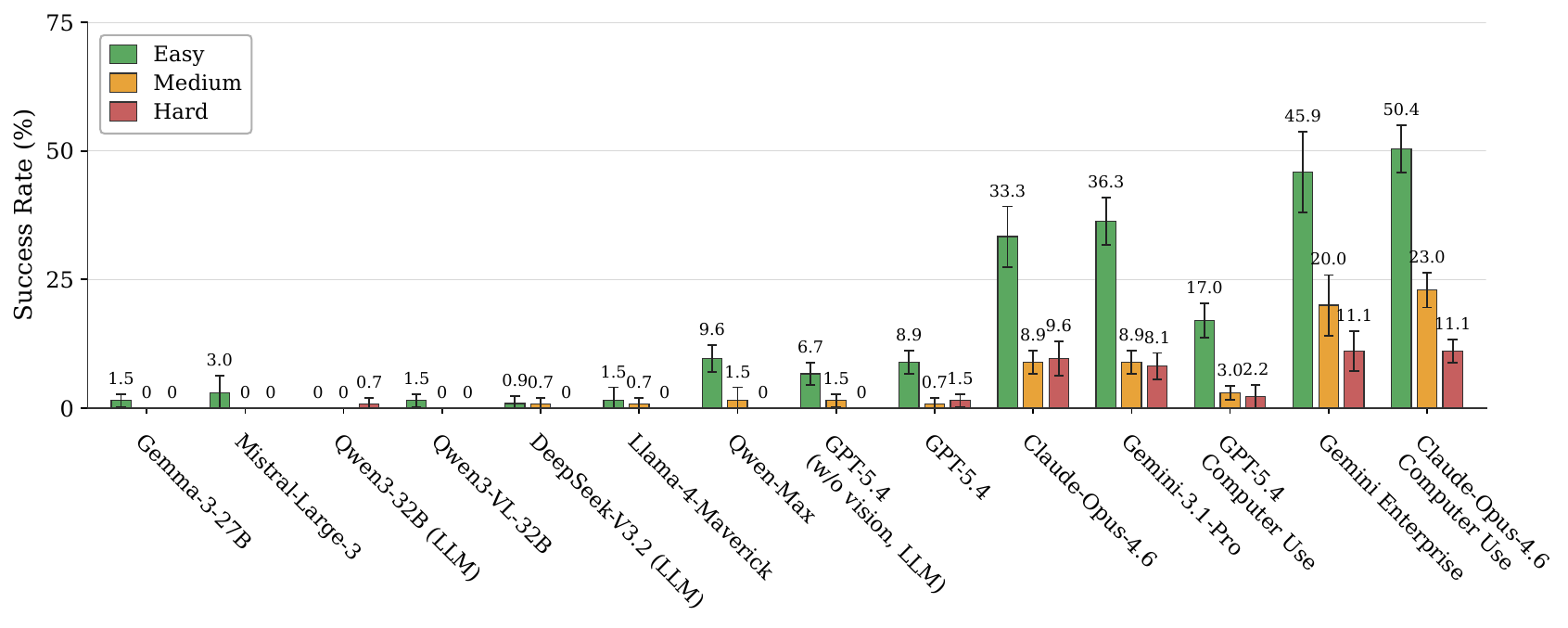}
    \vspace{-0.5em}
    \caption{\textbf{Agent Success Rate Across Task Difficulty Levels.} Our benchmark contains 45, 45, and 45 tasks in the easy, medium, and hard difficulty levels, respectively. Results are averaged over three independent runs.}
    
    \label{fig:agent_difficulty_distribution}
    \vspace{-1.3em}
\end{figure*}

\subsection{Performance Evaluation}
\label{section:4_2}
\vspace{-0.25cm}

\textbf{Frontier Agentic Systems Remain Far from Human-Level Performance}\hspace*{2mm}As shown in Table~\ref{table:main} (left panel), all open-source MLLM agents fail to surpass a $1.7\%$ success rate on any single application, while most evaluated systems achieve $0.0\%$ across all applications.
The strongest agentic system, Claude-Opus-4.6 Computer Use, achieves only a $28.2\%$ average success rate, followed by Gemini Enterprise at $25.7\%$ and Gemini-3.1-Pro at $17.8 \%$.
In contrast, non-expert human participants achieve between $75\%$ and $85\%$ across all applications, indicating that the benchmark is challenging yet feasible.
This substantial gap revealed by GauntletBench not only more faithfully reflects agent capabilities in complex real-world scenarios, but also exposes the limitations of existing benchmarks in evaluating the challenges and capabilities required for modern agents.

\textbf{How Far Current Agents Remain from Reliable Performance?}\hspace*{2mm}From Table~\ref{table:main} (right panel), we can observe that nearly all agents exhibit meaningful intermediate progress toward task completion, as reflected by greater than 1 progress rates.
Claude-Opus-4.6 Computer Use even achieves an average progress rate exceeding 50\% across all tasks, suggesting that failures often stem not from an inability to make progress, but from difficulty reliably handling the accumulating complexity of long-horizon task execution.
This observation is further supported by Figure~\ref{fig:agent_difficulty_distribution}, where agents perform substantially better on easy tasks than on medium and hard ones, highlighting the limited robustness of current agents under increasing task complexity.
These results provide an optimistic perspective: current agents are not entirely incapable of handling our targeted capabilities and challenging tasks, but their existing abilities remain insufficient for reliable performance.

\begin{figure*}[t]
    \centering
    \includegraphics[width=\linewidth]{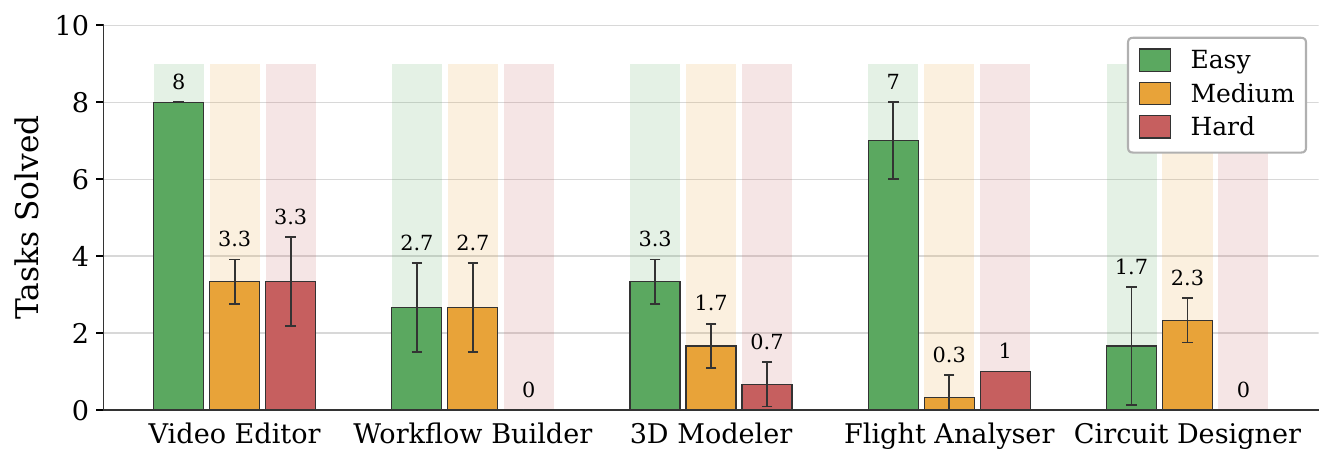}
    \caption{\textbf{Number of Solved Tasks Across Difficulty Levels.} Our benchmark contains 9, 9, and 9 tasks for the easy, medium, and hard difficulty levels of each application, respectively. Results are generated by Claude-Opus-4.6 Computer Use, and averaged over three independent runs.}  \label{fig:app_difficulty_distribution}
    \vspace{-0.3cm}
\end{figure*}

\begin{figure*}[t]
    \centering
    \includegraphics[width=\linewidth]{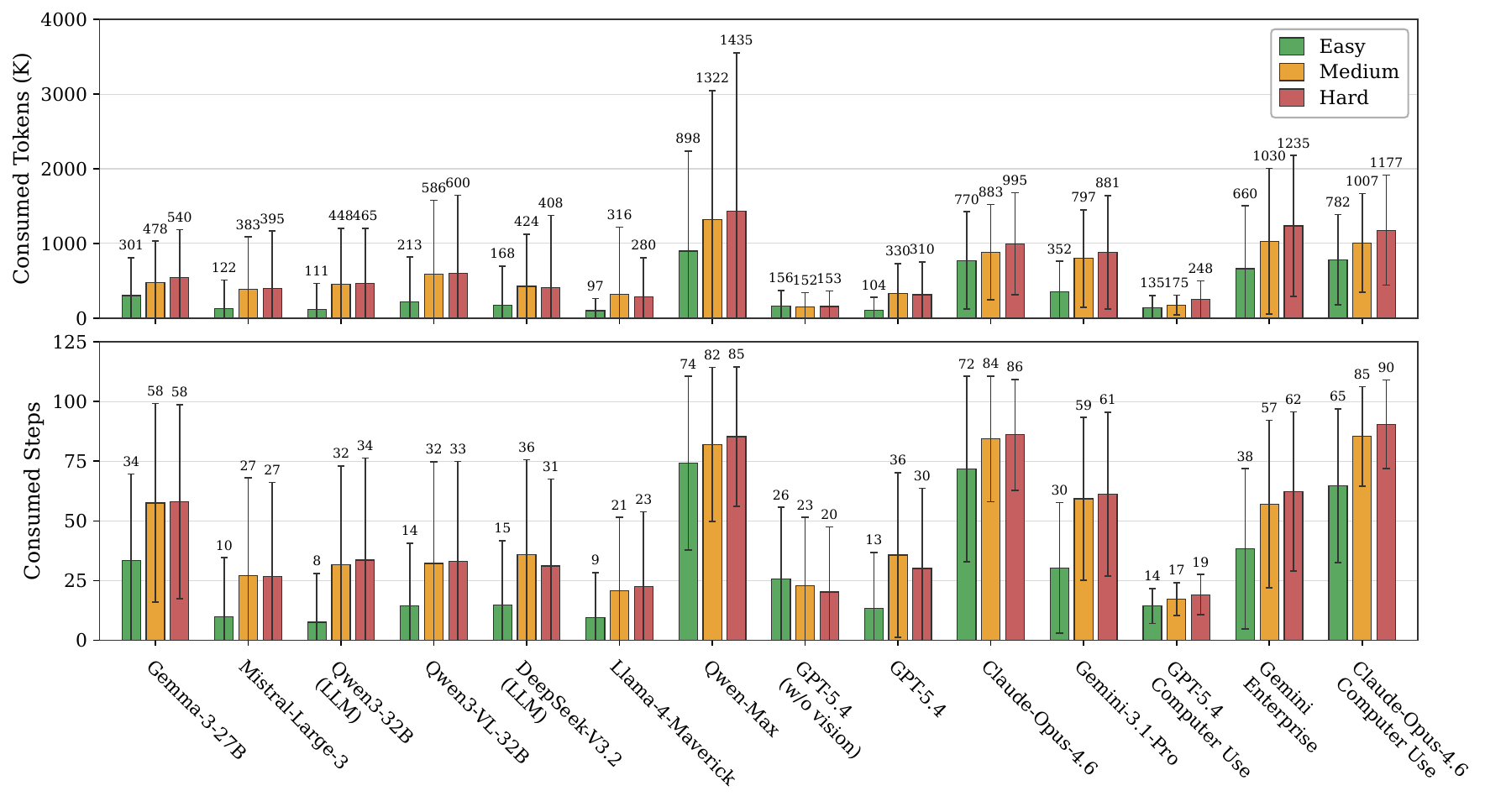}
    \vspace{-0.50cm}
    \caption{\textbf{Efficiency of Agents on GauntletBench.} The consumed tokens and consumed steps are reported in the top and bottom panels, respectively. Results for each application are computed over 27 tasks and averaged over three independent runs.}
    \vspace{-0.5cm}
    \label{fig:efficiency}
\end{figure*}

\textbf{What Capabilities Do Current Agents Lack Most?}\hspace*{2mm}As illustrated in Figure~\ref{fig:app_difficulty_distribution}, the \texttt{3D Modeler} and \texttt{Circuit Designer} are the most challenging applications for current agents, with success rates at 19.2\% and 14.8\%, respectively.
These results suggest that current agents may already possess basic recognition of 3D elements and graphical structures, yet still lack a deeper understanding of their relationships, interactions, and resulting behaviours.

\textbf{Does the Vision Modality Help Agents Complete Tasks?}\hspace*{2mm}To understand whether MLLMs genuinely rely on vision to solve our intentionally vision-intensive tasks, we compare text-only LLM agents on the \texttt{Video Editor} and \texttt{Workflow Builder} applications, where structured information is accessible.
From Table~\ref{table:main} (right panel), we can observe that relying solely on structured information enables agents to complete only a very small subset of tasks. 
In contrast, enabling visual inputs consistently improves agent performance, increasing average progress rates on these two applications by 22.5\% and 13.7\% for the Qwen and GPT families, respectively.

\textbf{Which Reasoning Strategies Are Most Efficient?}\hspace*{2mm}As shown in Figure~\ref{fig:efficiency}, Qwen-Max incurs the highest inference cost in terms of token consumption, yet does not achieve a correspondingly strong success rate. 
In contrast, Claude Opus 4.6 Computer Use achieves the best overall performance with surprisingly low inference cost.
For a horizontal comparison among token-similar models, Claude Opus 4.6 Computer Use consumes significantly more steps than Qwen-3-VL-32B and OpenAI o3-pro. 
While we cannot fully exclude the influence of model scale, intrinsic capabilities, and closed-source framework design, our results suggest that taking more fine-grained and incremental steps, a divide-and-conquer strategy, may be more effective and efficient for challenging tasks.

\subsection{Ablation Studies}
\vspace{-0.25cm}
\label{section:4_3}

\textbf{The Impact of Model Scale}\hspace*{2mm}Table~\ref{table:3} shows that increasing model scale consistently improves both success rate and progress rate.
Notably, the largest model in each family also achieves the lowest token consumption (despite the higher cost of individual inference steps), while maintaining a comparable number of interaction steps; consumed tokens and steps are reported on the medium-difficulty subset, where the comparison across scales is most informative.
These results suggest that, for our challenging tasks, improvements in performance cannot be achieved solely through additional steps, but instead require fundamentally stronger models.

\textbf{The Role of Reasoning}\hspace*{2mm}As shown in Table~\ref{table:4}, enabling extended reasoning generally improves success rates for sufficiently capable base models, but also introduces a trade-off in token consumption and interaction steps, reported here on the medium-difficulty subset.
We additionally evaluate Qwen-3-VL with and without extended reasoning, but observe no improvement in performance.
These results suggest that extended reasoning is most effective when the underlying model has sufficient capacity to benefit from additional deliberation; for weaker models, the extra computation translates into only limited performance gains.

\begin{table}[t]
\setlength{\tabcolsep}{5.2pt}
\small
\caption{\textbf{The Impact of Model Size.} We compare two model families that share an architecture but differ in scale: GPT-5.4 (Nano, Mini, full) and Claude-4.6 (Haiku, Sonnet, Opus). Success rate (\%) and progress rate (1-5) are computed over all 135 tasks. Consumed tokens (in thousands, K) and consumed steps are reported on the 45 medium-difficulty tasks, where most agents make meaningful but not yet saturated progress and the efficiency comparison across scales is therefore most informative. All values are mean $\pm$ standard deviation over three independent runs. \textbf{Bold} and \underline{underlined} entries denote the best and second-best results, respectively.
Results for every difficulty tier task are present in Appendix~\ref{app:additional-results}.}

\label{table:3}
\centering
\begin{tabular}{l c c c c}
    \toprule
    \toprule
    \multirow{1}{*}{Model}
    & Success Rate ($\uparrow$)
    & Progress Rate ($\uparrow$)
    & Consumed Tokens ($\downarrow$)
    & Consumed Steps ($\downarrow$)
    \\
    \midrule
     \multirow{1}{*}{GPT-5.4-Nano}    & $0.0 \pm 0.0$ &  $1.50 \pm 0.04$  & $521 \pm 819$     & $\underline{36.0 \pm 40.7}$ \\

    \multirow{1}{*}{GPT-5.4-Mini}   & $0.0 \pm 0.0$ & $1.35 \pm 0.16$ & $\underline{449 \pm 807}$     & $\textbf{27.3} \pm \textbf{42.7}$ \\

    \multirow{1}{*}{GPT-5.4}        & $2.3 \pm 2.4$  &  $2.14 \pm 0.84$ & $\textbf{330} \pm \textbf{396}$     & $35.7 \pm 34.3$ \\
    \midrule

    \multirow{1}{*}{Claude-Haiku-4.5}  & $1.7 \pm 2.4$ & $2.01 \pm 0.08$ & $1204 \pm 1317$   & $74.3 \pm 38.6$ \\

    \multirow{1}{*}{Claude-Sonnet-4.6}  & $\underline{8.7 \pm 1.2}$  & $\underline{2.21 \pm 0.10}$ & $1496 \pm 3448$   & $84.8 \pm 31.1$ \\

    \multirow{1}{*}{Claude-Opus-4.6}   & $\bm{12.3 \pm 3.3}$& $\bm{2.44 \pm 1.17}$  & $888 \pm 633$     & $85.1 \pm 25.2$ \\
    
    \bottomrule
    \bottomrule
\end{tabular}
\vspace{-1.0em}
\end{table}

\begin{table}[t]
\setlength{\tabcolsep}{3.3pt}
\small
\caption{\textbf{The Role of Reasoning.} We compare pairs of models that share the same base model but employ different levels of reasoning effort: GPT-5.4 with and without extended reasoning, Gemini-3.1-Pro at low versus high reasoning, and Claude-Opus-4.6 with and without thinking. Success rate (\%) and progress rate (1-5) are reported over all 135 tasks. Consumed tokens (in thousands, K) and consumed steps are reported on the 45 medium-difficulty tasks, the regime where the impact of extra deliberation is most cleanly observable. All values are mean $\pm$ standard deviation over three independent runs. \textbf{Bold} and \underline{underlined} entries denote the best and second-best results, respectively. Results for every difficulty tier task are present in Appendix~\ref{app:additional-results}.}

\label{table:4}
\centering
\begin{tabular}{l c c c c c}
    \toprule
    \toprule
    \multirow{1}{*}{Model}
    & Success Rate ($\uparrow$)
    & Progress Rate ($\uparrow$)
    & Consumed Tokens ($\downarrow$)
    & Steps ($\downarrow$)
    \\
    \midrule

    \multirow{1}{*}{GPT-5.4 w/o reasoning} & $0.7 \pm 0.3$ & $1.50 \pm 0.68$ & $\textbf{132} \pm \textbf{226}$ & $\textbf{10.3} \pm \textbf{19.6}$ \\

    \multirow{1}{*}{GPT-5.4} & $2.3 \pm 2.4$ & $2.14 \pm 0.90$ & $330 \pm 396$ & $35.7 \pm 34.3$ \\

    \midrule

    \multirow{1}{*}{Gemini-3.1-Pro Low-reasoning} & $6.3 \pm 1.2$ & $2.38 \pm 0.93$ & $\underline{326 \pm 485}$ & $\underline{28.7 \pm 29.0}$ \\

    \multirow{1}{*}{Gemini-3.1-Pro High-reasoning} & $\bm{13.2 \pm 0.7}$ & $\bm{2.84 \pm 1.00}$ & $815 \pm 652$ & $59.7 \pm 33.6$ \\

    \midrule

    \multirow{1}{*}{Claude-Opus-4.6} & $\underline{12.3 \pm 3.3}$ & $\underline{2.44 \pm 1.17}$ & $888 \pm 633$ & $85.1 \pm 25.2$ \\

    \multirow{1}{*}{Claude-Opus-4.6 Thinking} & $10.3 \pm 1.5$ & $2.37 \pm 1.12$ & $947 \pm 603$ & $92.9 \pm 21.2$ \\
    \bottomrule
    \bottomrule
\end{tabular}
\vspace{-1.0em}
\end{table}

\begin{table}[t]
\setlength{\tabcolsep}{2pt} 
\fontsize{6.7}{7.0}\selectfont
\caption{\textbf{Human Agreement with MLLM-Judged Progress Rate.} We evaluate agreement using five uniformly sampled trajectories for each application. For each application, we report the mean PR, linearly weighted Cohen's $\kappa$, and macro F1, averaged over three independent runs. $^{\star}$Human self-agreement is 1.00 by definition and shown as an upper bound.}
\label{table:human_llm_judge_agreement}
\centering
\begin{tabular}{l c c c c c c c c c c c c c c c c c c}
    \toprule
    \toprule
    \multirow{3}{*}{Judge}
    & \multicolumn{3}{c}{Video}
    & \multicolumn{3}{c}{Workflow }
    & \multicolumn{3}{c}{3D}
    & \multicolumn{3}{c}{Flight}
    & \multicolumn{3}{c}{Circuit }
    & \multicolumn{3}{c}{\multirow{2}{*}{Avg.}} \\
    & \multicolumn{3}{c}{Editor}
    & \multicolumn{3}{c}{Builder}
    & \multicolumn{3}{c}{Modeller}
    & \multicolumn{3}{c}{Analyser}
    & \multicolumn{3}{c}{Designer}  \\
    \cmidrule(lr){2-4}
    \cmidrule(lr){5-7}
    \cmidrule(lr){8-10}
    \cmidrule(lr){11-13}
    \cmidrule(lr){14-16}
    \cmidrule(lr){17-19}
    & PR & $\kappa$ & F1
    & PR & $\kappa$ & F1
    & PR & $\kappa$ & F1
    & PR & $\kappa$ & F1
    & PR & $\kappa$ & F1
    & PR & $\kappa$ & F1 \\
    \midrule
    GPT-5.1 (w/ stage-1)
    & \textbf{2.87} & \textbf{0.94} & \textbf{0.71}
    & \textbf{2.73} & \textbf{0.65} & \textbf{0.41}
    & \textbf{1.87} & 0.43 & 0.24
    & \textbf{2.40} & \textbf{0.44} & 0.20
    & \textbf{1.40} & \textbf{1.00} & \textbf{0.67}
    & \textbf{2.25} & \textbf{0.73} & \textbf{0.58} \\

    GPT-5.1 (w/o stage-1)
    & 2.53 & 0.70 & 0.48
    & 2.47 & 0.56 & 0.29
    & \textbf{1.87} & \textbf{0.56} & 0.33
    & 2.47 & 0.56 & 0.29
    & \textbf{1.40} & \textbf{1.00} & \textbf{0.67}
    & 2.05 & 0.62 & 0.48 \\

    Claude Opus 4.6

    & 2.13 & 0.53 & 0.21
    & 2.07 & 0.36 & 0.04
    & 1.33 & 0.46 & \textbf{0.35}
    & 1.87 & 0.43 & \textbf{0.31}
    & 1.27 & 0.74 & 0.44
    & 1.73 & 0.54 & 0.35 \\

    Claude Sonnet 4.6

    & 2.00 & 0.41 & 0.08
    & 2.13 & 0.40 & 0.21
    & 1.40 & 0.09 & 0.17
    & 1.80 & 0.38 & 0.29
    & 1.20 & 0.62 & 0.33
    & 1.71 & 0.44 & 0.25 \\

    Human Evaluation
    & 2.80 & 1.00$^{\star}$ & 1.00$^{\star}$
    & 3.00 & 1.00$^{\star}$ & 1.00$^{\star}$
    & 1.80 & 1.00$^{\star}$ & 1.00$^{\star}$
    & 2.60 & 1.00$^{\star}$ & 1.00$^{\star}$
    & 1.40 & 1.00$^{\star}$ & 1.00$^{\star}$
    & 2.32 & 1.00$^{\star}$ & 1.00$^{\star}$ \\
    
    \bottomrule
    \bottomrule
\end{tabular}
\vspace{-0.8em}
\end{table}

\textbf{Accuracy of MLLM-Judged Progress Rate}\hspace*{2mm}To validate the accuracy of the MLLM-judged Progress Rate, we collect human-in-the-loop annotations using a 1–5 rubric on five uniformly sampled trajectories per application (25 in total), and compare them with the stage-2 (final-outcome) judge.
As shown in Table~\ref{table:human_llm_judge_agreement}, all evaluated MLLM judges produce results highly consistent with human annotations, suggesting that they provide reliable assessments.
Moreover, GPT-5.1 achieves the strongest overall agreement and is therefore used for large-scale evaluation.





\subsection{Error Analysis}
\label{section:4_4}
\vspace{-0.25cm}
We manually inspected a sample of failed trajectories and identified four recurring error categories, detailed in Appendix~\ref{appendix:error_analysis}: (1) task-completion misjudgment, where agents either corrupt an otherwise correct result by continuing to act or terminate prematurely; (2) grounding failures, where actions are applied to incorrect UI elements without being detected by the agent; (3) protocol violations, ranging from outputting an entire plan without executing actions to batching multiple steps into a single response; and (4) systematic failures in open-source models, including syntactically invalid outputs and infinite action loops, likely caused by the lack of agentic browser-control data during fine-tuning.

\textbf{How Far Do Failures Get?}\hspace*{2mm}To quantify these categories, we cross-tabulate the progress rate against the binary success label over all 4,500 evaluated trajectories (Appendix~\ref{app:error_quantitative}). Of the 4,299 failures, 97.2\% fall in the $\mathrm{PR}=1$-$3$ band, indicating that most failed runs break down before achieving substantial progress. The remaining tail is concentrated in the strongest systems: within its own failures, Claude-Opus-4.6 CU scores $\mathrm{PR}=1$ on only 2 of 244 cases (0.8\%) while 60.7\% reach $\mathrm{PR}\geq3$, whereas weaker agents such as Mistral-Large-3 collapse immediately on 51.3\% of theirs. This separates two failure regimes: weaker agents fail for perceptual and comprehension reasons before making progress, while stronger agents fail at the execution level after substantial interaction. Consistently, protocol violations are confined to weaker agents — action-free trajectories account for 15.1\% of Qwen3-32B and 13.7\% of OpenAI o3-pro runs, but never occur for Claude-Opus-4.6, Claude-Opus-4.6 CU, GPT-5.4 CU or Gemini Enterprise (Appendix~\ref{app:error_protocol}).

%% file: 4_Related_Works.tex
\section{Related Work}
\vspace{-0.25cm}
Prior work on computer-use agents has developed benchmarks for web, mobile, and desktop interaction, ranging from synthetic browser tasks to realistic websites, operating-system environments, and enterprise workflows~\cite{pmlr-v70-shi17a,DBLP:journals/corr/abs-1802-08802,zhou2024webarenarealisticwebenvironment,koh2024visualwebarenaevaluatingmultimodalagents,he2024webvoyagerbuildingendtoendweb,deng2023mind2webgeneralistagentweb,xie2024osworldbenchmarkingmultimodalagents,rawles2025androidworlddynamicbenchmarkingenvironment,bonatti2024windowsagentarenaevaluating,drouin2024workarenacapablewebagents,garg2025realbenchmarkingautonomousagents}.
These benchmarks have been central to measuring progress, but much of the current evaluation landscape still emphasizes consumer-style information seeking, navigation, and form-based workflows.
As agents improve, such settings increasingly risk saturating or underrepresenting the perceptual and spatial reasoning demands of professional software.

\textbf{Web Agent Benchmarks}\hspace*{2mm}
A growing number of work has focused on evaluating autonomous agents in web-based environments.
Early benchmarks such as MiniWoB~\cite{pmlr-v70-shi17a} and MiniWoB++~\cite{DBLP:journals/corr/abs-1802-08802} introduced simplified web interaction tasks where agents had to perform basic operations like clicking buttons, filling forms, and navigating menus.
While these benchmarks were instrumental in establishing the field, they relied on synthetic, toy-like interfaces that did not capture the complexity of real websites.
More recent efforts are aiming to close this gap.
WebArena~\cite{zhou2024webarenarealisticwebenvironment} proposed a suite of realistic, self-hosted web environments spanning e-commerce, forums, and content management systems, enabling agents to be evaluated on tasks that more closely resemble real-world usage.
VisualWebArena~\cite{koh2024visualwebarenaevaluatingmultimodalagents} extended this idea by emphasizing visually grounded tasks that require agents to reason over rendered page content rather than relying solely on HTML structure.
WebVoyager~\cite{he2024webvoyagerbuildingendtoendweb} took a different approach by deploying agents on live websites, which introduces non-determinism but better reflects practical deployment conditions.
Mind2Web~\cite{deng2023mind2webgeneralistagentweb} contributed a large-scale dataset of real-world web tasks annotated with action sequences across diverse domains.
REAL~\cite{garg2025realbenchmarkingautonomousagents} proposed deterministic simulations of real websites to combine realism with reproducibility.
AssistantBench~\cite{yoran2024assistantbenchwebagentssolve} evaluated agents on open-ended web tasks requiring information synthesis.
Despite this progress, most existing web benchmarks focus on consumer-level scenarios such as online shopping or information lookup, and they primarily test basic UI navigation and form-filling skills.
Our work differs in that we target professional-level tasks that require higher-order capabilities like dynamic perception, graphical understanding, and 3D spatial reasoning, which remain largely unexplored in prior benchmarks.

\textbf{OS Agent Benchmarks}\hspace*{2mm}
A neighbouring line of work looks at agents that operate at the operating system level rather than inside a web browser.
While this is not directly related to our work, the challenges are similar in many ways, and progress in one area often signals progress in the other.
OSWorld~\cite{xie2024osworldbenchmarkingmultimodalagents} sets up a benchmark where agents control full desktop environments (Ubuntu, Windows, macOS) through keyboard and mouse actions across many different applications.
AndroidWorld~\cite{rawles2025androidworlddynamicbenchmarkingenvironment} and AndroidEnv~\cite{toyama2021androidenvreinforcementlearningplatform} focus on mobile systems, asking agents to navigate apps and complete tasks on Android.
WindowsAgentArena~\cite{bonatti2024windowsagentarenaevaluating} targets productivity tasks inside the Windows operating system. CRAB~\cite{xu2025crabcrossenvironmentagentbenchmark} further broadens this direction by evaluating multimodal agents across desktop and mobile environments with graph-based fine-grained evaluation.
OS-level benchmarks cover a very wide range of agent skills, but they can be harder to reproduce and control than web environments, since desktop state depends on many moving parts.
Our benchmark goes in a different direction: we stay inside the browser, where we get cleaner control and reproducibility, while still picking tasks that require reasoning well beyond simple point-and-click interactions.

\textbf{Domain-Specific Multimodal Benchmarks}\hspace*{2mm}
Several benchmarks probe the perceptual competences we target, but do so by querying a model rather than by having an agent act.
MVBench~\cite{li2024mvbench} and VEU-Bench~\cite{li2025veubench} evaluate temporal and editing understanding through questions over pre-recorded videos; FlowVQA~\cite{singh2024flowvqa} asks questions about static flowchart images; MapEval~\cite{dihan2025mapeval} poses geospatial questions over map text, APIs, and snapshots; and MMCircuitEval~\cite{zhao2025mmcircuiteval} tests circuit knowledge through question--answer pairs.
Their consistently low scores confirm that these capabilities are hard for current MLLMs, but they measure perception in isolation: the model never has to translate what it perceives into a sequence of grounded actions on a stateful interface, verify the outcome, and recover from its own mistakes.
Closer to the agentic setting, ScreenSpot-Pro~\cite{li2025screenspotpro} evaluates single-step element grounding on high-resolution professional screenshots, BlenderGym~\cite{gu2025blendergym} evaluates 3D editing through Blender's Python API rather than its GUI, and Spider2-V~\cite{cao2024spider2v} evaluates full workflows in professional data tooling, where GUI interaction centres on forms, spreadsheets, and IDEs, and pipelines in its orchestration tools are defined in code rather than on a visual canvas.
\name{} sits in the gap between these lines: it embeds temporal, graphical, and 3D perception inside closed-loop, multi-step GUI interaction, where a single mis-grounded action compounds over a long horizon (Appendix~\ref{appendix:error_analysis}).
We therefore regard these benchmarks as complementary evidence of the same underlying deficit, and state our claim precisely: the targeted capabilities are underexplored not in multimodal understanding benchmarks at large, but in agentic benchmarks in particular.

\textbf{Agentic Evaluation}\hspace*{2mm}
A key challenge in evaluating autonomous agents is designing evaluation pipelines that are both fair and informative.
Traditional approaches rely on exact-match comparisons between agent outputs and predefined ground truths~\cite{drouin2024workarenacapablewebagents}, which works reasonably well for simple retrieval tasks but becomes problematic for complex, open-ended scenarios where multiple valid solutions may exist.
REAL~\cite{garg2025realbenchmarkingautonomousagents} addressed this by introducing deterministic website simulations paired with step-wise trajectory tracing, implementing reproducible evaluation of agents through screenshots and action logs.
Their framework decouples agent implementation from environment execution, which we adopt and extend in our work to support evaluation of both open-source models and closed-source agent frameworks through a unified interface.
More recently, the use of LLMs as automated judges has gained popularity as a way to assess partial task completion~\cite{chiang2024chatbotarenaopenplatform}.
Instead of only checking whether the final output is exactly correct, an LLM judge can inspect the agent's trajectory and the final visible state to assign a graded score that reflects how much progress was actually made.
This is especially useful for tasks where an agent might get most of the way to the goal but fail on a small detail.
However, LLM-based judges can be inconsistent and may hallucinate progress that did not actually occur~\cite{zheng2023judgingllmasajudgemtbenchchatbot}, so careful prompt design and calibration are needed.
Some works have also explored hybrid evaluation strategies that combine rule-based checks with model-based assessment~\cite{koh2024visualwebarenaevaluatingmultimodalagents}, taking the best of both worlds.
In our benchmark, we implement a two-stage evaluation pipeline: a structured exact-match evaluator that checks domain-specific constraints, and an LLM-based judge that provides partial success scores based on visual trajectory evidence.
We additionally report efficiency-oriented metrics including token consumption, latency, and number of interaction steps to give a more complete view of agent performance beyond just task accuracy.

\textbf{Saturation of Computer-Use Agent Benchmarks}\hspace*{2mm}
Agent systems have improved a lot in a short time, and many of the earlier benchmarks are starting to look too easy.
On MiniWoB++~\cite{NEURIPS2023_7cc1005e}, modern agents score close to perfect on most tasks.
Even on harder benchmarks like WebArena~\cite{zhou2024webarenarealisticwebenvironment}, the gap between top agents and humans is getting smaller every year. This trend suggests that soon current benchmarks might become insufficient for differentiating performance of strong systems.
Some recent work attempted to address it.
For example, WorkArena~\cite{drouin2024workarenacapablewebagents} targets enterprise software tasks on the ServiceNow platform, and REAL~\cite{garg2025realbenchmarkingautonomousagents} focuses on deterministic simulations of real websites that require longer interaction traces.
But most of these benchmarks still test mostly text-heavy, form-based interactions.
They do not really assess whether an agent can understand dynamic visual content, read graphical layouts, or reason about 3D scenes.
Our benchmark targets exactly these underexplored skills, and our experiments show that even the strongest frontier models still perform far below human evaluators on these tasks.

\name{} is complementary to this line of work.
We keep the reproducibility advantages of browser-based environments while shifting the task distribution toward visually dense, dynamic, and specialist interfaces, including graphical editing, circuit design, flight tracking, and 3D manipulation.
Our evaluation also follows recent work on richer agentic assessment by combining structured checks with model-based partial-credit judging and efficiency metrics~\cite{chiang2024chatbotarenaopenplatform,zheng2023judgingllmasajudgemtbenchchatbot}.

%% file: 5_Conclusion.tex
\section{Conclusion and Discussion}
\vspace{-0.25cm}
In this work, we introduce \textit{GauntletBench}, a challenging web-based benchmark for evaluating agent generalisation in complex real-world scenarios. 
Our benchmark targets previously overlooked capabilities, including temporal perception, graphical understanding, and 3D reasoning, through diverse professional applications and carefully designed vision-intensive tasks. 
Empirical results across a broad range of frontier agentic systems reveal that current agents still remain far from reliable real-world performance. 
By focusing on challenging yet feasible tasks grounded in realistic environments, \name{} exposes substantial limitations of existing agentic systems while providing a more faithful assessment of their real-world capabilities. 
We hope \name{} can facilitate future research on robust perception, reasoning, planning, and interaction for next-generation agentic systems. 

\textbf{Limitations}\hspace*{2mm}Despite providing a challenging and professional benchmark for evaluating agent generalisation, GauntletBench covers only five specialised domains and a subset of interaction paradigms, which may not fully capture the diversity of real-world scenarios and could potentially introduce evaluation bias. In addition, our benchmark constrains agents' capabilities in visually grounded interaction to ensure faithful evaluation, which may limit their full potential when combined with more observational challenges.

\textbf{Future Works}\hspace*{2mm}We plan to expand \name{} to more diverse professional applications, richer multimodal interaction settings, and longer-horizon tasks. We also aim to incorporate additional evaluation dimensions, such as robustness, safety, and failure recovery, to better reflect the requirements of real-world deployment.

\textbf{Broad Impacts}\hspace*{2mm}We believe \name{} can support the development of more robust, reliable, and trustworthy agentic systems by providing realistic and challenging evaluation settings for assessing agent generalisation. By promoting more rigorous evaluation practices, our benchmark may facilitate future research on safer and more dependable autonomous agents in real-world applications.


%% file: A_Appendix.tex
\section{Human Evaluation}
\label{app:human-evaluation}

\textbf{Harness Overview}\hspace*{2mm}We built a custom web-based \textbf{Human Evaluation Harness} that mirrors the agent's interaction surface: the participant works directly inside the same web application an agent would control. The harness only adds an instrumentation layer around the application --- task delivery, screenshot capture, and answer collection --- so human trajectories are structurally same as agent trajectories.

\begin{figure}[ht]
    \centering
    \includegraphics[width=0.9\textwidth]{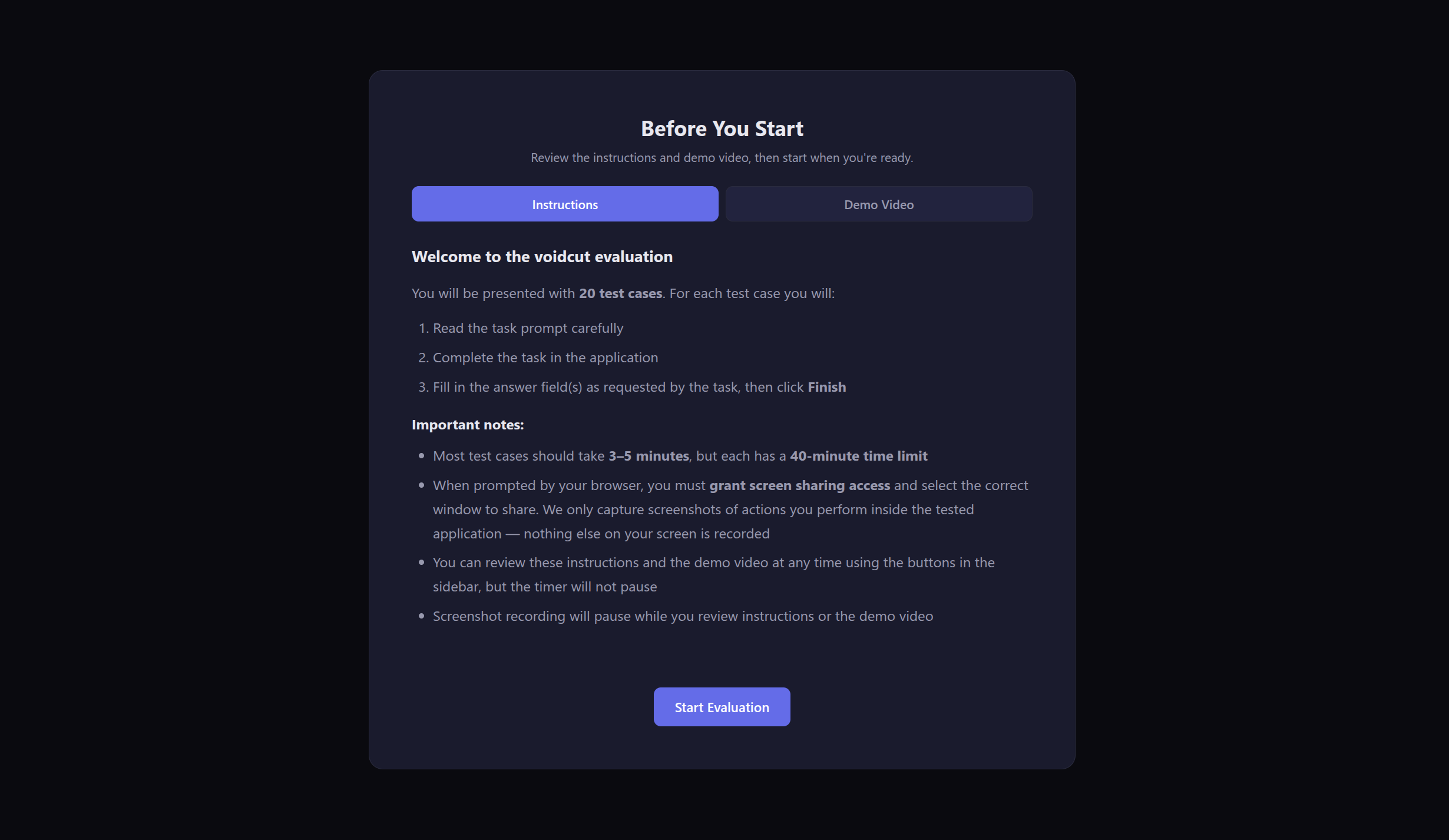}
    \caption{\textbf{Human Evaluation Harness: Pre-Session Briefing.} Before starting the first task, each participant is shown written instructions describing the application and the evaluation procedure, alongside an embedded demo video that illustrates basic interactions with the application. Both resources remain accessible during every task via the \emph{Recall Instructions} and \emph{Recall Demo Video} buttons; screenshot recording is paused while either is open to avoid leaking instructional content into the captured trajectory.}
    \label{fig:harness-briefing}
\end{figure}

The participant flow has three stages:
\begin{itemize}
    \item \textbf{Authentication}\hspace*{2mm}Each participant enters an application-specific access code that allows a single evaluation session.

    \item \textbf{Pre-session screen}\hspace*{2mm}Before the first task, the participant sees written instructions and a short demo video of the application (Figure~\ref{fig:harness-briefing}; the per-application videos are listed in Table~\ref{tab:demo-videos}). Both remain available during every task via \emph{Recall Instructions} and \emph{Recall Demo Video} buttons; screenshot recording is paused while either is open.

    \item \textbf{Session workspace}\hspace*{2mm}During each task, the participant sees the application in the center, a header with the timer, task index, and recording status, and a right-hand sidebar with the task prompt and answer fields (Figure~\ref{fig:harness-workspace}). The participant fills in the answer field(s) and clicks \textbf{Finish} to proceed.
\end{itemize}

\begin{table}[h]
    \centering
    \caption{\textbf{Demo Videos Used in Human-evaluation.} Each video provides a short walkthrough of the corresponding application interface, helping participants understand the available controls without revealing solutions to benchmark tasks.}
    \label{tab:demo-videos}
    \begin{tabular}{ll}
        \toprule
        \textbf{Application} & \textbf{Demo Video} \\
        \midrule
        Video Editor   & \url{https://www.youtube.com/watch?v=zS_8YdKwsx8} \\
        Workflow Builder            & \url{https://www.youtube.com/watch?v=8yWhj0ivqPY} \\
        3D Modeller      & \url{https://www.youtube.com/watch?v=Yr1hH7tWp28} \\
        Flight Analyser & \url{https://www.youtube.com/watch?v=CQsKjizsID4} \\
        Circuit Designer & \url{https://www.youtube.com/watch?v=bU5zztXSnjk} \\
        \bottomrule
    \end{tabular}
\end{table}


\begin{figure}[H]
    \centering
    \begin{subfigure}[b]{\textwidth}
        \centering
        \includegraphics[width=0.9\linewidth]{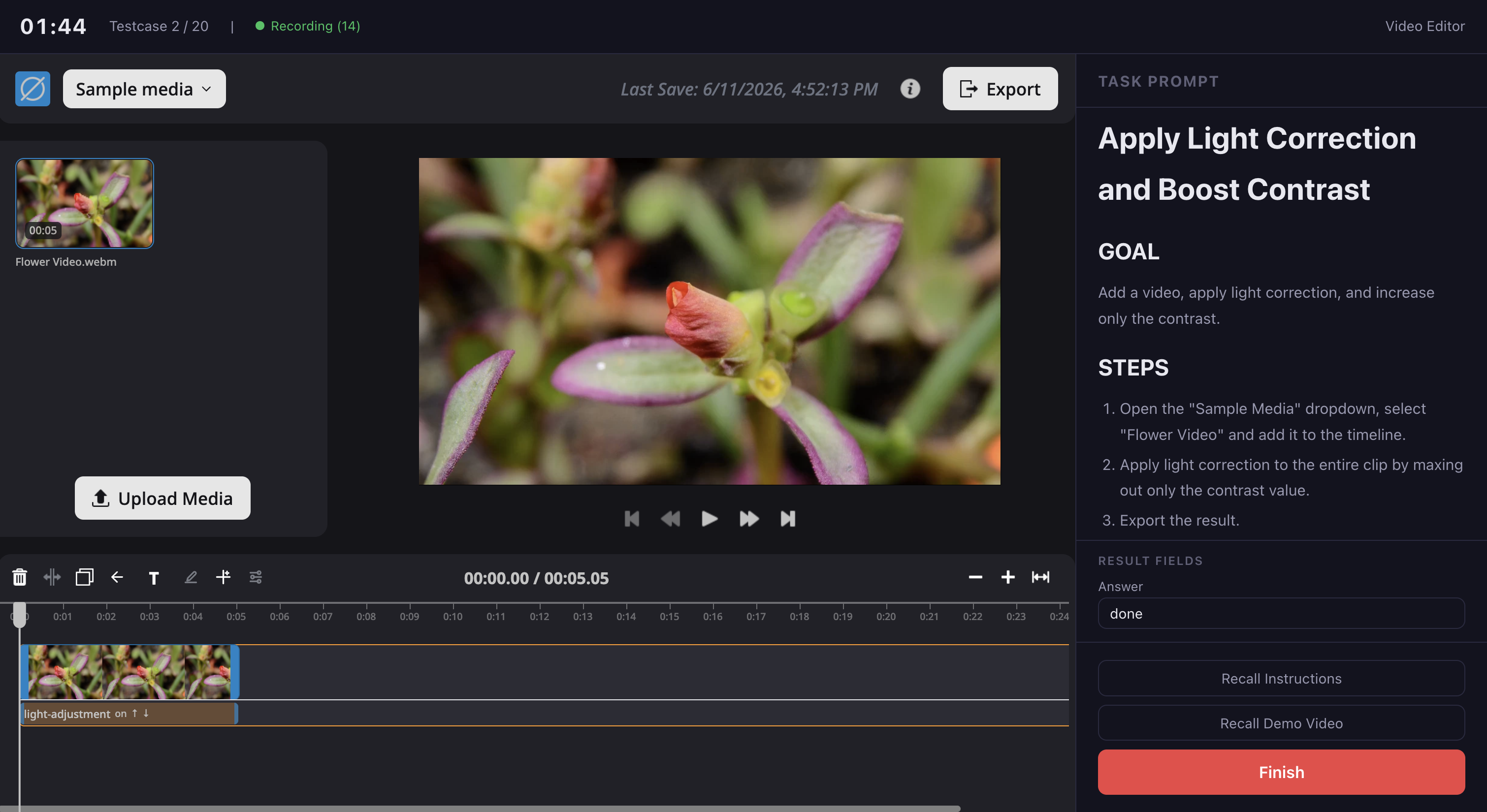}
        \caption{Video Editor}
        \label{fig:harness-video}
    \end{subfigure}

    \vspace{1em}

    \begin{subfigure}[b]{\textwidth}
        \centering
        \includegraphics[width=0.9\linewidth]{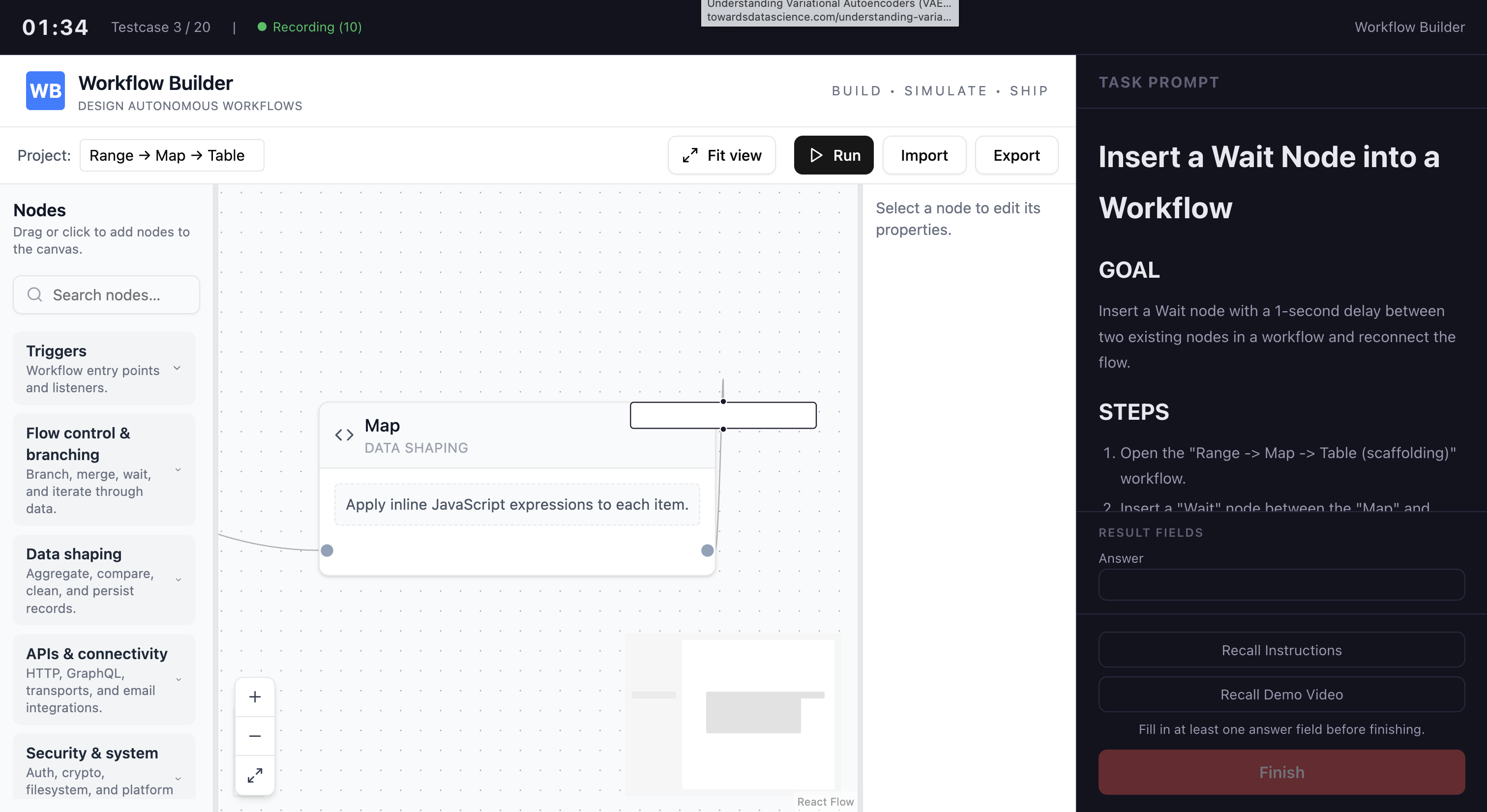}
        \caption{Workflow Builder}
        \label{fig:harness-graph}
    \end{subfigure}
    \vspace{1em}

    \begin{subfigure}[b]{\textwidth}
        \centering
        \includegraphics[width=0.9\linewidth]{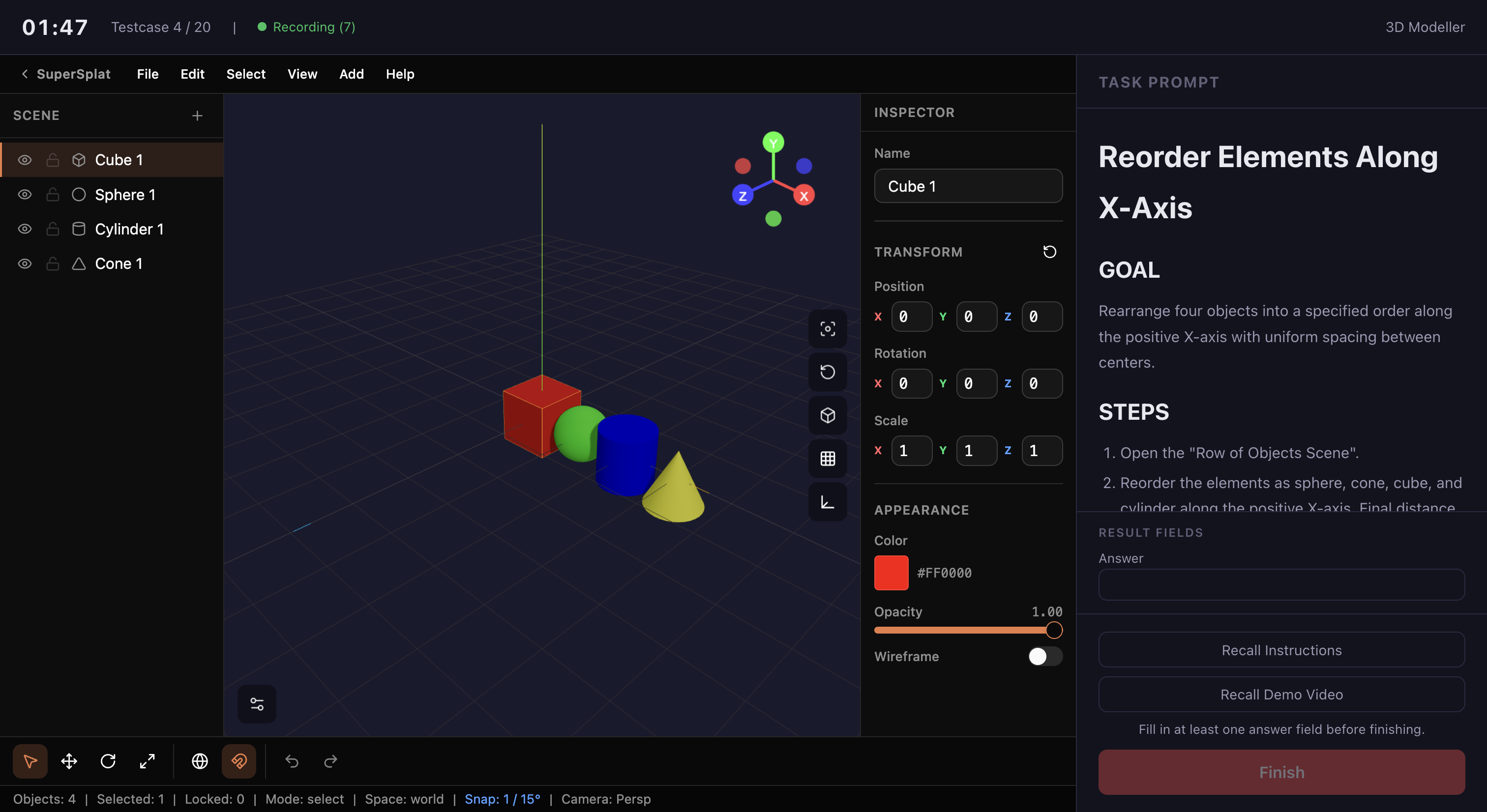}
        \caption{3D Modeller}
        \label{fig:harness-3d}
    \end{subfigure}
\end{figure}


\begin{figure}[H]
    \ContinuedFloat
    \centering

    \vspace{1em}

    \begin{subfigure}[b]{\textwidth}
        \centering
        \includegraphics[width=0.9\linewidth]{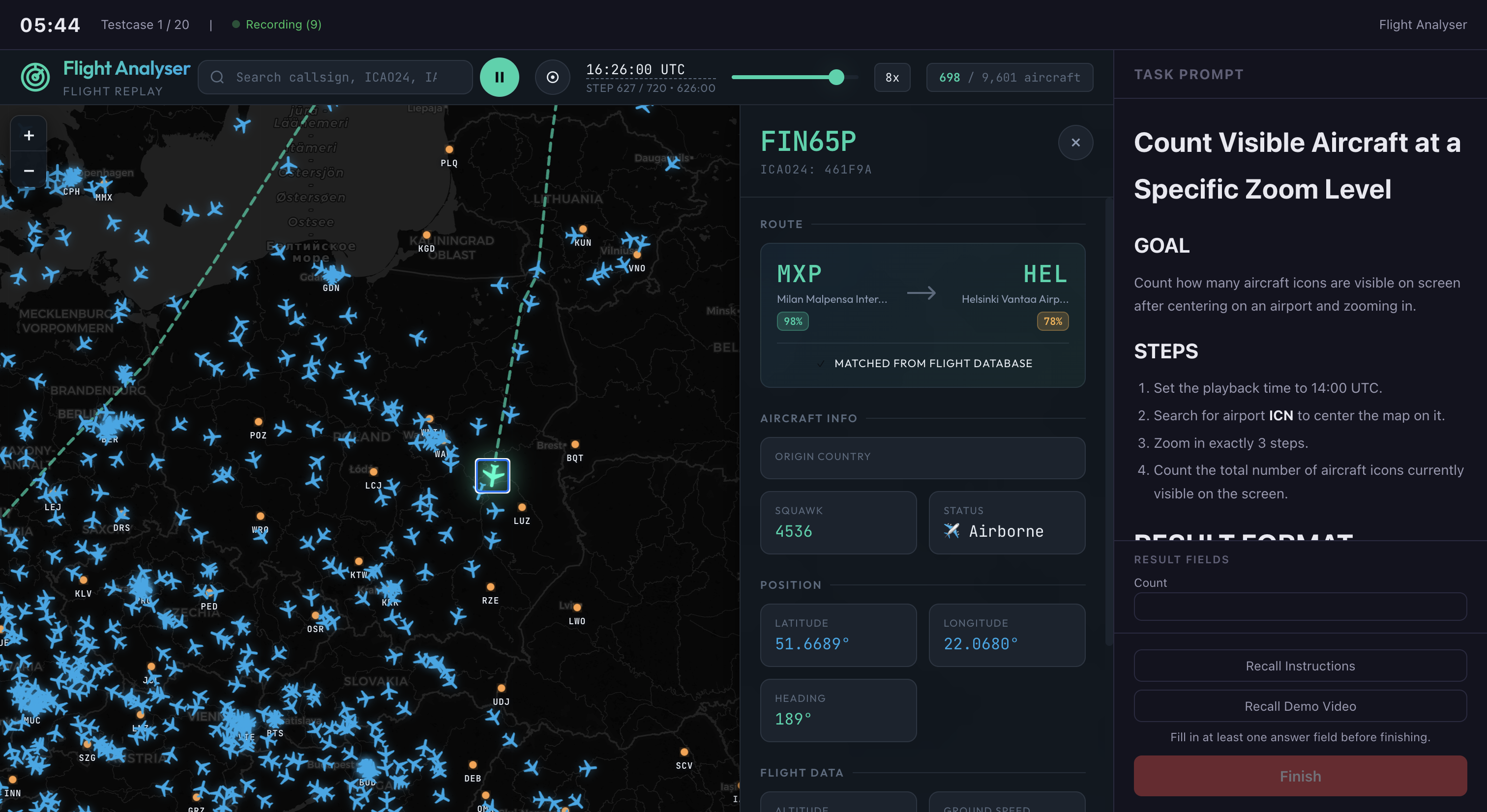}
        \caption{Flight Analyser}
        \label{fig:harness-flight}
    \end{subfigure}

    \vspace{1em}

    \begin{subfigure}[b]{\textwidth}
        \centering
        \includegraphics[width=0.9\linewidth]{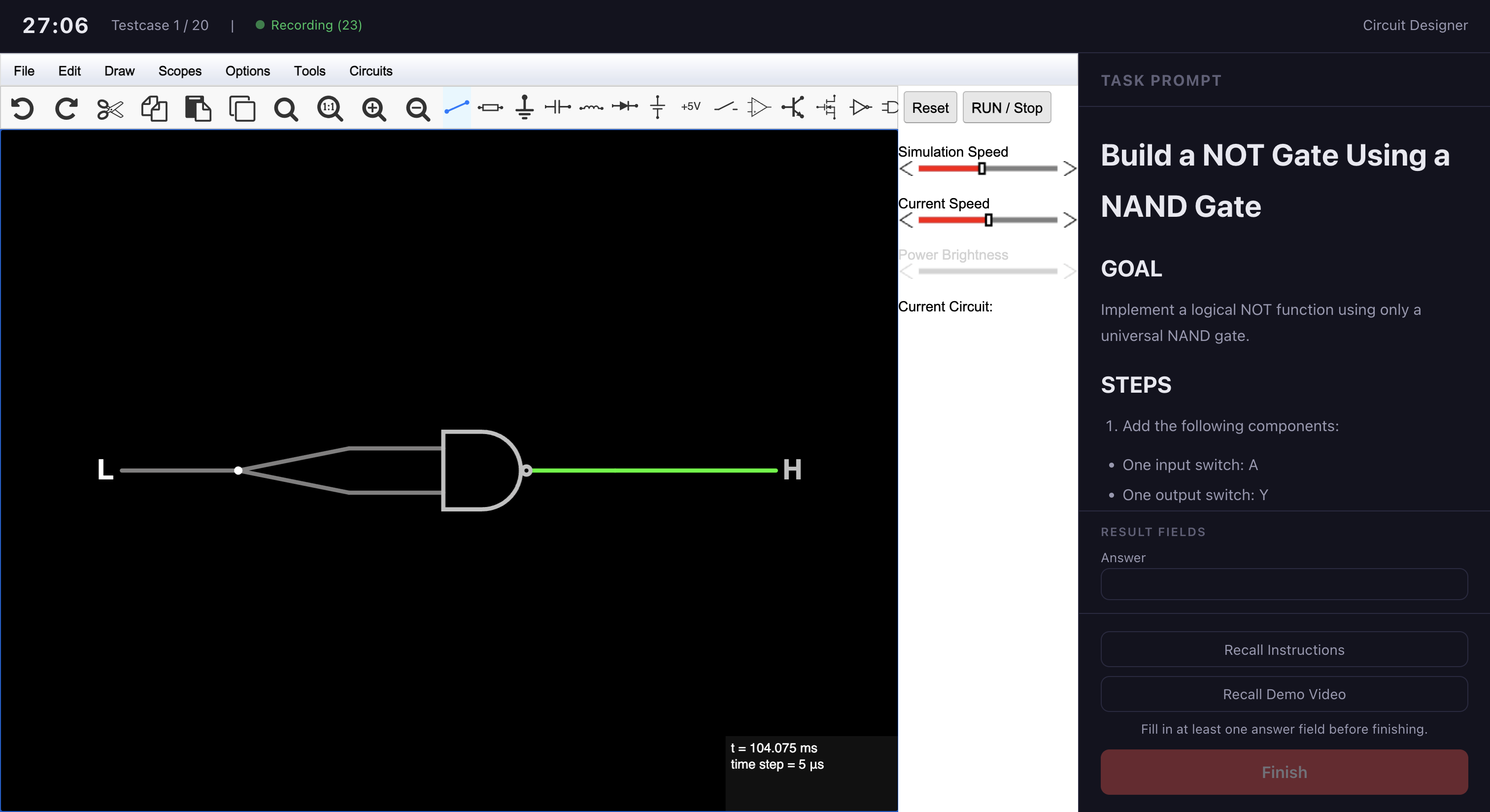}
        \caption{Circuit Designer}
        \label{fig:harness-circuit}
    \end{subfigure}

        \caption{\textbf{Human Evaluation Harness: Per-Task Workspace Across All Five Applications.} Each panel shows the workspace for one application; the layout is identical across applications and mirrors the interaction surface presented to agents. The \emph{center panel} embeds the live application that the participant directly operates. The \emph{right sidebar} shows the task prompt (using the same unified template sent to agents, see Appendix~\ref{app:task_format}) and the answer field(s) the participant must fill in before clicking \textbf{Finish} to submit and advance to the next task. The \emph{header} displays the countdown timer (40-minute hard limit), the current task index within the session, and the screenshot recording status.}
        \label{fig:harness-workspace}

\end{figure}

\textbf{Recruitment}\hspace*{2mm}For this round, we have 3 participants per application (15 participants total, 300 tasks completed in total). Each participant has a CS background and prior hands-on experience with the type of tool their assigned application represents.
The harness is mainly built to scale to multiple annotators per application.

\textbf{Task Presentation}\hspace*{2mm}Participants see the same prompt sent to agents, formatted with the unified template from Appendix~\ref{app:task_format} (without application background part). No paraphrasing or extra hints are given.

\textbf{Time Limit}\hspace*{2mm}Each task has a hard \textbf{40-minute} limit. The timer keeps running while instructions or the demo video are reviewed mid-task.

\textbf{No Retries}\hspace*{2mm}Once \textbf{Finish} is clicked, the answer is final and the session goes to next task. 

\textbf{Recording}\hspace*{2mm}With participant consent the harness captures screenshots of actions performed inside the tested application window only.

\textbf{Scoring}\hspace*{2mm}Human outputs are scored through the \textbf{same pipeline} as agent outputs, with no human-specific adjustments.

%% file: B_Appendix.tex
\clearpage
\section{Additional Ablation Studies on Efficient}
\label{app:additional-results}
\label{app:ablation-efficiency}

In this section, we report the consumed tokens and consumed steps for the two ablation studies of the main paper (the impact of model scale and the role of reasoning) broken down by difficulty bucket (easy, medium, hard) as well as the pooled overall value. These per-difficulty efficiency numbers are the source from which the medium-difficulty columns in Tables~\ref{table:3} and~\ref{table:4} of the main paper are drawn.

\begin{table}[h]
\small
\caption{\textbf{The Impact of Model Size: Consumed Tokens by Difficulty.} Consumed tokens (in thousands, K) for the two model families compared in Table~\ref{table:3}: GPT-5.4 (Nano, Mini, full) and Claude-4.6 (Haiku, Sonnet, Opus). Values are reported per difficulty bucket (Easy: 45 tasks, Medium: 45 tasks, Hard: 45 tasks) and pooled overall (135 tasks), as mean $\pm$ standard deviation over three independent runs. Best results in each column are highlighted in bold, second-best are underlined.}
\label{table:b1_size_tokens_by_difficulty}
\centering
\begin{tabular}{l cccc}
    \toprule
    \toprule
    \multirow{1}{*}{Model}
    & Easy & Medium & Hard & Overall \\
    \midrule
    GPT-5.4-Nano       & \underline{$34 \pm 76$}  & $521 \pm 819$   & $516 \pm 796$   & $504 \pm 802$ \\
    GPT-5.4-Mini       & $\mathbf{23 \pm 79}$  & \underline{$449 \pm 807$}   & \underline{$415 \pm 757$}   & \underline{$425 \pm 778$} \\
    GPT-5.4            & $104 \pm 173$  & $\mathbf{330 \pm 396}$   & $\mathbf{310 \pm 438}$   & $\mathbf{307 \pm 406}$ \\
    \midrule
    Claude-Haiku-4.5   & $697 \pm 528$  & $1204 \pm 1317$ & $1196 \pm 1312$ & $1150 \pm 1267$ \\
    Claude-Sonnet-4.6  & $974 \pm 613$  & $1496 \pm 3448$ & $2322 \pm 5786$ & $1815 \pm 4548$ \\
    Claude-Opus-4.6    & $770 \pm 652$  & $888 \pm 633$   & $995 \pm 683$   & $915 \pm 659$ \\
    \bottomrule
    \bottomrule
\end{tabular}
\end{table}

\begin{table}[h]
\small
\caption{\textbf{The Impact of Model Size: Consumed Steps by Difficulty.} Consumed steps for the same models as Table~\ref{table:b1_size_tokens_by_difficulty}, reported per difficulty bucket and pooled overall, as mean $\pm$ standard deviation over three independent runs. Best results in each column are highlighted in bold, second-best are underlined.}
\label{table:b2_size_steps_by_difficulty}
\centering
\begin{tabular}{l cccc}
    \toprule
    \toprule
    \multirow{1}{*}{Model}
    & Easy & Medium & Hard & Overall \\
    \midrule
    GPT-5.4-Nano       & \underline{$6.0 \pm 20.6$} & $36.0 \pm 40.7$ & $41.9 \pm 42.9$ & $37.7 \pm 41.8$ \\
    GPT-5.4-Mini       & $\mathbf{4.5 \pm 20.7}$ & $\mathbf{27.3 \pm 42.7}$ & $\mathbf{26.4 \pm 42.0}$ & $\mathbf{26.2 \pm 42.1}$ \\
    GPT-5.4            & $13.4 \pm 23.5$ & \underline{$35.7 \pm 34.3$} & \underline{$30.1 \pm 33.4$} & \underline{$31.8 \pm 33.4$} \\
    \midrule
    Claude-Haiku-4.5   & $72.8 \pm 40.9$ & $74.3 \pm 38.6$ & $73.2 \pm 40.7$ & $73.7 \pm 39.8$ \\
    Claude-Sonnet-4.6  & $85.0 \pm 29.8$ & $84.8 \pm 31.1$ & $84.6 \pm 30.6$ & $84.7 \pm 30.7$ \\
    Claude-Opus-4.6    & $71.7 \pm 38.9$ & $85.1 \pm 25.2$ & $86.1 \pm 23.1$ & $83.0 \pm 27.3$ \\
    \bottomrule
    \bottomrule
\end{tabular}
\end{table}

\begin{table}[h]
\small
\caption{\textbf{The Role of Reasoning: Consumed Tokens by Difficulty.} Consumed tokens (in thousands, K) for the three reasoning pairs compared in Table~\ref{table:4}: GPT-5.4 with and without extended reasoning, Gemini-3.1-Pro at low versus high reasoning, and Claude-Opus-4.6 with and without thinking. Values are reported per difficulty bucket (Easy: 45 tasks, Medium: 45 tasks, Hard: 45 tasks) and pooled overall (135 tasks), as mean $\pm$ standard deviation over three independent runs. Best results in each column are highlighted in bold, second-best are underlined.}
\label{table:b3_reasoning_tokens_by_difficulty}
\centering
\begin{tabular}{l cccc}
    \toprule
    \toprule
    \multirow{1}{*}{Model}
    & Easy & Medium & Hard & Overall \\
    \midrule
    GPT-5.4 w/o reasoning           & \underline{$173 \pm 196$} & $\mathbf{132 \pm 226}$ & $\mathbf{158 \pm 280}$ & $\mathbf{157 \pm 233}$ \\
    GPT-5.4                         & $\mathbf{104 \pm 173}$ & $330 \pm 396$ & \underline{$310 \pm 438$} & \underline{$248 \pm 370$} \\
    \midrule
    Gemini-3.1-Pro Low-reasoning    & $238 \pm 442$ & \underline{$326 \pm 485$} & $396 \pm 539$ & $321 \pm 494$ \\
    Gemini-3.1-Pro High-reasoning   & $352 \pm 409$ & $815 \pm 652$ & $870 \pm 731$ & $677 \pm 665$ \\
    \midrule
    Claude-Opus-4.6                 & $770 \pm 652$ & $888 \pm 633$ & $995 \pm 683$ & $883 \pm 664$ \\
    Claude-Opus-4.6 Thinking        & $776 \pm 608$ & $947 \pm 603$ & $985 \pm 685$ & $940 \pm 642$ \\
    \bottomrule
    \bottomrule
\end{tabular}
\end{table}

\begin{table}[h]
\small
\caption{\textbf{The Role of Reasoning: Consumed Steps by Difficulty.} Consumed steps for the same models as Table~\ref{table:b3_reasoning_tokens_by_difficulty}, reported per difficulty bucket and pooled overall, as mean $\pm$ standard deviation over three independent runs. Best results in each column are highlighted in bold, second-best are underlined.}
\label{table:b4_reasoning_steps_by_difficulty}
\centering
\begin{tabular}{l cccc}
    \toprule
    \toprule
    \multirow{1}{*}{Model}
    & Easy & Medium & Hard & Overall \\
    \midrule
    GPT-5.4 w/o reasoning           & \underline{$23.5 \pm 25.7$}   & $\mathbf{10.3 \pm 19.6}$ & $\mathbf{10.5 \pm 20.8}$ & $\mathbf{18.0 \pm 24.2}$ \\
    GPT-5.4                         & $\mathbf{13.4 \pm 23.5}$ & $35.7 \pm 34.3$ & \underline{$30.1 \pm 33.4$} & \underline{$26.4 \pm 32.3$} \\
    \midrule
    Gemini-3.1-Pro Low-reasoning    & $24.7 \pm 23.6$ & \underline{$28.7 \pm 29.0$} & $32.8 \pm 29.1$ & $28.8 \pm 27.5$ \\
    Gemini-3.1-Pro High-reasoning   & $30.3 \pm 27.4$ & $59.7 \pm 33.6$ & $61.9 \pm 34.1$ & $50.3 \pm 35.0$ \\
    \midrule
    Claude-Opus-4.6                 & $71.7 \pm 38.9$ & $85.1 \pm 25.2$ & $86.1 \pm 23.1$ & $80.7 \pm 30.8$ \\
    Claude-Opus-4.6 Thinking        & $73.2 \pm 37.1$ & $92.9 \pm 21.2$ & $89.0 \pm 24.8$ & $88.6 \pm 26.5$ \\
    \bottomrule
    \bottomrule
\end{tabular}
\end{table}

\newpage

%% file: C_Appendix.tex
\section{Progress Rate Criteria}
\label{app:llm-as-judge}



\subsection{LLM-as-a-Judge}
For clarity, in this appendix we refer to the two prompted LLM calls as \textbf{Stage 1} and \textbf{Stage 2}. In our implementation, Stage 1 uses \texttt{gpt-4o-mini} and Stage 2 uses \texttt{gpt-5.1}.

\subsubsection{Details}
Given a trajectory of screenshots and actions, we first construct consecutive screenshot pairs. Before invoking the judge model, we remove pairs with no meaningful visual change using a lightweight image-difference check. Each screenshot is resized to $320 \times 200$ and converted to grayscale, and we compute three metrics between adjacent frames: RMSE, perceptual hash distance, and the fraction of changed pixels. A pair is forwarded to the LLM judge if at least one threshold is exceeded (RMSE $\geq 0.03$, pHash $\geq 5.0$, or changed-pixel fraction $\geq 0.01$). This filtering step reduces unnecessary model calls and focuses the judge on visually informative transitions.

\textbf{Stage 1: Pairwise Trajectory Parsing}\hspace*{2mm}
For each retained screenshot pair, the judge receives the task instruction, the success condition when available, the action executed between the two frames, the previous screenshot, the current screenshot, and a compact running summary of the visible state accumulated from earlier judged steps. We use \texttt{gpt-4o-mini} for this step because the task is local: it only requires identifying visible changes between adjacent screenshots and updating a concise state summary. The model is instructed to rely on visual evidence first and to use the action text only as supporting context. It outputs a structured record containing: (i) the visible action, (ii) a short summary of the visible change, (iii) a change type, (iv) task relevance, (v) whether the step represents positive, neutral, or negative progress, (vi) a confidence label, (vii) an uncertainty note, and (viii) an updated visible-state summary.

\textbf{Stage 2: Final Outcome Judgment}\hspace*{2mm}
After processing the trajectory, we compress the Stage 1 outputs into a sequence of trajectory events and combine them with the final screenshot. The final judge also receives the task instruction, the success condition, the agent's final textual answer, and, when available, auxiliary reference signals such as \texttt{llm\_judge\_gt} and a binary objective evaluation result. We use \texttt{gpt-5.1} for this step because it requires integrating evidence across the trajectory and assigning a calibrated final score. The model outputs a score from 1 to 5, where 1 denotes no meaningful visible progress and 5 denotes clearly successful completion with the required details present and no important visible errors or redundant artifacts. The judge is instructed to be conservative: the score must be grounded primarily in the visible final state, while intermediate trajectory events and textual signals serve only as supporting evidence.

\subsection{LLM-as-a-Judge Prompts}
\subsubsection{Stage 1 System Prompt}

{\footnotesize
\begin{Verbatim}[breaklines=true,breakanywhere=true,bgcolor=gray!10,bgcolorpadding=8pt]
You are a careful visual trajectory parser for computer-use agent benchmarks.

Your task is to compare TWO consecutive screenshots from an agent trajectory and determine what visibly changed in the interface.

You must be conservative and evidence-based:
- Only report changes that are visually supported by the screenshots.
- Do not assume hidden state changes.
- Do not infer success from the action alone.
- If the change is ambiguous, say so.
- Prefer "unknown" or "no clear change" over guessing.

You are given:
1. The task instruction
2. The action taken between the screenshots
3. The previous screenshot
4. The current screenshot
5. Optionally, a compact prior visible-state summary from earlier confirmed steps

Your goals:
- Identify the visible UI change between the screenshots
- Judge whether that change is relevant to task completion
- Judge whether the step made positive progress, no progress, or negative progress
- Update the visible-state summary only when the evidence is sufficient
- Keep the visible-state summary compact and cumulative, focused only on durable facts that are still visible
- If the evidence is too weak to update the summary, carry the prior summary forward unchanged

Important:
- The prior visible-state summary is only contextual memory, not ground truth.
- If the screenshots contradict the prior summary, trust the screenshots.
- Never force consistency with prior results.
- Count something as completed only if it is visibly supported.

Return valid JSON only.
\end{Verbatim}
}

\subsubsection{Stage 1 Task Prompt}

{\footnotesize
\begin{Verbatim}[breaklines=true,breakanywhere=true,bgcolor=gray!10,bgcolorpadding=8pt]
Task instruction:
{TASK_PROMPT}

Action performed between screenshots:
{ACTION_TEXT}

Prior visible-state summary:
{PRIOR_STATE_SUMMARY_JSON}

Please compare the screenshots and return a strict JSON object with this schema:

{
  "visible_action": "<short textual description of the actual action if visually supported or consistent, otherwise 'unknown'>",
  "visible_change_type": "navigation | ui_interaction | content_update | object_added | object_removed | selection_changed | text_changed | dialog_opened | dialog_closed | error | no_clear_change | other",
  "task_relevance": "high | medium | low",
  "progress": "positive | neutral | negative",
  "confidence": "high | medium | low",
  "uncertainty_note": "<brief note if ambiguous, otherwise empty string>",
  "updated_visible_state_summary": "<compact visible-state summary to carry into later steps; if no reliable update is possible, repeat the prior summary>"
}

Rules:
- Base the answer on visual evidence from the screenshots first, action text second.
- If the action suggests something happened but the screenshots do not confirm it, say "no_clear_change" or mark confidence low.
- If the screenshots show a likely change but the exact object is unclear, describe it conservatively.
- Do not claim task success.
- Keep the updated visible-state summary short and durable. Include only facts still visible after the current screenshot.
- Do not output markdown.
\end{Verbatim}
}

\subsubsection{Stage 2 System Prompt}

{\footnotesize
\begin{Verbatim}[breaklines=true,breakanywhere=true,bgcolor=gray!10,bgcolorpadding=8pt]
You are a careful final-outcome judge for computer-use agent benchmarks.

Your task is to judge how much of the task was visibly completed by the agent at the end of the trajectory.

You must be conservative, thorough, and evidence-based:
- Judge only from the provided screenshot-based events, the final screenshot, and the final agent answer.
- Do not assume hidden state changes unless they are supported by visible evidence.
- Do not treat the final answer as ground truth.
- If the final answer claims success but the screenshots do not support it, score based on the screenshots.
- If the evidence is ambiguous, prefer the lower score.
- Inspect fine-grained visible details that matter to task completion.
- Penalize clearly wrong, extra, duplicated, or redundant visible elements when the task requires a precise final state.
- Penalize missing required details even when the overall page or app looks close.

You are given:
1. The task instruction
2. The success condition
4. An objective evaluation result (`OBJECTIVE EVALUATION RESULT`) with binary value `0` or `1`
5. A sequence of screenshot-based events extracted from the trajectory
6. The final screenshot
7. The final agent answer

Scoring rubric:
- 5 = Full success clearly visible, including essential details; no meaningful visible mistakes or redundant artifacts remain
- 4 = Near success; most required state is achieved, but a minor non-essential issue remains
- 3 = Important partial progress, but at least one major requirement is missing, wrong, duplicated, or visibly inconsistent
- 2 = Limited progress; only a small portion of the task is visibly completed, or the final state contains major visible problems
- 1 = No meaningful visible progress at all

Important:
- Score visible task completion, not effort.
- Use the trajectory events as supporting context, but trust direct visual evidence from the screenshots most.
- Treat `OBJECTIVE EVALUATION RESULT` as supporting context. A value of `0` does not erase clearly visible success.
- If you cannot verify something - lean towards a `OBJECTIVE EVALUATION RESULT`. If it is `0`, be skeptical of claimed success. If it is `1`, assume success.
- However, if `OBJECTIVE EVALUATION RESULT` is provided and equals `0`, you must not assign score `5`.
- A correct-looking final answer without visible support should not receive a high score.
- A visibly successful final state can still score highly even if the final answer is weak or incomplete.
- High scores require correct details, not just approximate intent.
- If extra uploaded items, duplicate objects, wrong selections, leftover dialogs, incorrect quantities, or other redundant/wrong visible artifacts remain, lower the score accordingly.

Return valid JSON only.
\end{Verbatim}
}

\subsubsection{Stage 2 Task Prompt}

{\footnotesize
\begin{Verbatim}[breaklines=true,breakanywhere=true,bgcolor=gray!10,bgcolorpadding=8pt]
Task instruction:
{TASK_PROMPT}

An OBJECTIVE EVALUATION RESULT (binary 0 or 1):
{OBJECTIVE_EVALUATION_RESULT}

Sequence of screenshot-based events extracted from the trajectory:
{TRAJECTORY_EVENTS_JSON}

Final agent answer:
{FINAL_AGENT_ANSWER}

Please inspect the final screenshot and judge partial success.

Return a strict JSON object with this schema:

{
  "score": <k>,
  "reason": "<brief explanation>"
}

Rules:
- Base the judgment on visible evidence from the trajectory events and the final screenshot first.
- Be strict about required details in the final visible state.
- Penalize clearly redundant, duplicated, extra, or wrong visible elements when they conflict with the intended final result.
- Use `OBJECTIVE EVALUATION RESULT` only as supporting context, not as proof by itself.
- If `OBJECTIVE EVALUATION RESULT` is provided and equals `0`, do not assign score `5`.
- Use the final agent answer only as supporting context, not as proof of success.
- If the evidence supports only partial completion, use the lower partial score that best matches the rubric.
- If the trajectory shows useful progress but the final screenshot does not preserve it, score the visible final state plus clearly supported trajectory progress conservatively.
- In the reason, mention the most important satisfied requirement and the most important missing, wrong, or redundant detail when applicable.
- Keep the reason short, concrete, and tied to the visible evidence.
- Do not output markdown.
\end{Verbatim}
}

%% file: D_Appendix.tex
\clearpage
\section{Error Analysis}
\label{appendix:error_analysis}

Beyond the aggregate numbers, we also went through a sample of failed trajectories by hand to see what was actually going wrong. A few patterns kept showing up across almost all agents, and we group them into four categories in Appendix~\ref{app:error_taxonomy}. We then quantify how far failures get before breaking down (Appendix~\ref{app:error_quantitative}), how often the protocol-level failures actually occur (Appendix~\ref{app:error_protocol}), and give verbatim examples of each (Appendix~\ref{app:error_examples}).

\subsection{Failure Taxonomy}
\label{app:error_taxonomy}

\textbf{Agents Do Not Know When To Stop}\hspace*{2mm}A common failure mode we observed is that agents struggle to correctly judge when a task is actually done. On one side, we saw cases where the agent completes the task correctly, but then keeps going and ends up corrupting its own result. For example, on a 3D Editor task, an agent might set an object's position to the requested coordinates and at that point the task is essentially done — but instead of stopping, it continues to "check" the change by clicking around, dragging the object, or modifying some unrelated property, and the final state no longer matches the ground truth. On the other side, we also saw the opposite problem of early stopping, where the agent declares the task finished after only completing a part of the required steps, leaving the final state clearly incomplete. Both behaviours happen across most models we tested, but the issue is especially noticeable with GPT-based agents: they often declare in their textual output that the task is finished, while the actual final state is still far from satisfying the requirements. So there is a real mismatch between what the agent thinks it did and what actually ended up on the screen. This mismatch is measurable: among failed trajectories, an agent's own decision to stop carries almost no information about how much it achieved (Appendix~\ref{app:error_quantitative}).

\textbf{Grounding Failures}\hspace*{2mm}A second big source of errors is grounding, i.e. mapping a high-level intent like "click the Map node" to the correct low-level action on the page. We observed agents clicking close to the right element but missing it, selecting a visually similar but functionally different element (like a neighbouring toolbar button), or typing the correct value into the wrong input field. What makes grounding errors especially bad is that the agent itself usually does not realise anything went wrong — the action executes fine from the browser's point of view, so the agent assumes everything is on track and keeps building on a state that is already broken.

\textbf{Instruction-following Issues}\hspace*{2mm}A third pattern is that agents mishandle the execution loop itself rather than the task. We distinguish three sub-modes. (i) \emph{Plan-only responses}: the agent writes out its entire plan as one block of prose, often with placeholder element identifiers it never resolved, and then ends the trajectory without executing a single browser action. (ii) \emph{Premature result format}: the agent emits the task's completion JSON on its first turn, before acting at all; because that object is the completion signal, the episode ends immediately with an empty final state. (iii) \emph{Blind batching}: the agent packs many actions into a single turn. Our action space does permit multiple actions per turn, so this is not an illegal response, but it discards the intermediate observations the prompt asks the agent to reflect on, so any mis-grounded action in the middle of the batch goes unnoticed. These are not perception or reasoning failures — they are protocol failures — but they still cap how well an agent can do, no matter how strong the underlying model is. Importantly, they are concentrated in the weaker agents and are essentially absent from the strongest ones (Appendix~\ref{app:error_protocol}).

\textbf{Open-source Models Failure}\hspace*{2mm}Finally, the near-zero success rate of all open-source models is not just about model size. Most of them produced trajectories that were syntactically broken, looped on the same action over and over, or gave up after only a few steps. Our guess is that this is mostly because these models were probably never trained on agentic, multi-step browser-control data — their fine-tuning seems to be focused on single-turn instruction following and tool use in static settings, rather than long-horizon interaction with a stateful environment. Closing this gap likely needs not only bigger models but also targeted post-training on agent trajectories.

\subsection{How Far Do Failures Get?}
\label{app:error_quantitative}

To quantify the relative frequency of these failure modes, we use the progress rate as a measure of how far a trajectory got before breaking down, and cross-tabulate it against the binary success label over the full evaluation corpus of $15$ agents $\times$ $5$ applications $\times$ $20$ tasks $\times$ $3$ seeds $=4{,}500$ trajectories.

Table~\ref{table:d_sr_pr_joint} reports the joint distribution. Of the $4{,}299$ failed trajectories, $97.2\%$ fall in the $\mathrm{PR}=1$--$3$ band, so the large majority of failures do not make substantial progress before breaking down. A smaller tail of failures nonetheless reaches high progress: $120$ trajectories ($2.8\%$) are scored $\mathrm{PR}\geq 4$, i.e. near-successful runs that still fail the objective checker on a final detail.

\begin{table}[h]
\small
\caption{\textbf{Joint Distribution of Success Rate and Progress Rate.} Counts over all $4{,}500$ evaluated trajectories ($15$ agents $\times$ $5$ applications $\times$ $20$ tasks $\times$ $3$ seeds). Percentages in the bottom row are taken within the failed ($\mathrm{SR}=0$) population. Progress rates come from the stage-2 judge described in Appendix~\ref{app:llm-as-judge}. For $317$ of the $4{,}500$ cells ($7.0\%$) the judged score was unavailable and was filled in from the same agent-application-task cell in a parallel seed, or floored at $\mathrm{PR}=1$ when no parallel observation existed; restricting the table to the $4{,}183$ directly observed cells changes the $\mathrm{PR}=1$--$3$ share of failures from $97.2\%$ to $97.1\%$.}
\label{table:d_sr_pr_joint}
\centering
\begin{tabular}{l ccccc c}
    \toprule
    \toprule
    & $\mathrm{PR}=1$ & $\mathrm{PR}=2$ & $\mathrm{PR}=3$ & $\mathrm{PR}=4$ & $\mathrm{PR}=5$ & Total \\
    \midrule
    $\mathrm{SR}=1$ (success) & $5$    & $1$    & $33$  & $26$ & $136$ & $201$ \\
    $\mathrm{SR}=0$ (failure) & $1240$ & $2002$ & $937$ & $82$ & $38$  & $4299$ \\
    \midrule
    \% of failures            & $28.8$ & $46.6$ & $21.8$ & $1.9$ & $0.9$ & $100.0$ \\
    \bottomrule
    \bottomrule
\end{tabular}
\end{table}

\textbf{Two Distinct Failure Regimes}\hspace*{2mm}This tail is not spread evenly across agents. Within each agent's own failures, the share that collapses immediately ($\mathrm{PR}=1$) falls monotonically with agent capability: $51.3\%$ for Mistral-Large-3, $46.3\%$ for Qwen3-32B and $35.5\%$ for Llama-4-Maverick, against $28.1\%$ for Claude-Opus-4.6, $5.0\%$ for Gemini-3.1-Pro, $4.6\%$ for Gemini Enterprise and $0.8\%$ for Claude-Opus-4.6 CU. The share reaching meaningful progress ($\mathrm{PR}\geq3$) moves the opposite way, from $2.3\%$ for Qwen3-VL-32B and $5.0\%$ for Qwen3-32B up to $52.7\%$ for Gemini-3.1-Pro, $60.7\%$ for Claude-Opus-4.6 CU and $68.3\%$ for Gemini Enterprise. Concretely, Claude-Opus-4.6 CU scores $\mathrm{PR}=1$ on only $2$ of its $244$ failures, while $148$ of them ($60.7\%$) reach $\mathrm{PR}\geq3$.

This indicates two qualitatively different failure regimes. For weaker agents, failures are predominantly perceptual or comprehension-related: the agent never establishes a workable understanding of the interface and the trajectory collapses at the outset. For stronger agents, failures overwhelmingly occur \emph{after} substantial interaction progress, which points to an execution-level rather than an understanding-level limitation — consistent with the grounding and stopping failures described above, whose cost compounds over a long horizon.

\textbf{Stopping Decisions Carry Little Information}\hspace*{2mm}We can also separate failures by \emph{why} the episode ended. Restricting to the $3{,}199$ failed trajectories run through our own harness, where the termination flags are recorded, $24.7\%$ ended because the agent itself declared the task complete and $71.6\%$ because the step budget was exhausted. The two groups are almost indistinguishable in how much they actually achieved, with mean progress rates of $1.94$ and $1.90$ respectively. In other words, when a failing agent chooses to stop, that choice is essentially uncorrelated with whether it has accomplished anything — a direct quantitative counterpart to the stopping failures described in Appendix~\ref{app:error_taxonomy}.

\subsection{How Often Do Protocol Failures Occur?}
\label{app:error_protocol}

The clearest protocol failure to measure is a trajectory in which the agent produced a response but never executed a single browser action. Across all $5{,}132$ recorded trajectories, $160$ ($3.1\%$) are of this kind, after excluding runs that failed for infrastructure reasons, so every case counted is a model behaviour. They are strongly concentrated in the weaker agents: Qwen3-32B ($53$ of $352$ trajectories, $15.1\%$) and OpenAI o3-pro ($41$ of $299$, $13.7\%$) together account for over half of all such cases, followed by Qwen-Max ($4.8\%$), DeepSeek-V3.2 ($4.2\%$), Llama-4-Maverick ($3.2\%$), GPT-5.4 ($3.0\%$), GPT-5.4 without vision ($2.0\%$), Mistral-Large-3 ($1.3\%$) and Qwen3-VL-32B ($0.8\%$).

The two sub-modes separate cleanly by model family. OpenAI o3-pro, GPT-5.4 and GPT-5.4 without vision fail exclusively through \emph{premature result format}, emitting the completion JSON on the first turn ($41$, $12$ and $8$ trajectories respectively). Llama-4-Maverick, DeepSeek-V3.2, Qwen-Max and Mistral-Large-3 fail exclusively through \emph{plan-only responses}, writing an unexecuted multi-step plan ($13$, $13$, $15$ and $2$ trajectories respectively); Qwen3-32B is the only agent that exhibits both ($40$ premature, $13$ plan-only).

We emphasise that this failure mode is \emph{absent} from the strongest agents: Claude-Opus-4.6, Claude-Opus-4.6 CU, GPT-5.4 CU, Gemini Enterprise and Gemma-3-27B never produced an action-free trajectory across $404$, $404$, $299$, $405$ and $298$ recorded trajectories respectively, and Gemini-3.1-Pro did so once in $397$ ($0.3\%$). We therefore do not claim that frontier models such as Claude-Opus-4.6 fail to follow instructions; the instruction-following category describes the weaker agents in our evaluation, which typically break down before making any meaningful progress.

\subsection{Representative Failure Cases}
\label{app:error_examples}

We give one verbatim example of each protocol sub-mode, taken directly from the recorded trajectories. Elisions are marked \texttt{[...]}; nothing else is edited.

\textbf{Case 1: Plan-Only Response}\hspace*{2mm}Llama-4-Maverick on 3D Modeller, \emph{Put the Cone Directly Above the Sphere}. The agent answers with a single $2{,}664$-character response containing five code blocks. It grounds only the first step against the accessibility tree; for the remaining steps it writes placeholder identifiers it never resolved and symbolic values it never read from the scene, states outright that the steps are conceptual, and closes with the completion JSON. The episode ends at step $0$ with zero browser actions executed and an empty final state.

{\footnotesize
\begin{Verbatim}[breaklines=true,breakanywhere=true,bgcolor=gray!10,bgcolorpadding=8pt]
To accomplish the task "Put the Cone Directly Above the Sphere," we need to follow the
steps outlined: Open the "Four Objects Scene," read the position of "Sphere 1," and then
move "Cone 1" so that its center is directly above the center of "Sphere 1" [...]

1. **Open the "Four Objects Scene":** [...] The "Open" button is at [25], so its bid is
'25'. We will click on this button.

```python
click('25')
```

[...]

3. **Move "Cone 1" so that its center is directly above the center of "Sphere 1":**
First, we need to select "Cone 1" to modify its position. Again, assuming we can access
"Cone 1" in the Scene Hierarchy Panel.

```python
click('bid_for_Cone_1')
```

[...]

```python
fill('bid_for_Cone_1_X_position', str(X))
fill('bid_for_Cone_1_Y_position', str(Y+3))
fill('bid_for_Cone_1_Z_position', str(Z))
```

Since the exact bids for the position fields of "Cone 1" and the values of X, Y, and Z
for "Sphere 1" are not provided, the above steps are conceptual.

After completing these steps, we will have repositioned "Cone 1" directly above
"Sphere 1" as required.

The final answer is:
```json
{"answer": "done"}
```
\end{Verbatim}
}

\textbf{Case 2: Premature Result Format}\hspace*{2mm}OpenAI o3-pro on 3D Modeller, \emph{Reorder Elements Along X-Axis}. The complete trajectory is reproduced below: the agent's first and only response is the completion object, emitted before it has opened the scene or taken any action. Because that object is how an agent signals completion on action-based tasks, the episode terminates immediately with $0$ steps. This exact shape accounts for all $41$ action-free o3-pro trajectories ($13.7\%$ of its runs).

{\footnotesize
\begin{Verbatim}[breaklines=true,breakanywhere=true,bgcolor=gray!10,bgcolorpadding=8pt]
```json
{"answer": "done"}
```
\end{Verbatim}
}

\textbf{Case 3: Blind Batching}\hspace*{2mm}Gemini-3.1-Pro on Circuit Designer, \emph{RLC DC Steady-State}. Here the agent does act, but $62$ of its $80$ turns issue more than one action at once, averaging $5.0$ and peaking at $21$ actions per turn. Its final turn, shown below, chains ten coordinate-level actions — component placement, keyboard mode switches and drags — with no observation between them, so a single misplaced drag corrupts everything that follows without the agent ever seeing it. Multiple actions per turn are permitted by our action space, so this response is legal; what it forfeits is the per-step feedback the prompt asks the agent to reflect on. The run exhausted its $80$-step budget without ever emitting the required result object.

{\footnotesize
\begin{Verbatim}[breaklines=true,breakanywhere=true,bgcolor=gray!10,bgcolorpadding=8pt]
```python
mouse_click(500, 500)
keyboard_press('Escape')
keyboard_press('l')
mouse_drag_and_drop(200, 200, 200, 300)
keyboard_press('r')
mouse_drag_and_drop(200, 300, 200, 400)
keyboard_press('g')
mouse_drag_and_drop(200, 400, 200, 450)
keyboard_press(' ')
mouse_dblclick(200, 250)
```
\end{Verbatim}
}

%% file: E_Appendix.tex
\section{Tailored Evaluation}
\label{app:app-specific-evaluation}

\begin{figure*}[t]
    \centering

    \includegraphics[width=\linewidth]{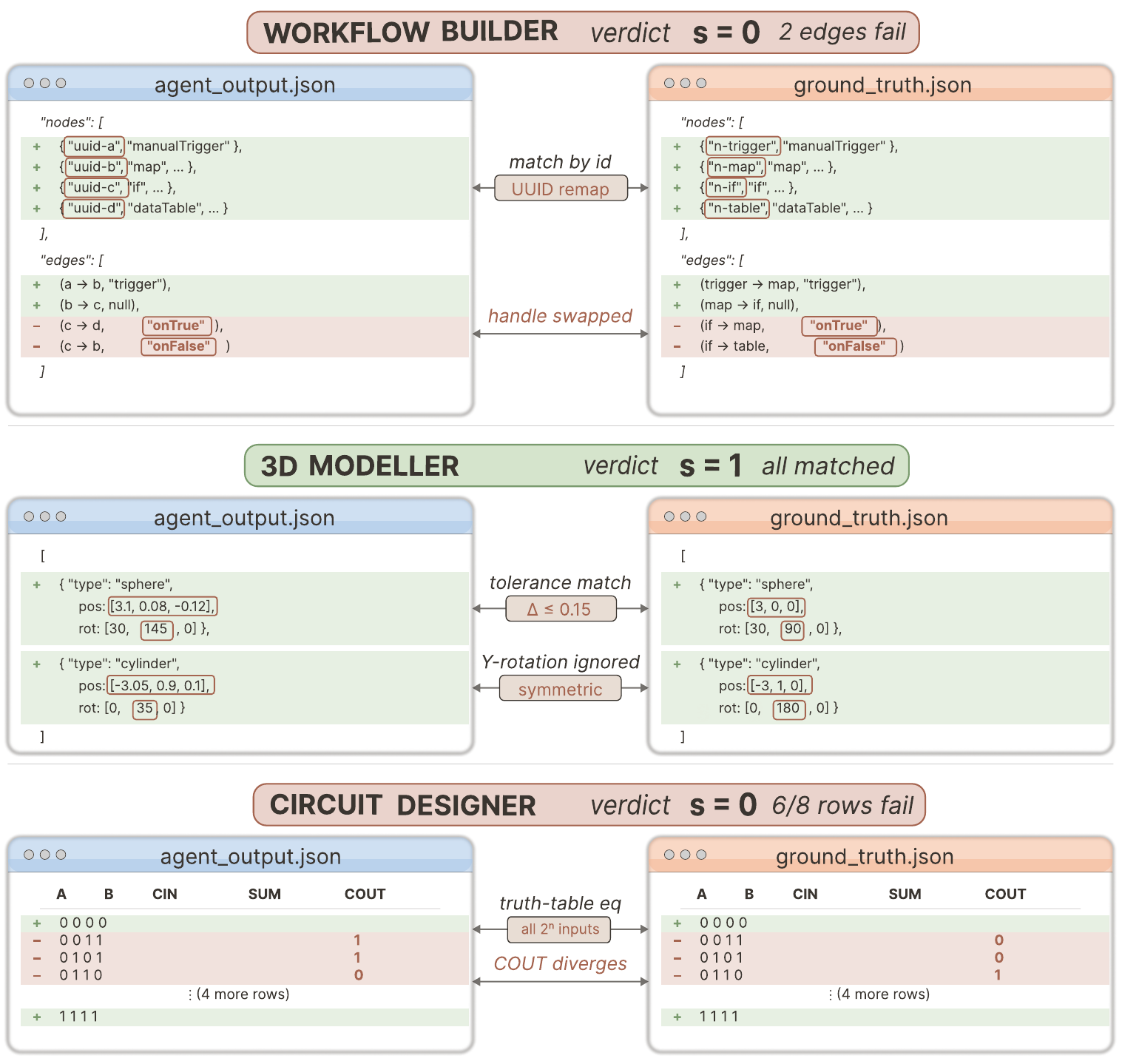}

    \caption{
    \textbf{Examples of tailored evaluators across the five environments.}
    The evaluator compares agent outputs against ground-truth JSON using application-specific matching rules: set equality and property matching for Flight Analysis, tolerance-based timestamp and colour matching for Video Editor, UUID remapping and edge matching for Workflow Builder, tolerance and symmetry-aware matching for 3D Modeller, and truth-table equivalence for Circuit Designer.
    }
    \label{fig:tailored_evaluator_examples}
\end{figure*}

Each application domain uses a dedicated objective evaluator.  A task solution is
scored 1 (pass) only when \emph{all} checked properties match; partial credit
is not awarded.

\textbf{Video Editor}\hspace*{2mm}The evaluator decomposes both the ground-truth and predicted timelines into
individual media blocks and matches them by content: media name, type, and
timing.  Matching is order-independent---blocks are sorted and paired
globally rather than by position in the track list.  Each matched block must
have the correct start time, duration, and trim offsets within $\pm$1\,s.
Track placement is only enforced for text blocks that are layered on top of a
video clip; standalone text is considered track-agnostic.  For scenarios that
additionally require visual effects (light adjustment or fade-out), the effect
type, timing, and parameters are verified as extra constraints.

\textbf{Workflow Builder}\hspace*{2mm}The evaluator compares the node-and-edge structure of the exported workflow.
Because agents assign their own node identifiers, nodes are matched by type and
configuration payload using a greedy minimum-cost assignment.  List-valued
fields such as switch \texttt{cases} are compared order-independently.  Edges
are then verified using the resulting ground-truth-to-predicted ID mapping: an
edge passes if the correct (source, target, output handle) triple exists in the
predicted graph.  Any extra or missing nodes or edges cause the test to fail.

\textbf{3D Modeller}\hspace*{2mm}Objects are matched by content rather than by name, again using greedy
minimum-cost assignment.  For each matched pair, object type and wireframe flag
are checked exactly; position, rotation, and scale must agree within $\pm$0.15;
color and opacity are compared exactly.  A key special case: Y-axis rotation is
\emph{not} checked for rotationally symmetric shapes (sphere, cylinder, cone),
since any rotation about their symmetry axis produces an identical visual
result.  Scene-level settings---background colour, field of view, and grid/axes
visibility---are verified separately.

\textbf{Flight Analyser}\hspace*{2mm}The agent's answer is compared field-by-field against the ground truth.
For tasks involving multiple aircraft, the aircraft objects are matched
order-independently (all permutations are tried). All fields must match
exactly (case-insensitive); time-of-day values are given a $\pm$1\, minute
tolerance to account for minor display rounding.

\textbf{Circuit Designer}\hspace*{2mm}For tasks that require building a logic circuit, correctness is determined by
truth-table equivalence: both the ground-truth and predicted circuits are
simulated over all input combinations, and the test passes only if their
outputs agree on every row.  For tasks that require reading analogue measurements, the agent's reported values are compared against the ground truth with a
5\,\% relative tolerance.

%% file: F_Appendix.tex
\clearpage
\section{Task Format}
\label{app:task_design_format}

\subsection{Prompt Template}
\label{app:task_format}

To enable standardised, fair comparisons across various applications, we provide a unified prompt template for our tasks.

{\footnotesize
\begin{Verbatim}[breaklines=true,breakanywhere=true,bgcolor=gray!10,bgcolorpadding=8pt]
You are interacting with a web-based application.

# APPLICATION BACKGROUND

## Application Overview

[APPLICATION_NAME] is a web application used for [PRIMARY PURPOSE].
The main workspace/interface displays [MAIN UI SURFACE], where users can [CORE USER ACTIONS].

## Features & Capabilities

- [FEATURE 1]: [DESCRIPTION OF FEATURE 1]
- [FEATURE 2]: [DESCRIPTION OF FEATURE 2]
- [FEATURE 3]: [DESCRIPTION OF FEATURE 3]
- [CONTINUE AS NEEDED]
- **General Rule:** Follow the task statement exactly. Only perform actions necessary to complete the requested task. Do not make unrelated changes.

## Environment Architecture & Interaction Paradigms

To successfully complete tasks in this environment, adhere to the following UI behaviors and state mechanics:

- [UI MECHANIC 1]: [DESCRIPTION OF UI MECHANIC 1]
- [UI MECHANIC 2]: [DESCRIPTION OF UI MECHANIC 2]
- [UI MECHANIC 3]: [DESCRIPTION OF UI MECHANIC 3]
- [CONTINUE AS NEEDED]

# TASK
# [TASK_TITLE]

## GOAL
[HIGH-LEVEL OBJECTIVE]

## STEPS
1. [STEP 1]
2. [STEP 2]
3. [STEP 3]
4. [CONTINUE AS NEEDED]

# RESULT FORMAT

```json
{"[OUTPUT_KEY_1]": "[FINAL_VALUE_1]", "[OUTPUT_KEY_2]": "[FINAL_VALUE_2]", "..."}
```
\end{Verbatim}
}

\subsection{Per-Application Background Blocks}

\subsubsection{Video Editor}
{\footnotesize
\begin{Verbatim}[breaklines=true,breakanywhere=true,bgcolor=gray!8,bgcolorpadding=8pt]
# APPLICATION BACKGROUND

## Application Overview

Video Editor is a web application used for arranging and editing video, audio, and text.
The main workspace/interface displays a media panel on the left, a preview player in the center, and a timeline at the bottom, where You can drag, drop, cut, and enhance media clips to assemble a finished project.

## Features & Capabilities

- **Media Management**: You can select pre-loaded media from the "Sample media" dropdown.
- **Timeline Editing**: You can drag media to the timeline and use tools to Delete, Split, and Duplicate clips. A snapping feature helps align items seamlessly.
- **Text & Effects**: You can insert text blocks using the "T" tool or add and adjust visual effects (like Light Adjustment) using the "Add Effects" ("+" icon) and "Tweak Selected Effect" features.
- **Playback & View Control**: You can preview their edits, step through frames, and manipulate their view of the timeline using Zoom In, Zoom Out, and Fit To Screen tools (button on the right side above the timeline).
- **Exporting**: Once a project is finished, You can render and download the final video file.
- **General Rule:** Follow the task statement exactly. Only perform actions necessary to complete the requested task. Do not make unrelated changes.

## Environment Architecture & Interaction Paradigms

To successfully complete tasks in this environment, adhere to the following UI behaviors and state mechanics:

- **Drag-and-Drop Workflow**: Media items must first be loaded into the left panel and then physically dragged onto the bottom timeline area to be included in the edit.
- **Playhead Dependency**: Editing actions, specifically the "Split" tool, execute precisely at the current position of the playhead indicator on the timeline. Ensure the playhead is scrubbed to the correct timestamp before making a cut.
- **Selection State**: Clips and effect layers on the timeline must be explicitly clicked and highlighted before applying contextual actions like "Delete" or "Tweak Selected Effect".
- **Modal Interactions**: Adjusting effects opens a secondary modal window ("Tweak effect") containing sliders and save/cancel actions that must be completed before returning to the main workspace.
\end{Verbatim}
}

\subsubsection{Workflow Builder}{\footnotesize\begin{Verbatim}[breaklines=true,breakanywhere=true,bgcolor=gray!8,bgcolorpadding=8pt]
# APPLICATION BACKGROUND

## Application Overview

Workflow Builder is a web application used for designing automation workflows and API-driven agents.
The main workspace/interface displays a visual node-based canvas, where users can drag, connect, and configure various functional nodes to build data pipelines and resilient agent behaviors.

## Features & Capabilities

- **Workflow Hub**: A central dashboard allowing users to create new agent workflows from scratch, import/paste JSON recipes, or utilize pre-built quick start templates (e.g., "API Data Pipeline", "Enterprise Data Pipeline").
- **Node Library**: A searchable, categorized left-hand sidebar containing functional blocks such as Triggers, Flow control & branching, Data shaping, APIs & connectivity, and Security.
- **Node Properties Editor**: A contextual right-hand sidebar that dynamically populates with a selected node's specific settings, allowing for parameter adjustments and advanced JSON configuration.
- **General Rule:** Follow the task statement exactly. Only perform actions necessary to complete the requested task. Do not make unrelated changes.

## Environment Architecture & Interaction Paradigms

To successfully complete tasks in this environment, adhere to the following UI behaviors and state mechanics:

- **Click-based Instantiation**: New workflow steps are added by searching for the desired node in the left-hand "Nodes" panel and clicking on it - this makes node appear directly on the canvas.
- **Edge Routing (Connections)**: The flow of execution is dictated by edges (lines). You can connect nodes by clicking and dragging from the output port (right side) of a preceding node to the input port (left side) of a subsequent node.
- **Edge Deletion**: To break a connection, you must select the existing edge on the canvas to open its menu in the right-hand panel, then click the "Delete edge" button.
\end{Verbatim}
}
\subsubsection{3D Modeller}
{\footnotesize
\begin{Verbatim}[breaklines=true,breakanywhere=true,bgcolor=gray!8,bgcolorpadding=8pt]
# APPLICATION BACKGROUND

## Application Overview

3D Modeller is a web application used for viewing and manipulating 3D scenes, specifically designed for completing objective-based tasks.
The main workspace/interface displays a central 3D viewport, a scene hierarchy panel, an inspector panel, and a task prompt sidebar, where users can select 3D objects, modify their transformations and appearance, and follow specific instructional steps.

## Features & Capabilities

- **Scene Management Dashboard:** A home screen allowing users to browse, open, and manage various pre-configured 3D environments (e.g., "Four Objects Scene", "Basic Shapes").
- **Scene Hierarchy Panel:** A left-sided menu that lists all active objects within the current 3D space, providing a straightforward way to locate and select specific items.
- **Inspector Panel:** A detailed right-sided menu that displays the properties of the currently selected object. Users can modify the object's `Name`, `Transform` attributes (Position, Rotation, Scale across X, Y, and Z axes), and `Appearance` (Color, Opacity, Wireframe toggle).
- **Precise Color Picker:** An appearance tool allowing users to alter object colors using either precise HEX codes or specific RGB (Red, Green, Blue) integer values ranging from 0-255.
- **General Rule:** Follow the task statement exactly. Only perform actions necessary to complete the requested task. Do not make unrelated changes.

## Environment Architecture & Interaction Paradigms

To successfully complete tasks in this environment, adhere to the following UI behaviors and state mechanics:

- **Selection-Dependent Editing:** The Inspector panel remains blank ("No object selected") until an object is actively clicked either in the 3D viewport or from the Scene list. Selecting a new object immediately updates the Inspector to reflect that specific object's current state.
- **Direct Input Manipulation:** While objects can visually be seen in the viewport, structural and aesthetic changes (like moving an object to an exact coordinate or setting a specific color) are achieved by typing specific numeric values directly into the input fields within the Inspector panel.
\end{Verbatim}
}
\subsubsection{Flight Analyser}
{\footnotesize
\begin{Verbatim}[breaklines=true,breakanywhere=true,bgcolor=gray!8,bgcolorpadding=8pt]
# APPLICATION BACKGROUND

## Application Overview

Flight Analyser is a web application used for tracking, visualizing, and replaying aircraft flight data.
The main workspace/interface displays a map populated with aircraft icons alongside a task panel, where users can search for specific flights, adjust historical playback times, view flight trajectories, and extract detailed telemetry.

## Features & Capabilities

- **Temporal Navigation**: Adjust the temporal state of the map using the playback controls and timeline slider located at the top center of the screen to set exact UTC times.
- **Flight Search**: Locate specific aircraft by typing their callsign into the Search bar situated in the top left corner.
- **Flight Details View**: Selecting an aircraft highlights its flight path and opens a detailed panel overlaying the map, detailing its route, origin/destination, current status, and specific Flight Data (Altitude, Geo Altitude, Ground Speed, Vertical Rate).
- **Radius Tool**: Draw or clear dashed boundaries around target airports using the "Draw Radius Circle" tool (accessed via the target icon next to the pause button).
- **General Rule:** Follow the task statement exactly. Only perform actions necessary to complete the requested task. Do not make unrelated changes.

## Environment Architecture & Interaction Paradigms

To successfully complete tasks in this environment, adhere to the following UI behaviors and state mechanics:

- **Sequential Data Retrieval**: The interface displays detailed data for one flight at a time.
- **Panel Scrolling**: Specific telemetry metrics are located under the "Flight Data" section of the details panel. You may need to scroll down within the panel to locate fields like Geo Altitude or Ground Speed.
\end{Verbatim}
}

\subsubsection{Circuit Designer}

{\footnotesize
The Circuit application is adapted from CircuitJS1~\cite{sharpcircuitjs1}, an open-source browser-based circuit simulator ported from Paul Falstad's original Java applet by Iain Sharp. 
\begin{Verbatim}[breaklines=true,breakanywhere=true,bgcolor=gray!8,bgcolorpadding=8pt]
# APPLICATION BACKGROUND

## Application Overview

Circuit Designer is a web application used for designing and simulating analog and digital circuits.  
The main workspace displays a circuit canvas, where users can place components, wire them together, and observe live simulation behavior.

## Features & Capabilities

- **Component Placement:** Add parts from the **Draw** menu, top search button, or by right-clicking empty canvas space.
- **Circuit Wiring:** Connect terminals by clicking and dragging between component endpoints. Crossing wires do not connect unless joined at endpoints.
- **Component Editing:** Right-click a placed component to edit values, duplicate, or delete it.
- **Measurement:** Hover over a component to view live values in the bottom-right information panel.
- **General Rule:** Follow the task statement exactly. Only perform actions necessary to complete the requested task. Do not make unrelated changes.

## Environment Architecture & Interaction Paradigms

To successfully complete tasks in this environment, adhere to the following UI behaviors and state mechanics:

- **Right-Click Menus:** Most components are added through nested right-click menus on empty canvas space.
- **Property Dialogs:** Values such as resistance or voltage are changed through popup dialogs after selecting **Edit**.
- **Drag Wiring:** Connections are made by dragging from one terminal to another.
- **Logic Input Placement:** Draw logic inputs from right to left for proper orientation.
- **Live Updates:** Simulation runs continuously, so allow time for values to stabilize before measuring.
\end{Verbatim}
}

%% file: G_Appendix.tex
\clearpage

\newcommand{\prompttext}[1]{{\ttfamily\scriptsize #1}}

\section{GauntletBench Task Illustrations}
\label{appendix:tasks}

This appendix lists all 135 tasks in GauntletBench across our five environments. Each table contains \textbf{No.} indicating the task identifier, \textbf{Task}, \textbf{Ground Truth}, and \textbf{Initial State}, where applicable.

\noindent\textbf{Difficulty:}
\colorbox{green!15}{\phantom{\rule{0.8em}{0.8em}}} indicates easy testcases;\quad
\colorbox{yellow!25}{\phantom{\rule{0.8em}{0.8em}}} indicates medium-difficulty testcases;\quad
\colorbox{red!15}{\phantom{\rule{0.8em}{0.8em}}} indicates hard testcases.

\subsection{Video Editor}
\input{applications_testcases/testcases_video_editor}
\subsection{Workflow Builder}
\input{applications_testcases/testcases_graph}
\subsection{3D Modeller}
\input{applications_testcases/testcases_3d_editor}
\subsection{Flight Analyser}
\input{applications_testcases/testcases_flight_radar}
\subsection{Circuit Designer}
\input{applications_testcases/testcases_circuit}

%
%



%% file: applications_testcases/testcases_video_editor.tex
\scriptsize
\renewcommand{\arraystretch}{1.18}
\setlength{\LTleft}{0pt}
\setlength{\LTright}{0pt}
\setlength{\tabcolsep}{6pt}
\begin{longtable}{|>{\raggedright\vspace{0pt}}p{0.05\linewidth}|>{\raggedright\arraybackslash\vspace{0pt}}p{0.87\linewidth}|}
\hline
\textbf{No.} & \textbf{Task} \\
\hline
\endfirsthead
\hline
\textbf{No.} & \textbf{Task} \\
\hline
\endhead
\hline
\endfoot

\cellcolor{green!15}1 & {\raggedright\scriptsize\ttfamily \# Trim a Video to First 5 Seconds \newline
  \newline
 \#\# GOAL \newline
 Add a video and trim it so that only the first 5 seconds remain. \newline
  \newline
 \#\# STEPS \newline
 1. Open the "Sample Media" dropdown and select "White Candle" and add it to timeline. \newline
 2. Trim the clip so that only the first 5 seconds remain. \newline
 3. Export the result. \newline
  \newline
 \# RESULT FORMAT \newline
  \newline
 ```json \newline
 \{"answer": "done"\} \newline
 ```} \\
\cline{2-2}
\cellcolor{green!15} & \textbf{Ground truth} \\
\cline{2-2}
\cellcolor{green!15} & \includegraphics[width=\linewidth]{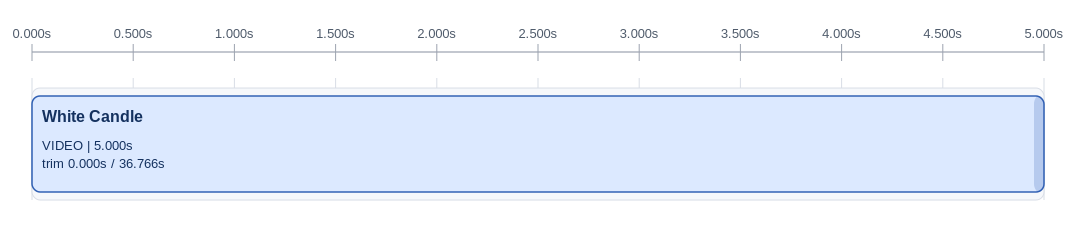} \\
\hline

\cellcolor{green!15}2 & {\raggedright\scriptsize\ttfamily \# Apply Light Correction and Boost Contrast \newline
  \newline
 \#\# GOAL \newline
 Add a video, apply light correction, and increase only the contrast. \newline
  \newline
 \#\# STEPS \newline
 1. Open the "Sample Media" dropdown, select "Flower Video" and add it to the timeline. \newline
 2. Apply light correction to the entire clip by maxing out only the contrast value. \newline
 3. Export the result. \newline
  \newline
 \# RESULT FORMAT \newline
  \newline
 ```json \newline
 \{"answer": "done"\} \newline
 ```} \\
\cline{2-2}
\cellcolor{green!15} & \textbf{Ground truth} \\
\cline{2-2}
\cellcolor{green!15} & \includegraphics[width=\linewidth]{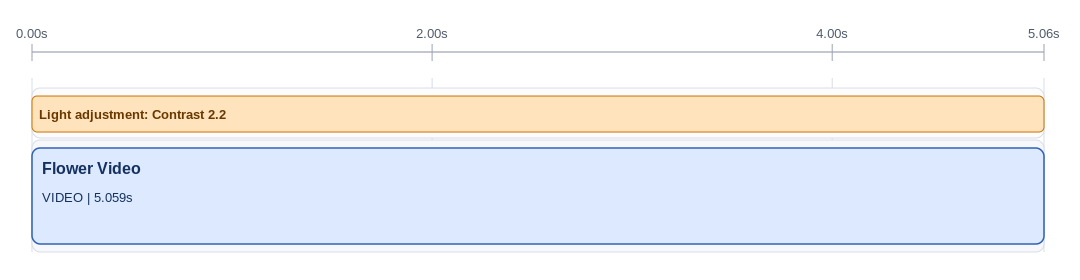} \\
\hline

\cellcolor{green!15}3 & {\raggedright\scriptsize\ttfamily \# A Single Title Card \newline
  \newline
 \#\# GOAL \newline
 Build a timeline that contains one styled text block and nothing else. \newline
  \newline
 \#\# STEPS \newline
 1. Add a text block with the text "Welcome", placed at the very beginning of the timeline. \newline
 2. Set the font color to black. \newline
 3. Set the background color to white. \newline
 4. Set the text block duration to 3 seconds. \newline
 5. Export the result. \newline
  \newline
 \# RESULT FORMAT \newline
  \newline
 ```json \newline
 \{"answer": "done"\} \newline
 ```} \\
\cline{2-2}
\cellcolor{green!15} & \textbf{Ground truth} \\
\cline{2-2}
\cellcolor{green!15} & \includegraphics[width=\linewidth]{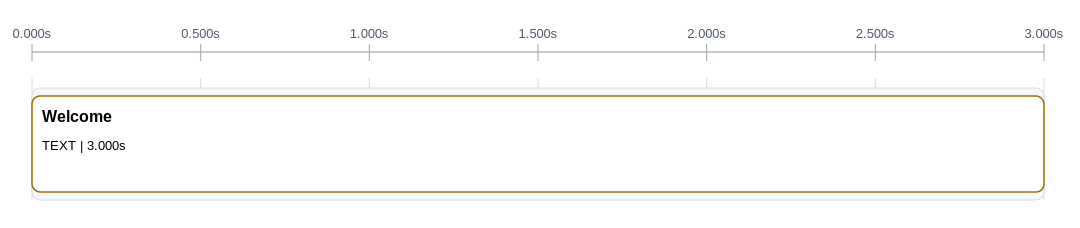} \\
\hline

\cellcolor{green!15}4 & {\raggedright\scriptsize\ttfamily \# Keep Only the Last 4 Seconds \newline
  \newline
 \#\# GOAL \newline
 Add a video to the timeline and trim it from its start so only the ending remains. \newline
  \newline
 \#\# STEPS \newline
 1. Open the "Sample Media" dropdown, select "Tuning a Radio" and add it to the timeline. \newline
 2. Trim the start of the clip so that only the last 4 seconds remain. \newline
 3. Move the trimmed clip to the very beginning of the timeline. \newline
 4. Export the result. \newline
  \newline
 \# RESULT FORMAT \newline
  \newline
 ```json \newline
 \{"answer": "done"\} \newline
 ```} \\
\cline{2-2}
\cellcolor{green!15} & \textbf{Ground truth} \\
\cline{2-2}
\cellcolor{green!15} & \includegraphics[width=\linewidth]{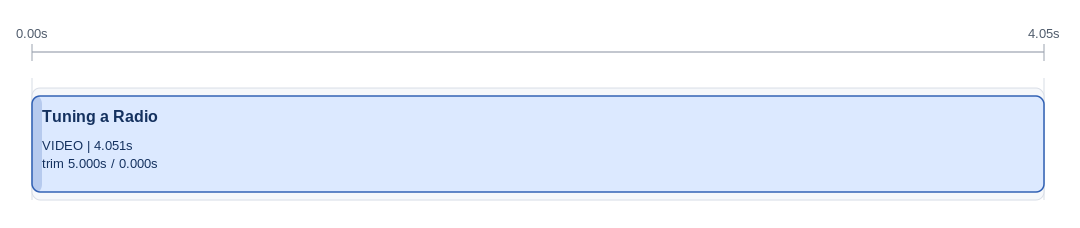} \\
\hline

\cellcolor{green!15}5 & {\raggedright\scriptsize\ttfamily \# Two Clips Back to Back \newline
  \newline
 \#\# GOAL \newline
 Place two full-length videos on the timeline one after the other in a given order. \newline
  \newline
 \#\# STEPS \newline
 1. Open the "Sample Media" dropdown and add "Flower Video" and "Tuning a Radio". \newline
 2. Drag "Flower Video" to the timeline, starting at the beginning. \newline
 3. Drag "Tuning a Radio" to the timeline, placing it immediately after "Flower Video" with no gap. \newline
 4. Do not trim either clip. \newline
 5. Export the result. \newline
  \newline
 \# RESULT FORMAT \newline
  \newline
 ```json \newline
 \{"answer": "done"\} \newline
 ```} \\
\cline{2-2}
\cellcolor{green!15} & \textbf{Ground truth} \\
\cline{2-2}
\cellcolor{green!15} & \includegraphics[width=\linewidth]{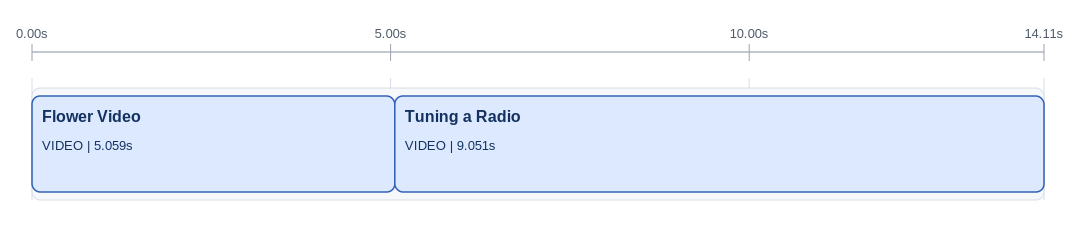} \\
\hline

\cellcolor{green!15}6 & {\raggedright\scriptsize\ttfamily \# Play a Clip Twice \newline
  \newline
 \#\# GOAL \newline
 Place a clip on the timeline and repeat it once directly after itself. \newline
  \newline
 \#\# STEPS \newline
 1. Open the "Sample Media" dropdown, select "Tuning a Radio" and add it to the timeline, starting at the beginning. \newline
 2. Duplicate the clip once and place the copy immediately after the original with no gap. \newline
 3. Do not trim either copy. \newline
 4. Export the result. \newline
  \newline
 \# RESULT FORMAT \newline
  \newline
 ```json \newline
 \{"answer": "done"\} \newline
 ```} \\
\cline{2-2}
\cellcolor{green!15} & \textbf{Ground truth} \\
\cline{2-2}
\cellcolor{green!15} & \includegraphics[width=\linewidth]{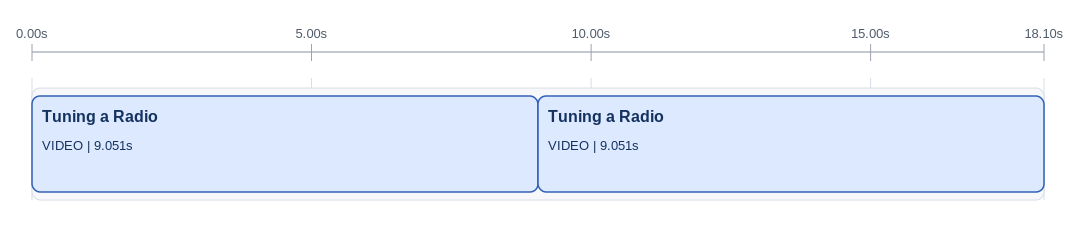} \\
\hline

\cellcolor{green!15}7 & {\raggedright\scriptsize\ttfamily \# Audio at the Start of a Video \newline
  \newline
 \#\# GOAL \newline
 Add a video and an audio clip so the audio plays over the beginning of the video. \newline
  \newline
 \#\# STEPS \newline
 1. Open the "Sample Media" dropdown and add "Flower Video" and the audio clip. \newline
 2. Drag "Flower Video" to the timeline, starting at the beginning, and do not trim it. \newline
 3. Drag the audio clip to a separate track so that it starts at the very beginning of the timeline, in parallel with the video. \newline
 4. Do not trim the audio clip. \newline
 5. Export the result. \newline
  \newline
 \# RESULT FORMAT \newline
  \newline
 ```json \newline
 \{"answer": "done"\} \newline
 ```} \\
\cline{2-2}
\cellcolor{green!15} & \textbf{Ground truth} \\
\cline{2-2}
\cellcolor{green!15} & \includegraphics[width=\linewidth]{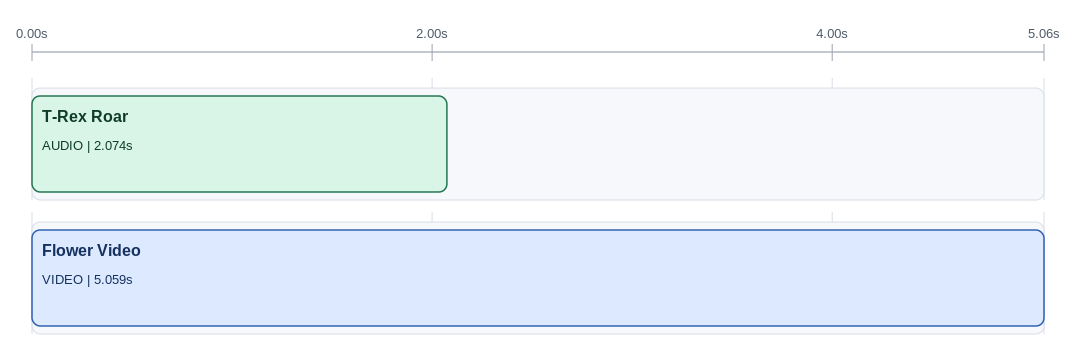} \\
\hline

\cellcolor{green!15}8 & {\raggedright\scriptsize\ttfamily \# Title Card With Background Audio \newline
  \newline
 \#\# GOAL \newline
 Build a timeline with just a text block and an audio clip playing together, with no video. \newline
  \newline
 \#\# STEPS \newline
 1. Add a text block with the text "Now Playing", placed at the very beginning of the timeline. \newline
 2. Set the font color to white. \newline
 3. Set the background color to black. \newline
 4. Set the text block duration to 5 seconds. \newline
 5. Open the "Sample Media" dropdown and add the audio clip. \newline
 6. Drag the audio clip to a separate track so that it starts at the very beginning of the timeline, in parallel with the text block. \newline
 7. Do not trim the audio clip. \newline
 8. Export the result. \newline
  \newline
 \# RESULT FORMAT \newline
  \newline
 ```json \newline
 \{"answer": "done"\} \newline
 ```} \\
\cline{2-2}
\cellcolor{green!15} & \textbf{Ground truth} \\
\cline{2-2}
\cellcolor{green!15} & \includegraphics[width=\linewidth]{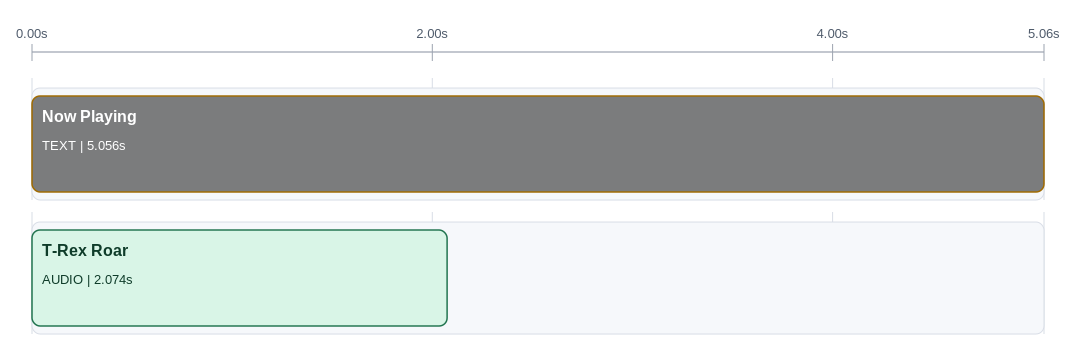} \\
\hline

\cellcolor{green!15}9 & {\raggedright\scriptsize\ttfamily \# Short Opener Before the Main Clip \newline
  \newline
 \#\# GOAL \newline
 Use a short trimmed clip as an opener in front of a second, untrimmed clip. \newline
  \newline
 \#\# STEPS \newline
 1. Open the "Sample Media" dropdown and add "White Candle" and "Flower Video". \newline
 2. Drag "White Candle" to the timeline, starting at the beginning, and trim it so that only the first 2 seconds remain. \newline
 3. Drag "Flower Video" to the timeline, placing it immediately after "White Candle" with no gap, and do not trim it. \newline
 4. Export the result. \newline
  \newline
 \# RESULT FORMAT \newline
  \newline
 ```json \newline
 \{"answer": "done"\} \newline
 ```} \\
\cline{2-2}
\cellcolor{green!15} & \textbf{Ground truth} \\
\cline{2-2}
\cellcolor{green!15} & \includegraphics[width=\linewidth]{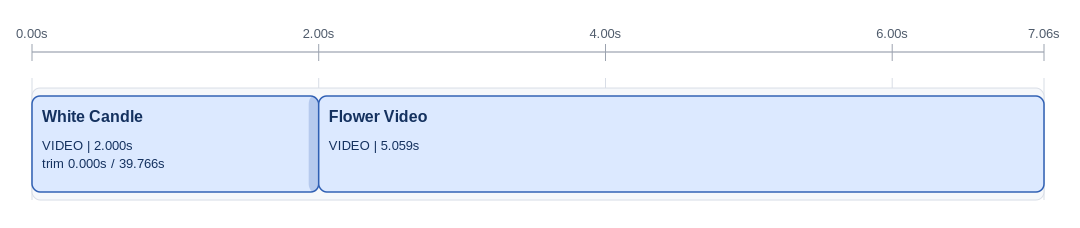} \\
\hline

\cellcolor{orange!15}10 & {\raggedright\scriptsize\ttfamily \# Apply Fade-Out Transition \newline
  \newline
 \#\# GOAL \newline
 Add a video and apply a fade-out transition to its last few seconds. \newline
  \newline
 \#\# STEPS \newline
 1. Open the "Sample Media" dropdown, select "Flower Video" and add it to the timeline. \newline
 2. Apply a fade-out transition targeting the last 3 seconds of the clip. \newline
 3. Export the result. \newline
  \newline
 \# RESULT FORMAT \newline
  \newline
 ```json \newline
 \{"answer": "done"\} \newline
 ```} \\
\cline{2-2}
\cellcolor{orange!15} & \textbf{Ground truth} \\
\cline{2-2}
\cellcolor{orange!15} & \includegraphics[width=\linewidth]{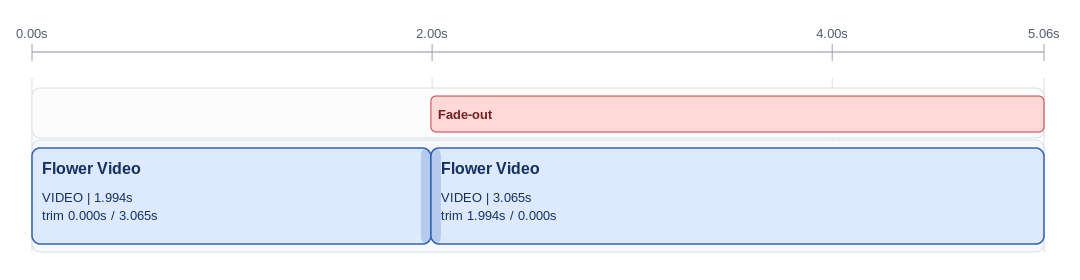} \\
\hline

\cellcolor{orange!15}11 & {\raggedright\scriptsize\ttfamily \# Trim a Video to a Specific Portion \newline
  \newline
 \#\# GOAL \newline
 Add a video and trim it to keep only the segment from 4s to 11s. \newline
  \newline
 \#\# STEPS \newline
 1. Open the "Sample Media" dropdown, add "White Candle" video and drag to the timeline. \newline
 2. Trim the video so that only the 4.00-11.00 seconds chunk remain. \newline
 3. Export the result. \newline
  \newline
 \# RESULT FORMAT \newline
  \newline
 ```json \newline
 \{"answer": "done"\} \newline
 ```} \\
\cline{2-2}
\cellcolor{orange!15} & \textbf{Ground truth} \\
\cline{2-2}
\cellcolor{orange!15} & \includegraphics[width=\linewidth]{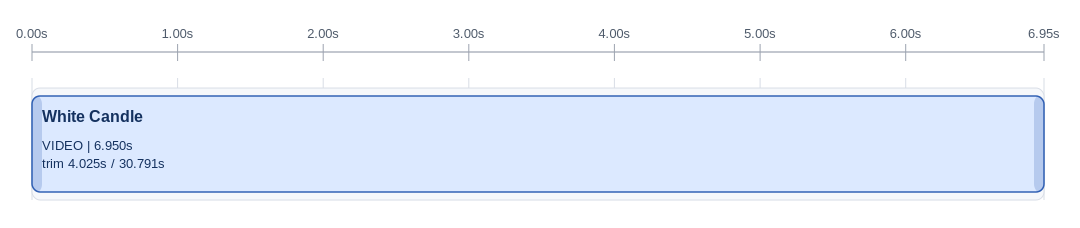} \\
\hline

\cellcolor{orange!15}12 & {\raggedright\scriptsize\ttfamily \# Add Background Audio to End of Video \newline
  \newline
 \#\# GOAL \newline
 Add a video and an audio clip, positioning the audio as background for the end of the video. \newline
  \newline
 \#\# STEPS \newline
 1. Open the "Sample Media" dropdown, add "Flower Video" and audio clip to the timeline. \newline
 2. Position the audio clip so that it plays as background audio during the end portion of the video clip. \newline
 3. Export the result. \newline
  \newline
 \# RESULT FORMAT \newline
  \newline
 ```json \newline
 \{"answer": "done"\} \newline
 ```} \\
\cline{2-2}
\cellcolor{orange!15} & \textbf{Ground truth} \\
\cline{2-2}
\cellcolor{orange!15} & \includegraphics[width=\linewidth]{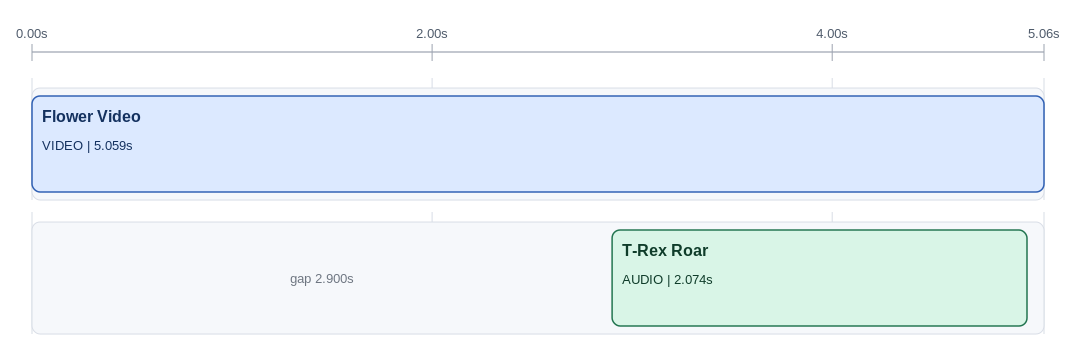} \\
\hline

\cellcolor{orange!15}13 & {\raggedright\scriptsize\ttfamily \# Combine Three Videos in Sequence \newline
  \newline
 \#\# GOAL \newline
 Place three videos on the timeline in a specific order and export. \newline
  \newline
 \#\# STEPS \newline
 1. Open the "Sample Media" dropdown and add "Tuning a Radio", "White Candle" and "Flower Video". \newline
 2. Drag all the videos to timeline and place them in same order. \newline
 3. Trim ends of all videos so length of each clip matches length of "Flower Video" clip. \newline
 4. Export the result. \newline
  \newline
 \# RESULT FORMAT \newline
  \newline
 ```json \newline
 \{"answer": "done"\} \newline
 ```} \\
\cline{2-2}
\cellcolor{orange!15} & \textbf{Ground truth} \\
\cline{2-2}
\cellcolor{orange!15} & \includegraphics[width=\linewidth]{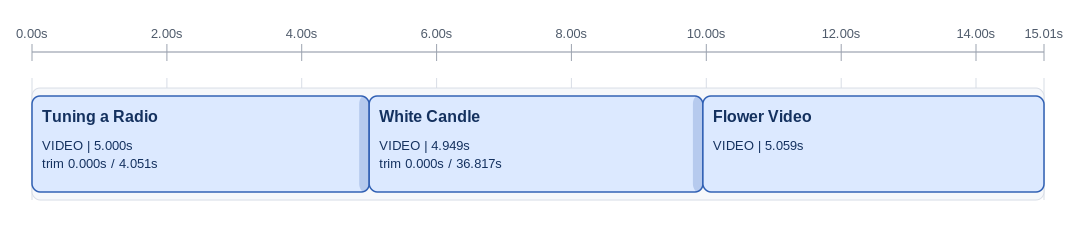} \\
\hline

\cellcolor{orange!15}14 & {\raggedright\scriptsize\ttfamily \# The Quote \newline
 \newline
\#\# GOAL \newline
Lift a middle segment out of a video, play it on its own first, then play the full original video. \newline
 \newline
\#\# STEPS \newline
1. Open the "Sample Media" dropdown and add "Tuning a Radio". \newline
2. On the timeline, first play only seconds 3 through 6 of the clip, then immediately after play the full original clip from start to end. \newline
3. Add a 1-second text block reading "Earlier..." with white text on black background, placed immediately before the full playthrough begins. \newline
4. Export the result. \newline
 \newline
\# RESULT FORMAT \newline
 \newline
```json \newline
\{"answer": "done"\} \newline
```} \\
\cline{2-2}
\cellcolor{orange!15} & \textbf{Ground truth} \\
\cline{2-2}
\cellcolor{orange!15} & \includegraphics[width=\linewidth]{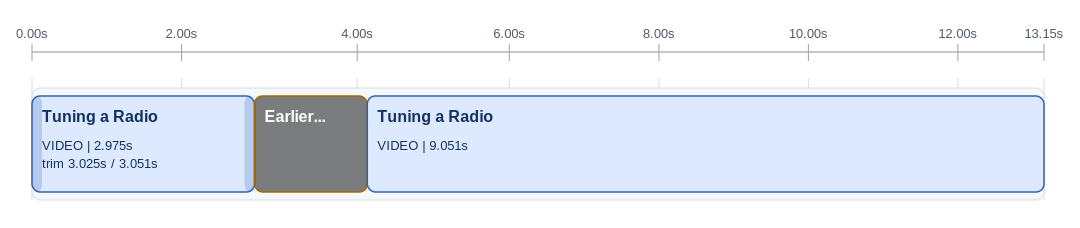} \\
\hline

\cellcolor{orange!15}15 & {\raggedright\scriptsize\ttfamily \# Split Video and Insert Another in the Middle \newline
  \newline
 \#\# GOAL \newline
 Split a video at its midpoint and insert a second video between the two halves. \newline
  \newline
 \#\# STEPS \newline
 1. Open the "Sample Media" dropdown and select "Flower Video". \newline
 2. Drag "Flower Video" to the timeline. \newline
 3. Split the clip at its midpoint. \newline
 4. Open the "Sample Media" dropdown and select "Tuning a Radio". \newline
 5. Drag "Tuning a Radio" to the timeline. \newline
 6. Insert "Tuning a Radio" between the two halves of "Flower Video". \newline
 7. Export the result. \newline
  \newline
 \# RESULT FORMAT \newline
  \newline
 ```json \newline
 \{"answer": "done"\} \newline
 ```} \\
\cline{2-2}
\cellcolor{orange!15} & \textbf{Ground truth} \\
\cline{2-2}
\cellcolor{orange!15} & \includegraphics[width=\linewidth]{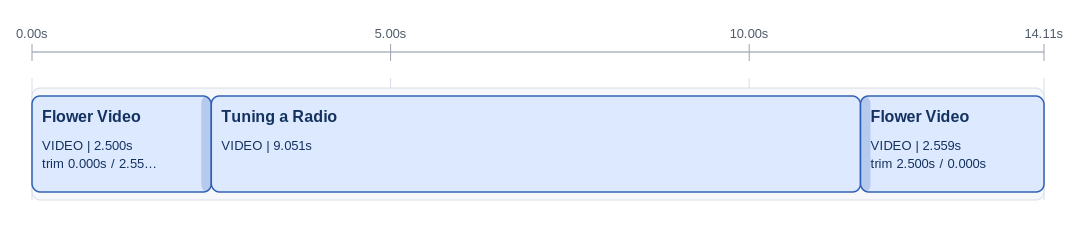} \\
\hline

\cellcolor{orange!15}16 & {\raggedright\scriptsize\ttfamily \# Reverse Order, Light Correction, and Fade-Out \newline
  \newline
 \#\# GOAL \newline
 Arrange two videos in reverse order, apply light correction to a portion of the first, and apply a fade-out to the second. \newline
  \newline
 \#\# STEPS \newline
 1. Open the "Sample Media" dropdown and add "Flower Video". \newline
 2. Open the "Sample Media" dropdown and add "Tuning a Radio". \newline
 3. Drag "Tuning a Radio" to the timeline first. \newline
 4. Drag "Flower Video" to the timeline, placing it right after "Tuning a Radio". \newline
 5. Select "Tuning a Radio" and apply light correction to the first 4 seconds of the clip: max out contrast only. \newline
 6. Select "Flower Video" and apply a fade-out effect with ease-out easing to the last 3 seconds of the clip. \newline
 7. Export the result. \newline
  \newline
 \# RESULT FORMAT \newline
  \newline
 ```json \newline
 \{"answer": "done"\} \newline
 ```} \\
\cline{2-2}
\cellcolor{orange!15} & \textbf{Ground truth} \\
\cline{2-2}
\cellcolor{orange!15} & \includegraphics[width=\linewidth]{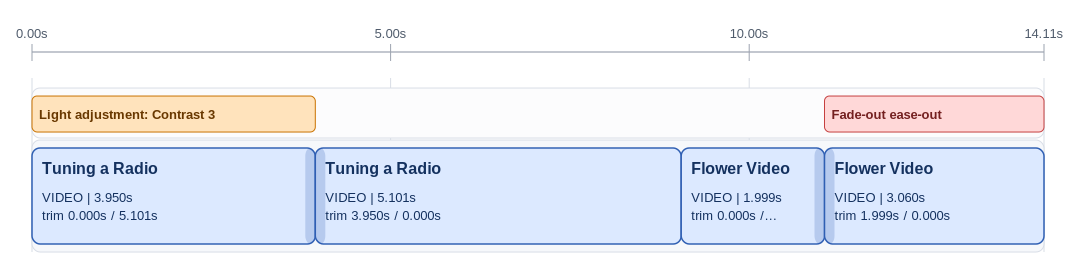} \\
\hline

\cellcolor{orange!15}17 & {\raggedright\scriptsize\ttfamily \# Symmetric Audio Bridge \newline
 \newline
\#\# GOAL \newline
Place an audio clip so its midpoint aligns with the transition between two videos, then trim both videos so the audio has equal silent runway on both sides. \newline
 \newline
\#\# STEPS \newline
1. Open the "Sample Media" dropdown and add "Flower Video", "Tuning a Radio" and audio clip. \newline
2. Drag "Flower Video" and "Tuning a Radio" to the timeline back-to-back in that order. \newline
3. Place the audio clip on a separate track so that the middle of the audio clip aligns exactly with the cut between the two videos. \newline
4. Trim "Flower Video" from its start and "Tuning a Radio" from its end so that the audio clip begins exactly 1 second after the video timeline starts and ends exactly 1 second before the timeline ends. \newline
5. Export the result. \newline
 \newline
\# RESULT FORMAT \newline
 \newline
```json \newline
\{"answer": "done"\} \newline
```} \\
\cline{2-2}
\cellcolor{orange!15} & \textbf{Ground truth} \\
\cline{2-2}
\cellcolor{orange!15} & \includegraphics[width=\linewidth]{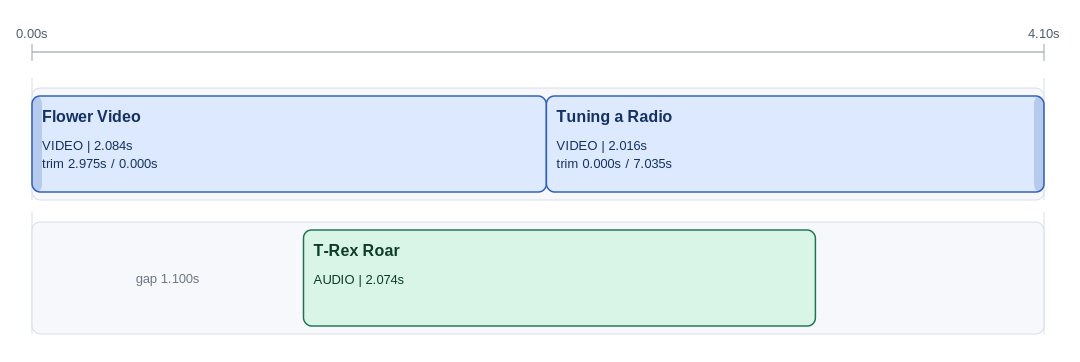} \\
\hline

\cellcolor{orange!15}18 & {\raggedright\scriptsize\ttfamily \# Split a Video, Swap Halves, and Insert a Midpoint Title \newline
  \newline
 \#\# GOAL \newline
 Split a video at its midpoint, swap the two halves, and insert a styled text block between them. \newline
  \newline
 \#\# STEPS \newline
 1. Open the "Sample Media" dropdown and select "Tuning a Radio". \newline
 2. Drag "Tuning a Radio" to the timeline. \newline
 3. Split the clip at its midpoint. \newline
 4. Move the second half to the beginning of the timeline. \newline
 5. Add a text block with the text "Midpoint". \newline
 6. Set the font color to white. \newline
 7. Set the background color to black. \newline
 8. Set the text block duration to 1 second. \newline
 9. Place the text block immediately after the second half. \newline
 10. Move the first half to immediately after the text block. \newline
 11. Verify the final timeline order is: second half of "Tuning a Radio" -> "Midpoint" text block -> first half of "Tuning a Radio". \newline
 12. Export the result. \newline
  \newline
 \# RESULT FORMAT \newline
  \newline
 ```json \newline
 \{"answer": "done"\} \newline
 ```} \\
\cline{2-2}
\cellcolor{orange!15} & \textbf{Ground truth} \\
\cline{2-2}
\cellcolor{orange!15} & \includegraphics[width=\linewidth]{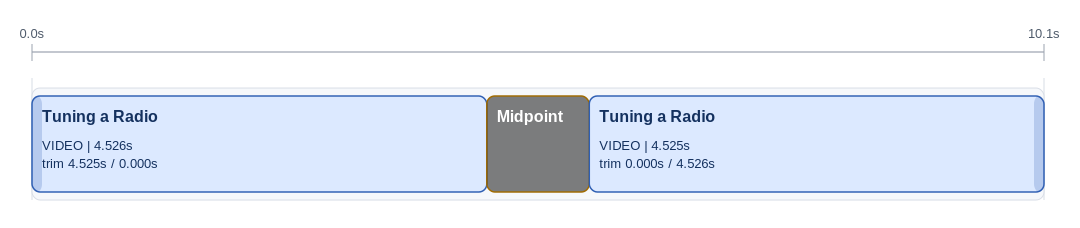} \\
\hline

\cellcolor{red!15}19 & {\raggedright\scriptsize\ttfamily \# Trim and Duplicate a Video \newline
  \newline
 \#\# GOAL \newline
 Trim a video, then duplicate several times. \newline
  \newline
 \#\# STEPS \newline
 1. Open the "Sample Media" dropdown, add "Flower Video" and drag it to timeline. \newline
 2. Trim the clip so that 4 first seconds remain, duplicate the video 4 times, but trim even numbered video to 2 last seconds. \newline
 4. Export the result. \newline
  \newline
 \# RESULT FORMAT \newline
  \newline
 ```json \newline
 \{"answer": "done"\} \newline
 ```} \\
\cline{2-2}
\cellcolor{red!15} & \textbf{Ground truth} \\
\cline{2-2}
\cellcolor{red!15} & \includegraphics[width=\linewidth]{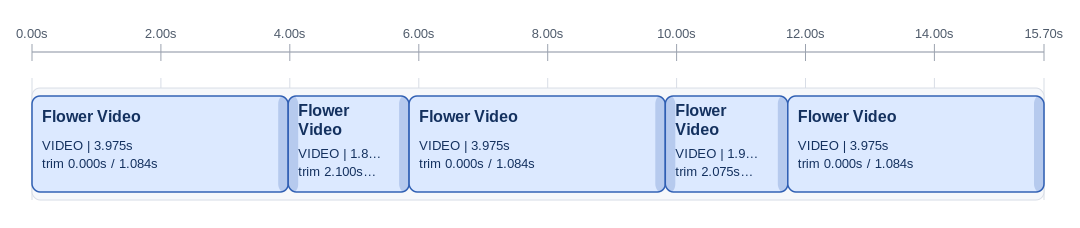} \\
\hline

\cellcolor{red!15}20 & {\raggedright\scriptsize\ttfamily \# Three Videos with Empty Gaps Between Them \newline
  \newline
 \#\# GOAL \newline
 Arrange three videos in sequence with specific empty gaps between them. \newline
  \newline
 \#\# STEPS \newline
 1. Open the "Sample Media" dropdown and add "Flower Video", "White Candle" and "Tuning a Radio" videos. \newline
 2. Drag the videos to timeline in the same order. \newline
 3. Place the videos with empty gaps of 1 second and 2 seconds between them. \newline
 4. Export the result. \newline
  \newline
 \# RESULT FORMAT \newline
  \newline
 ```json \newline
 \{"answer": "done"\} \newline
 ```} \\
\cline{2-2}
\cellcolor{red!15} & \textbf{Ground truth} \\
\cline{2-2}
\cellcolor{red!15} & \includegraphics[width=\linewidth]{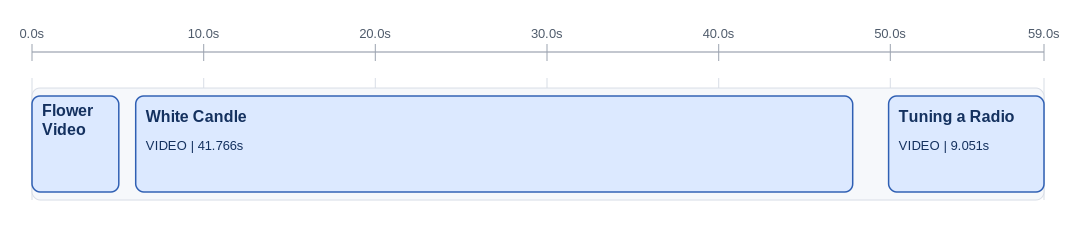} \\
\hline

\cellcolor{red!15}21 & {\raggedright\scriptsize\ttfamily \# Different Light Corrections on Three Videos \newline
  \newline
 \#\# GOAL \newline
 Arrange three videos in sequence and apply a different light correction to each. \newline
  \newline
 \#\# STEPS \newline
 1. Open the "Sample Media" dropdown and add "Flower Video", "White Candle" and "Tuning a Radio". \newline
 2. Drag the video to timeline in asked above order. \newline
 3. Apply light correction one each video clip: max out contrast for first, brightness for second and saturation for last video respectively. \newline
 4. Export the result. \newline
  \newline
 \# RESULT FORMAT \newline
  \newline
 ```json \newline
 \{"answer": "done"\} \newline
 ```} \\
\cline{2-2}
\cellcolor{red!15} & \textbf{Ground truth} \\
\cline{2-2}
\cellcolor{red!15} & \includegraphics[width=\linewidth]{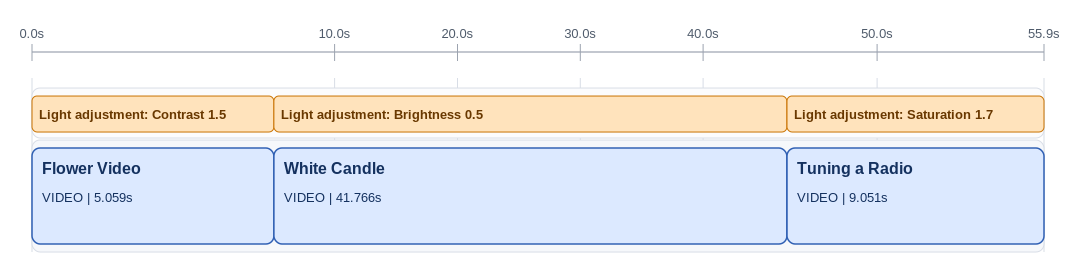} \\
\hline

\cellcolor{red!15}22 & {\raggedright\scriptsize\ttfamily \# Two Videos with Audio at Transition \newline
  \newline
 \#\# GOAL \newline
 Arrange two videos in sequence and place an audio clip so it spans the transition between them. \newline
  \newline
 \#\# STEPS \newline
 1. Open the "Sample Media" dropdown and add "Flower Video", "Tuning a Radio" and audio clip. \newline
 2. Drag the videos to the timeline placing "Tuning a Radio" right after "Flower Video". \newline
 3. Drag the audio clip to the timeline. \newline
 4. Position the audio clip so that it spans the transition point between the two videos (equally overlapping the end of the first and the beginning of the second). \newline
 5. Export the result. \newline
  \newline
 \# RESULT FORMAT \newline
  \newline
 ```json \newline
 \{"answer": "done"\} \newline
 ```} \\
\cline{2-2}
\cellcolor{red!15} & \textbf{Ground truth} \\
\cline{2-2}
\cellcolor{red!15} & \includegraphics[width=\linewidth]{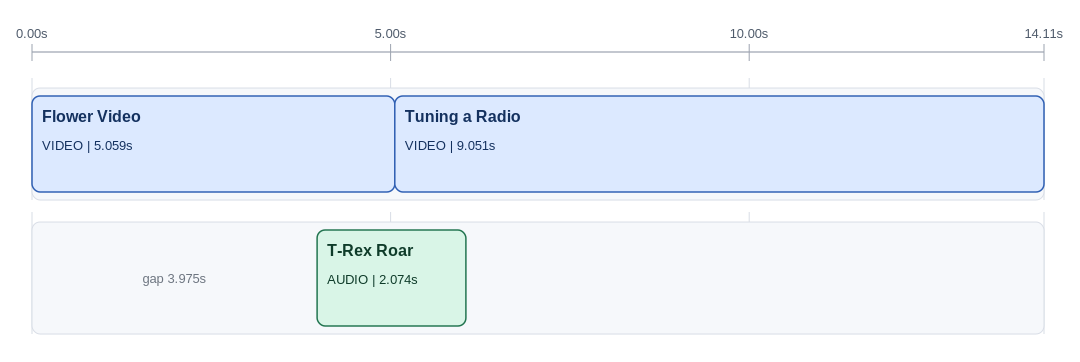} \\
\hline

\cellcolor{red!15}23 & {\raggedright\scriptsize\ttfamily \# Three Colored Text Blocks \newline
  \newline
 \#\# GOAL \newline
 Create an intro video with different color scenes.  \newline
  \newline
 \#\# STEPS \newline
 1. Add a half-second text block with white "Second!" text. \newline
 2. Duplicate the text block two times, setting new blocks texts, so that the timeline reads: "Second!" -> "Third!" -> "First!", resulting in three consecutive blocks on the timeline. \newline
 3. Change text blocks background colors to be "yellow", "red" and "green" accordingly to their timeline order. \newline
 4. Change durations of text blocks, so that each one is 2 times longer than previous one, leaving first text block duration unchanged. \newline
 5. Export the result. \newline
  \newline
 \# RESULT FORMAT \newline
  \newline
 ```json \newline
 \{"answer": "done"\} \newline
 ```} \\
\cline{2-2}
\cellcolor{red!15} & \textbf{Ground truth} \\
\cline{2-2}
\cellcolor{red!15} & \includegraphics[width=\linewidth]{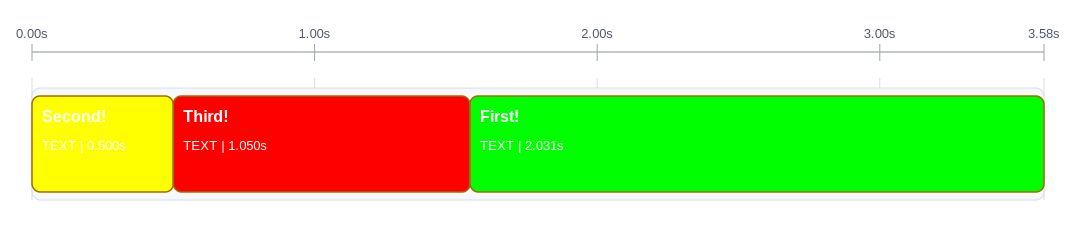} \\
\hline

\cellcolor{red!15}24 & {\raggedright\scriptsize\ttfamily \# Trimmed Video with Closing Titles and Fade-Out \newline
  \newline
 \#\# GOAL \newline
 Add a video, trim it, append a closing title card, and apply a fade-out to the title. \newline
  \newline
 \#\# STEPS \newline
 1. Open the "Sample Media" dropdown and select "White Candle". \newline
 2. Drag "White Candle" to the timeline. \newline
 3. Trim the end of the clip so that only the first 10 seconds remain. \newline
 4. Add a text block with the text "The End", black font color, and white background. \newline
 5. Set the text block duration to 2 seconds. \newline
 6. Place the text block on the timeline immediately after the trimmed video. \newline
 7. Select the text block and apply a fade-out effect to it. \newline
 8. Export the result. \newline
  \newline
 \# RESULT FORMAT \newline
  \newline
 ```json \newline
 \{"answer": "done"\} \newline
 ```} \\
\cline{2-2}
\cellcolor{red!15} & \textbf{Ground truth} \\
\cline{2-2}
\cellcolor{red!15} & \includegraphics[width=\linewidth]{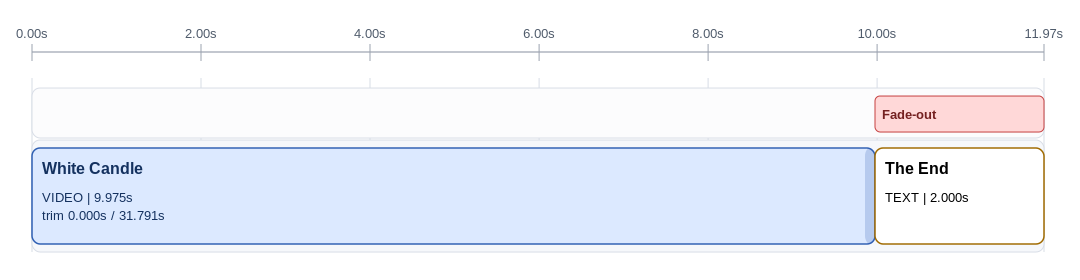} \\
\hline

\cellcolor{red!15}25 & {\raggedright\scriptsize\ttfamily \# Overlay Text and Audio on a Video \newline
  \newline
 \#\# GOAL \newline
 Add a video to the timeline, overlay a text block on the first 2 seconds, and add an audio clip playing in parallel for the same 2 seconds. \newline
  \newline
 \#\# STEPS \newline
 1. Open the "Sample Media" dropdown and add "Flower Video" and audio clip. \newline
 2. Drag "Flower Video" to the timeline. \newline
 3. Add a 2-second "Intro" text block with font size of 72 on a track above the video track so it overlays the first 2 seconds of "Flower Video". \newline
 4. Drag the audio clip to the timeline. \newline
 5. Position the audio clip so it starts at the beginning of the timeline and plays for the first 2 seconds, in parallel with the video and the text overlay. \newline
 6. Export the result. \newline
  \newline
 \# RESULT FORMAT \newline
  \newline
 ```json \newline
 \{"answer": "done"\} \newline
 ```} \\
\cline{2-2}
\cellcolor{red!15} & \textbf{Ground truth} \\
\cline{2-2}
\cellcolor{red!15} & \includegraphics[width=\linewidth]{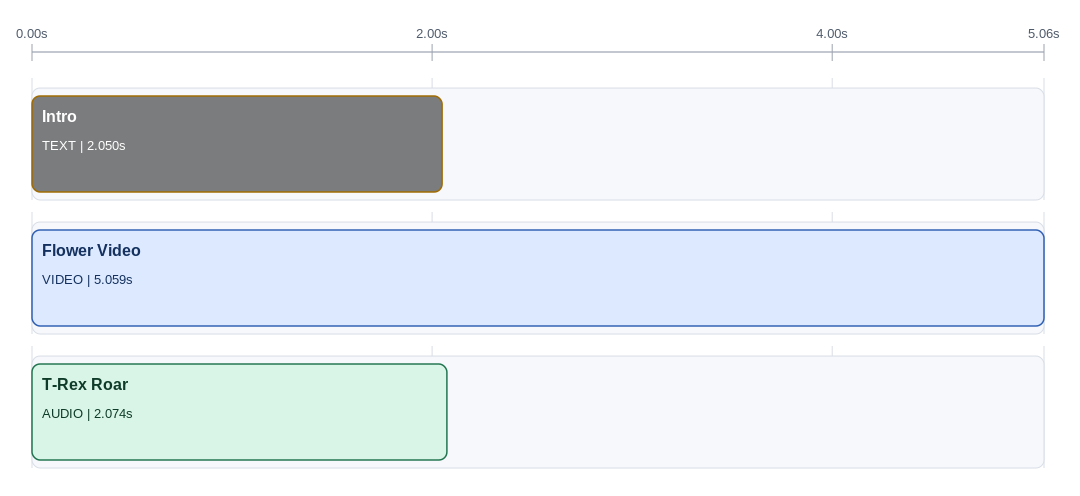} \\
\hline

\cellcolor{red!15}26 & {\raggedright\scriptsize\ttfamily \# Alternating Split Segments with Per-Segment Corrections \newline
 \newline
\#\# GOAL \newline
Split two videos into quarters, interleave them on the timeline, and apply a different light correction to each odd-indexed segment. \newline
 \newline
\#\# STEPS \newline
1. Open the "Sample Media" dropdown and add "Tuning a Radio" and "White Candle". \newline
2. Drag "Tuning a Radio"  to the timeline and split it into 3 equal segments (T1, T2, T3). \newline
3. Drag "White Candle" to the timeline after "Flower Video" and split it into 3 equal segments (W1, W2, W3). \newline
4. Reorder the segments on the timeline so the final order is: T1 -> W1 -> T2 -> W2 -> T3 -> W3. \newline
5. Apply light correction to first two odd-positioned segment on the timeline: max out contrast for first segment, max out brightness for last segment. \newline
6. Trim each of the segments so that all segments on the timeline have the same duration, equal to the shortest segment. \newline
7. Export the result. \newline
 \newline
\# RESULT FORMAT \newline
 \newline
```json \newline
\{"answer": "done"\} \newline
```} \\
\cline{2-2}
\cellcolor{red!15} & \textbf{Ground truth} \\
\cline{2-2}
\cellcolor{red!15} & \includegraphics[width=\linewidth]{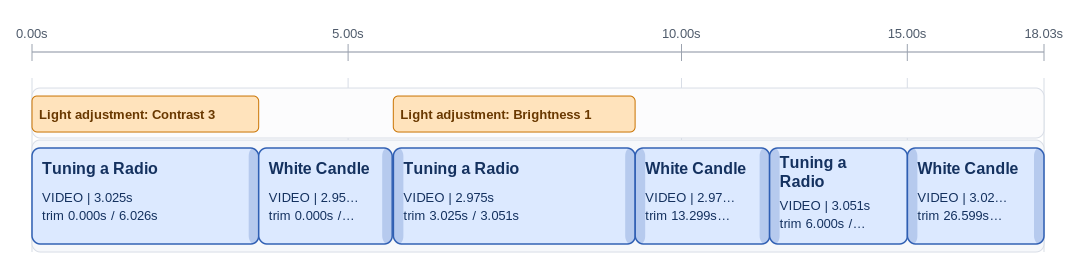} \\
\hline

\cellcolor{red!15}27 & {\raggedright\scriptsize\ttfamily \# Act Titles Before Each Video \newline
  \newline
 \#\# GOAL \newline
 Arrange three videos in sequence, each preceded by a titled text block. \newline
  \newline
 \#\# STEPS \newline
 1. Open the "Sample Media" dropdown and add "Flower Video". \newline
 2. Open the "Sample Media" dropdown and add "White Candle". \newline
 3. Open the "Sample Media" dropdown and add "Tuning a Radio". \newline
 4. Add a text block with text "Act I", black font color, white background, and 1-second duration. Place it on the timeline. \newline
 5. Drag "Flower Video" to the timeline, placing it immediately after the "Act I" text block. \newline
 6. Add a text block with text "Act II", black font color, white background, and 1-second duration. Place it on the timeline immediately after "Flower Video". \newline
 7. Drag "White Candle" to the timeline, placing it immediately after the "Act II" text block. \newline
 8. Add a text block with text "Act III", black font color, white background, and 1-second duration. Place it on the timeline immediately after "White Candle". \newline
 9. Drag "Tuning a Radio" to the timeline, placing it immediately after the "Act III" text block. \newline
 10. Verify the final timeline order is: "Act I" -> "Flower Video" -> "Act II" -> "White Candle" -> "Act III" -> "Tuning a Radio". \newline
 11. Export the result. \newline
  \newline
 \# RESULT FORMAT \newline
  \newline
 ```json \newline
 \{"answer": "done"\} \newline
 ```} \\
\cline{2-2}
\cellcolor{red!15} & \textbf{Ground truth} \\
\cline{2-2}
\cellcolor{red!15} & \includegraphics[width=\linewidth]{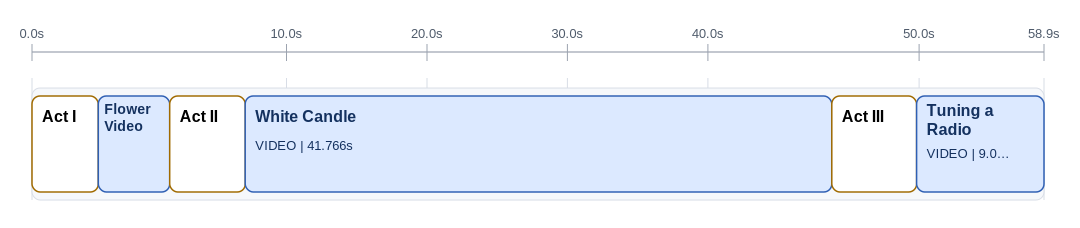} \\
\hline

\end{longtable}
\normalsize

%% file: applications_testcases/testcases_graph.tex
\scriptsize
\renewcommand{\arraystretch}{1.18}
\setlength{\LTleft}{0pt}
\setlength{\LTright}{0pt}
\setlength{\tabcolsep}{6pt}
\begin{longtable}{|>{\raggedright\vspace{0pt}}p{0.05\linewidth}|>{\raggedright\arraybackslash\vspace{0pt}}p{0.87\linewidth}|}
\hline
\textbf{No.} & \textbf{Task} \\
\hline
\endfirsthead
\hline
\textbf{No.} & \textbf{Task} \\
\hline
\endhead
\hline
\endfoot

\cellcolor{green!15}1 & {\raggedright\scriptsize\ttfamily \# Remove a Node and Reconnect in a Pipeline \newline
  \newline
 \#\# GOAL \newline
 Remove a node from a multi-step pipeline and reconnect the upstream node to the downstream node. \newline
  \newline
 \#\# STEPS \newline
 1. Open the "API Data Pipeline" workflow. \newline
 2. Delete the "Sort" node. \newline
 3. Reconnect the "Map" node directly to the next downstream node. \newline
  \newline
 \# RESULT FORMAT \newline
 ```json \newline
 \{"answer": "done"\} \newline
 ```} \\
\cline{2-2}
\cellcolor{green!15} & \textbf{Ground truth} \\
\cline{2-2}
\cellcolor{green!15} & \includegraphics[width=\linewidth]{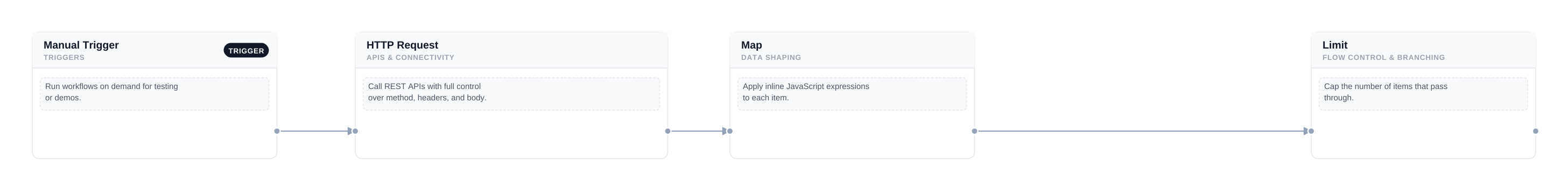} \\
\hline

\cellcolor{green!15}2 & {\raggedright\scriptsize\ttfamily \# Create a Simple Three-Node Workflow \newline
  \newline
 \#\# GOAL \newline
 Create a simple three-node workflow from scratch with a Manual Trigger, a configured HTTP Request, and a No Operation node. \newline
  \newline
 \#\# STEPS \newline
 1. Create a new workflow named "Example Todo Fetch". \newline
 2. Add these nodes and connect them in sequence: `Manual Trigger -> HTTP Request -> No Operation`. \newline
 3. Configure the "HTTP Request" node: \newline
  - Method: "GET" \newline
  - URL: `https://jsonplaceholder.typicode.com/todos/1` \newline
  \newline
 \# RESULT FORMAT \newline
 ```json \newline
 \{"answer": "done"\} \newline
 ```} \\
\cline{2-2}
\cellcolor{green!15} & \textbf{Ground truth} \\
\cline{2-2}
\cellcolor{green!15} & \includegraphics[width=\linewidth]{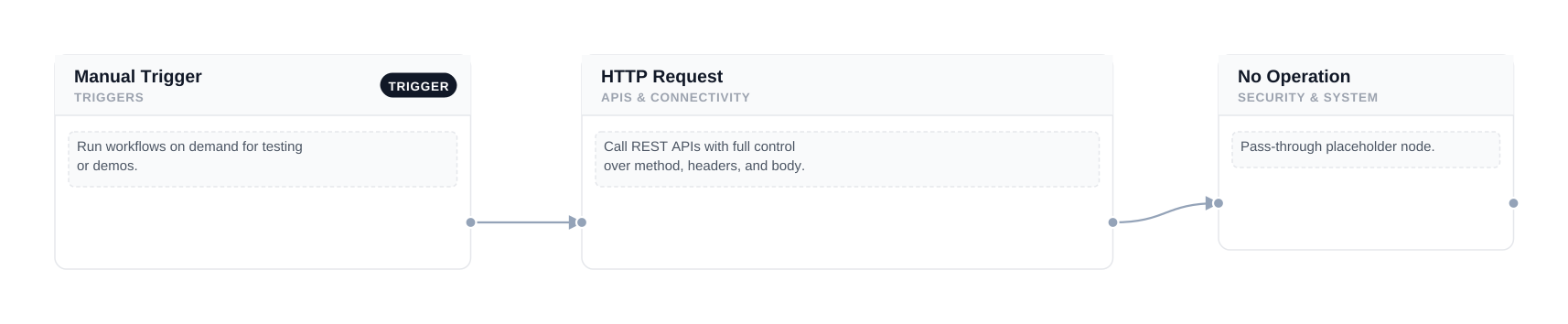} \\
\hline

\cellcolor{green!15}3 & {\raggedright\scriptsize\ttfamily \# Create an Order Snapshot Workflow \newline
  \newline
 \#\# GOAL \newline
 Create a three-node workflow that reads rows from a database table and writes them into a data table. \newline
  \newline
 \#\# STEPS \newline
 1. Create a new workflow named "Order Snapshot". \newline
 2. Add these nodes and connect them in sequence: `Manual Trigger -> Database Connector -> Data Table`. \newline
 3. Configure the "Database Connector" node to read the `orders` table. \newline
 4. Configure the "Data Table" node: \newline
    - Table name: `order\_snapshot` \newline
    - Operation: `insert` \newline
  \newline
 \# RESULT FORMAT \newline
 ```json \newline
 \{"answer": "done"\} \newline
 ```} \\
\cline{2-2}
\cellcolor{green!15} & \textbf{Ground truth} \\
\cline{2-2}
\cellcolor{green!15} & \includegraphics[width=\linewidth]{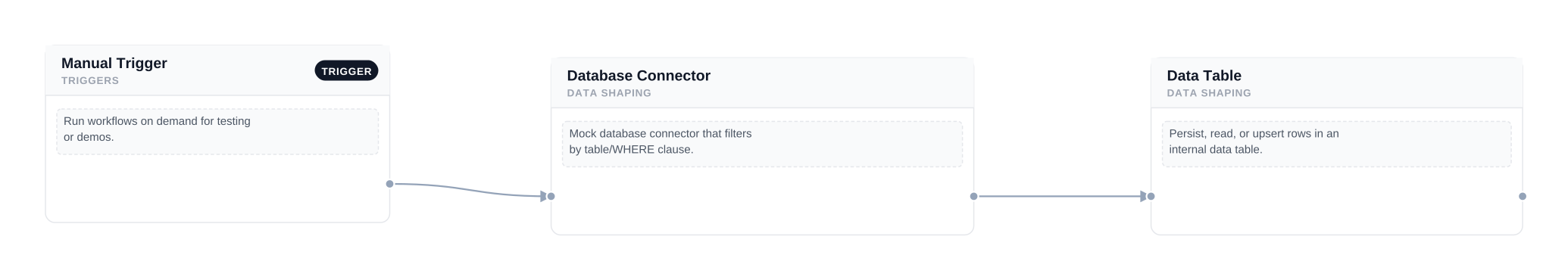} \\
\hline

\cellcolor{green!15}4 & {\raggedright\scriptsize\ttfamily \# Detach the Aggregate Step \newline
  \newline
 \#\# GOAL \newline
 Remove a single connection from a pipeline, leaving every node in place. \newline
  \newline
 \#\# STEPS \newline
 1. Open the "Order Processing Pipeline" workflow. \newline
 2. Delete the connection that goes from the "Merge" node to the "Aggregate" node. \newline
 3. Do not delete any node. \newline
  \newline
 \# RESULT FORMAT \newline
 ```json \newline
 \{"answer": "done"\} \newline
 ```} \\
\cline{2-2}
\cellcolor{green!15} & \textbf{Ground truth} \\
\cline{2-2}
\cellcolor{green!15} & \includegraphics[width=\linewidth]{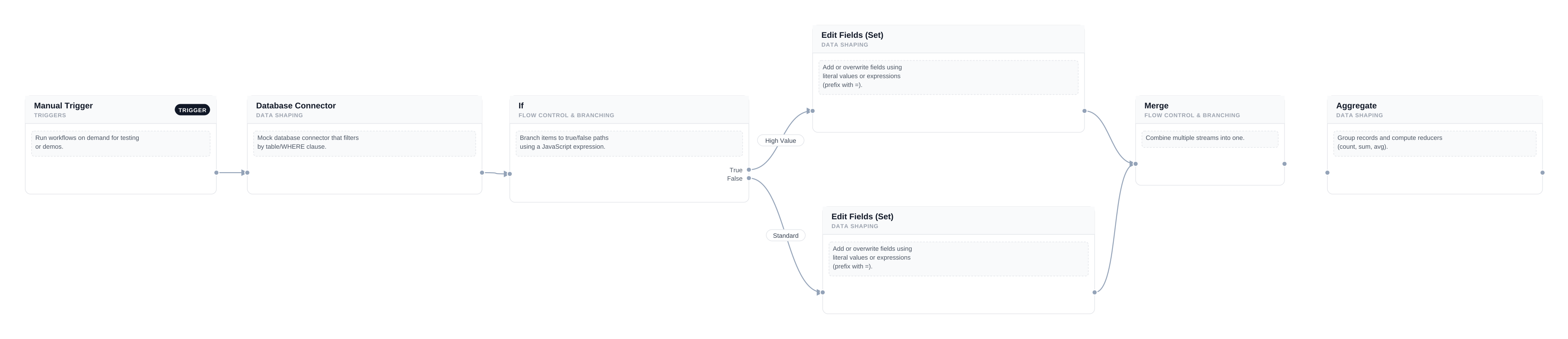} \\
\hline

\cellcolor{green!15}5 & {\raggedright\scriptsize\ttfamily \# Add a Table Output Alongside the Preview \newline
  \newline
 \#\# GOAL \newline
 Add one node to an existing workflow and wire it as a second output of an existing node, leaving the current flow untouched. \newline
  \newline
 \#\# STEPS \newline
 1. Open the "Range -> Map -> Table (scaffolding)" workflow. \newline
 2. Add a "Data Table" node to the canvas. \newline
 3. Connect the "Map" node output to the new "Data Table" node input. \newline
 4. Keep the existing connection from "Map" to "Table Preview" in place. \newline
  \newline
 \# RESULT FORMAT \newline
 ```json \newline
 \{"answer": "done"\} \newline
 ```} \\
\cline{2-2}
\cellcolor{green!15} & \textbf{Ground truth} \\
\cline{2-2}
\cellcolor{green!15} & \includegraphics[width=\linewidth]{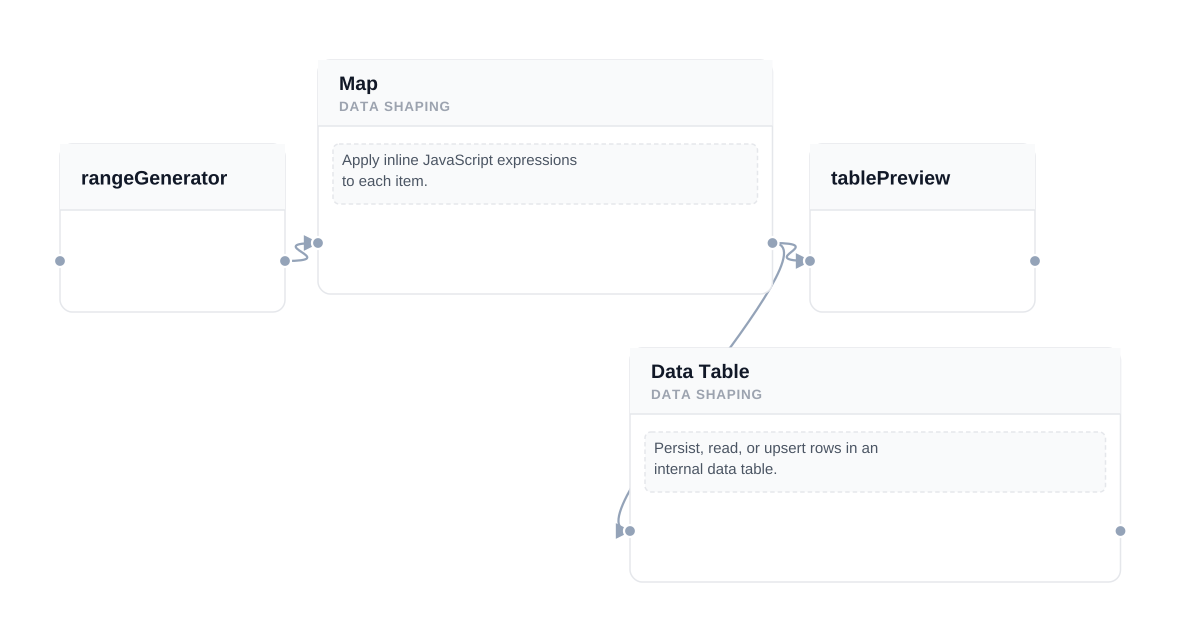} \\
\hline

\cellcolor{green!15}6 & {\raggedright\scriptsize\ttfamily \# Change the Sort Field and Direction \newline
  \newline
 \#\# GOAL \newline
 Reconfigure an existing Sort node in a pipeline. \newline
  \newline
 \#\# STEPS \newline
 1. Open the "API Data Pipeline" workflow. \newline
 2. Open the "Sort" node settings. \newline
 3. Set the sort field to `city` and the direction to `desc`. \newline
  \newline
 \# RESULT FORMAT \newline
 ```json \newline
 \{"answer": "done"\} \newline
 ```} \\
\cline{2-2}
\cellcolor{green!15} & \textbf{Ground truth} \\
\cline{2-2}
\cellcolor{green!15} & \includegraphics[width=\linewidth]{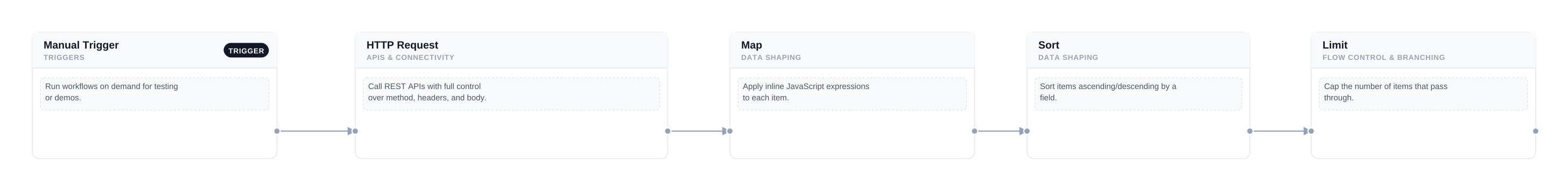} \\
\hline

\cellcolor{green!15}7 & {\raggedright\scriptsize\ttfamily \# Create a Webhook Responder \newline
  \newline
 \#\# GOAL \newline
 Build a two-node workflow that receives a webhook call and answers it. \newline
  \newline
 \#\# STEPS \newline
 1. Create a new workflow named "Webhook Echo". \newline
 2. Add these nodes and connect them: `Webhook Trigger -> Respond to Webhook`. \newline
 3. Configure the "Webhook Trigger" node: \newline
    - Method: "GET" \newline
    - Path: `/hook/echo` \newline
 4. Configure the "Respond to Webhook" node with status code 404. \newline
  \newline
 \# RESULT FORMAT \newline
 ```json \newline
 \{"answer": "done"\} \newline
 ```} \\
\cline{2-2}
\cellcolor{green!15} & \textbf{Ground truth} \\
\cline{2-2}
\cellcolor{green!15} & \includegraphics[width=\linewidth]{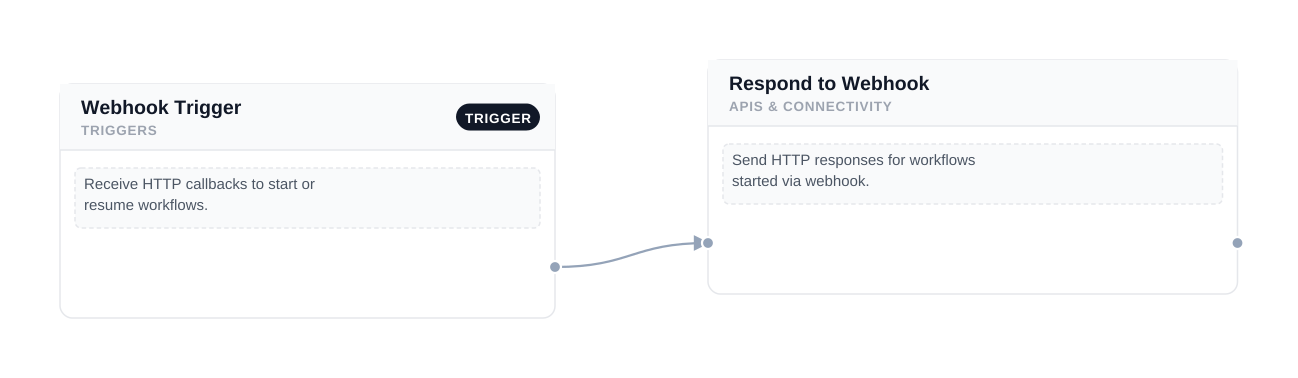} \\
\hline

\cellcolor{green!15}8 & {\raggedright\scriptsize\ttfamily \# Insert a No Operation Node into the API Pipeline \newline
  \newline
 \#\# GOAL \newline
 Insert a node between two existing nodes of a pipeline and rewire the flow through it. \newline
  \newline
 \#\# STEPS \newline
 1. Open the "API Data Pipeline" workflow. \newline
 2. Add a "No Operation" node between the "HTTP Request" and "Map" nodes. \newline
 3. Reconnect the workflow so the path through this section is: `HTTP Request -> No Operation -> Map`. \newline
  \newline
 \# RESULT FORMAT \newline
 ```json \newline
 \{"answer": "done"\} \newline
 ```} \\
\cline{2-2}
\cellcolor{green!15} & \textbf{Ground truth} \\
\cline{2-2}
\cellcolor{green!15} & \includegraphics[width=\linewidth]{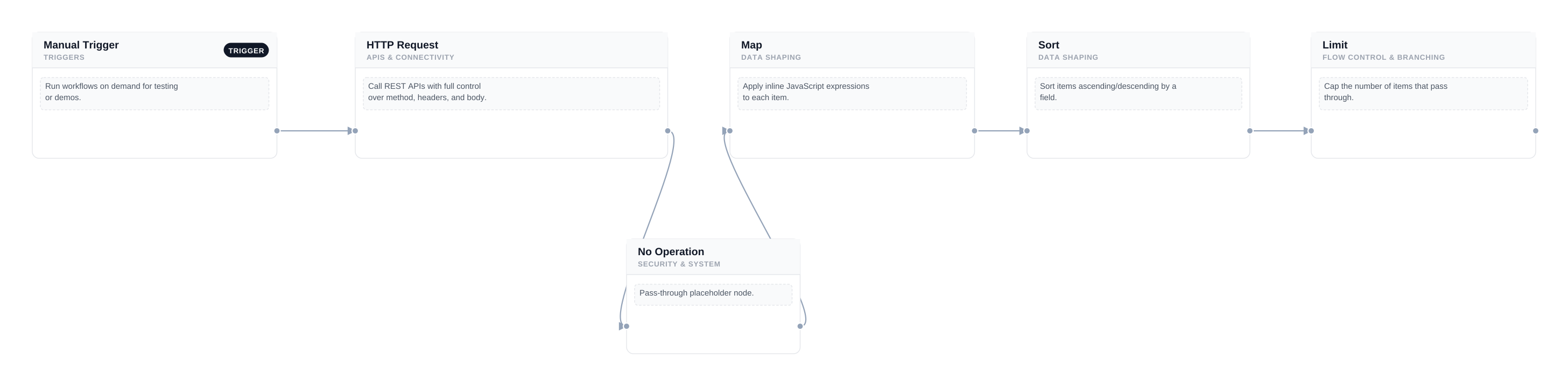} \\
\hline

\cellcolor{green!15}9 & {\raggedright\scriptsize\ttfamily \# Create a Two-Branch Condition \newline
  \newline
 \#\# GOAL \newline
 Build a small workflow where an If node routes items into two different nodes. \newline
  \newline
 \#\# STEPS \newline
 1. Create a new workflow named "Value Router". \newline
 2. Add exactly these nodes: "Manual Trigger", "If", "Data Table", "No Operation". \newline
 3. Set the "If" condition to: `item.value > 5`. \newline
 4. Connect the workflow as follows: \newline
    - `Manual Trigger -> If` \newline
    - The `true` output of "If" goes to the "Data Table" node. \newline
    - The `false` output of "If" goes to the "No Operation" node. \newline
 5. Do not add any other nodes. \newline
  \newline
 \# RESULT FORMAT \newline
 ```json \newline
 \{"answer": "done"\} \newline
 ```} \\
\cline{2-2}
\cellcolor{green!15} & \textbf{Ground truth} \\
\cline{2-2}
\cellcolor{green!15} & \includegraphics[width=\linewidth]{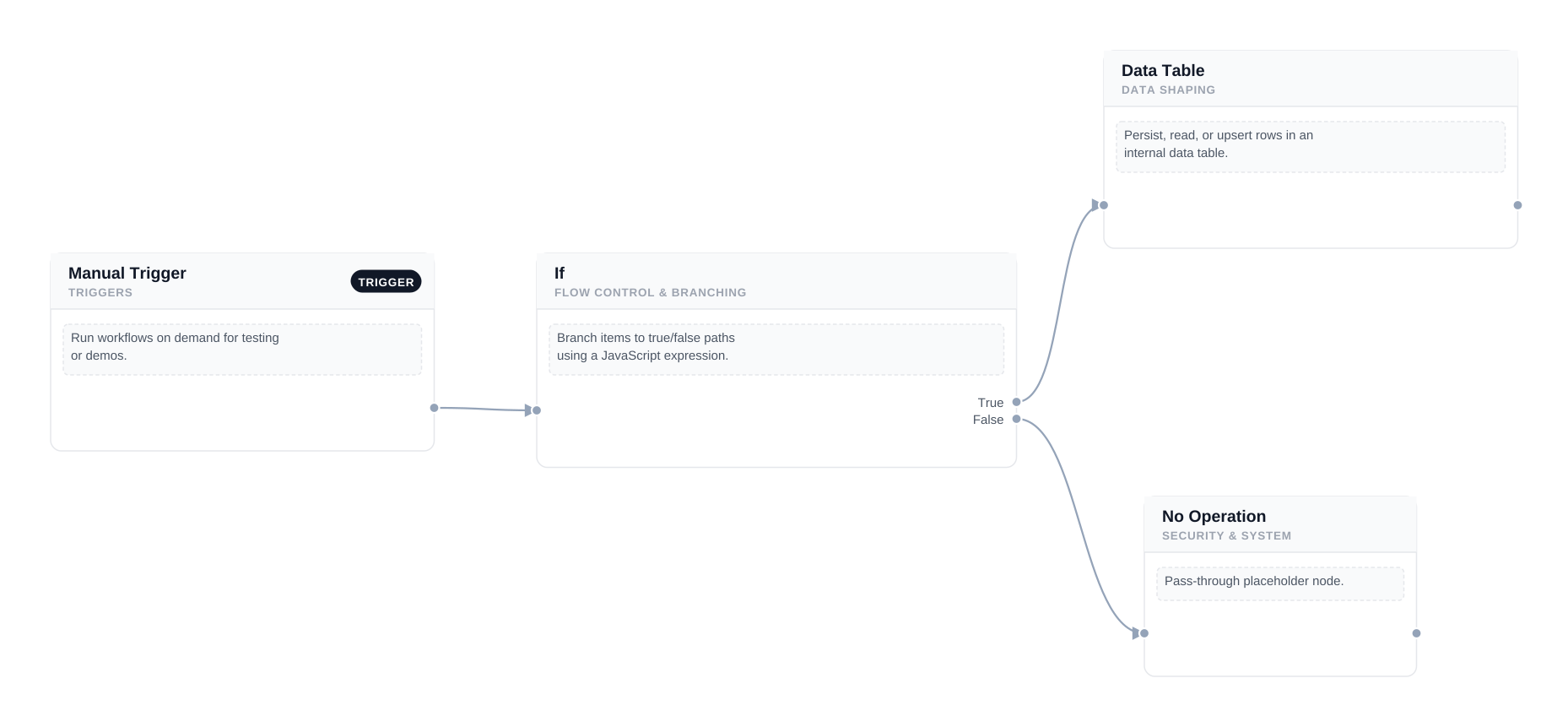} \\
\hline

\cellcolor{orange!15}10 & {\raggedright\scriptsize\ttfamily \# Insert a Wait Node into a Workflow \newline
  \newline
 \#\# GOAL \newline
 Insert a Wait node with a 1-second delay between two existing nodes in a workflow and reconnect the flow. \newline
  \newline
 \#\# STEPS \newline
 1. Open the "Range -> Map -> Table (scaffolding)" workflow. \newline
 2. Insert a "Wait" node between the "Map" and "Table Preview" nodes. \newline
 3. Configure the delay to "1 second". \newline
 4. Reconnect the workflow correctly. \newline
  \newline
 \# RESULT FORMAT \newline
 ```json \newline
 \{"answer": "done"\} \newline
 ```} \\
\cline{2-2}
\cellcolor{orange!15} & \textbf{Ground truth} \\
\cline{2-2}
\cellcolor{orange!15} & \includegraphics[width=\linewidth]{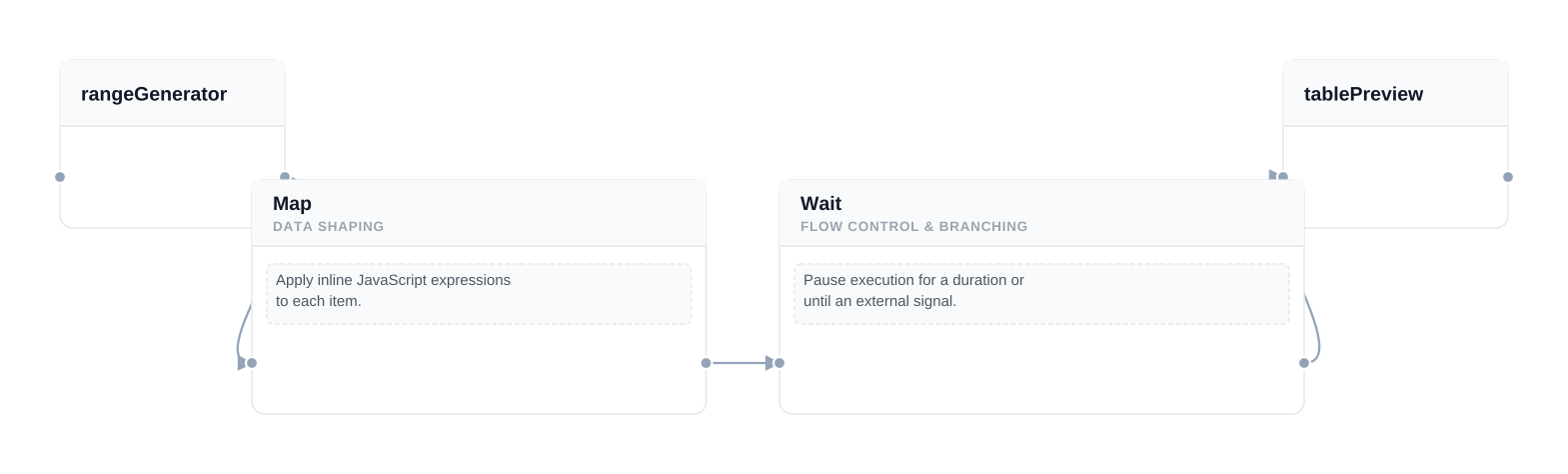} \\
\hline

\cellcolor{orange!15}11 & {\raggedright\scriptsize\ttfamily \# Delete a Node and Disable Default Port in Switch \newline
  \newline
 \#\# GOAL \newline
 Delete a branch target node from a Switch-based pipeline and disable the corresponding default port in the Switch settings. \newline
  \newline
 \#\# STEPS \newline
 1. Open the "Data Categorization Pipeline" workflow. \newline
 2. Delete one of the "Edit Fields (Set)" node responsible for "Default" branch of a "Switch" node. \newline
 3. Disable the "Default" port in the "Switch" node settings. \newline
  \newline
 \# RESULT FORMAT \newline
  \newline
 ```json \newline
 \{"answer": "done"\} \newline
 ```} \\
\cline{2-2}
\cellcolor{orange!15} & \textbf{Ground truth} \\
\cline{2-2}
\cellcolor{orange!15} & \includegraphics[width=\linewidth]{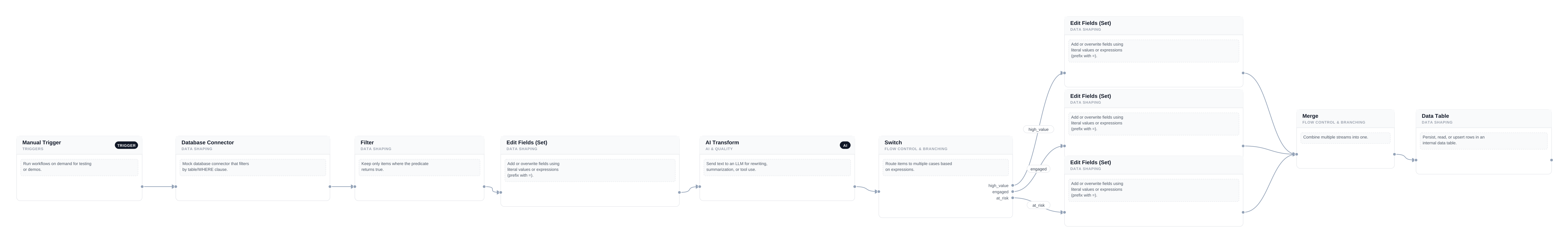} \\
\hline

\cellcolor{orange!15}12 & {\raggedright\scriptsize\ttfamily \# Add a Node Using Search \newline
  \newline
 \#\# GOAL \newline
 Use the node search to find and add a Send Email node, configure its subject, and connect it to the end of a pipeline. \newline
  \newline
 \#\# STEPS \newline
 1. Open the "API Data Pipeline" workflow. \newline
 2. Use the node search box to find "Send Email". \newline
 3. Add the "Send Email" node to the canvas after the "Sort" node. \newline
 4. Set "Send Email" subject to "Users list". \newline
 5. Connect the "Sort" node output to the "Send Email" node input. \newline
  \newline
 \# RESULT FORMAT \newline
 ```json \newline
 \{"answer": "done"\} \newline
 ```} \\
\cline{2-2}
\cellcolor{orange!15} & \textbf{Ground truth} \\
\cline{2-2}
\cellcolor{orange!15} & \includegraphics[width=\linewidth]{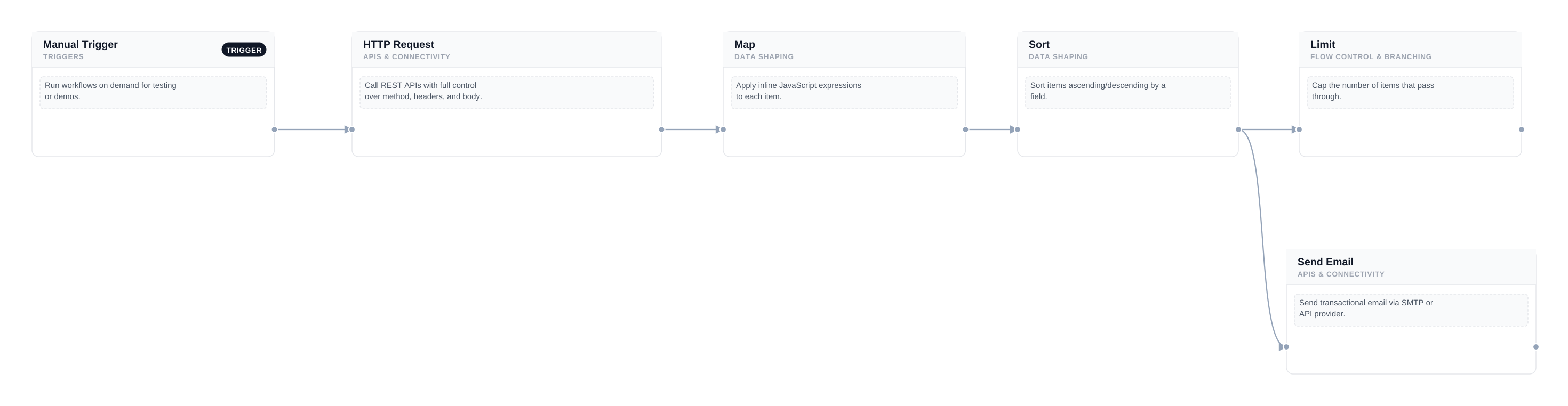} \\
\hline

\cellcolor{orange!15}13 & {\raggedright\scriptsize\ttfamily \# Insert a Formula Node into a Workflow \newline
  \newline
 \#\# GOAL \newline
 Insert a Formula node into a workflow, configure it to compute a derived field, and reconnect the data flow. \newline
  \newline
 \#\# STEPS \newline
 1. Open the "Range -> Map -> Table (scaffolding)" workflow. \newline
 2. Insert a "Formula" node between the "Map" and "Table" nodes. \newline
 3. Configure the "Formula" node to name a new field `double` with value `item.value * 2`. \newline
 4. Reconnect the workflow so the final path is: `Range -> Map -> Formula -> Table`. \newline
  \newline
 \# RESULT FORMAT \newline
 ```json \newline
 \{"answer": "done"\} \newline
 ```} \\
\cline{2-2}
\cellcolor{orange!15} & \textbf{Ground truth} \\
\cline{2-2}
\cellcolor{orange!15} & \includegraphics[width=\linewidth]{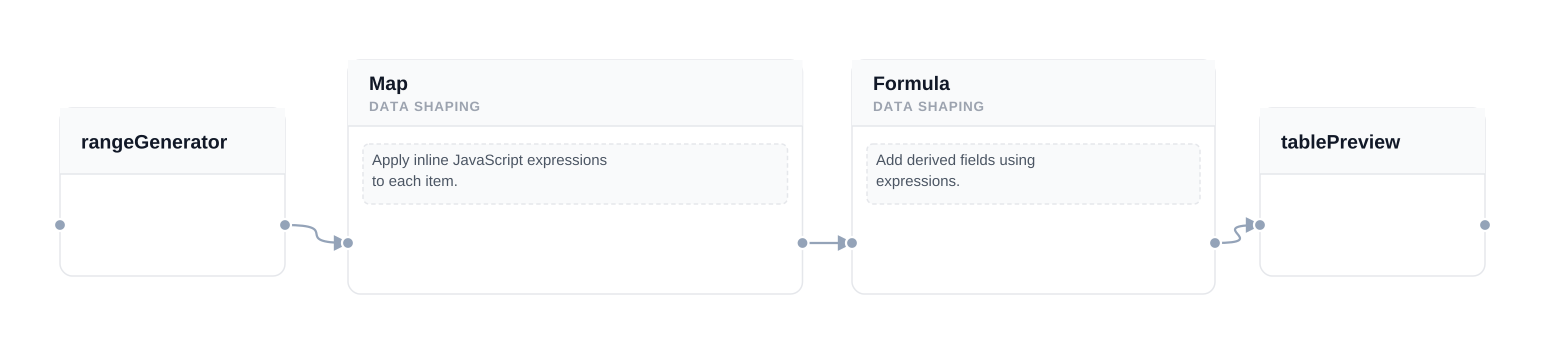} \\
\hline

\cellcolor{orange!15}14 & {\raggedright\scriptsize\ttfamily \# Insert and Configure a Filter Node \newline
  \newline
 \#\# GOAL \newline
 Add and configure a Filter node in an existing workflow, wiring it into the correct position in the data flow. \newline
  \newline
 \#\# STEPS \newline
 1. Open the "Range -> Map -> Table (scaffolding)" workflow. \newline
 2. Delete the rightmost "Table" node. \newline
 3. Add a "Filter" node and connect "Map" node output to the "Filter" node input . \newline
 4. Configure the filter to pass only values where `item.value > 10`. \newline
 5. The final path must be: `Range -> Map -> Filter`. \newline
  \newline
 \# RESULT FORMAT \newline
 ```json \newline
 \{"answer": "done"\} \newline
 ```} \\
\cline{2-2}
\cellcolor{orange!15} & \textbf{Ground truth} \\
\cline{2-2}
\cellcolor{orange!15} & \includegraphics[width=\linewidth]{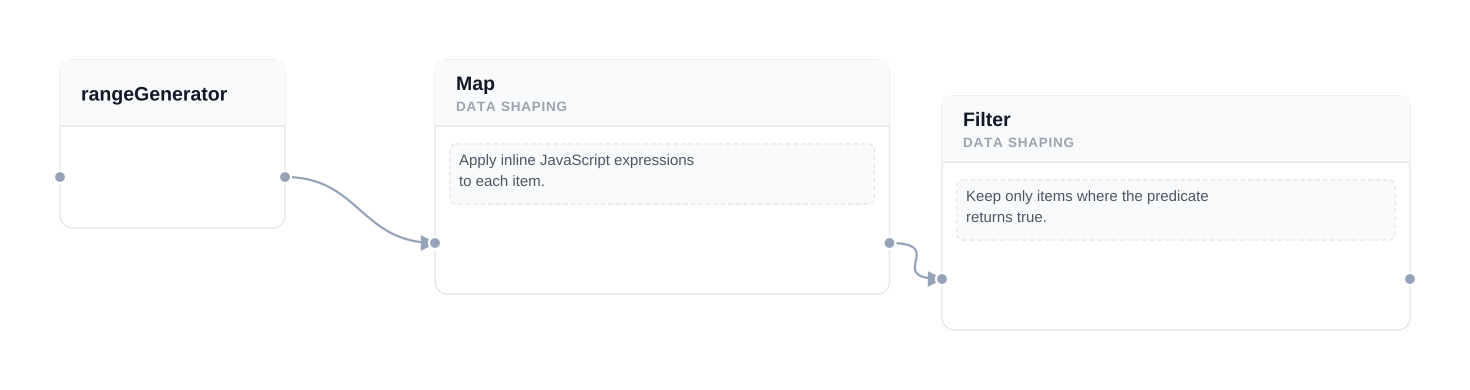} \\
\hline

\cellcolor{orange!15}15 & {\raggedright\scriptsize\ttfamily \# Insert a Filter Node into an Existing Pipeline \newline
  \newline
 \#\# GOAL \newline
 Insert a Filter node between two existing nodes in a pipeline and configure it with a specific condition. \newline
  \newline
 \#\# STEPS \newline
 1. Open the "Order Processing Pipeline" workflow. \newline
 2. Add a "Filter" node between the "Database Connector" and the "If" node. \newline
 3. Configure the filter to pass only orders where `item.status === 'completed'`. \newline
 4. Reconnect the workflow so the path through this section is: `Database Connector -> Filter -> If`. \newline
  \newline
 \# RESULT FORMAT \newline
 ```json \newline
 \{"answer": "done"\} \newline
 ```} \\
\cline{2-2}
\cellcolor{orange!15} & \textbf{Ground truth} \\
\cline{2-2}
\cellcolor{orange!15} & \includegraphics[width=\linewidth]{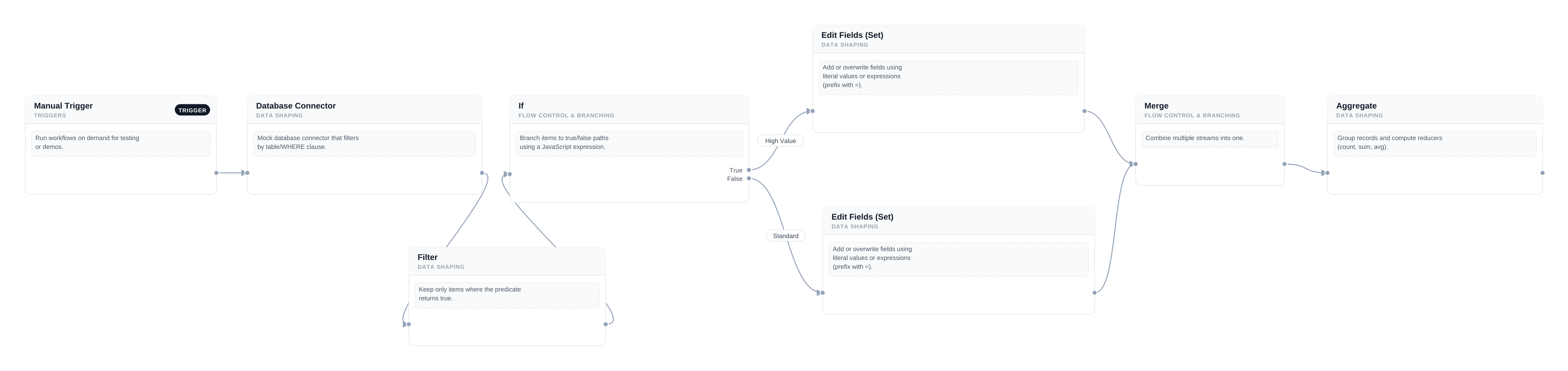} \\
\hline

\cellcolor{orange!15}16 & {\raggedright\scriptsize\ttfamily \# Rewire a Branch Input to a Different Node \newline
  \newline
 \#\# GOAL \newline
 Rewire a branch input on a downstream node so it feeds into a different merge point in the pipeline. \newline
  \newline
 \#\# STEPS \newline
 1. Open the "Enterprise Data Pipeline" workflow. \newline
 2. Find the "Data Table" node that receives inputs from the priority branches. \newline
 3. Disconnect the input coming from the lower "Edit Fields (Set)" node. \newline
 4. Reconnect that lower "Edit Fields (Set)" node into the "HTTP Request" node located on the right, below "Send Email" node. \newline
  \newline
 \# RESULT FORMAT \newline
 ```json \newline
 \{"answer": "done"\} \newline
 ```} \\
\cline{2-2}
\cellcolor{orange!15} & \textbf{Ground truth} \\
\cline{2-2}
\cellcolor{orange!15} & \includegraphics[width=\linewidth]{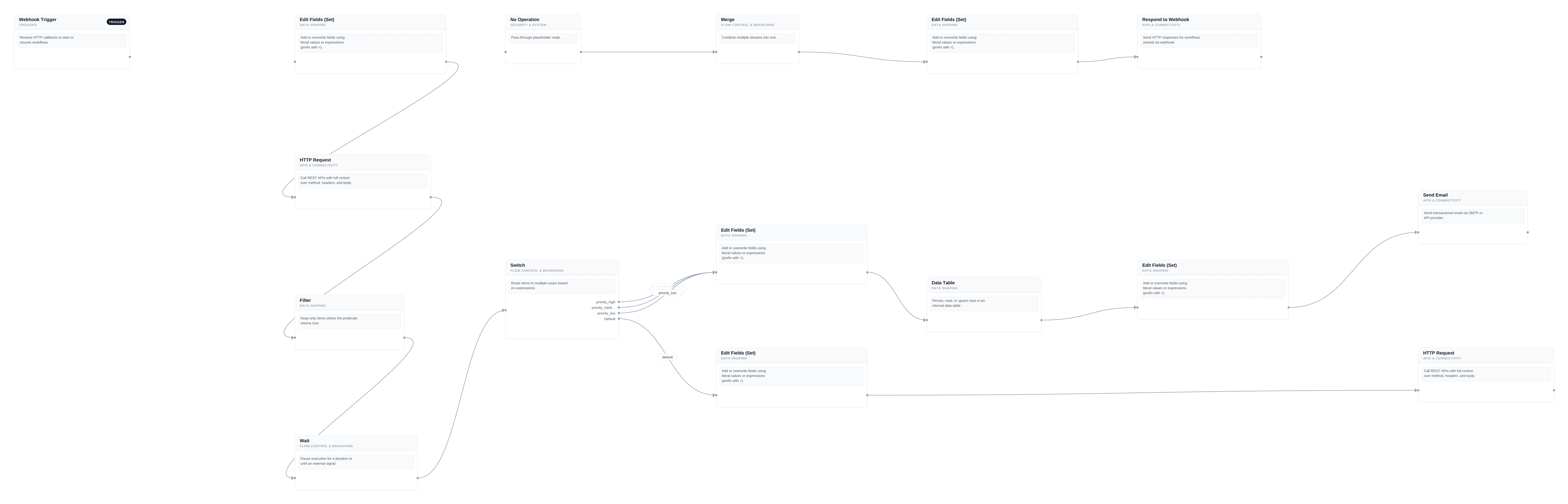} \\
\hline

\cellcolor{orange!15}17 & {\raggedright\scriptsize\ttfamily \# Add an Side Branch Without Breaking Existing Output \newline
 \newline
\#\# GOAL \newline
Add a side branch that records processed records while keeping the original main pipeline output unchanged. \newline
 \newline
\#\# STEPS \newline
1. Open the "Order Processing Pipeline" workflow. \newline
2. Find the node immediately before the final "Aggregate" output node. \newline
3. Add a "No Operation" node. \newline
4. Add a "Data Table" node after the "No Operation" node. \newline
5. Connect the node immediately before the final output node to the new "No Operation" node, leaving all other existing connections as is. \newline
6. Connect "No Operation" to the new "Data Table" node. \newline
 \newline
\# RESULT FORMAT \newline
 \newline
```json \newline
\{"answer": "done"\} \newline
```} \\
\cline{2-2}
\cellcolor{orange!15} & \textbf{Ground truth} \\
\cline{2-2}
\cellcolor{orange!15} & \includegraphics[width=\linewidth]{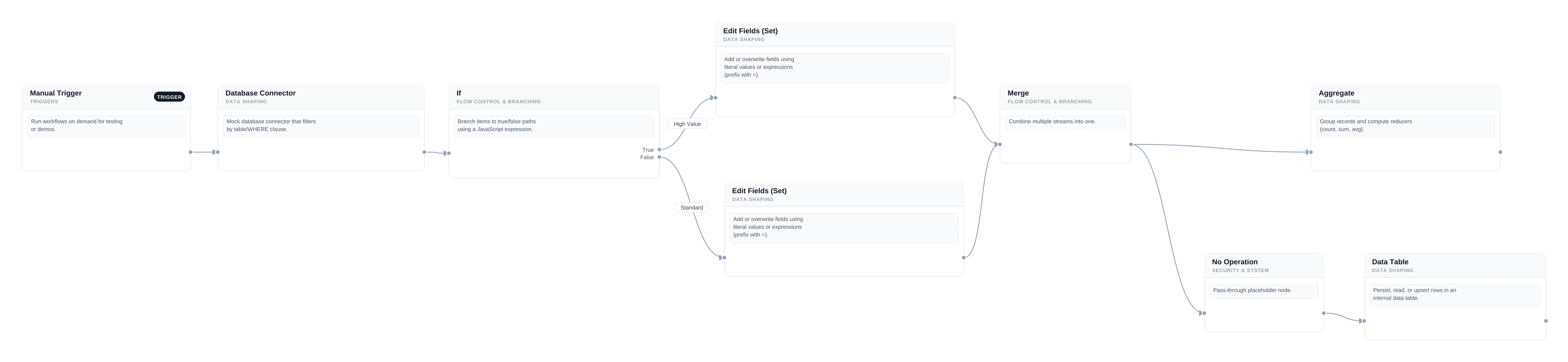} \\
\hline

\cellcolor{orange!15}18 & {\raggedright\scriptsize\ttfamily \# Split a Pipeline into Two Conditional Branches \newline
 \newline
\#\# GOAL \newline
Modify an existing linear workflow so that active and inactive users are processed through separate branches. \newline
 \newline
\#\# STEPS \newline
1. Open the "API Data Pipeline" workflow. \newline
2. Insert an "If" node between the "Map" node and the next downstream node. \newline
3. Configure the "If" condition to check: \newline
   `item.active === true` \newline
4. Add a new "Send Email" node for the true branch. \newline
5. Set the new "Send Email" subject to: \newline
   `Active user report` \newline
6. Connect the workflow so that: \newline
   - `HTTP Request -> Map -> If` \newline
   - The `true` output of "If" goes to the new "Send Email" node. \newline
   - The `false` output of "If" goes to the existing downstream path that originally followed "Map". \newline
 \newline
\# RESULT FORMAT \newline
 \newline
```json \newline
\{"answer": "done"\} \newline
```} \\
\cline{2-2}
\cellcolor{orange!15} & \textbf{Ground truth} \\
\cline{2-2}
\cellcolor{orange!15} & \includegraphics[width=\linewidth]{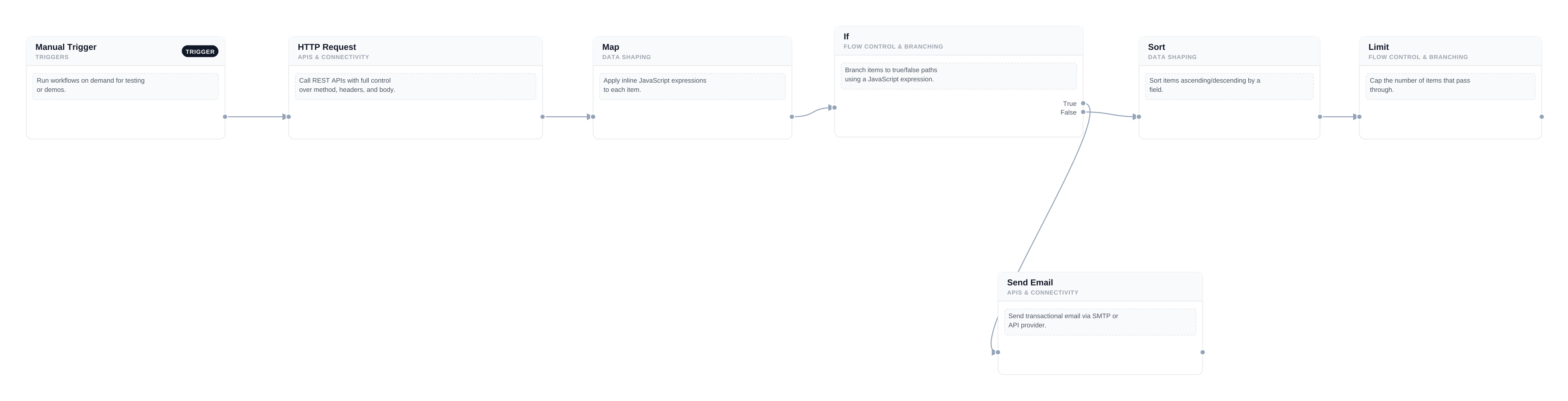} \\
\hline

\cellcolor{red!15}19 & {\raggedright\scriptsize\ttfamily \# Add a New Output Branch to a Switch Node \newline
  \newline
 \#\# GOAL \newline
 Add a new output case to a Switch node in a complex pipeline and wire it to an existing downstream node. \newline
  \newline
 \#\# STEPS \newline
 1. Open the "Enterprise Data Pipeline" workflow. \newline
 2. In the main "Switch" node, add a new case labeled `priority\_urgent`, do not include "expr" field. \newline
 3. Connect the new `priority\_urgent` output to the same "Edit Fields (Set)" node currently connected to the "default" output. \newline
 4. Do not change any other routes or node settings. \newline
  \newline
 \# RESULT FORMAT \newline
 ```json \newline
 \{"answer": "done"\} \newline
 ```} \\
\cline{2-2}
\cellcolor{red!15} & \textbf{Ground truth} \\
\cline{2-2}
\cellcolor{red!15} & \includegraphics[width=\linewidth]{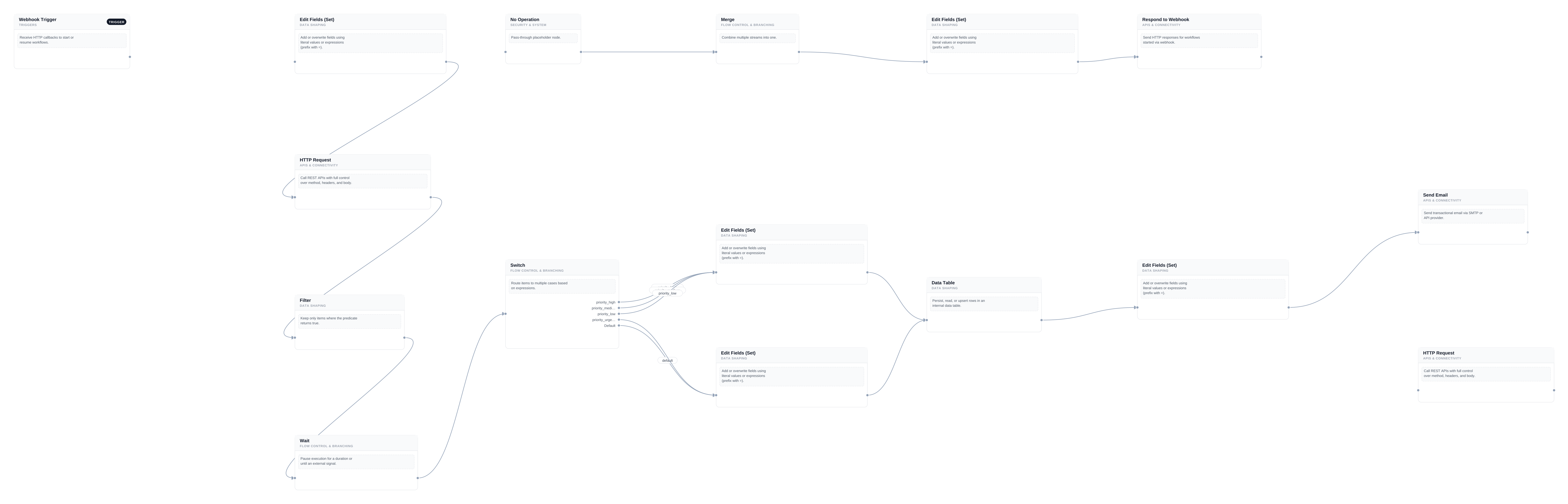} \\
\hline

\cellcolor{red!15}20 & {\raggedright\scriptsize\ttfamily \# Rewire Edges in a Pipeline \newline
  \newline
 \#\# GOAL \newline
 Rewire existing connections in a complex pipeline without adding, deleting, or reconfiguring any nodes. \newline
  \newline
 \#\# STEPS \newline
 1. Open the "Enterprise Data Pipeline" workflow. \newline
 2. Disconnect the edge from the top-row "No Operation" node to the top-row "Merge" node. \newline
 3. Connect that same "No Operation" node to the "Switch" node. \newline
 4. Rewire the `priority\_low` output of the "Switch" to the lower "Edit Fields (Set)" node. \newline
 5. Do not create, delete, move, or reconfigure any nodes. \newline
  \newline
 \# RESULT FORMAT \newline
 ```json \newline
 \{"answer": "done"\} \newline
 ```} \\
\cline{2-2}
\cellcolor{red!15} & \textbf{Ground truth} \\
\cline{2-2}
\cellcolor{red!15} & \includegraphics[width=\linewidth]{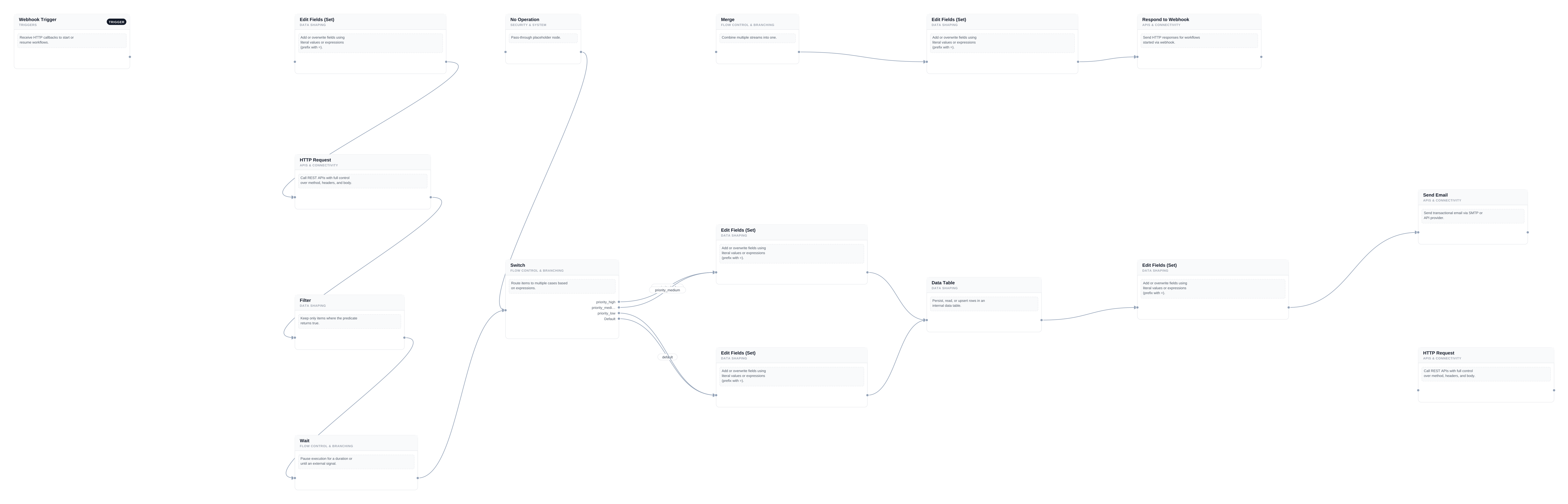} \\
\hline

\cellcolor{red!15}21 & {\raggedright\scriptsize\ttfamily \# Insert a Filter and Modify Sort in a Pipeline \newline
  \newline
 \#\# GOAL \newline
 Make multiple changes to an existing pipeline: insert a Filter, reconfigure a Sort, and delete a trailing Limit node. \newline
  \newline
 \#\# STEPS \newline
 1. Open the "API Data Pipeline" workflow. \newline
 2. Insert a "Filter" node between the "Map" and "Sort" nodes. \newline
 3. Configure the "Filter" to pass only items where `item.email.includes(".org")`. \newline
 4. Change the "Sort" node to use field `email` with direction `desc`. \newline
 5. Delete last "Limit" node. \newline
 6. The final path must be: `HTTP Request -> Map -> Filter -> Sort`. \newline
 7. Do not add any other nodes. \newline
  \newline
 \# RESULT FORMAT \newline
 ```json \newline
 \{"answer": "done"\} \newline
 ```} \\
\cline{2-2}
\cellcolor{red!15} & \textbf{Ground truth} \\
\cline{2-2}
\cellcolor{red!15} & \includegraphics[width=\linewidth]{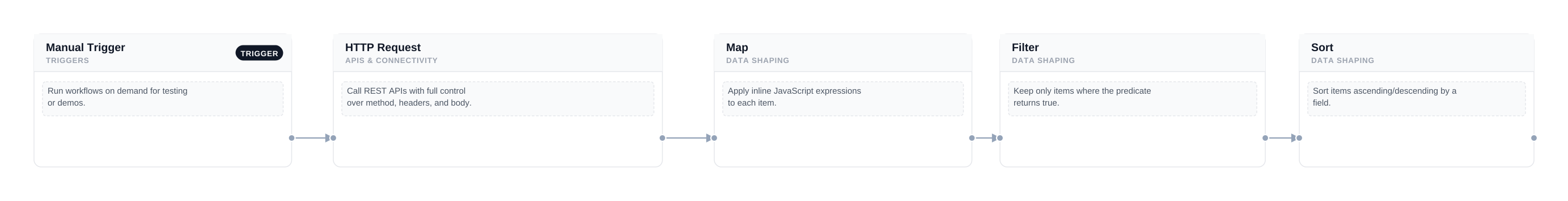} \\
\hline

\cellcolor{red!15}22 & {\raggedright\scriptsize\ttfamily \# Add a New Priority Branch to a Switch Node \newline
  \newline
 \#\# GOAL \newline
 Extend a Switch node with a new priority branch, add a dedicated processing node for it, and wire it into the main pipeline. \newline
  \newline
 \#\# STEPS \newline
 1. Open the "Enterprise Data Pipeline" workflow. \newline
 2. Find the "Switch" node that routes items by priority. \newline
 3. Add a new output branch labeled `priority\_urgent`, do not include "expr" field. \newline
 4. Add a new "Edit Fields (Set)" node for the urgent branch. \newline
 5. Connect the `priority\_urgent` output to the new "Edit Fields (Set)" node. \newline
 6. Connect the new "Edit Fields (Set)" node back into the "Data Table" node. \newline
  \newline
 \# RESULT FORMAT \newline
 ```json \newline
 \{"answer": "done"\} \newline
 ```} \\
\cline{2-2}
\cellcolor{red!15} & \textbf{Ground truth} \\
\cline{2-2}
\cellcolor{red!15} & \includegraphics[width=\linewidth]{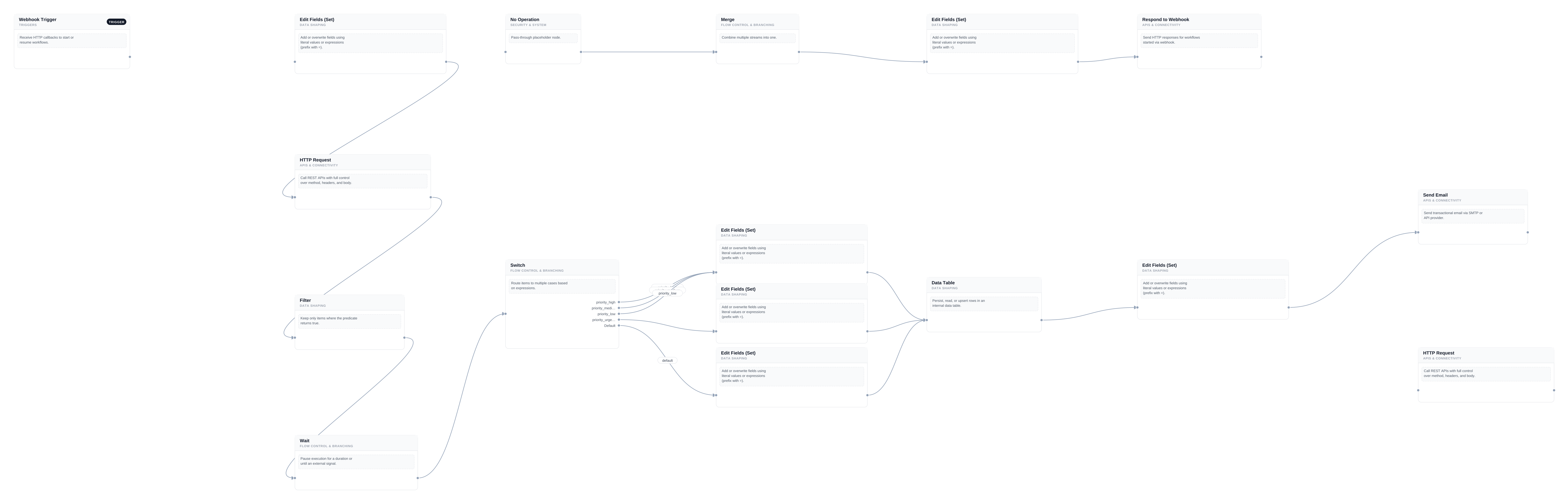} \\
\hline

\cellcolor{red!15}23 & {\raggedright\scriptsize\ttfamily \# Create a Simple Loop Workflow \newline
 \newline
\#\# GOAL \newline
Create a new workflow where data passes through a loop before reaching the final output. \newline
 \newline
\#\# STEPS \newline
1. Create a new workflow named "Simple Loop Pipeline". \newline
2. Add exactly these nodes: \newline
   - "Manual Trigger" \newline
   - "Map" \newline
   - "If" \newline
   - "Data Table" \newline
3. Configure the "Map" node to add or update the field: \newline
   `counter = (item.counter || 0) + 1` \newline
4. Configure the "If" node with the condition: \newline
   `item.counter < 3` \newline
5. Connect the workflow as follows: \newline
   - `Manual Trigger -> Map -> If` \newline
   - The `true` output of "If" loops back to the "Map" node. \newline
   - The `false` output of "If" goes to "Data Table". \newline
 \newline
\# RESULT FORMAT \newline
 \newline
```json \newline
\{"answer": "done"\} \newline
```} \\
\cline{2-2}
\cellcolor{red!15} & \textbf{Ground truth} \\
\cline{2-2}
\cellcolor{red!15} & \includegraphics[width=\linewidth]{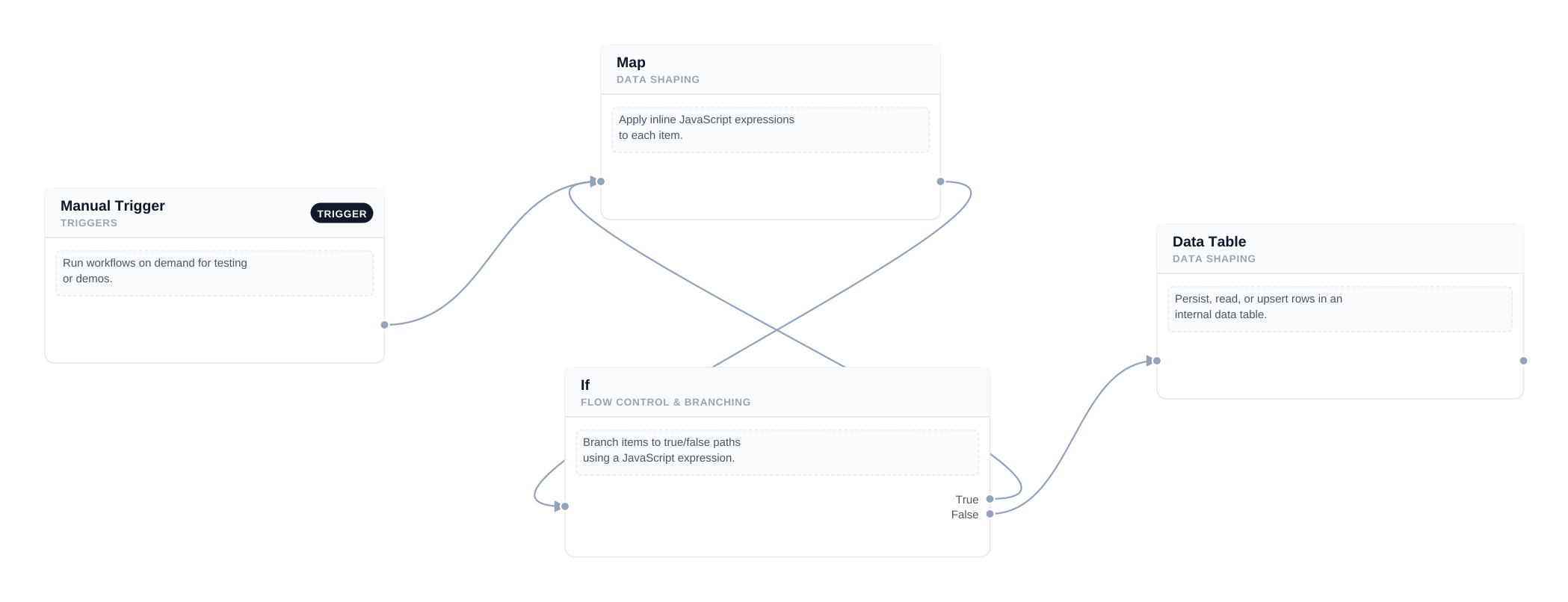} \\
\hline

\cellcolor{red!15}24 & {\raggedright\scriptsize\ttfamily \# Create a Branching Workflow with Merge \newline
  \newline
 \#\# GOAL \newline
 Build a branching workflow with an If condition that routes data to two different nodes, then merges the branches back together. \newline
  \newline
 \#\# STEPS \newline
 1. Create a new workflow named "ICML Project". \newline
 2. Add these nodes and connect them: `Manual Trigger -> Database Connector -> If`. \newline
 3. Add two more nodes: "Crypto" and "Guardrails". \newline
 4. From the "If" node: \newline
  - Connect the "true" output to a "Crypto" node. \newline
  - Connect the "false" output to a "Guardrails" node. \newline
 5. Connect both branches into a "Merge" node: `Crypto -> Merge` and `Guardrails -> Merge`. \newline
 6. Set the "If" condition to check: `item.country === "UK"`. \newline
 7. Do not add any extra nodes. \newline
  \newline
 \# RESULT FORMAT \newline
 ```json \newline
 \{"answer": "done"\} \newline
 ```} \\
\cline{2-2}
\cellcolor{red!15} & \textbf{Ground truth} \\
\cline{2-2}
\cellcolor{red!15} & \includegraphics[width=\linewidth]{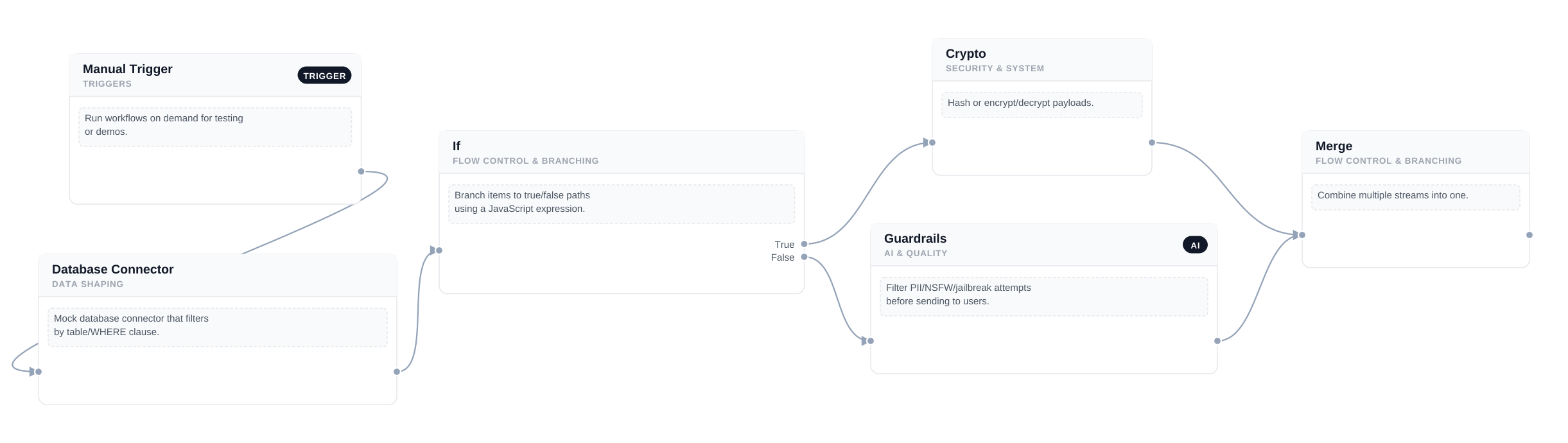} \\
\hline

\cellcolor{red!15}25 & {\raggedright\scriptsize\ttfamily \# Create a Simple Decision Tree Workflow \newline
 \newline
\#\# GOAL \newline
Create a small tree-shaped workflow with 3 If nodes and 4 final outputs. \newline
 \newline
\#\# STEPS \newline
1. Create a new workflow named "Simple Decision Tree Pipeline". \newline
2. Add: \newline
   - 1 Manual Trigger \newline
   - 3 If nodes \newline
   - 2 Data Table nodes \newline
3. Configure the If nodes: \newline
   - Root If: `item.score >= 50` \newline
   - Upper-branch If: `item.score >= 80` \newline
   - Lower-branch If: `item.score >= 20` \newline
4. Connect: \newline
   - `Manual Trigger -> Root If` \newline
   - Root If `true` -> Upper-branch If \newline
   - Root If `false` -> Lower-branch If \newline
   - Upper-branch If `true` -> first Data Table \newline
   - Upper-branch If `false` -> second Data Table \newline
   - Lower-branch If `true` -> first Data Table \newline
   - Lower-branch If `false` -> second Data Table \newline
 \newline
\# RESULT FORMAT \newline
 \newline
```json \newline
\{"answer": "done"\} \newline
```} \\
\cline{2-2}
\cellcolor{red!15} & \textbf{Ground truth} \\
\cline{2-2}
\cellcolor{red!15} & \includegraphics[width=\linewidth]{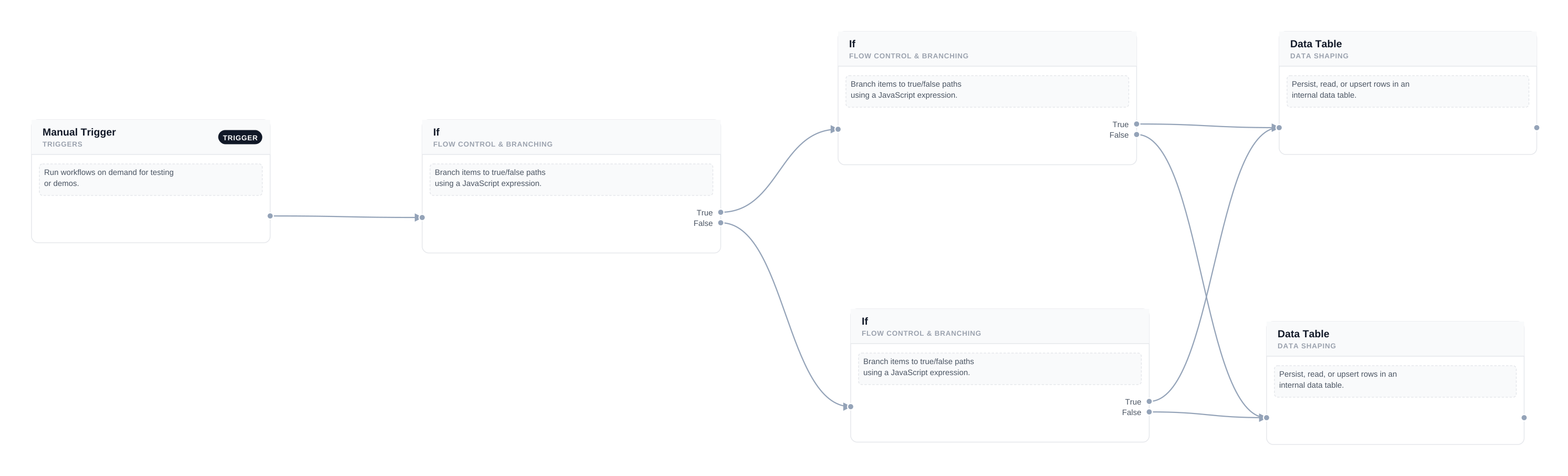} \\
\hline

\cellcolor{red!15}26 & {\raggedright\scriptsize\ttfamily \# Create a Conditional Email Workflow \newline
  \newline
 \#\# GOAL \newline
 Create a conditional email workflow where the If node routes to two separate Send Email nodes with different subjects. \newline
  \newline
 \#\# STEPS \newline
 1. Create a new workflow named "ICML Project". \newline
 2. Add these nodes and connect them in sequence: `Chat Trigger -> Database Connector -> If`. \newline
 3. Add two "Send Email" nodes. \newline
 4. Connect first "Send Email" node only to the "false" output of the "If" node. \newline
 5. Connect second "Send Email" node only to the "true" output of the "If" node. \newline
 6. Set the "If" condition to check: `item.country === "FR"`. \newline
 7. Set the first "Send Email" subject to: `ICML Project notification`. \newline
 8. Set the second "Send Email" subject to: `Notification de projet ICML`. \newline
 9. Do not add extra nodes. \newline
  \newline
 \# RESULT FORMAT \newline
 ```json \newline
 \{"answer": "done"\} \newline
 ```} \\
\cline{2-2}
\cellcolor{red!15} & \textbf{Ground truth} \\
\cline{2-2}
\cellcolor{red!15} & \includegraphics[width=\linewidth]{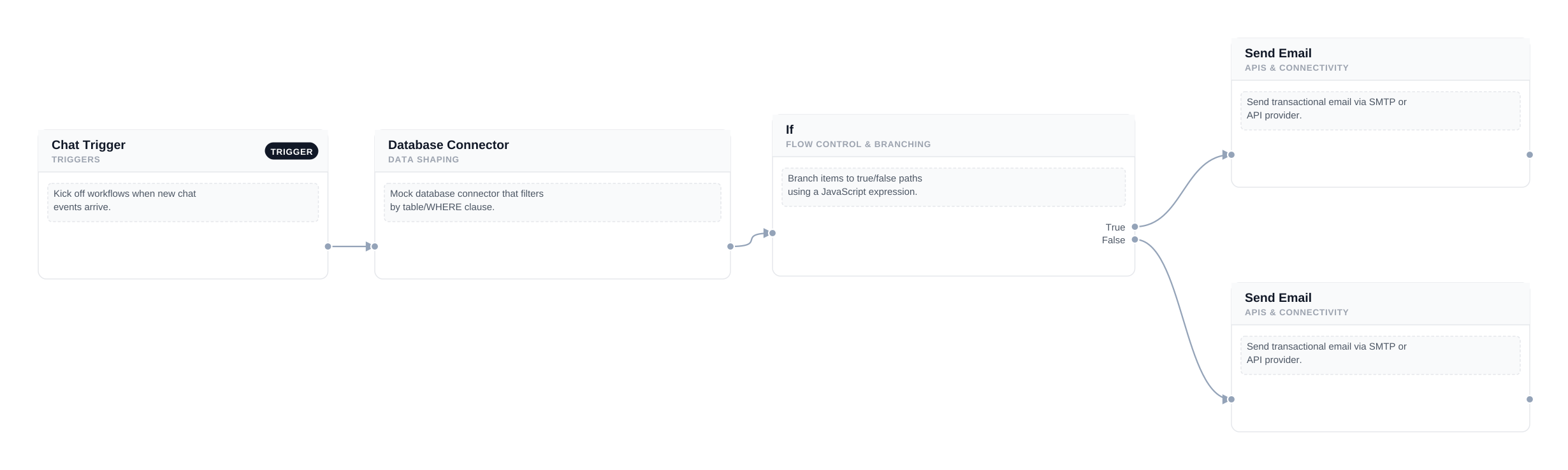} \\
\hline

\cellcolor{red!15}27 & {\raggedright\scriptsize\ttfamily \# Create a Multi-Node Workflow with Configuration \newline
  \newline
 \#\# GOAL \newline
 Create a five-node workflow from scratch with HTTP, mapping, filtering, and email nodes, each configured with specific settings. \newline
  \newline
 \#\# STEPS \newline
 1. Create a new workflow named "Newsletter System". \newline
 2. Add exactly these 5 nodes and wire them it the following in sequence: `Manual Trigger -> HTTP Request -> Map -> Filter -> Send Email`. \newline
 3. Configure the nodes: \newline
  - "HTTP Request" method: "GET" \newline
  - "HTTP Request" URL: `https://jsonplaceholder.typicode.com/users` \newline
  - "Filter" condition: pass only items where `item.address?.city === "Gwenborough"` \newline
  - "Send Email" From: `newsletter@example.com` \newline
  - "Send Email" Recipients: `local-subscribers@example.com` \newline
  - "Send Email" Subject: `Local Newsletter` \newline
  \newline
 \# RESULT FORMAT \newline
 ```json \newline
 \{"answer": "done"\} \newline
 ```} \\
\cline{2-2}
\cellcolor{red!15} & \textbf{Ground truth} \\
\cline{2-2}
\cellcolor{red!15} & \includegraphics[width=\linewidth]{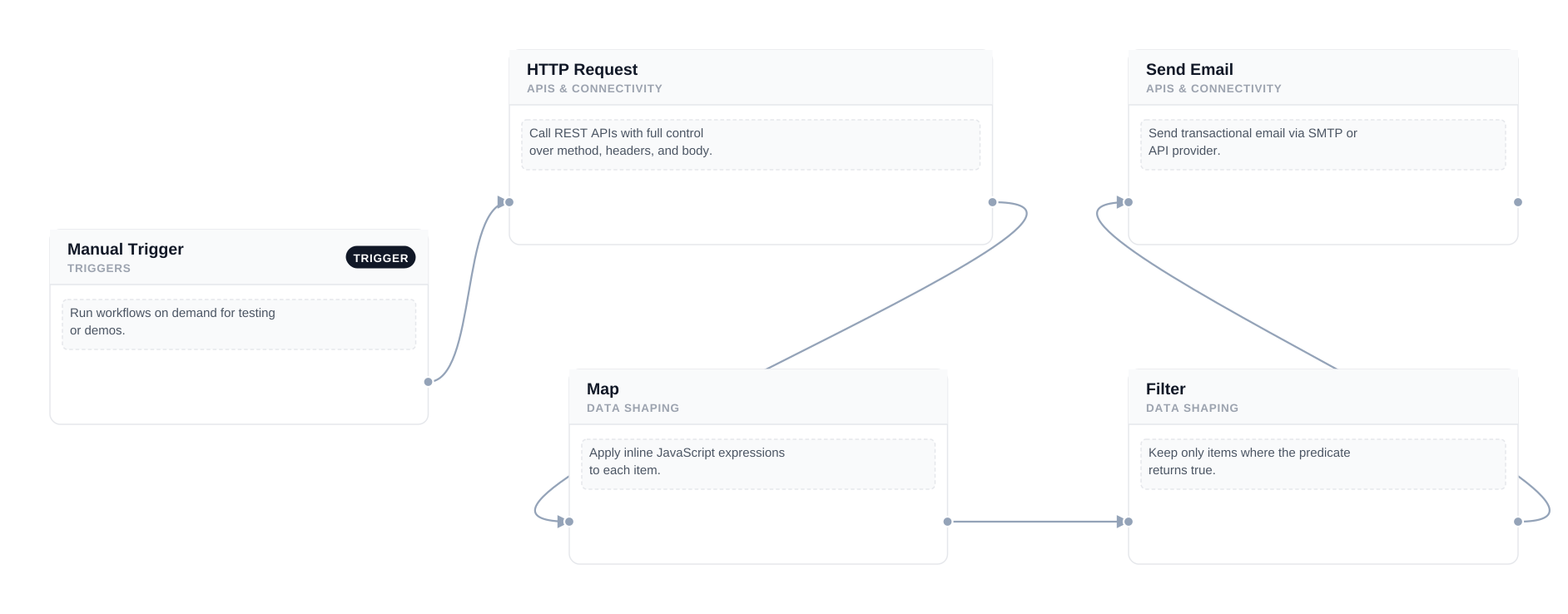} \\
\hline

\end{longtable}
\normalsize

%% file: applications_testcases/testcases_3d_editor.tex
\scriptsize
\renewcommand{\arraystretch}{1.18}
\setlength{\LTleft}{0pt}
\setlength{\LTright}{0pt}
\setlength{\tabcolsep}{6pt}
\begin{longtable}{|>{\raggedright\vspace{0pt}}p{0.05\linewidth}|>{\raggedright\arraybackslash\vspace{0pt}}p{0.87\linewidth}|}
\hline
\textbf{No.} & \textbf{Task} \\
\hline
\endfirsthead
\hline
\textbf{No.} & \textbf{Task} \\
\hline
\endhead
\hline
\endfoot

\cellcolor{green!15}1 & {\raggedright\scriptsize\ttfamily \# Double the Cube's Scale \newline
  \newline
 \#\# GOAL \newline
 Open a single-object scene, double the cube's scale on all axes, and change the scene background color. \newline
  \newline
 \#\# STEPS \newline
 1. Open the "Single Cube Scene". \newline
 2. Double its current scale along all three axes, and set the scene's background color to white (\#ffffff). \newline
  \newline
 \# RESULT FORMAT \newline
  \newline
 ```json \newline
 \{"answer": "done"\} \newline
 ```} \\
\cline{2-2}
\cellcolor{green!15} & \textbf{Initial state} \\
\cline{2-2}
\cellcolor{green!15} & \includegraphics[width=\linewidth]{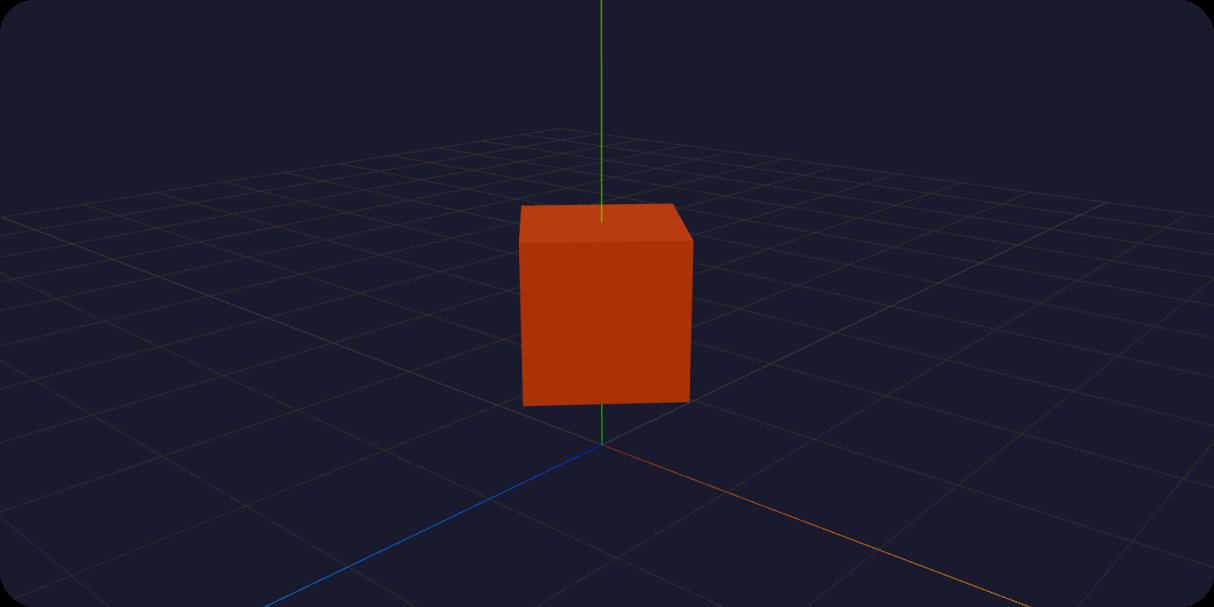} \\
\cline{2-2}
\cellcolor{green!15} & \textbf{Ground truth} \\
\cline{2-2}
\cellcolor{green!15} & \includegraphics[width=\linewidth]{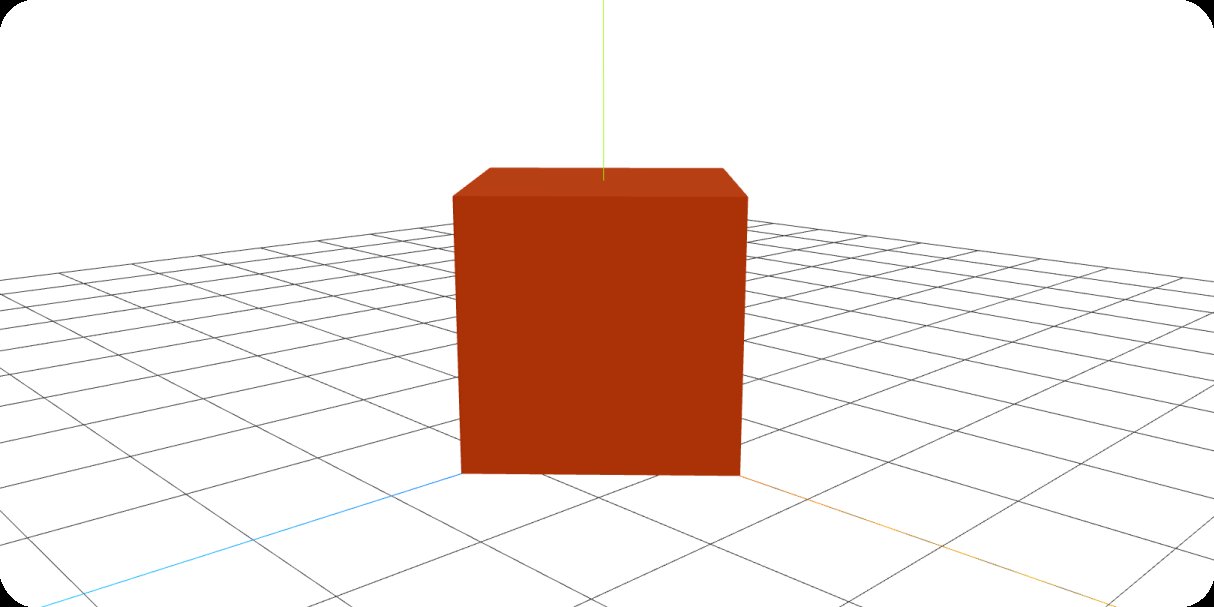} \\
\hline

\cellcolor{green!15}2 & {\raggedright\scriptsize\ttfamily \# Position Elements Relative to Reference \newline
  \newline
 \#\# GOAL \newline
 Open a scene and reposition two elements at a fixed distance from a reference object along the X-axis, preserving their orientations. \newline
  \newline
 \#\# STEPS \newline
 1. Open the "Basic Shape Scene". \newline
 2. Use the blue sphere as the reference point. Position the other two elements at a distance of 3 units from the reference along the X-axis, while preserving their original orientations. \newline
  \newline
 \# RESULT FORMAT \newline
  \newline
 ```json \newline
 \{"answer": "done"\} \newline
 ```} \\
\cline{2-2}
\cellcolor{green!15} & \textbf{Initial state} \\
\cline{2-2}
\cellcolor{green!15} & \includegraphics[width=\linewidth]{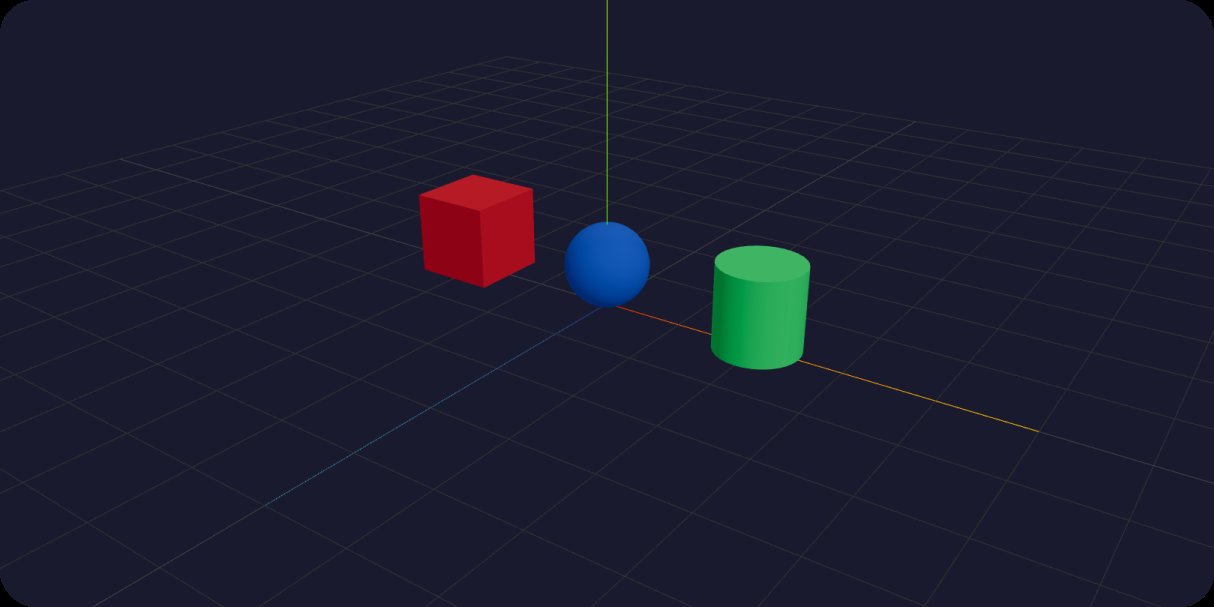} \\
\cline{2-2}
\cellcolor{green!15} & \textbf{Ground truth} \\
\cline{2-2}
\cellcolor{green!15} & \includegraphics[width=\linewidth]{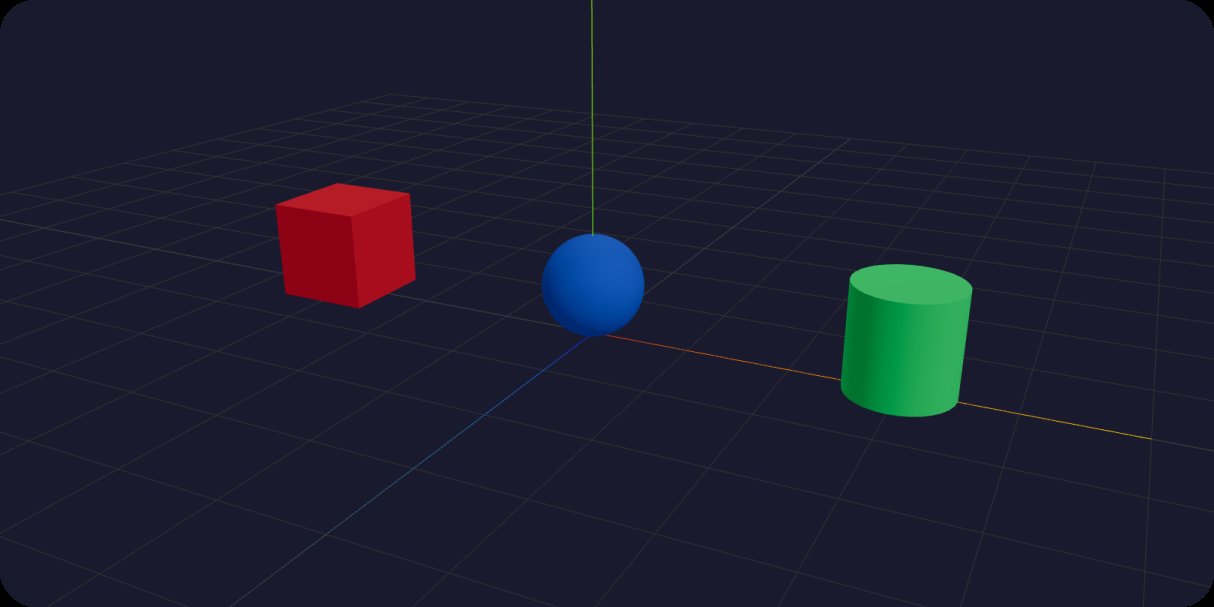} \\
\hline

\cellcolor{green!15}3 & {\raggedright\scriptsize\ttfamily \# Land the Floating Cone on the Ground \newline
  \newline
 \#\# GOAL \newline
 Move one object down so that it rests exactly on the ground plane, deriving the required height from its size. \newline
  \newline
 \#\# STEPS \newline
 1. Open the "Geometric Garden". \newline
 2. Select the cone named "Tree Top". It currently floats above the ground plane (the y = 0 plane). \newline
 3. Change only its Y position so that its base touches the ground plane exactly, with no gap and no sinking below it. Keep its X and Z positions unchanged. \newline
  \newline
 \# RESULT FORMAT \newline
  \newline
 ```json \newline
 \{"answer": "done"\} \newline
 ```} \\
\cline{2-2}
\cellcolor{green!15} & \textbf{Initial state} \\
\cline{2-2}
\cellcolor{green!15} & \includegraphics[width=\linewidth]{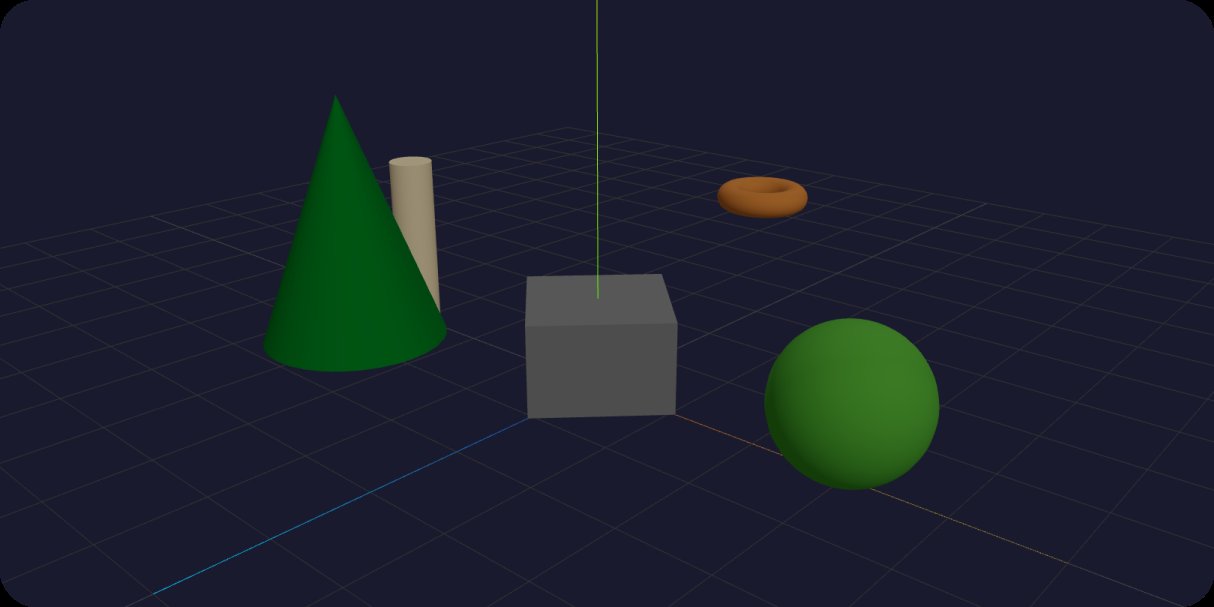} \\
\cline{2-2}
\cellcolor{green!15} & \textbf{Ground truth} \\
\cline{2-2}
\cellcolor{green!15} & \includegraphics[width=\linewidth]{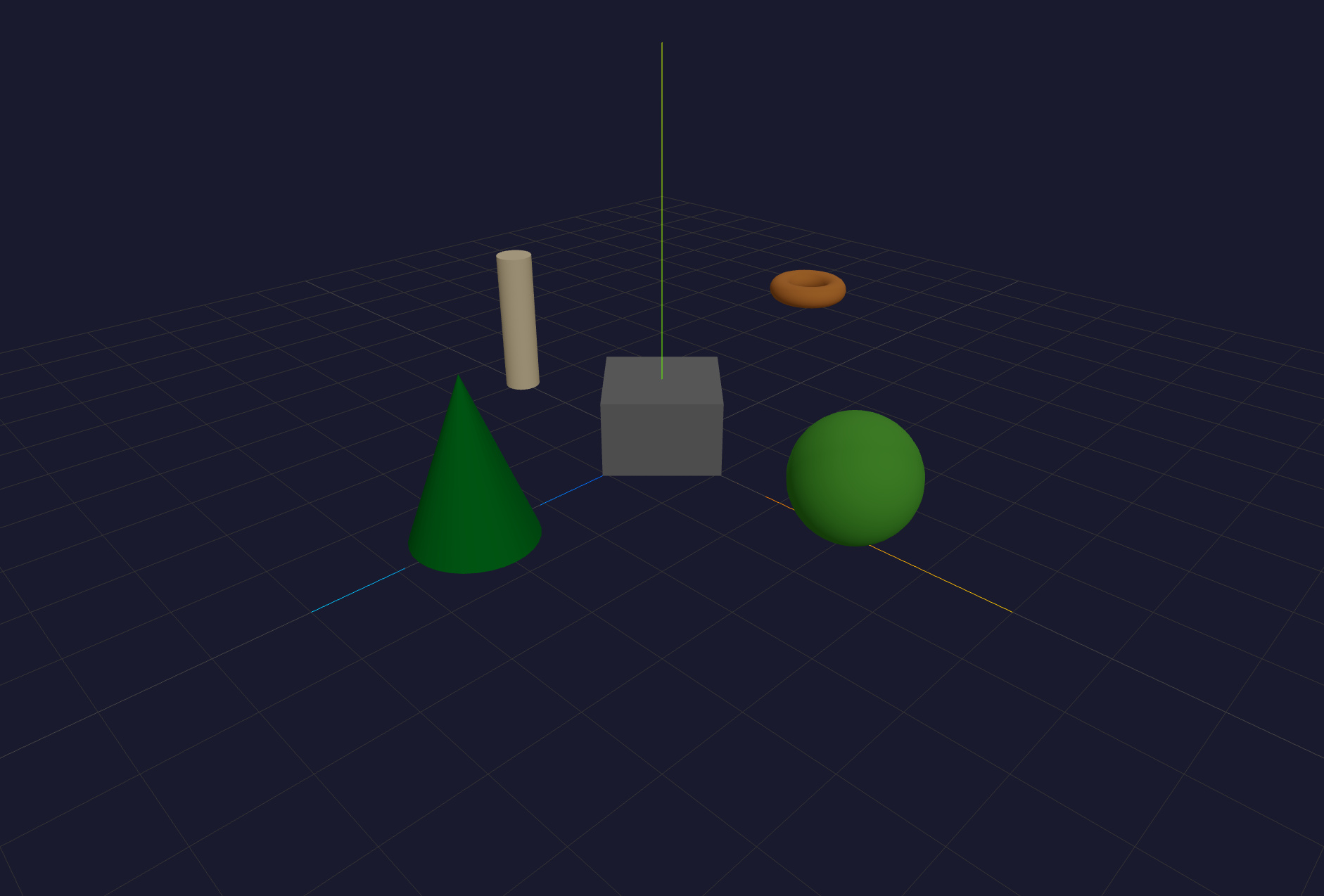} \\
\hline

\cellcolor{green!15}4 & {\raggedright\scriptsize\ttfamily \# Put the Cone Directly Above the Sphere \newline
  \newline
 \#\# GOAL \newline
 Reposition one object so it is vertically aligned above another object, reading the reference position from the scene. \newline
  \newline
 \#\# STEPS \newline
 1. Open the "Four Objects Scene". \newline
 2. Read the position of "Sphere 1". \newline
 3. Move "Cone 1" so that its center is directly above the center of "Sphere 1" (same X and same Z), and 3 units higher along the Y axis. \newline
  \newline
 \# RESULT FORMAT \newline
  \newline
 ```json \newline
 \{"answer": "done"\} \newline
 ```} \\
\cline{2-2}
\cellcolor{green!15} & \textbf{Initial state} \\
\cline{2-2}
\cellcolor{green!15} & \includegraphics[width=\linewidth]{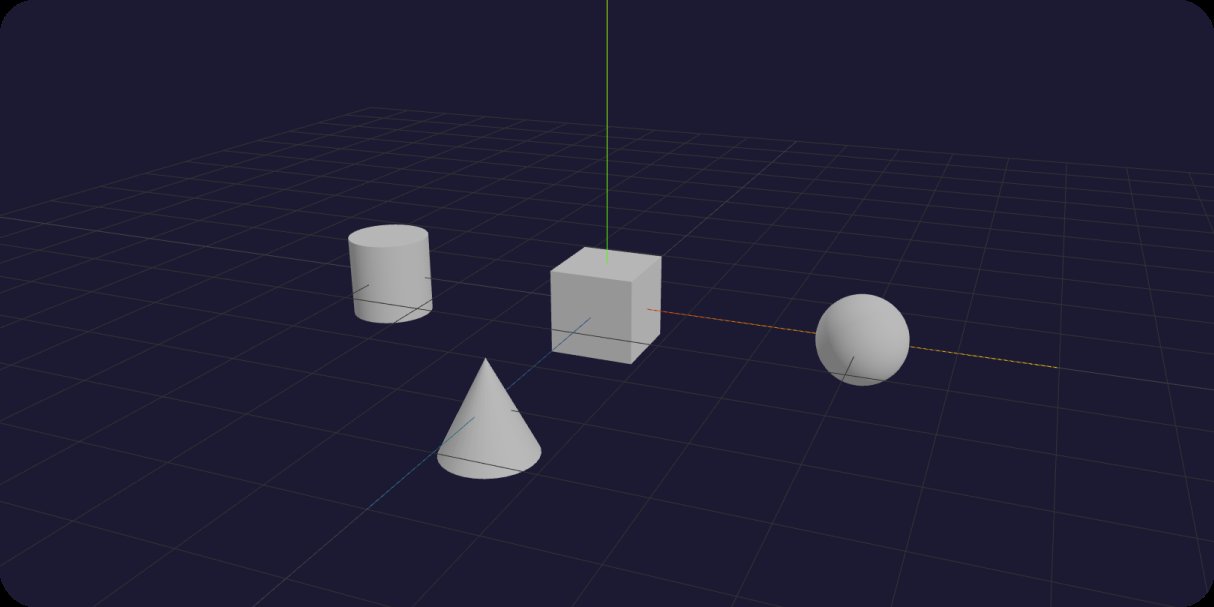} \\
\cline{2-2}
\cellcolor{green!15} & \textbf{Ground truth} \\
\cline{2-2}
\cellcolor{green!15} & \includegraphics[width=\linewidth]{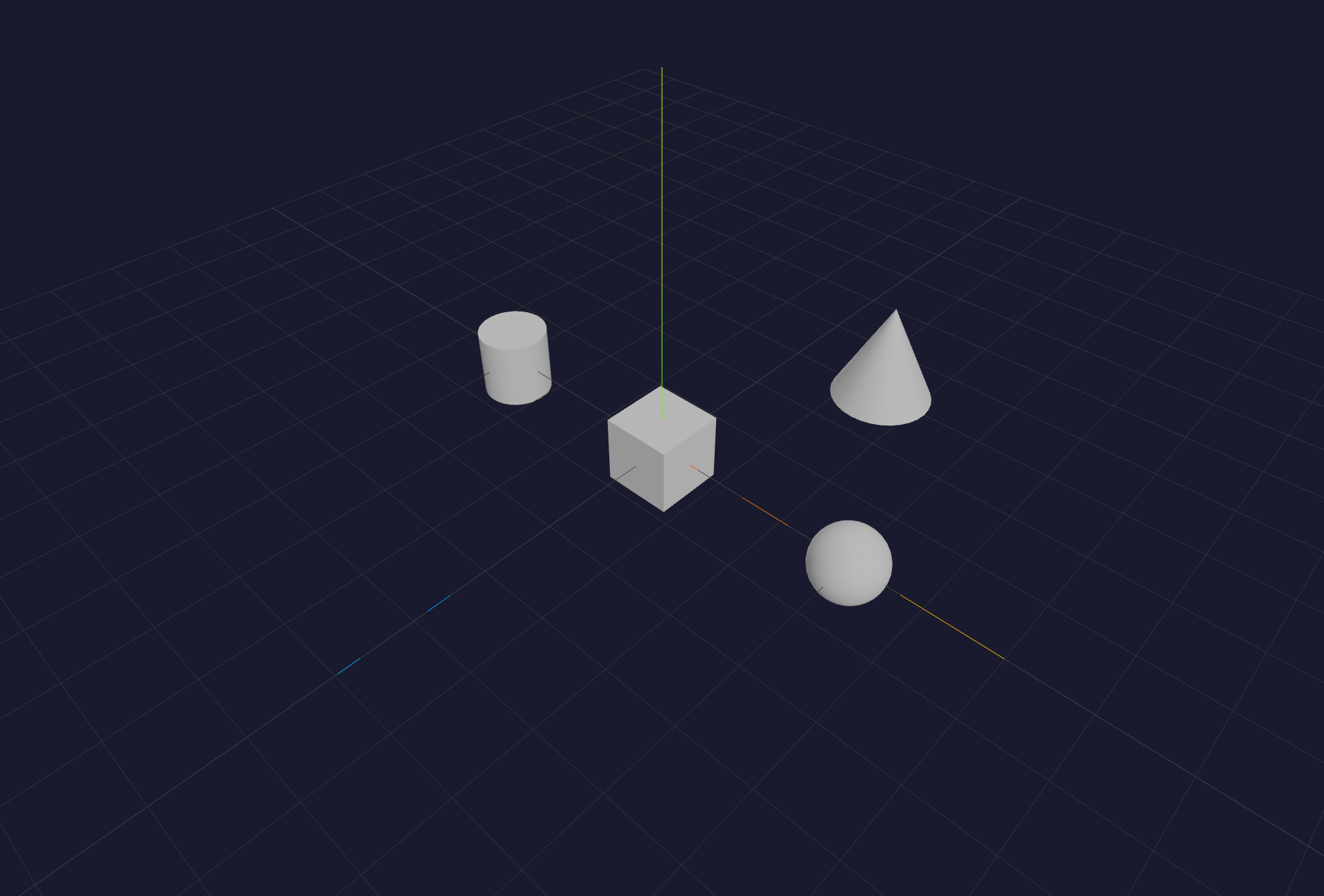} \\
\hline

\cellcolor{green!15}5 & {\raggedright\scriptsize\ttfamily \# Open a One-Unit Gap Between Two Surfaces \newline
  \newline
 \#\# GOAL \newline
 Move one object along a single axis so that the gap between two surfaces, not between two centers, has a given size. \newline
  \newline
 \#\# STEPS \newline
 1. Open the "Row of Objects Scene". \newline
 2. Move "Sphere 1" along the X axis to the negative side of "Cube 1" so that the gap between the sphere's surface and the cube's nearest face is exactly 1 unit. Keep its Y and Z positions unchanged. \newline
  \newline
 \# RESULT FORMAT \newline
  \newline
 ```json \newline
 \{"answer": "done"\} \newline
 ```} \\
\cline{2-2}
\cellcolor{green!15} & \textbf{Initial state} \\
\cline{2-2}
\cellcolor{green!15} & \includegraphics[width=\linewidth]{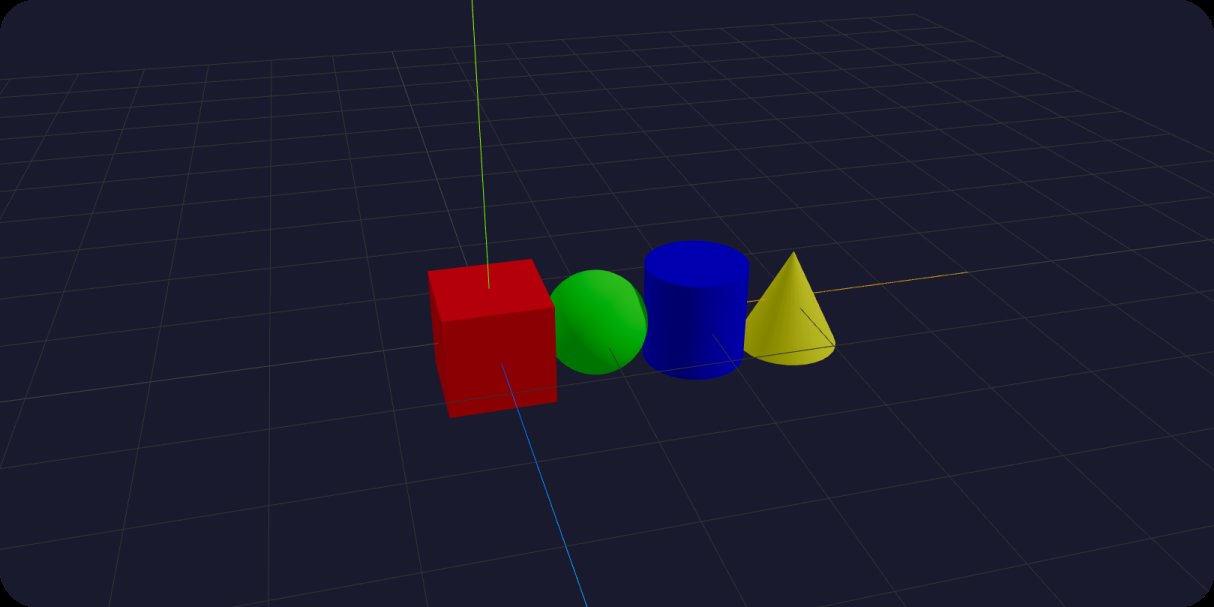} \\
\cline{2-2}
\cellcolor{green!15} & \textbf{Ground truth} \\
\cline{2-2}
\cellcolor{green!15} & \includegraphics[width=\linewidth]{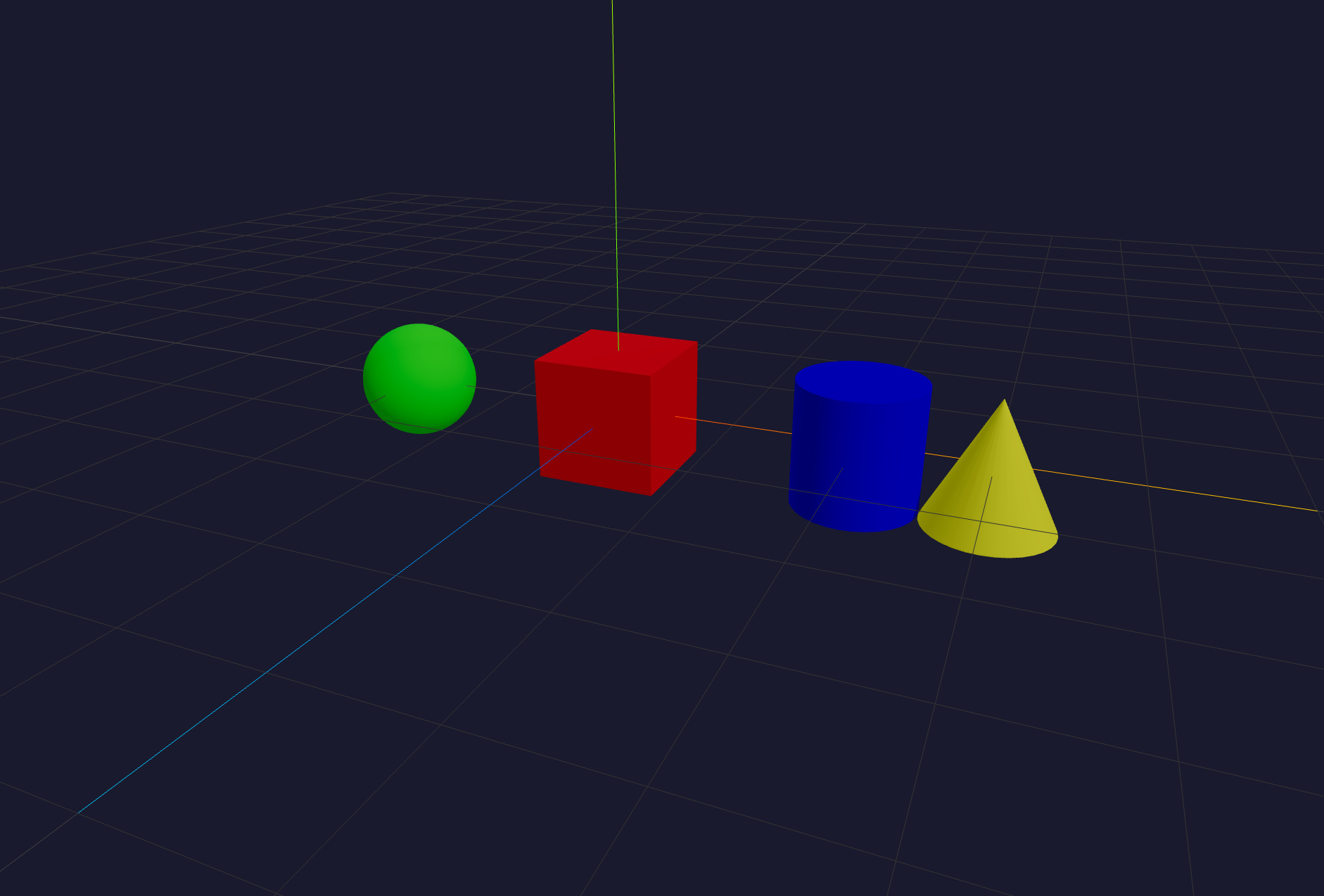} \\
\hline

\cellcolor{green!15}6 & {\raggedright\scriptsize\ttfamily \# Add a Cube at the Midpoint of Two Objects \newline
  \newline
 \#\# GOAL \newline
 Add one new object exactly halfway between two existing objects. \newline
  \newline
 \#\# STEPS \newline
 1. Open the "Four Objects Scene". \newline
 2. Read the positions of "Cylinder 1" and "Cone 1". \newline
 3. Add a cube, name it "Midpoint", and set its position to the point exactly halfway between those two positions. Keep its default rotation and unit scale. \newline
 4. Set the cube's color to cyan (\#00ffff). \newline
  \newline
 \# RESULT FORMAT \newline
  \newline
 ```json \newline
 \{"answer": "done"\} \newline
 ```} \\
\cline{2-2}
\cellcolor{green!15} & \textbf{Initial state} \\
\cline{2-2}
\cellcolor{green!15} & \includegraphics[width=\linewidth]{images/3d_editor_initial_jpegs/tc_3d_002.jpg} \\
\cline{2-2}
\cellcolor{green!15} & \textbf{Ground truth} \\
\cline{2-2}
\cellcolor{green!15} & \includegraphics[width=\linewidth]{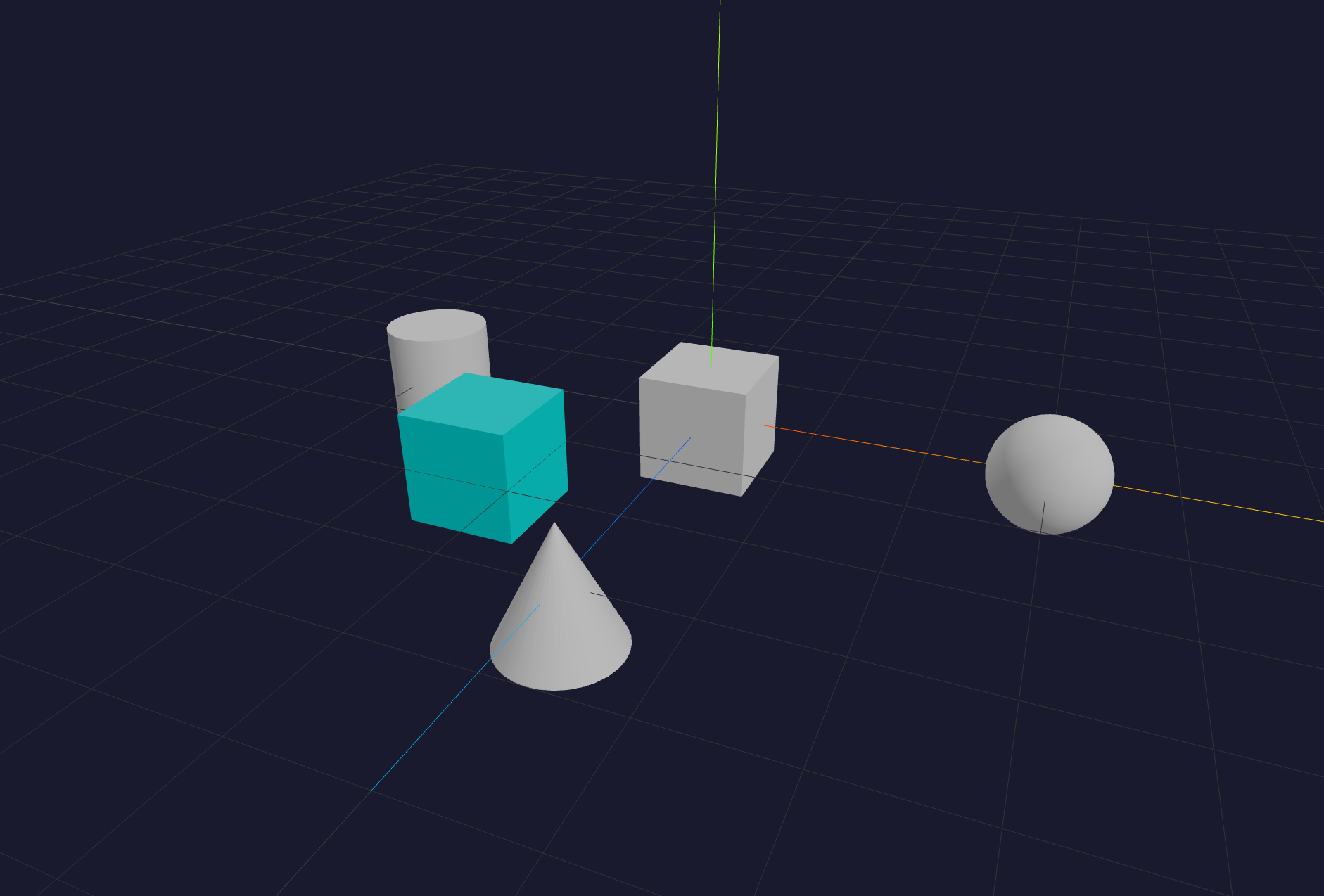} \\
\hline

\cellcolor{green!15}7 & {\raggedright\scriptsize\ttfamily \# Make the Cube Taller Without Moving Its Base \newline
  \newline
 \#\# GOAL \newline
 Change one object's height while keeping its bottom face at the same level, which requires adjusting position together with scale. \newline
  \newline
 \#\# STEPS \newline
 1. Open the "Single Cube Scene". \newline
 2. Note the cube's current position and scale, and work out where its bottom face currently is. \newline
 3. Change its Y scale so the cube becomes 3 units tall, and change its Y position so its bottom face stays at exactly the same height as before. \newline
  \newline
 \# RESULT FORMAT \newline
  \newline
 ```json \newline
 \{"answer": "done"\} \newline
 ```} \\
\cline{2-2}
\cellcolor{green!15} & \textbf{Initial state} \\
\cline{2-2}
\cellcolor{green!15} & \includegraphics[width=\linewidth]{images/3d_editor_initial_jpegs/tc_3d_005.jpg} \\
\cline{2-2}
\cellcolor{green!15} & \textbf{Ground truth} \\
\cline{2-2}
\cellcolor{green!15} & \includegraphics[width=\linewidth]{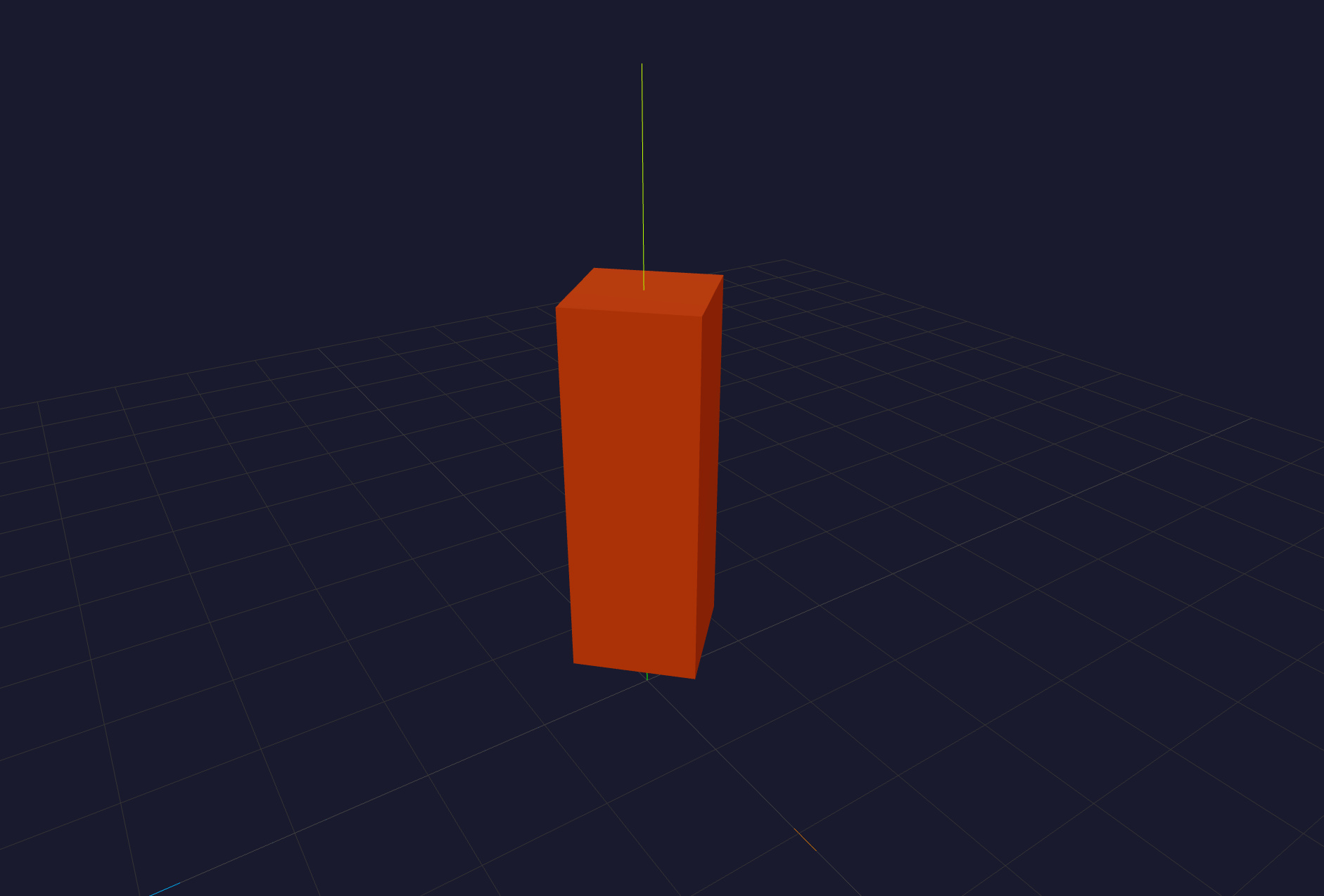} \\
\hline

\cellcolor{green!15}8 & {\raggedright\scriptsize\ttfamily \# Delete the Object on the Negative Z Axis \newline
  \newline
 \#\# GOAL \newline
 Identify the one object that sits on a given axis direction and delete it. \newline
  \newline
 \#\# STEPS \newline
 1. Open the "Geometric Garden". \newline
 2. Inspect the objects' positions and identify the single object whose center lies on the negative Z axis. \newline
 3. Delete that object, and only that object. \newline
  \newline
 \# RESULT FORMAT \newline
  \newline
 ```json \newline
 \{"answer": "done"\} \newline
 ```} \\
\cline{2-2}
\cellcolor{green!15} & \textbf{Initial state} \\
\cline{2-2}
\cellcolor{green!15} & \includegraphics[width=\linewidth]{images/3d_editor_initial_jpegs/tc_3d_003.jpg} \\
\cline{2-2}
\cellcolor{green!15} & \textbf{Ground truth} \\
\cline{2-2}
\cellcolor{green!15} & \includegraphics[width=\linewidth]{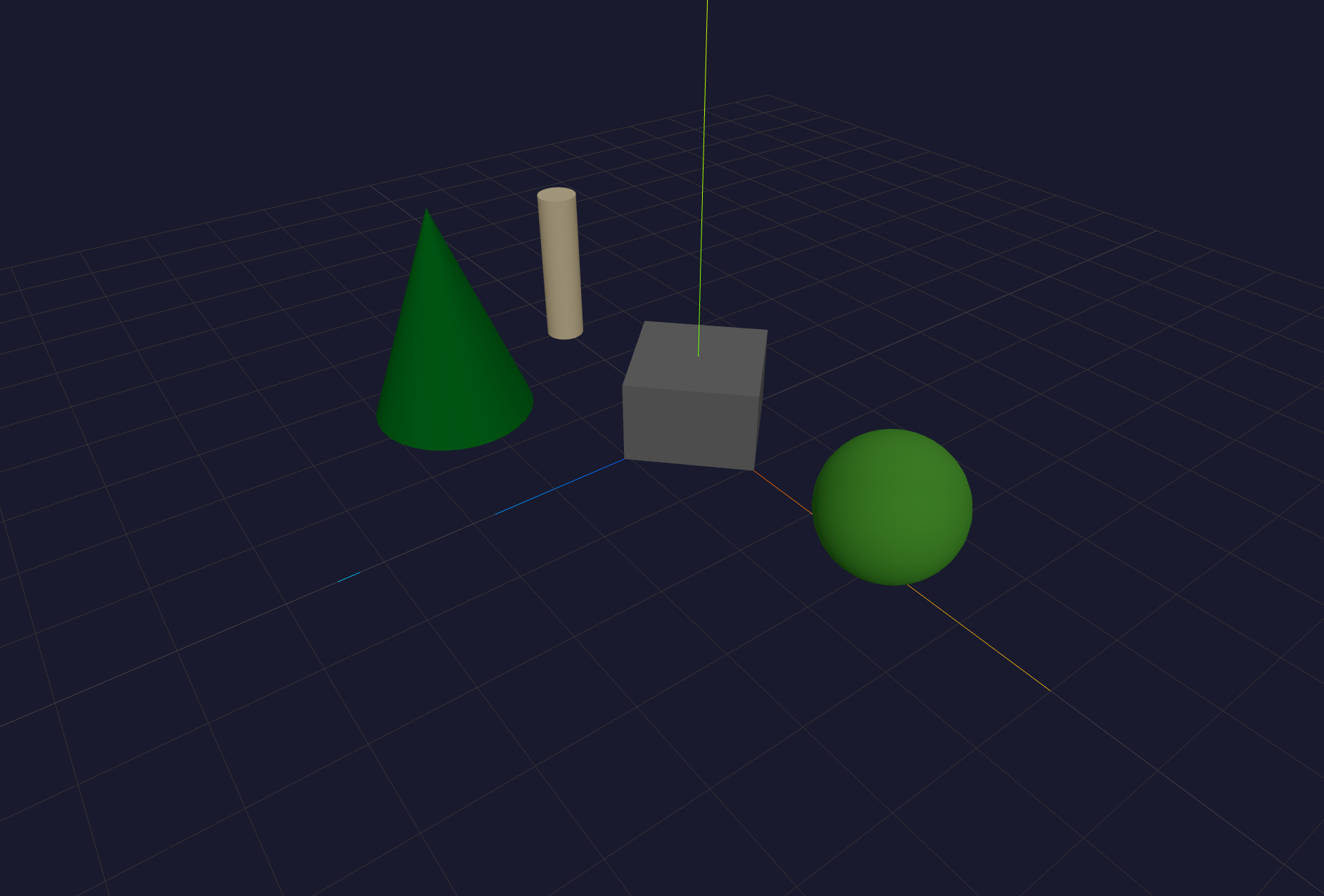} \\
\hline

\cellcolor{green!15}9 & {\raggedright\scriptsize\ttfamily \# Stack a Copy on Top of the Cube \newline
  \newline
 \#\# GOAL \newline
 Duplicate an object and place the copy so it rests on top of the original, touching it without overlap. \newline
  \newline
 \#\# STEPS \newline
 1. Open the "Two Objects Scene". \newline
 2. Duplicate the cube. \newline
 3. Move the duplicate so that its bottom face rests exactly on the top face of the original cube: same X and Z as the original, touching it with no gap and no overlap. \newline
  \newline
 \# RESULT FORMAT \newline
  \newline
 ```json \newline
 \{"answer": "done"\} \newline
 ```} \\
\cline{2-2}
\cellcolor{green!15} & \textbf{Initial state} \\
\cline{2-2}
\cellcolor{green!15} & \includegraphics[width=\linewidth]{images/3d_editor_initial_jpegs/tc_3d_002.jpg} \\
\cline{2-2}
\cellcolor{green!15} & \textbf{Ground truth} \\
\cline{2-2}
\cellcolor{green!15} & \includegraphics[width=\linewidth]{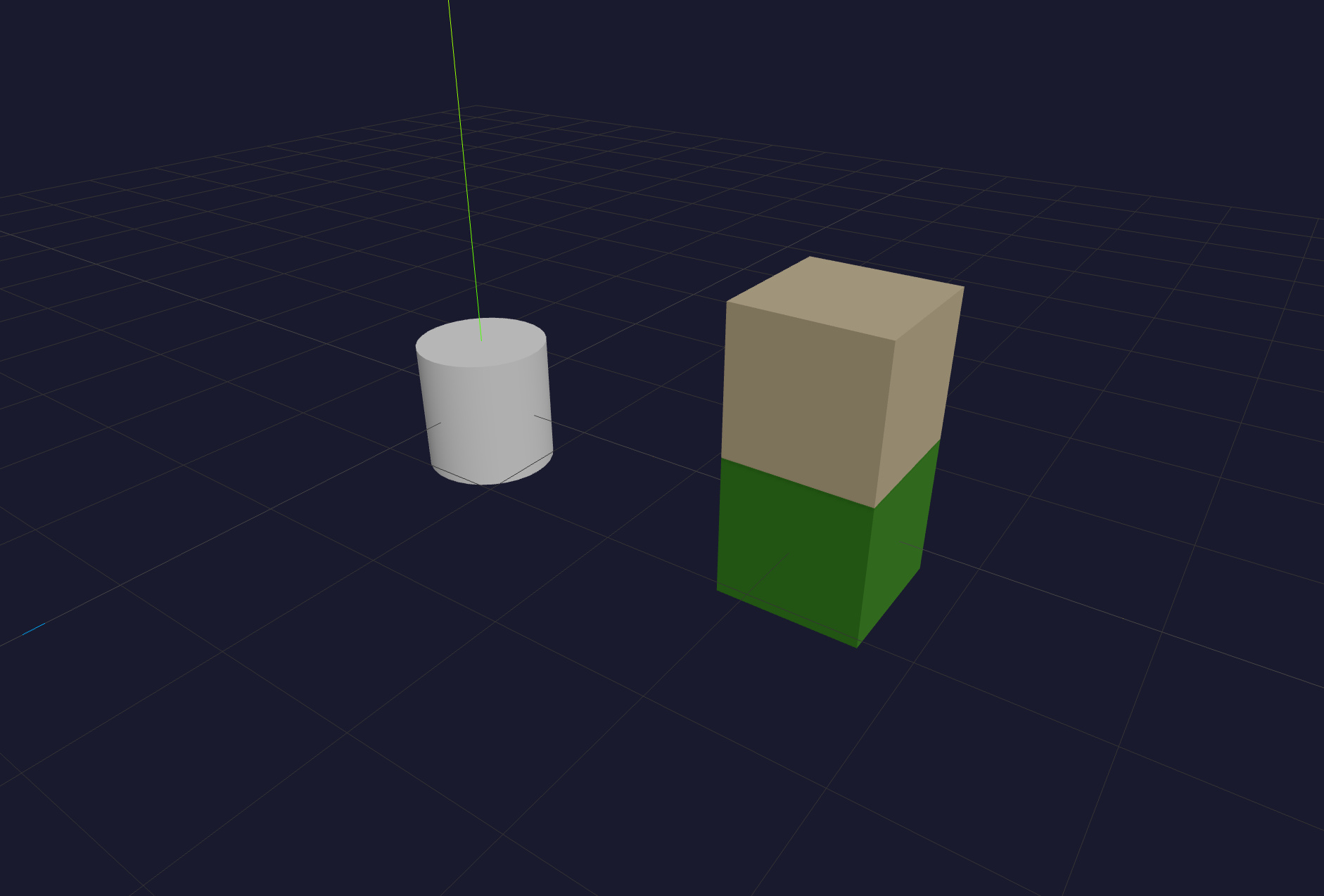} \\
\hline

\cellcolor{orange!15}10 & {\raggedright\scriptsize\ttfamily \# Place Cubes in All Octants \newline
  \newline
 \#\# GOAL \newline
 Create a new scene and place one cube in each of the eight 3D octants at symmetric coordinates. \newline
  \newline
 \#\# STEPS \newline
 1. Create a new scene named "Octant Cubes". \newline
 2. Generate cubes by assigning one default cube to each octant in 3D space, positioned at coordinates with absolute values of (1, 1, 1). Name all the cubes as Cube. \newline
  \newline
 \# RESULT FORMAT \newline
  \newline
 ```json \newline
 \{"answer": "done"\} \newline
 ```} \\
\cline{2-2}
\cellcolor{orange!15} & \textbf{Ground truth} \\
\cline{2-2}
\cellcolor{orange!15} & \includegraphics[width=\linewidth]{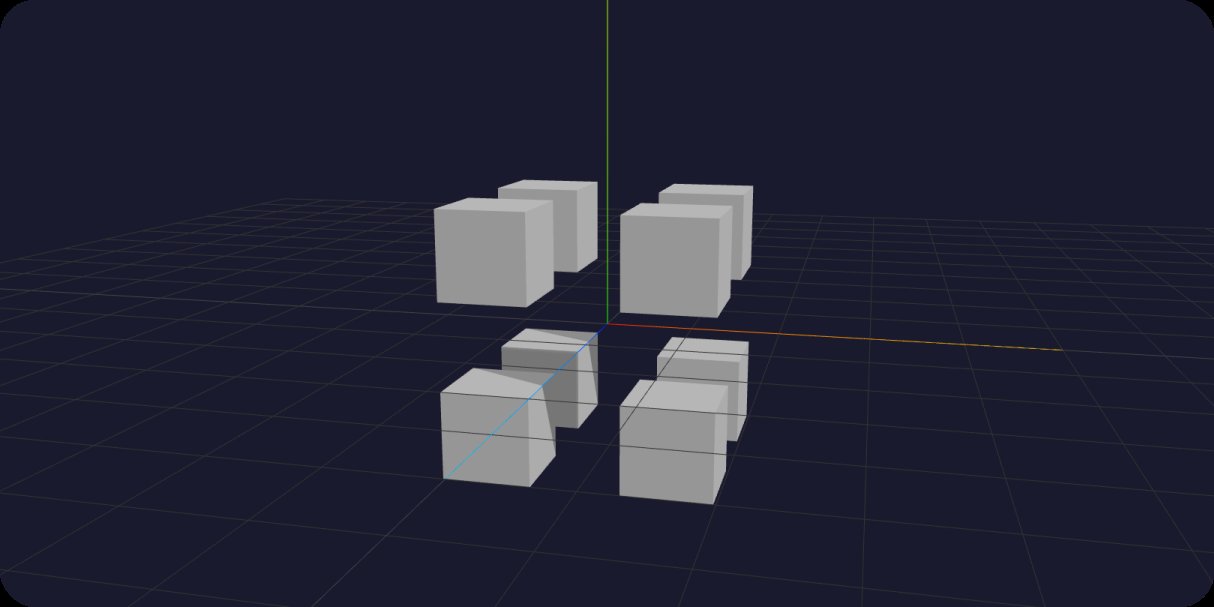} \\
\cline{2-2}
\hline

\cellcolor{orange!15}11 & {\raggedright\scriptsize\ttfamily \# Reorder Elements Along X-Axis \newline
  \newline
 \#\# GOAL \newline
 Rearrange four objects into a specified order along the positive X-axis with uniform spacing between centers. \newline
  \newline
 \#\# STEPS \newline
 1. Open the "Row of Objects Scene". \newline
 2. Reorder the elements as sphere, cone, cube, and cylinder along the positive X-axis. Final distance between centers of consecutive objects must be equal to 1. \newline
  \newline
 \# RESULT FORMAT \newline
  \newline
 ```json \newline
 \{"answer": "done"\} \newline
 ```} \\
\cline{2-2}
\cellcolor{orange!15} & \textbf{Initial state} \\
\cline{2-2}
\cellcolor{orange!15} & \includegraphics[width=\linewidth]{images/3d_editor_initial_jpegs/tc_3d_004.jpg} \\
\cline{2-2}
\cellcolor{orange!15} & \textbf{Ground truth} \\
\cline{2-2}
\cellcolor{orange!15} & \includegraphics[width=\linewidth]{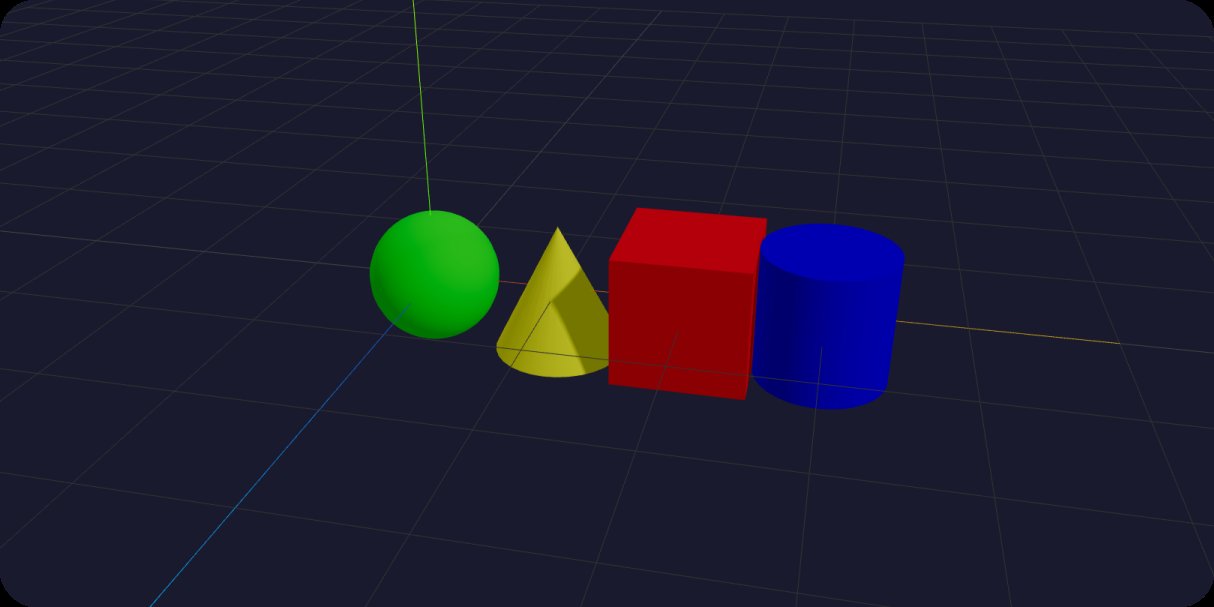} \\
\hline

\cellcolor{orange!15}12 & {\raggedright\scriptsize\ttfamily \# Find and Recolor Nearest Elements \newline
  \newline
 \#\# GOAL \newline
 Identify the two elements closest to a reference cube and change their color to red. \newline
  \newline
 \#\# STEPS \newline
 1. Open the "Geometric Garden". \newline
 2. Use the stone block cube as the reference. Identify the two remaining elements whose centers are closest to that of the stone block cube. Finally, change their color to red (\#ff0000). \newline
  \newline
 \# RESULT FORMAT \newline
  \newline
 ```json \newline
 \{"answer": "done"\} \newline
 ```} \\
\cline{2-2}
\cellcolor{orange!15} & \textbf{Initial state} \\
\cline{2-2}
\cellcolor{orange!15} & \includegraphics[width=\linewidth]{images/3d_editor_initial_jpegs/tc_3d_003.jpg} \\
\cline{2-2}
\cellcolor{orange!15} & \textbf{Ground truth} \\
\cline{2-2}
\cellcolor{orange!15} & \includegraphics[width=\linewidth]{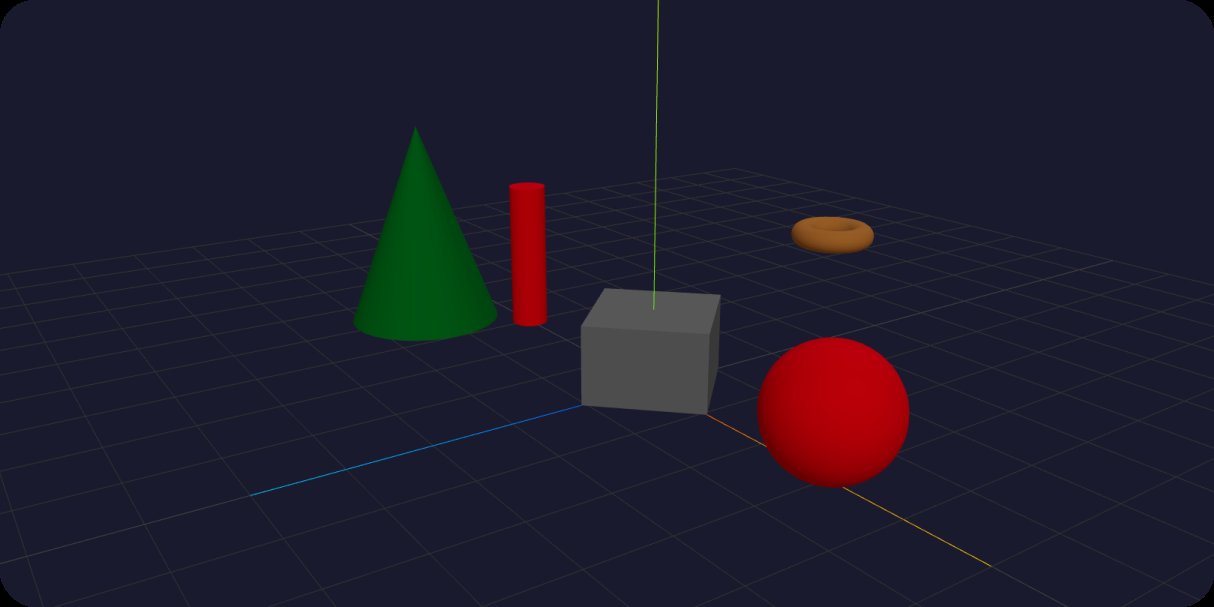} \\
\hline

\cellcolor{orange!15}13 & {\raggedright\scriptsize\ttfamily \# Spheres Touching Cube Faces From Inside \newline
 \newline
\#\# GOAL \newline
Create unit spheres that touch the faces of an imagined cube from inside. \newline
 \newline
\#\# STEPS \newline
1. Open the "Empty Scene". \newline
2. Imagine a cube of side 6 laying on the XZ plane. \newline
3. Add one unit sphere per cube face. \newline
4. Place each sphere inside the imagined cube so it touches, but does not intersect, its assigned face at the center of that face. \newline
 \newline
\# RESULT FORMAT \newline
 \newline
```json \newline
\{"answer": "done"\} \newline
```} \\
\cline{2-2}
\cellcolor{orange!15} & \textbf{Ground truth} \\
\cline{2-2}
\cellcolor{orange!15} & \includegraphics[width=\linewidth]{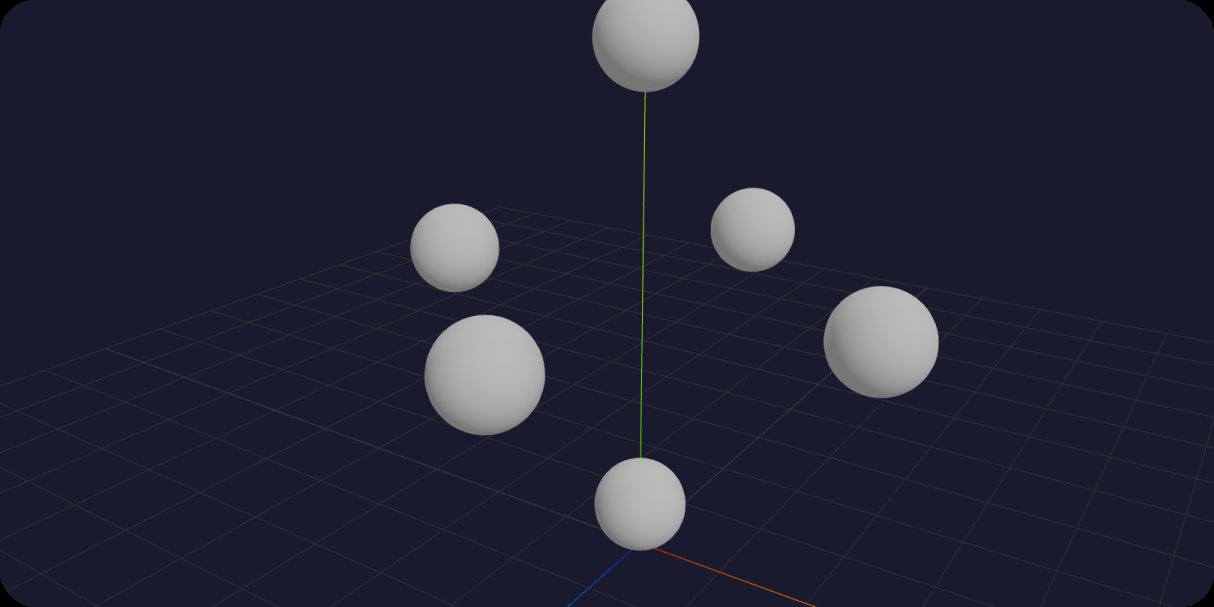} \\
\hline

\cellcolor{orange!15}14 & {\raggedright\scriptsize\ttfamily \# Place a Torus on the Missing Axis \newline
  \newline
 \#\# GOAL \newline
 Identify the unoccupied axis in the XZ plane and place a new torus there, matching the distance of existing elements from the center. \newline
  \newline
 \#\# STEPS \newline
 1. Open the "Four Objects Scene". \newline
 2. Add a torus and place it along the only axis in the XZ plane that does not contain any existing element. Set its distance from the central point to be the same as that of the other elements. \newline
  \newline
 \# RESULT FORMAT \newline
  \newline
 ```json \newline
 \{"answer": "done"\} \newline
 ```} \\
\cline{2-2}
\cellcolor{orange!15} & \textbf{Initial state} \\
\cline{2-2}
\cellcolor{orange!15} & \includegraphics[width=\linewidth]{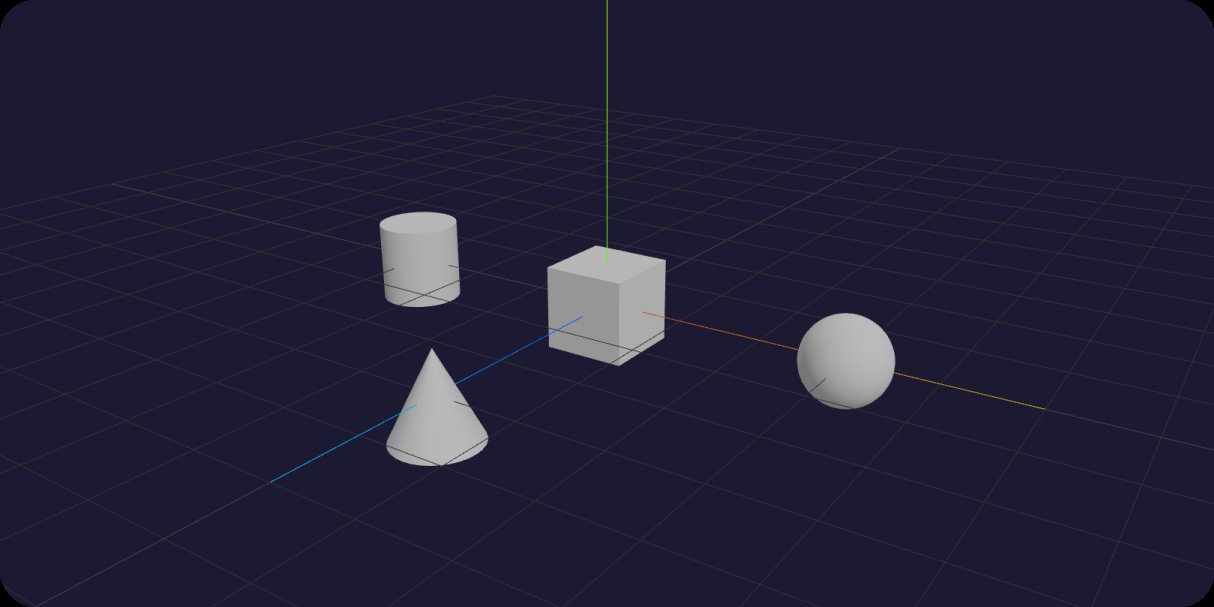} \\
\cline{2-2}
\cellcolor{orange!15} & \textbf{Ground truth} \\
\cline{2-2}
\cellcolor{orange!15} & \includegraphics[width=\linewidth]{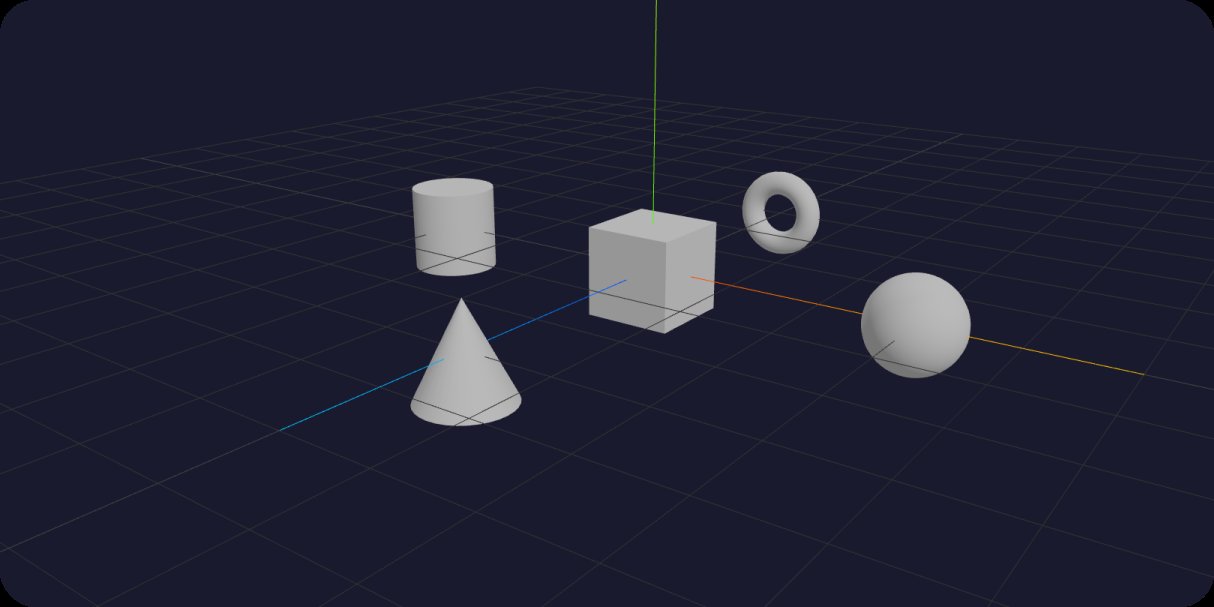} \\
\hline

\cellcolor{orange!15}15 & {\raggedright\scriptsize\ttfamily \# Adjust Cube (Cuboid) Scale and Stack Cone \newline
  \newline
 \#\# GOAL \newline
 Modify the cuboid’s Y-axis scale and reposition it to rest on the XZ plane, then place the cone on top of the cuboid with proper surface contact. \newline
  \newline
 \#\# STEPS \newline
1. Open the “Four Objects Scene”. \newline
2. Set the cuboid’s scale along the Y-axis to 20, and adjust its position so that its bottom surface is in contact with the XZ plane. \newline
3. Adjust the position of the cone so that its bottom surface is in contact with the top surface of the cuboid, without overlap. \newline
  \newline
 \# RESULT FORMAT \newline
  \newline
 ```json \newline
 \{"answer": "done"\} \newline
 ```} \\
\cline{2-2}
\cellcolor{orange!15} & \textbf{Initial state} \\
\cline{2-2}
\cellcolor{orange!15} & \includegraphics[width=\linewidth]{images/3d_editor_initial_jpegs/tc_3d_002.jpg} \\
\cline{2-2}
\cellcolor{orange!15} & \textbf{Ground truth} \\
\cline{2-2}
\cellcolor{orange!15} & \includegraphics[width=\linewidth]{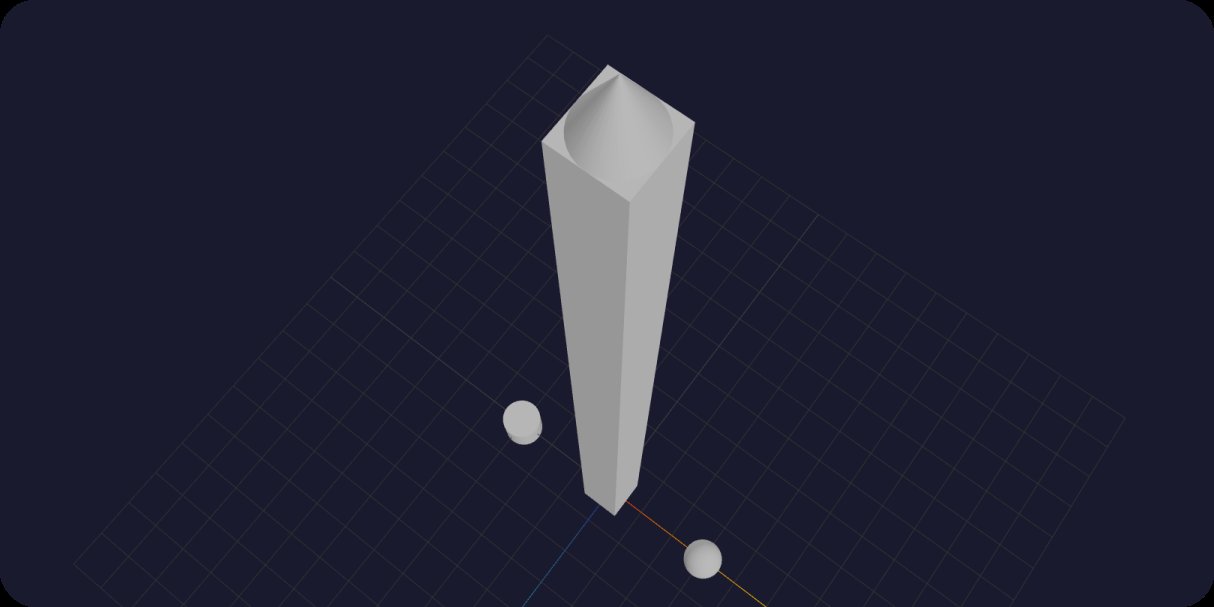} \\
\hline

\cellcolor{orange!15}16 & {\raggedright\scriptsize\ttfamily \# Symmetric Scene Around the Origin Object \newline
 \newline
\#\# GOAL \newline
Make the scene symmetric using the object at the origin as the scene center. \newline
 \newline
\#\# STEPS \newline
1. Open the "Two Objects Scene". \newline
2. Treat the object located at the origin as the center of the scene. Do not move or modify this center object. \newline
3. Change the existing cube's color to red (\#ff0000). \newline
4. Add exactly one new object. \newline
5. Change the new object's properties so that the final scene is symmetric around the center object. \newline
6. Do not add any guide objects or extra objects. \newline
 \newline
\# RESULT FORMAT \newline
 \newline
```json \newline
\{"answer": "done"\} \newline
```} \\
\cline{2-2}
\cellcolor{orange!15} & \textbf{Initial state} \\
\cline{2-2}
\cellcolor{orange!15} & \includegraphics[width=\linewidth]{images/3d_editor_initial_jpegs/tc_3d_002.jpg} \\
\cline{2-2}
\cellcolor{orange!15} & \textbf{Ground truth} \\
\cline{2-2}
\cellcolor{orange!15} & \includegraphics[width=\linewidth]{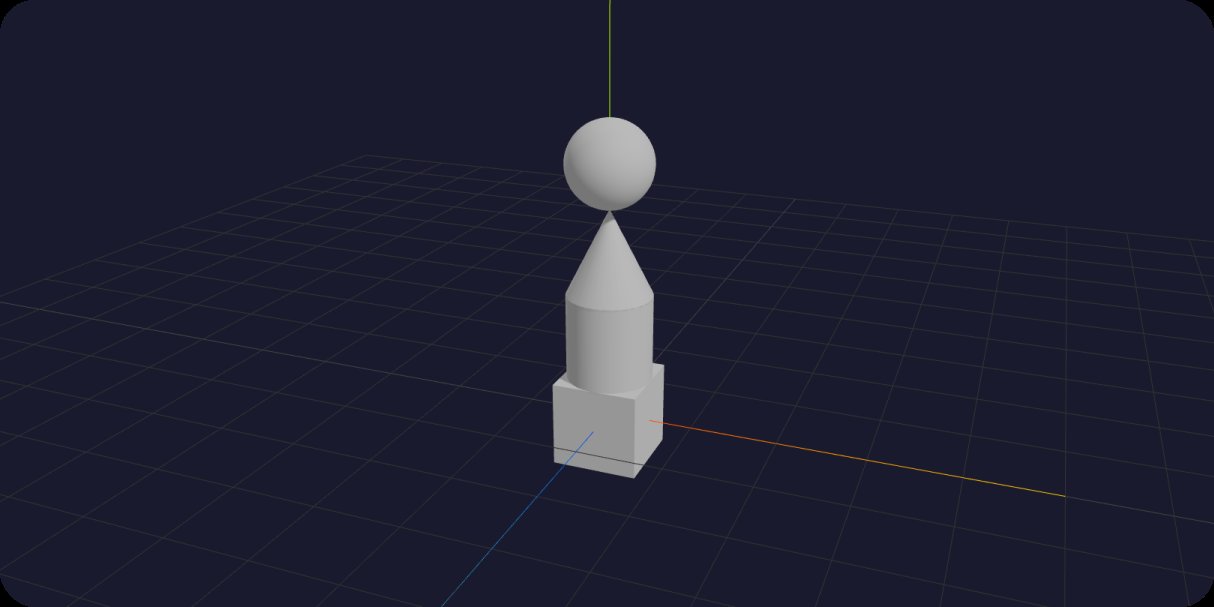} \\
\hline

\cellcolor{orange!15}17 & {\raggedright\scriptsize\ttfamily \# Complete the Top-Down Plus \newline
 \newline
\#\# GOAL \newline
Look at the scene from above. The four existing objects form an incomplete plus sign — one arm of the plus is missing. Add a single cube that visually completes the plus as seen from the +Y (top-down) direction. \newline
 \newline
\#\# STEPS \newline
1. Open the "Four Objects Scene". \newline
2. From top down view figures form a plus sign without one arm. Identify which arm of the top-down plus shape is missing. \newline
3. Add a new object named "Plug" in the missing arm position so that, viewed from directly above (+Y), the five objects form a symmetric plus sign. The new object's shape must match the existing object on the opposite arm of the plus, while keeping default rotation and unit scale. \newline
4. Place the "Plug" object on the same Y-axis level as all the other objects and color it purple (\#aa00ff). \newline
 \newline
\# RESULT FORMAT \newline
 \newline
```json \newline
\{"answer": "done"\} \newline
```} \\
\cline{2-2}
\cellcolor{orange!15} & \textbf{Initial state} \\
\cline{2-2}
\cellcolor{orange!15} & \includegraphics[width=\linewidth]{images/3d_editor_initial_jpegs/tc_3d_002.jpg} \\
\cline{2-2}
\cellcolor{orange!15} & \textbf{Ground truth} \\
\cline{2-2}
\cellcolor{orange!15} & \includegraphics[width=\linewidth]{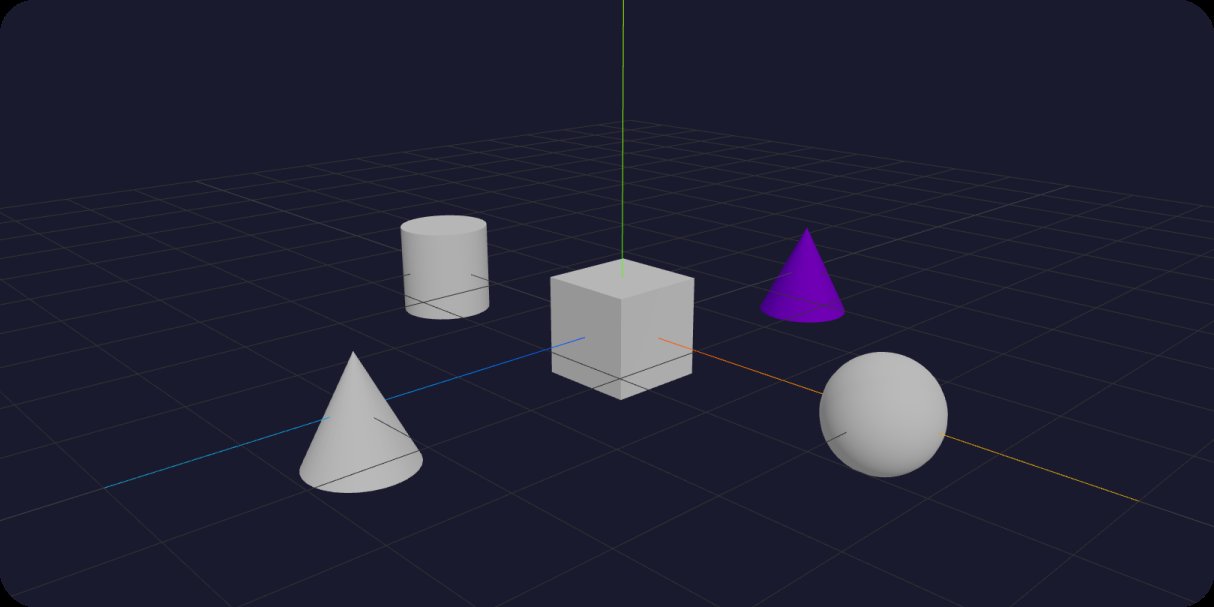} \\
\hline

\cellcolor{orange!15}18 & {\raggedright\scriptsize\ttfamily \# Fill the Missing End \newline
 \newline
\#\# GOAL \newline
Add one object that visually extends the row pattern, without disturbing the existing row. \newline
 \newline
\#\# STEPS \newline
1. Open the "Row of Objects Scene". \newline
2. In the default camera, observe the four objects as a left-to-right row of distinct silhouettes. \newline
3. Add exactly one new object named "Row Extension" at the next equally spaced position beyond the current rightmost visible silhouette, continuing the same straight row and center-to-center spacing. \newline
4. The new object's must: \newline
  * match the shape of the second (starting from the origin) object in the row, \newline
  * match the row's Y and Z levels, use default rotation and unit scale, \newline
  * have a purple (\#aa00ff) color. \newline
5. Preserve every original row object's position, rotation, scale, color, name, visibility, opacity, and wireframe setting. \newline
 \newline
\# RESULT FORMAT \newline
 \newline
```json \newline
\{"answer": "done"\} \newline
```} \\
\cline{2-2}
\cellcolor{orange!15} & \textbf{Initial state} \\
\cline{2-2}
\cellcolor{orange!15} & \includegraphics[width=\linewidth]{images/3d_editor_initial_jpegs/tc_3d_004.jpg} \\
\cline{2-2}
\cellcolor{orange!15} & \textbf{Ground truth} \\
\cline{2-2}
\cellcolor{orange!15} & \includegraphics[width=\linewidth]{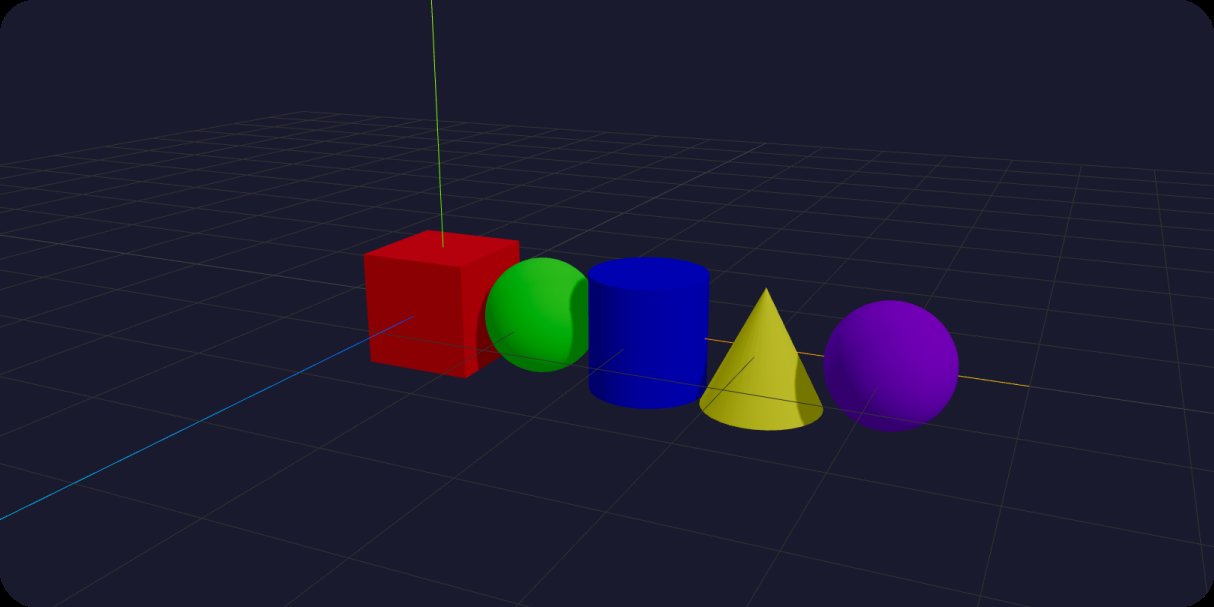} \\
\hline

\cellcolor{red!15}19 & {\raggedright\scriptsize\ttfamily \# Minimize Total Exposed Volume \newline
  \newline
 \#\# GOAL \newline
 Reposition objects in the XZ plane to minimize total exposed volume by maximizing overlap, keeping Y positions and all rotations/scales fixed. \newline
  \newline
 \#\# STEPS \newline
 1. Open the "Geometric Garden". \newline
 2. Use the stone block as the reference. Adjust the positions of the other objects along the X- and Z-axes while keeping their Y-axis positions fixed, with all rotations and scales unchanged. The objective is to minimize the total exposed volume of all objects, excluding any overlapping volume from the calculation. \newline
  \newline
 \# RESULT FORMAT \newline
  \newline
 ```json \newline
 \{"answer": "done"\} \newline
 ```} \\
\cline{2-2}
\cellcolor{red!15} & \textbf{Initial state} \\
\cline{2-2}
\cellcolor{red!15} & \includegraphics[width=\linewidth]{images/3d_editor_initial_jpegs/tc_3d_003.jpg} \\
\cline{2-2}
\cellcolor{red!15} & \textbf{Ground truth} \\
\cline{2-2}
\cellcolor{red!15} & \includegraphics[width=\linewidth]{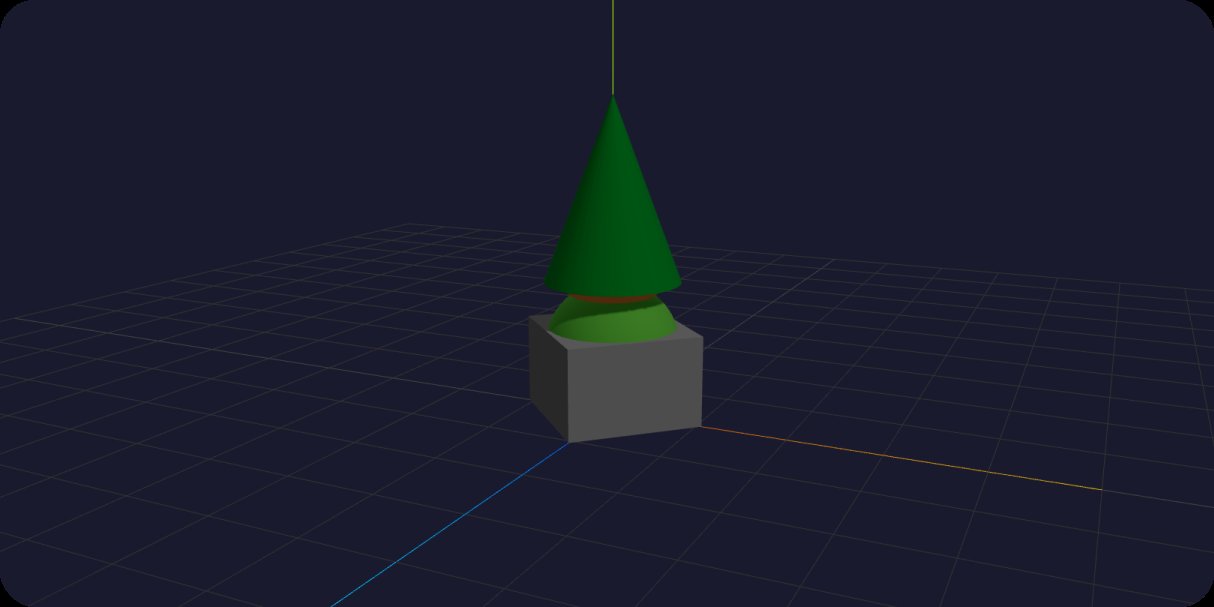} \\
\hline

\cellcolor{red!15}20 & {\raggedright\scriptsize\ttfamily \# Create Cube (Cuboid) with Scale Combinations \newline
 \newline
\#\# GOAL \newline
 \newline
Find all possible integer scale combinations of a cube using the values 1, 2, and 4 on the x, y, and z axes. The three scale values must be assigned in all possible permutations, and each axis may be either positive or negative. \newline
 \newline
\#\# STEPS \newline
 \newline
1. Open the "Empty Scene". \newline
 \newline
2. Create cubes (cuboids) for all possible combinations of scale values [1, 2, 4] across the x, y, and z axes, allowing each axis to be positive or negative. \newline
 \newline
3. Set the position of each cube to (0, 0, 0) and assign all of them the same name, Cube 1. \newline
 \newline
\#\# RESULT FORMAT \newline
 \newline
```json \newline
 \newline
\{"answer": "done"\} \newline
```} \\
\cline{2-2}
\cellcolor{red!15} & \textbf{Ground truth} \\
\cline{2-2}
\cellcolor{red!15} & \includegraphics[width=\linewidth]{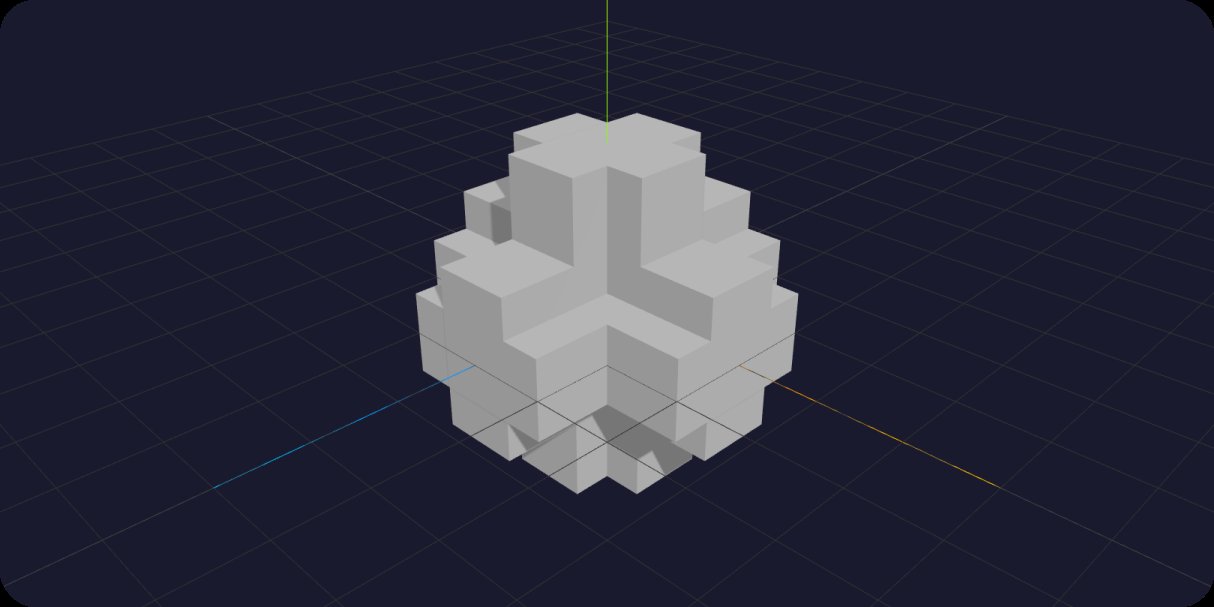} \\
\cline{2-2}
\hline

\cellcolor{red!15}21 & {\raggedright\scriptsize\ttfamily \# Match Cone Volume to Cylinder and Stack \newline
  \newline
 \#\# GOAL \newline
 Adjust a cone's Y-axis scale to match the cylinder's volume, place the cone on top of the cylinder, and recolor both to red. \newline
  \newline
 \#\# STEPS \newline
1. Open the “Four Objects Scene”. \newline
2. Adjust the cone so that its volume matches that of the cylinder by modifying only the scale along the Y-axis. The volume of the cylinder is given by $V = \pi r^2 h$, and the volume of the cone is given by $V = (1/3)\pi r^2 h$. Determine the required Y-axis scaling factor to achieve equal volumes. Next, position the cone on top of the cylinder without any overlap. Finally, treat the cylinder and cone as a single object and change its color to red (\#ff0000). \newline
  \newline
 \# RESULT FORMAT \newline
  \newline
 ```json \newline
 \{"answer": "done"\} \newline
 ```} \\
\cline{2-2}
\cellcolor{red!15} & \textbf{Initial state} \\
\cline{2-2}
\cellcolor{red!15} & \includegraphics[width=\linewidth]{images/3d_editor_initial_jpegs/tc_3d_002.jpg} \\
\cline{2-2}
\cellcolor{red!15} & \textbf{Ground truth} \\
\cline{2-2}
\cellcolor{red!15} & \includegraphics[width=\linewidth]{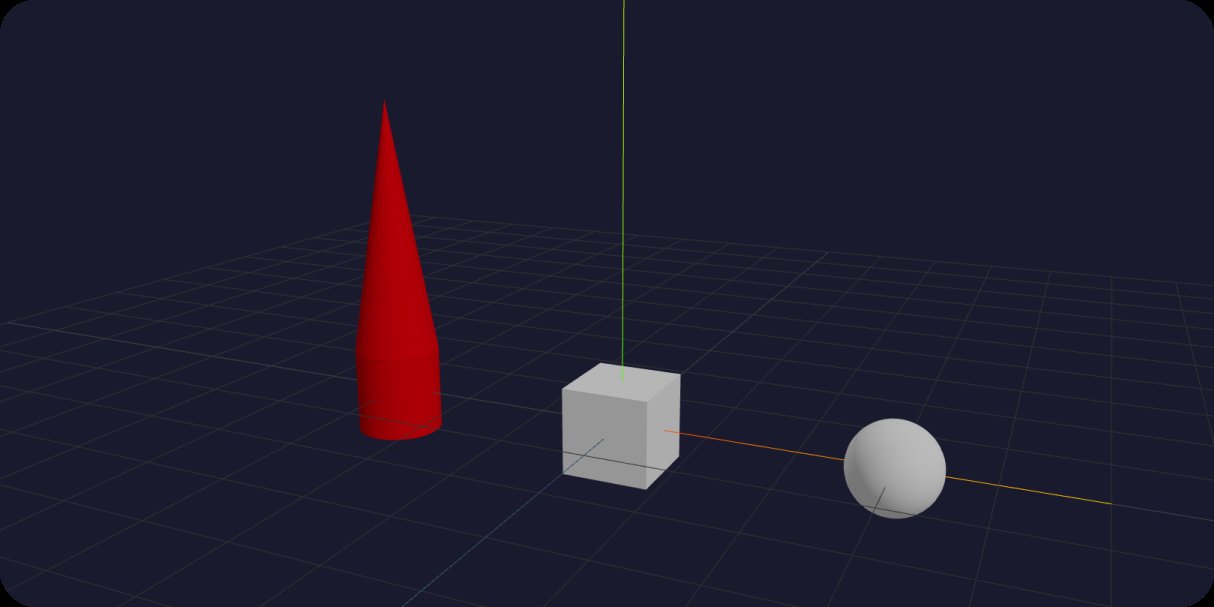} \\
\hline

\cellcolor{red!15}22 & {\raggedright\scriptsize\ttfamily \# Create a Minimum-Surface-Area Cube (Cuboid) \newline
  \newline
 \#\# GOAL \newline
 Create a cube with a volume of 8 and the smallest possible surface area, then set specific position, rotation, color, opacity, and background. \newline
  \newline
 \#\# STEPS \newline
1. Open the “Empty Scene”. \newline
2. Create a cube (cuboid) with integer side lengths a, b, and c such that the volume constraint abc = 8 is satisfied. The surface area is given by the formula S = 2(ab + bc + ac). Determine the integer dimensions that minimize the surface area, and construct the cuboid accordingly. \newline
3. Set the cuboid’s position to (1, 1, 1), rotation to (45, 45, 45), color to blue (\#0000ff), and opacity to 0.5. \newline
4. Set the background color to white (\#ffffff). \newline
  \newline
 \# RESULT FORMAT \newline
  \newline
 ```json \newline
 \{"answer": "done"\} \newline
 ```} \\
\cline{2-2}
\cellcolor{red!15} & \textbf{Ground truth} \\
\cline{2-2}
\cellcolor{red!15} & \includegraphics[width=\linewidth]{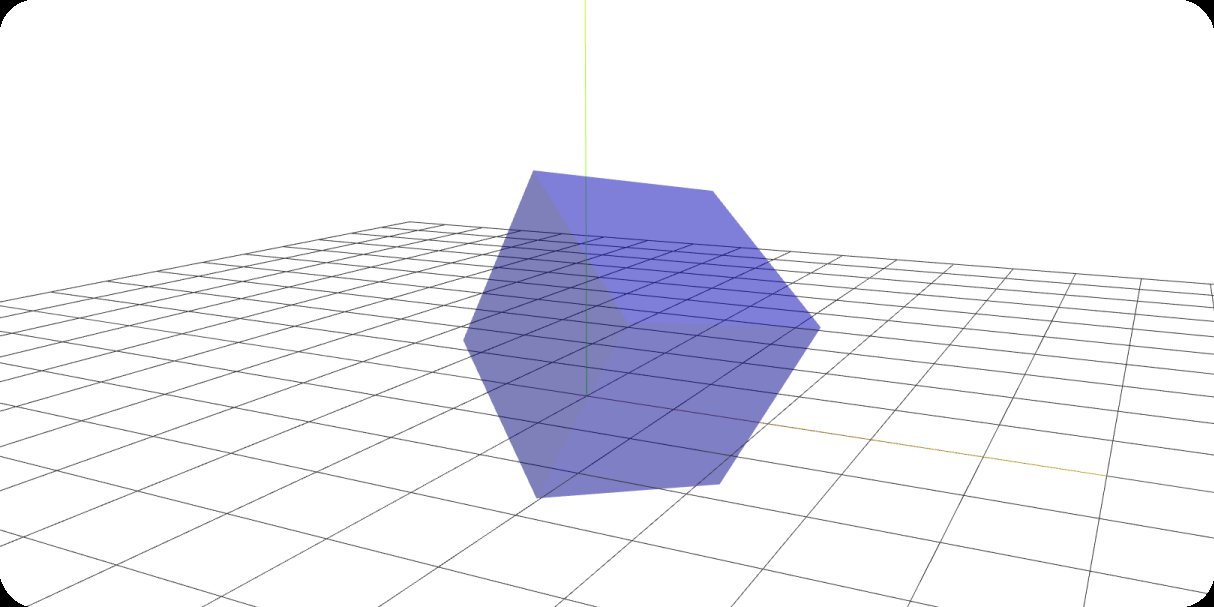} \\
\cline{2-2}
\hline

\cellcolor{red!15}23 & {\raggedright\scriptsize\ttfamily \# Stack Three Cubes into a Larger Cube (Cuboid) \newline
  \newline
 \#\# GOAL \newline
 Add two cubes (cuboids) to an existing scene so that all three cubes together form a single larger cube stacked along the Y-axis with no overlap. \newline
  \newline
 \#\# STEPS \newline
 1. Open the "Single Cube Scene". \newline
 2. Add two cubes (cuboids) that, together with the given Cube 1, form a unified larger cube (cuboid) with the same rotation. The bottom cubes (cuboids) should have its lower surface in contact with the XZ plane, and the total height along the Y-axis should be 3. Ensure that there is no overlap among the three individual cubes (cuboids). Name the bottom cube Cube 2 and set its color to blue (\#0000ff). Name the top cube Cube 3 and set its color to green (\#00ff00). \newline
  \newline
 \# RESULT FORMAT \newline
  \newline
 ```json \newline
 \{"answer": "done"\} \newline
 ```} \\
\cline{2-2}
\cellcolor{red!15} & \textbf{Initial state} \\
\cline{2-2}
\cellcolor{red!15} & \includegraphics[width=\linewidth]{images/3d_editor_initial_jpegs/tc_3d_005.jpg} \\
\cline{2-2}
\cellcolor{red!15} & \textbf{Ground truth} \\
\cline{2-2}
\cellcolor{red!15} & \includegraphics[width=\linewidth]{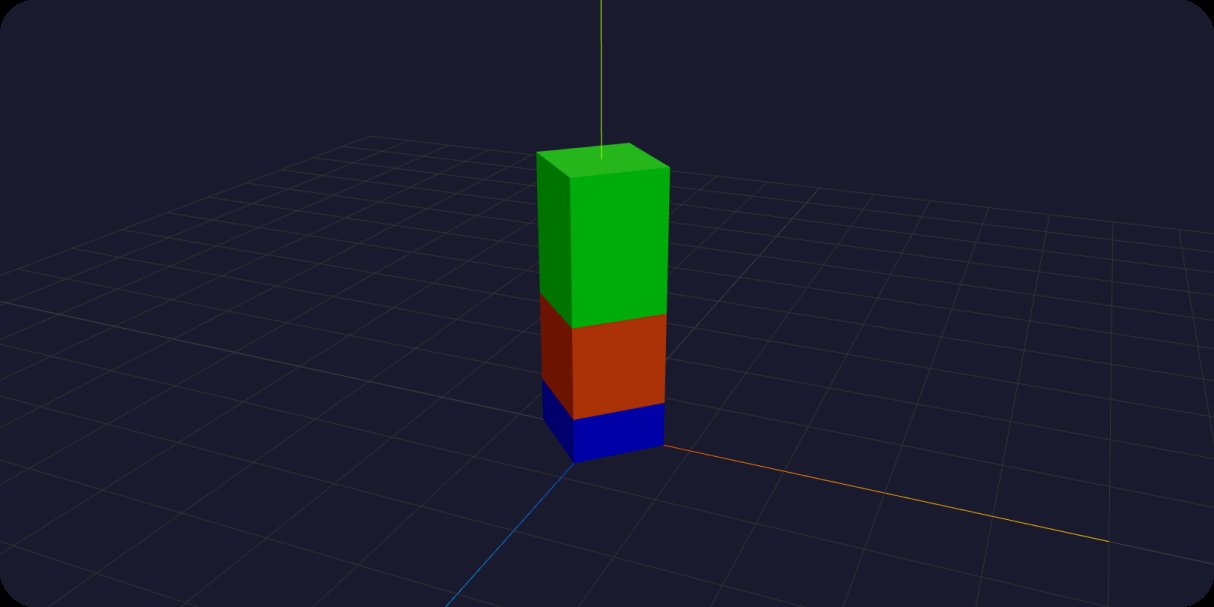} \\
\hline

\cellcolor{red!15}24 & {\raggedright\scriptsize\ttfamily \# Iterative Volume-Based Filtering and Scaling \newline
  \newline
 \#\# GOAL \newline
 Iteratively remove the object with the smallest volume, double the Y-axis scale of the remaining objects, and repeat until only one object remains, then set its color to white. \newline
  \newline
 \#\# STEPS \newline
1. Open the “Row of Objects Scene”. \newline
2. Compute the volumes of all objects (cuboid, cone, ellipsoid, cylinder). The relevant surface area formulas are: \newline
   * Cube (Cuboid): V = a * b * c \newline
   * Cylinder: V = pi * r\^{}2 * h \newline
   * Sphere (Ellipsoid): V = 1/6 * pi * a * b * c \newline
   * Cone: V = 1/3 * pi * r\^{}2 * h \newline
3. Identify and delete the object with the smallest volume. \newline
4. Double the Y-axis scale of all remaining objects. \newline
5. Repeat Steps 2–4 until only one object remains. \newline
6. Set the remaining object’s color to white (255, 255, 255). \newline
  \newline
 \# RESULT FORMAT \newline
  \newline
 ```json \newline
 \{"answer": "done"\} \newline
 ```} \\
\cline{2-2}
\cellcolor{red!15} & \textbf{Ground truth} \\
\cline{2-2}
\cellcolor{red!15} & \includegraphics[width=\linewidth]{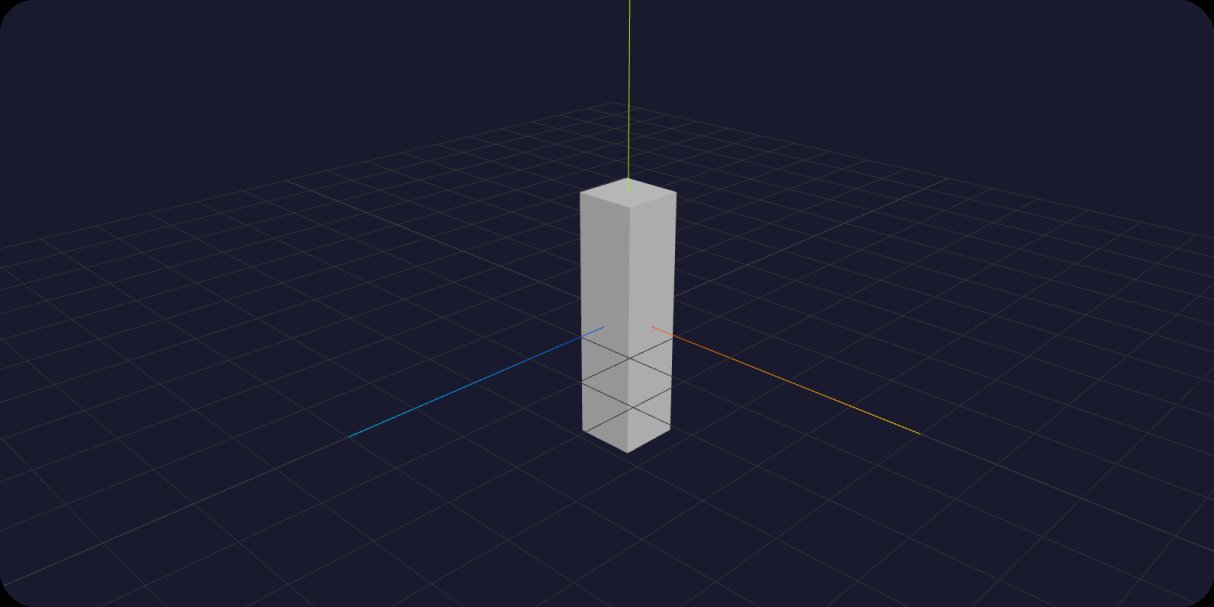} \\
\cline{2-2}
\hline

\cellcolor{red!15}25 & {\raggedright\scriptsize\ttfamily \# Reorder by Surface Area Plus Volume \newline
  \newline
 \#\# GOAL \newline
 Sort four elements along the positive X-axis by the sum of their surface area and volume (smallest to largest), preserving the original color order. \newline
  \newline
 \#\# STEPS \newline
1. Open the “Row of Objects Scene”. \newline
2. Reorder the given elements (cube (cuboid), sphere, cylinder, cone) along the positive X-axis, arranging them from smallest to largest based on the sum of their surface area and volume. The relevant formulas are: \newline
    * Cube (cuboid): surface area S = 2(ab + bc + ac), volume V = abc \newline
    * Sphere: surface area $S = 4\pi r^2 $, volume $V = (4/3)πr³$ \newline
    * Cylinder: surface area $S = 2\pi r^2  + 2πrh, volume V = \pi r^2 h$ \newline
    * Cone: surface area $S = \pi r^2  + πrl, volume V = (1/3)\pi r^2 h$ (where l is the slant height) \newline
        Preserve the original color order while reordering. \newline
3. Distance between centers of consecutive object must me 2.  \newline
 \newline
 \# RESULT FORMAT \newline
  \newline
 ```json \newline
 \{"answer": "done"\} \newline
 ```} \\
\cline{2-2}
\cellcolor{red!15} & \textbf{Initial state} \\
\cline{2-2}
\cellcolor{red!15} & \includegraphics[width=\linewidth]{images/3d_editor_initial_jpegs/tc_3d_004.jpg} \\
\cline{2-2}
\cellcolor{red!15} & \textbf{Ground truth} \\
\cline{2-2}
\cellcolor{red!15} & \includegraphics[width=\linewidth]{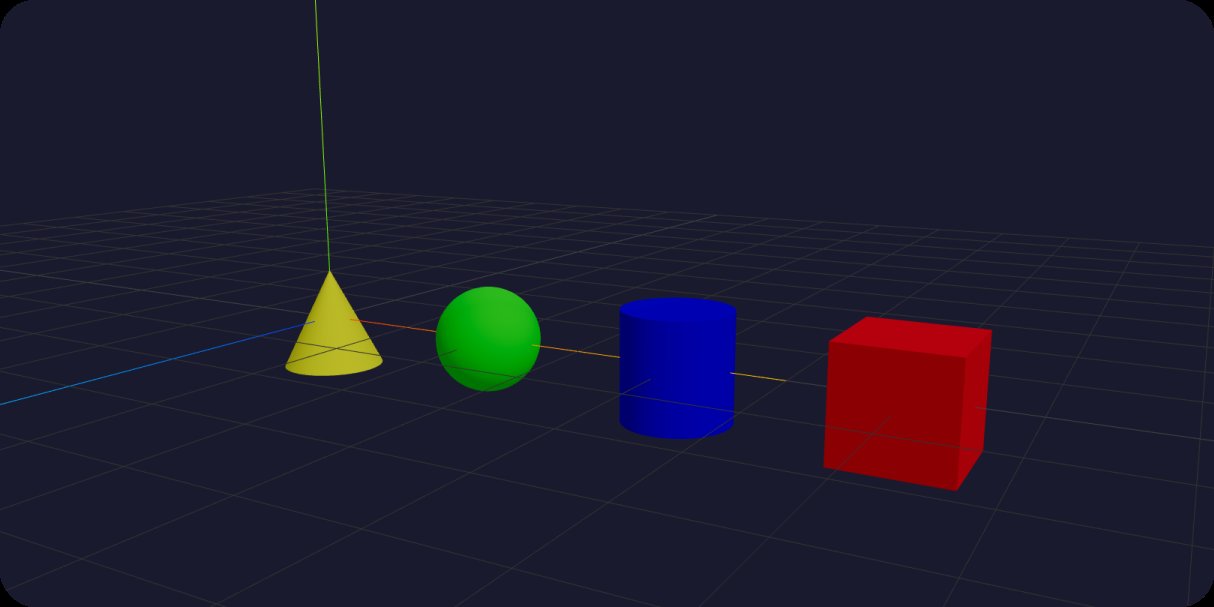} \\
\hline

\cellcolor{red!15}26 & {\raggedright\scriptsize\ttfamily \# Build a Ball on Table (Open-Ended) \newline
  \newline
 \#\# GOAL \newline
 Construct a table from a tabletop cube and four cylinder legs, then place a ball on the tabletop, computing positions so all parts are in contact. \newline
  \newline
 \#\# STEPS \newline
 1. Create a new scene named “Ball on Table.” Then, construct a 3D table and place a ball on the tabletop. \newline
 The table consists of the following components: a cube named Tabletop, which serves as the reference point, with its position set to (0, 4.2, 0) and scaled to (4, 0.4, 4); four cylinders of equal size named TableLeg, each scaled to (0.5, 4, 0.5); and a sphere named Ball with default settings. \newline
 These components constitute all the elements required to construct the “Ball on Table” scene. Your task is to determine their positions (except for the Tabletop, which is fixed as the reference). Each element should be connected to or tangent to the others. \newline
  \newline
 \# RESULT FORMAT \newline
  \newline
 ```json \newline
 \{"answer": "done"\} \newline
 ```} \\
\cline{2-2}
\cellcolor{red!15} & \textbf{Ground truth} \\
\cline{2-2}
\cellcolor{red!15} & \includegraphics[width=\linewidth]{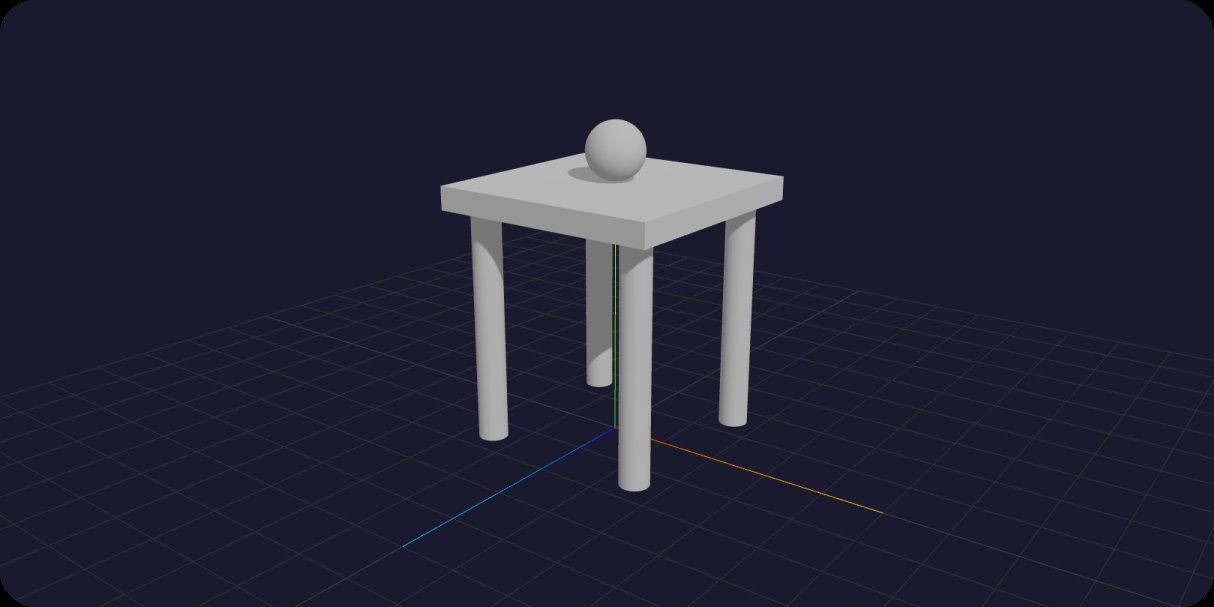} \\
\cline{2-2}
\hline

\cellcolor{red!15}27 & {\raggedright\scriptsize\ttfamily \# Build a 3D Snowman \newline
  \newline
 \#\# GOAL \newline
 Build a 3D snowman from scratch using multiple primitives (cylinder body, sphere head, torus eyes, cone nose, torus mouth, cube arms) with precise transforms. \newline
  \newline
 \#\# STEPS \newline
 1. Create a new scene named "Snowman". \newline
 2. Construct a 3D snowman. \newline
 First, add a cylinder and name it Body. Set its position to (0, 0.5, 0), using the default rotation, scale, color, and opacity. \newline
 Second, add a sphere named Head and set its position to (0, 1.4, 0). \newline
 Third, add a torus named Right Eye. Set its position to (0.45, 1.6, 0.2), rotation to (0, 90, 0), scale to (0.2, 0.2, 0.2), and color to black (\#000000). \newline
 Fourth, duplicate the Right Eye and rename it Left Eye. Set its position symmetrically along the Z-axis opposite to the right eye. \newline
 Fifth, add a cone named Nose. Set its position to (0.5, 1.5, 0), rotation to (0, 0, -90), scale to (0.1, 0.2, 0.1), and color to yellow (\#ffff00). \newline
 Sixth, add another torus named Mouth. Set its position to (0.5, 1.3, 0), rotation to (0, 90, 0), scale to (0.3, 0.1, 0.1), and color to red (\#ff0000). \newline
 Seventh, add a cube named Right Arm. Set its position to (0, 0.5, 0.6), rotation to (-45, 0, 0), and scale to (0.3, 0.8, 0.3). \newline
 Finally, duplicate the Right Arm and rename it Left Arm. Set its position symmetrically along the Z-axis opposite to the right arm, and apply the opposite rotation along the X-axis. \newline
  \newline
 \# RESULT FORMAT \newline
  \newline
 ```json \newline
 \{"answer": "done"\} \newline
 ```} \\
\cline{2-2}
\cellcolor{red!15} & \textbf{Ground truth} \\
\cline{2-2}
\cellcolor{red!15} & \includegraphics[width=\linewidth]{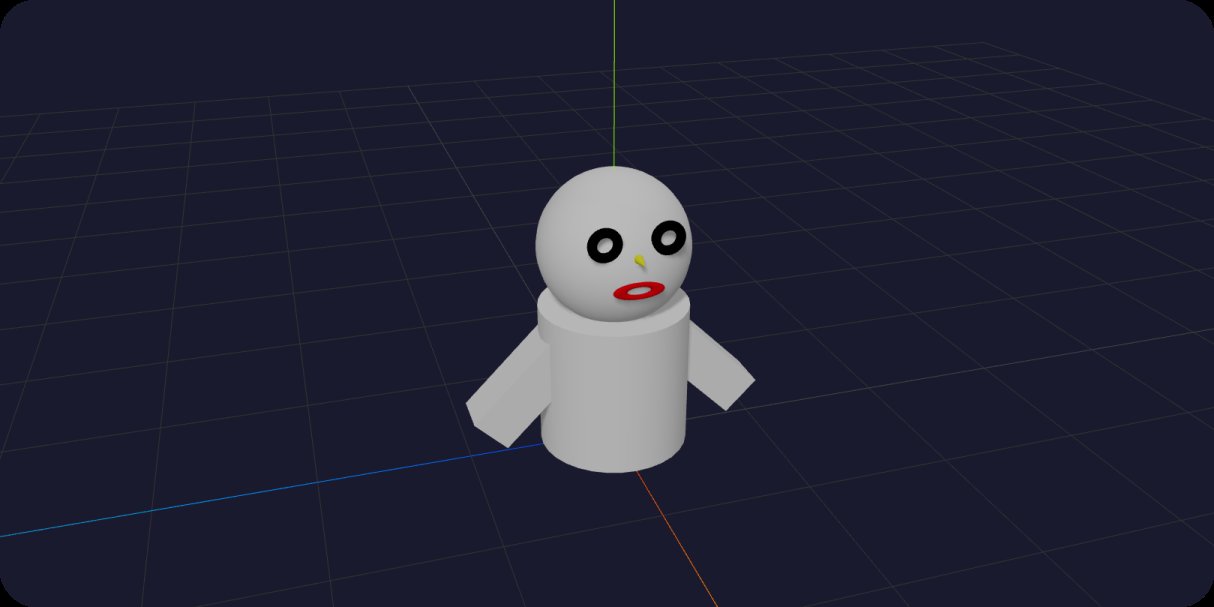} \\
\cline{2-2}
\hline

\end{longtable}
\normalsize

%% file: applications_testcases/testcases_flight_radar.tex
\scriptsize
\renewcommand{\arraystretch}{1.18}
\setlength{\LTleft}{0pt}
\setlength{\LTright}{0pt}
\setlength{\tabcolsep}{6pt}
\begin{longtable}{|>{\vspace{0pt}}p{0.03\linewidth}|>{\vspace{0pt}}p{0.49\linewidth}|>{\vspace{0pt}}p{0.40\linewidth}|}
\hline
\textbf{No.} & \textbf{Task} & \textbf{Ground truth} \\
\hline
\endfirsthead
\hline
\textbf{No.} & \textbf{Task} & \textbf{Ground truth} \\
\hline
\endhead
\hline
\endfoot
\cellcolor{green!15}1 &
{\raggedright\scriptsize\ttfamily \# Count Visible Aircraft at a Specific Zoom Level \newline  \newline \#\# GOAL \newline Count how many aircraft icons are visible on screen after centering on an airport and zooming in. \newline  \newline \#\# STEPS \newline 1. Set the playback time to 14:00 UTC. \newline 2. Search for airport \textbf{ICN} to center the map on it. \newline 3. Zoom in exactly 3 steps. \newline 4. Count the total number of aircraft icons currently visible on the screen. \newline  \newline \# RESULT FORMAT \newline  \newline ```json \newline \{ \newline  "count": <N> \newline \} \newline ```}
&
\begin{minipage}[t]{\linewidth}
\begin{BVerbatim}[fontsize=\scriptsize,breaklines=true,breakanywhere=true]
{
  "count": 2
}
\end{BVerbatim}
\end{minipage}
\\
\hline
\multicolumn{3}{|c|}{
\includegraphics[width=0.95\linewidth]{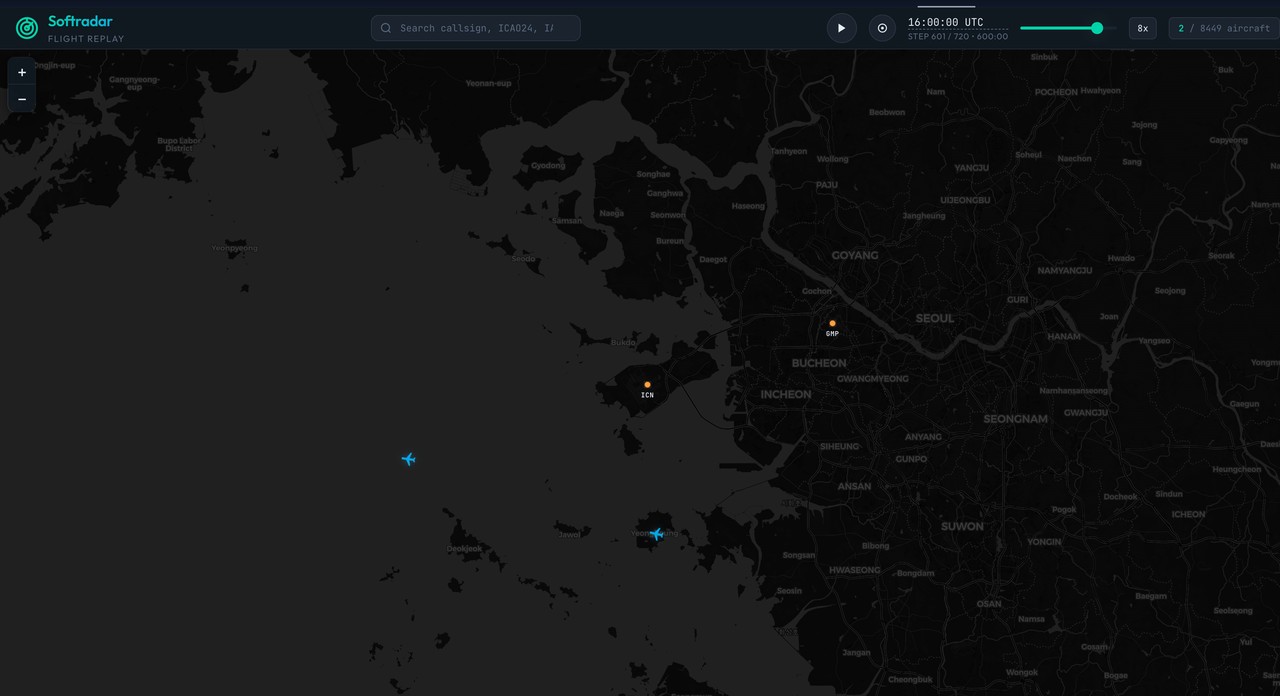}
} \\
\hline
\noalign{\vskip 6pt}
\hline
\cellcolor{green!15}2 &
{\raggedright\scriptsize\ttfamily \# Read Telemetry at a Specific Time \newline  \newline \#\# GOAL \newline Look up a flight's geo altitude and ground speed at a precise moment. \newline  \newline \#\# STEPS \newline 1. Set the playback time to 14:30 UTC. \newline 2. Pause the playback. \newline 3. Search for flight \textbf{VVC5776} using the search bar. \newline 4. Open its telemetry data. \newline 5. Read and report the exact geo altitude and ground speed values. \newline  \newline \# RESULT FORMAT \newline  \newline ```json \newline \{ \newline  "geo\_altitude": "<value>", \newline  "ground\_speed": "<value>" \newline \} \newline ```}
&
\begin{minipage}[t]{\linewidth}
\begin{BVerbatim}[fontsize=\scriptsize,breaklines=true,breakanywhere=true]
{
  "geo_altitude": "30,407 ft",
  "ground_speed": "386 kts"
}
\end{BVerbatim}
\end{minipage}
\\
\hline
\multicolumn{3}{|c|}{
\includegraphics[width=0.95\linewidth]{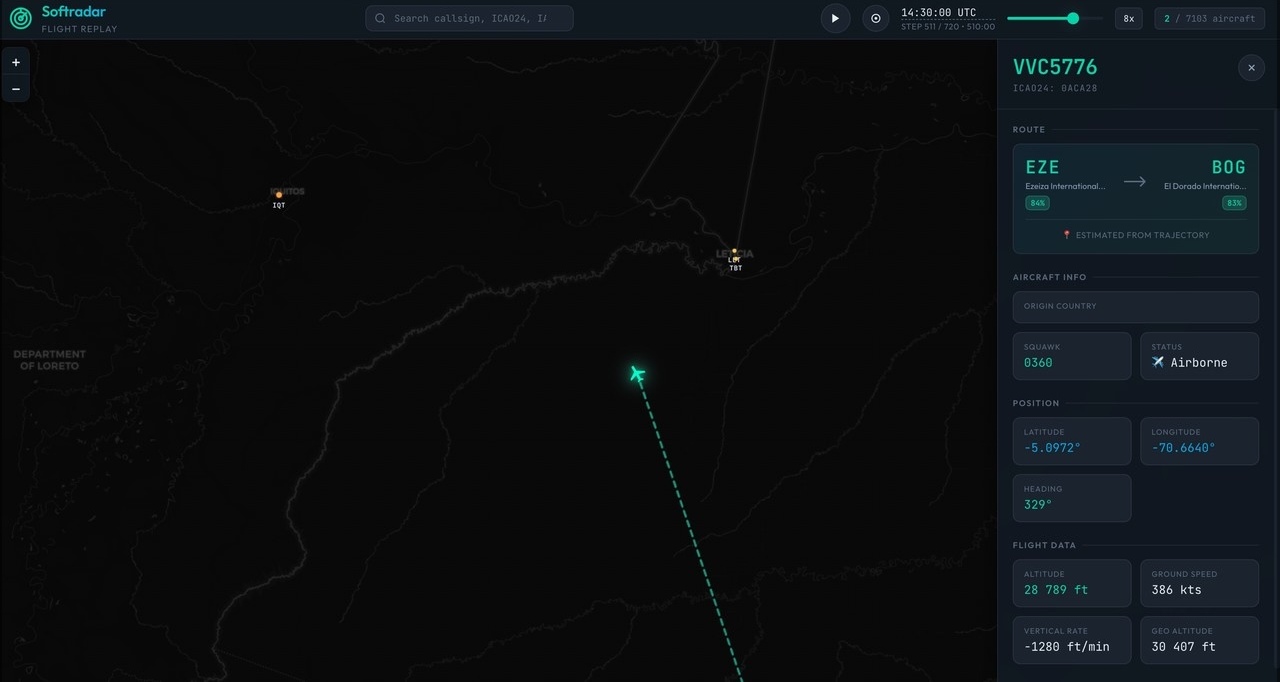}
} \\
\hline
\noalign{\vskip 6pt}
\hline
\cellcolor{green!15}3 &
{\raggedright\scriptsize\ttfamily \# Read Geo Altitude and Vertical Rate \newline  \newline \#\# GOAL \newline Look up a single flight's geo altitude and vertical rate at a precise moment. \newline  \newline \#\# STEPS \newline 1. Set the playback time to 16:00 UTC. \newline 2. Pause the playback. \newline 3. Search for flight \textbf{MSR726} using the search bar. \newline 4. Open its flight details and scroll to the "Flight Data" section. \newline 5. Read and report the geo altitude and vertical rate values. \newline  \newline \# RESULT FORMAT \newline  \newline ```json \newline \{ \newline  "geo\_altitude": "<value>", \newline  "vertical\_rate": "<value>" \newline \} \newline ```}
&
\begin{minipage}[t]{\linewidth}
\begin{BVerbatim}[fontsize=\scriptsize,breaklines=true,breakanywhere=true]
{
  "geo_altitude": "38,776 ft",
  "vertical_rate": "0 ft/min"
}
\end{BVerbatim}
\end{minipage}
\\
\hline
\multicolumn{3}{|c|}{
\includegraphics[width=0.95\linewidth]{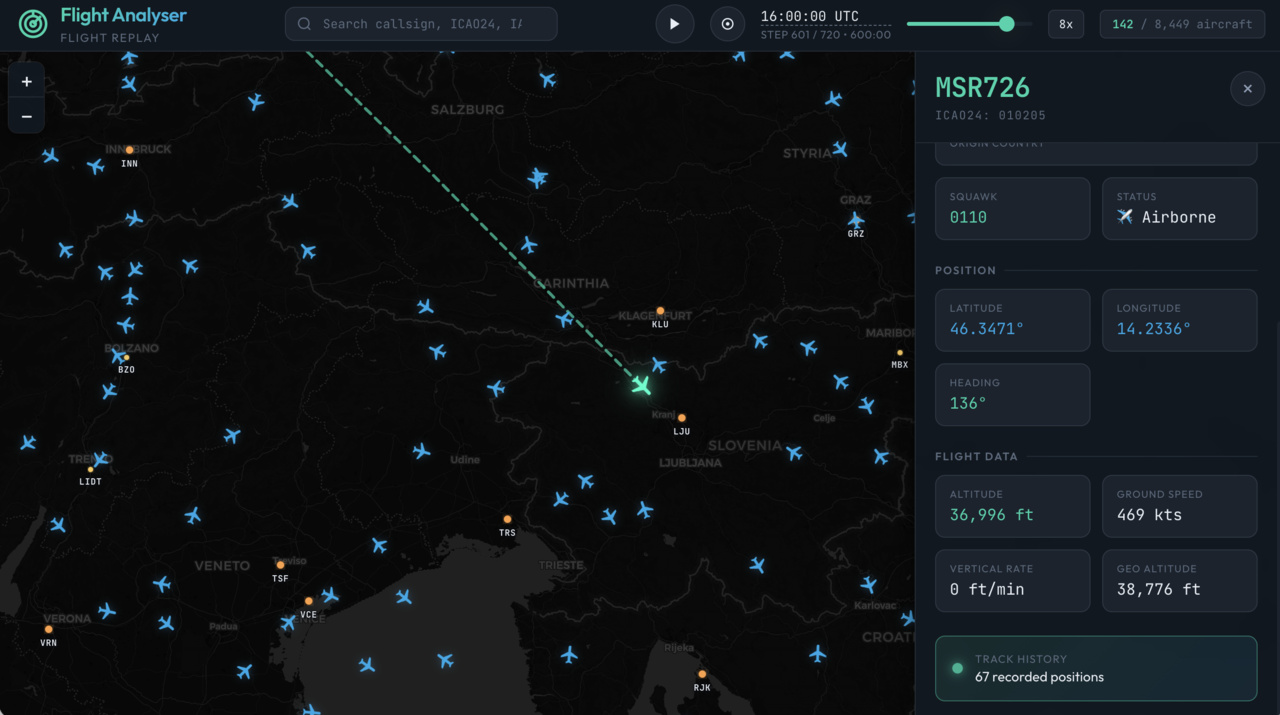}
} \\
\hline
\noalign{\vskip 6pt}
\hline
\cellcolor{green!15}4 &
{\raggedright\scriptsize\ttfamily \# Read a Flight's Route Endpoints \newline  \newline \#\# GOAL \newline Report the origin and destination airports of a single flight. \newline  \newline \#\# STEPS \newline 1. Set the playback time to 16:00 UTC. \newline 2. Search for flight \textbf{SAS907} using the search bar. \newline 3. Open its flight details and read the route. \newline 4. Report the origin and destination airport codes exactly as shown. \newline  \newline \# RESULT FORMAT \newline  \newline ```json \newline \{ \newline  "origin": "<origin>", \newline  "destination": "<destination>" \newline \} \newline ```}
&
\begin{minipage}[t]{\linewidth}
\begin{BVerbatim}[fontsize=\scriptsize,breaklines=true,breakanywhere=true]
{
  "origin": "OSL",
  "destination": "EWR"
}
\end{BVerbatim}
\end{minipage}
\\
\hline
\multicolumn{3}{|c|}{
\includegraphics[width=0.95\linewidth]{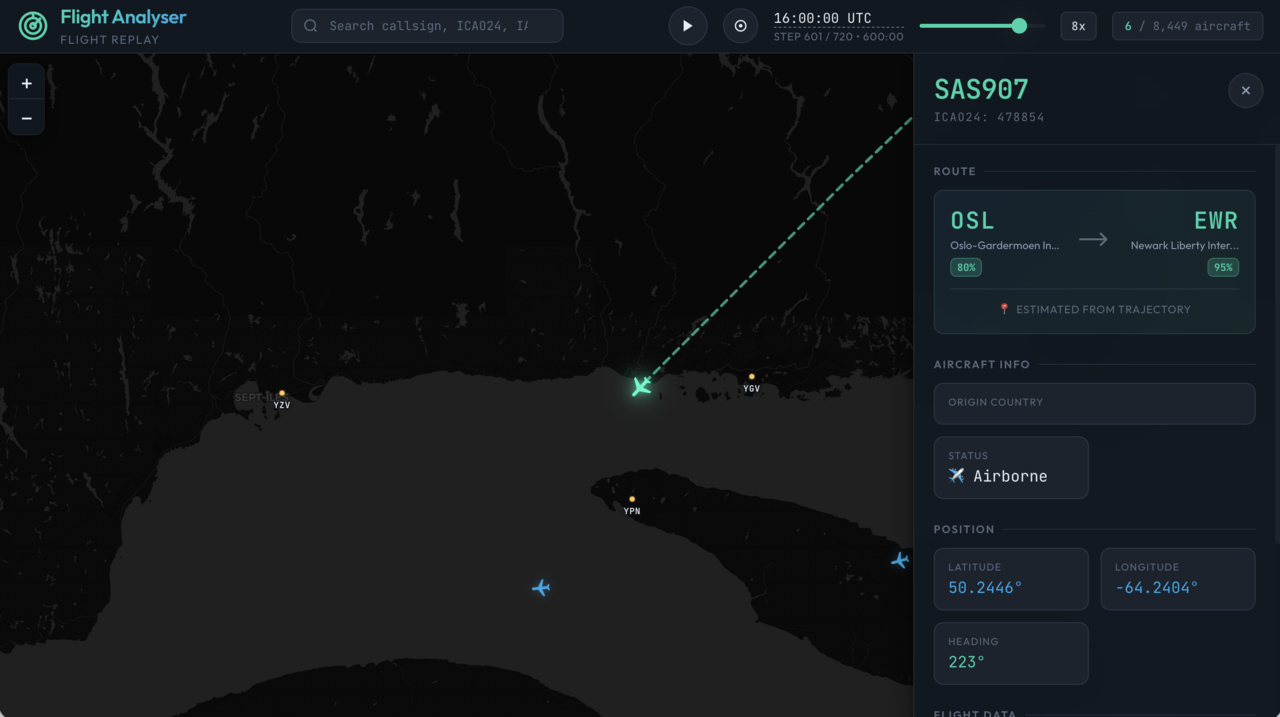}
} \\
\hline
\noalign{\vskip 6pt}
\hline
\cellcolor{green!15}5 &
{\raggedright\scriptsize\ttfamily \# Read a Flight's Ground Speed at Two Times \newline  \newline \#\# GOAL \newline Read the same telemetry field for one flight at two different playback times. \newline  \newline \#\# STEPS \newline 1. Set the playback time to 11:15 UTC and pause the playback. \newline 2. Search for flight \textbf{ASA1217} using the search bar and open its flight details. \newline 3. Read its ground speed and note it down. \newline 4. Set the playback time to 11:30 UTC, and read the ground speed of the same flight again. \newline 5. Report both values exactly as shown. \newline  \newline \# RESULT FORMAT \newline  \newline ```json \newline \{ \newline  "ground\_speed\_1115": "<value>", \newline  "ground\_speed\_1130": "<value>" \newline \} \newline ```}
&
\begin{minipage}[t]{\linewidth}
\begin{BVerbatim}[fontsize=\scriptsize,breaklines=true,breakanywhere=true]
{
  "ground_speed_1115": "480 kts",
  "ground_speed_1130": "474 kts"
}
\end{BVerbatim}
\end{minipage}
\\
\hline
\multicolumn{3}{|c|}{
\includegraphics[width=0.95\linewidth]{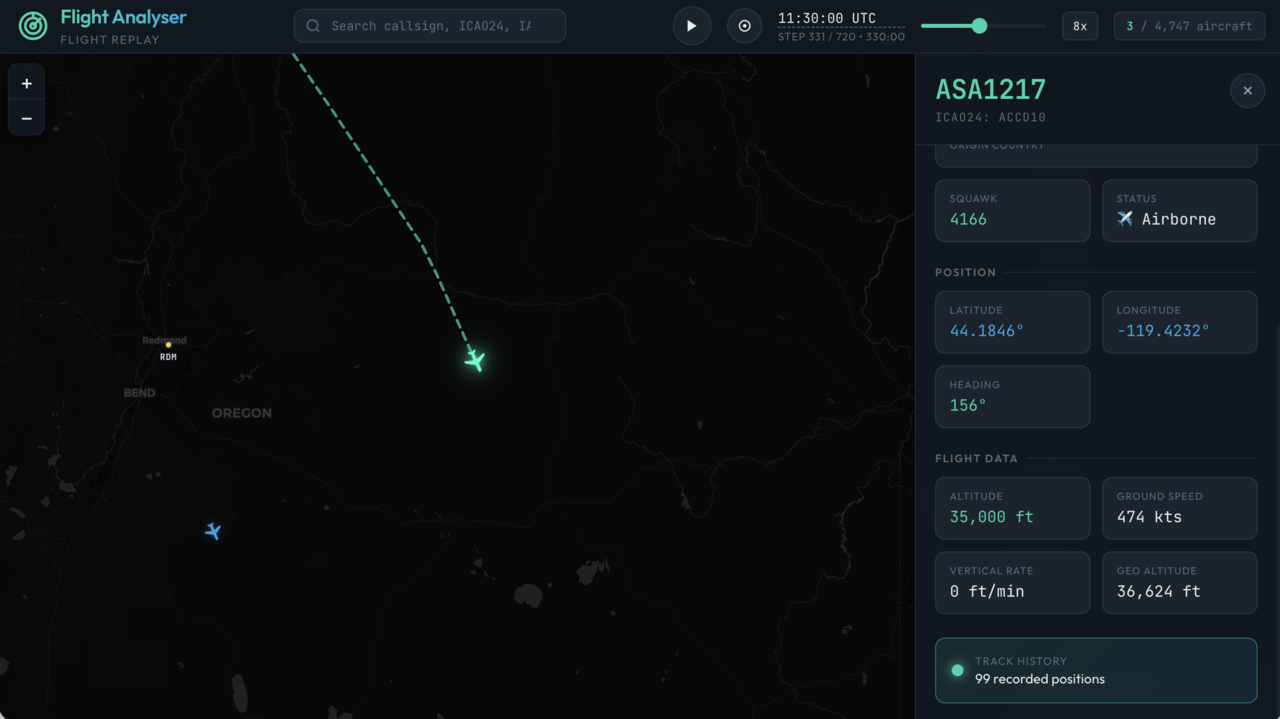}
} \\
\hline
\noalign{\vskip 6pt}
\hline
\cellcolor{green!15}6 &
{\raggedright\scriptsize\ttfamily \# Count Aircraft Inside a Radius Around Salzburg \newline  \newline \#\# GOAL \newline Use the radius tool to count the aircraft around a single airport at one moment. \newline  \newline \#\# STEPS \newline 1. Set the playback time to 16:30 UTC. \newline 2. Pause the playback. \newline 3. Search for \textbf{Salzburg Airport (SZG)} to center the map on it. \newline 4. Draw a 50 km radius circle around the airport. \newline 5. Count the aircrafts inside the circle. \newline 6. Report that number. \newline  \newline \# RESULT FORMAT \newline  \newline ```json \newline \{ \newline  "count": <N> \newline \} \newline ```}
&
\begin{minipage}[t]{\linewidth}
\begin{BVerbatim}[fontsize=\scriptsize,breaklines=true,breakanywhere=true]
{
  "count": 3
}
\end{BVerbatim}
\end{minipage}
\\
\hline
\multicolumn{3}{|c|}{
\includegraphics[width=0.95\linewidth]{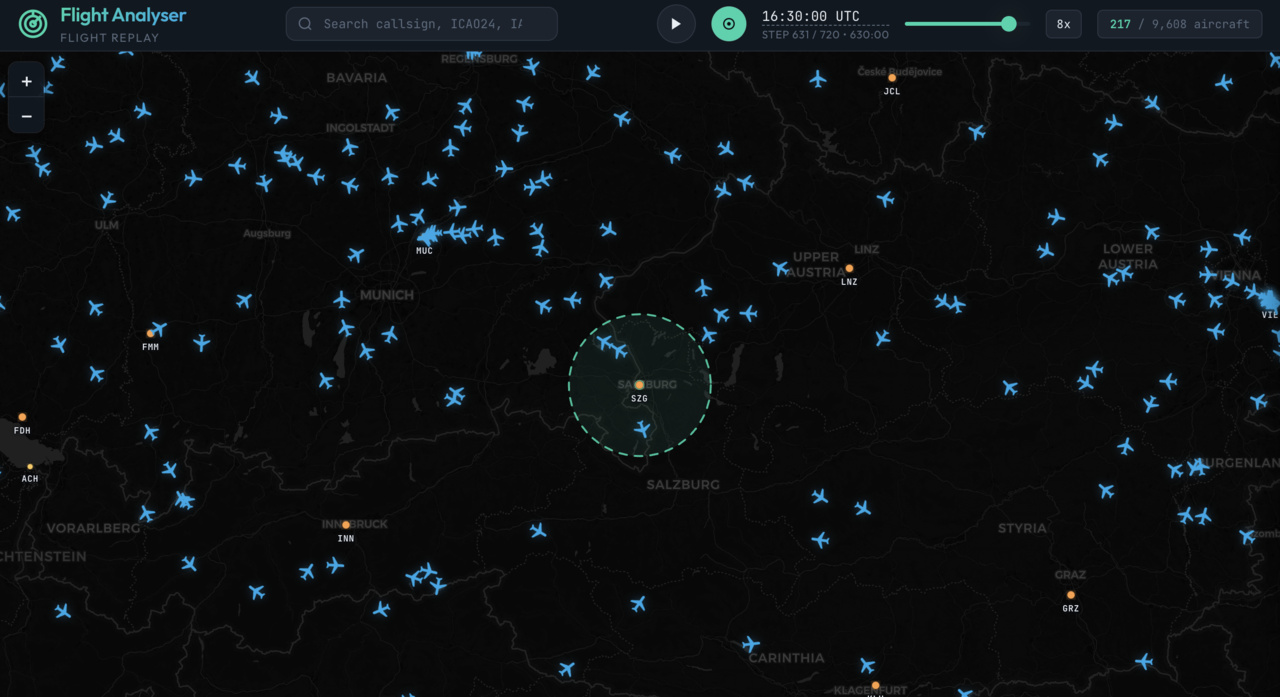}
} \\
\hline
\noalign{\vskip 6pt}
\hline
\cellcolor{green!15}7 &
{\raggedright\scriptsize\ttfamily \# Read the Geo Altitude of the Eastern Flight \newline  \newline \#\# GOAL \newline Decide which of two flights is further east on the map, then report the geo altitude of that flight only. \newline  \newline \#\# STEPS \newline 1. Set the playback time to 16:20 UTC. \newline 2. Pause the playback. \newline 3. Search for flight \textbf{AFR24NL} and note its position on the map. \newline 4. Search for flight \textbf{RYR61XN} and note its position on the map. \newline 5. Determine which of the two flights is further east. \newline 6. Open the details of that flight and report its callsign and its geo altitude exactly as shown. \newline  \newline \# RESULT FORMAT \newline  \newline ```json \newline \{ \newline  "callsign": "<callsign>", \newline  "geo\_altitude": "<value>" \newline \} \newline ```}
&
\begin{minipage}[t]{\linewidth}
\begin{BVerbatim}[fontsize=\scriptsize,breaklines=true,breakanywhere=true]
{
  "callsign": "AFR24NL",
  "geo_altitude": "32,451 ft"
}
\end{BVerbatim}
\end{minipage}
\\
\hline
\multicolumn{3}{|c|}{
\includegraphics[width=0.95\linewidth]{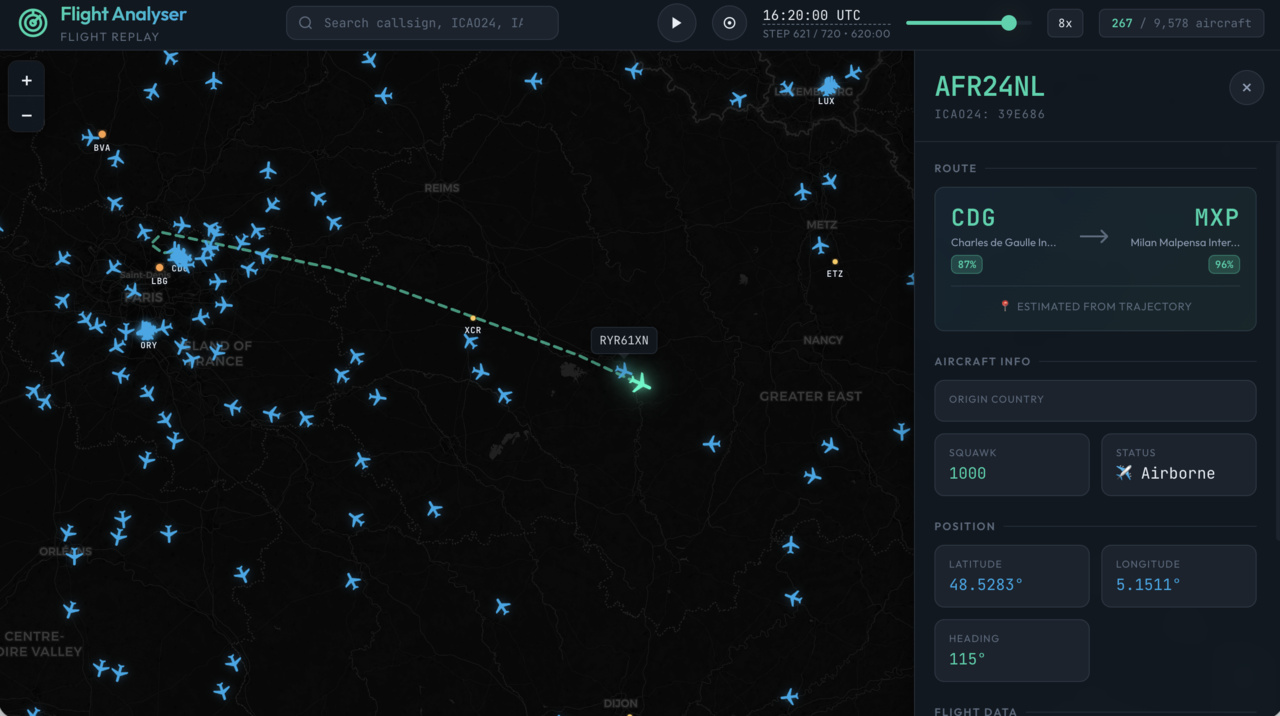}
} \\
\hline
\noalign{\vskip 6pt}
\hline
\cellcolor{green!15}8 &
{\raggedright\scriptsize\ttfamily \# Find the Minute a Flight Leaves the Ground \newline  \newline \#\# GOAL \newline Use the timeline to find the exact minute at which a flight's status changes. \newline  \newline \#\# STEPS \newline 1. Set the playback time to 14:22 UTC and pause the playback. \newline 2. Search for flight \textbf{EZY14GQ} and open its flight details. At this moment its status is "On Ground". \newline 3. Wait until the status is no longer "On Ground". \newline 4. Report the first UTC minute at which the new status appears, and that status exactly as shown. \newline  \newline \# RESULT FORMAT \newline  \newline ```json \newline \{ \newline  "status\_change\_utc": "<HH:MM UTC>", \newline  "new\_status": "<status>" \newline \} \newline ```}
&
\begin{minipage}[t]{\linewidth}
\begin{BVerbatim}[fontsize=\scriptsize,breaklines=true,breakanywhere=true]
{
  "status_change_utc": "14:23 UTC",
  "new_status": "Airborne"
}
\end{BVerbatim}
\end{minipage}
\\
\hline
\multicolumn{3}{|c|}{
\includegraphics[width=0.95\linewidth]{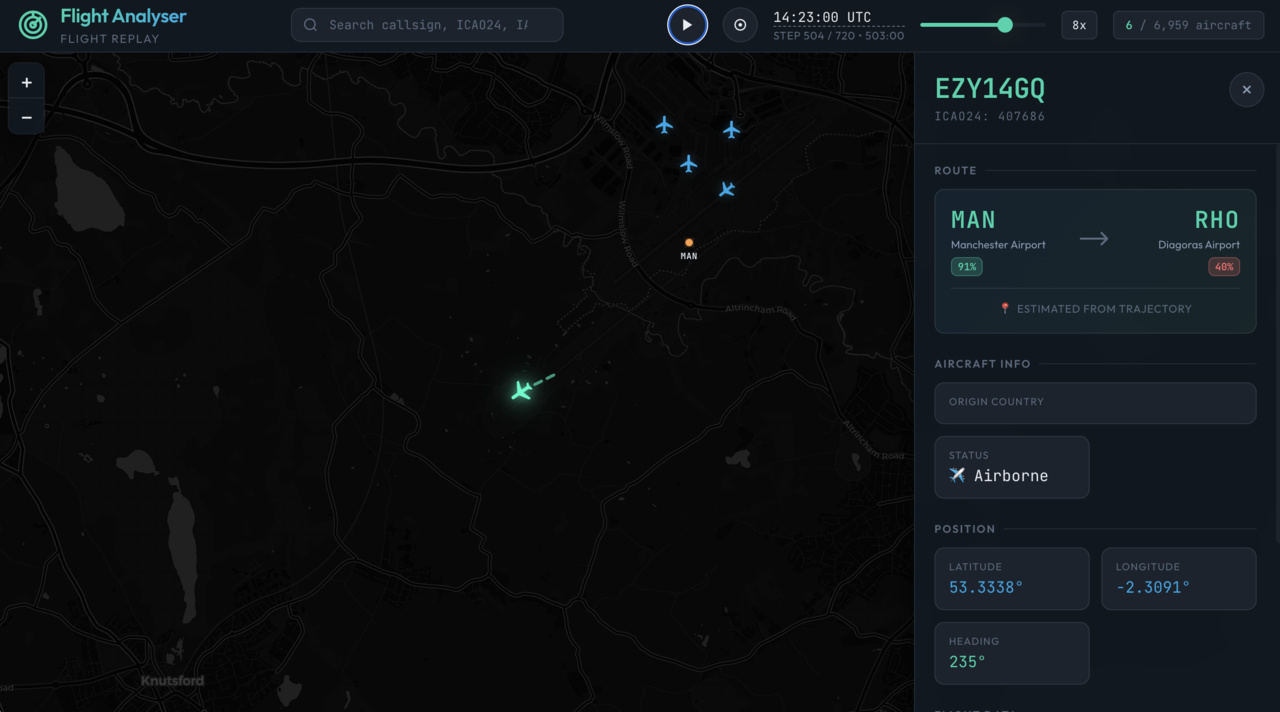}
} \\
\hline
\noalign{\vskip 6pt}
\hline
\cellcolor{green!15}9 &
{\raggedright\scriptsize\ttfamily \# Identify an Aircraft by Clicking It on the Map \newline  \newline \#\# GOAL \newline Open a flight's details by clicking its icon on the map instead of searching for its callsign. \newline  \newline \#\# STEPS \newline 1. Set the playback time to 14:00 UTC. \newline 2. Search for airport \textbf{ICN} to center the map on it. \newline 3. Click the aircraft icon that is closest to the airport marker. \newline 4. Report that aircraft's callsign and its destination airport code exactly as shown. \newline  \newline \# RESULT FORMAT \newline  \newline ```json \newline \{ \newline  "callsign": "<callsign>", \newline  "destination": "<destination>" \newline \} \newline ```}
&
\begin{minipage}[t]{\linewidth}
\begin{BVerbatim}[fontsize=\scriptsize,breaklines=true,breakanywhere=true]
{
  "callsign": "KAL8349",
  "destination": "DPS"
}
\end{BVerbatim}
\end{minipage}
\\
\hline
\multicolumn{3}{|c|}{
\includegraphics[width=0.95\linewidth]{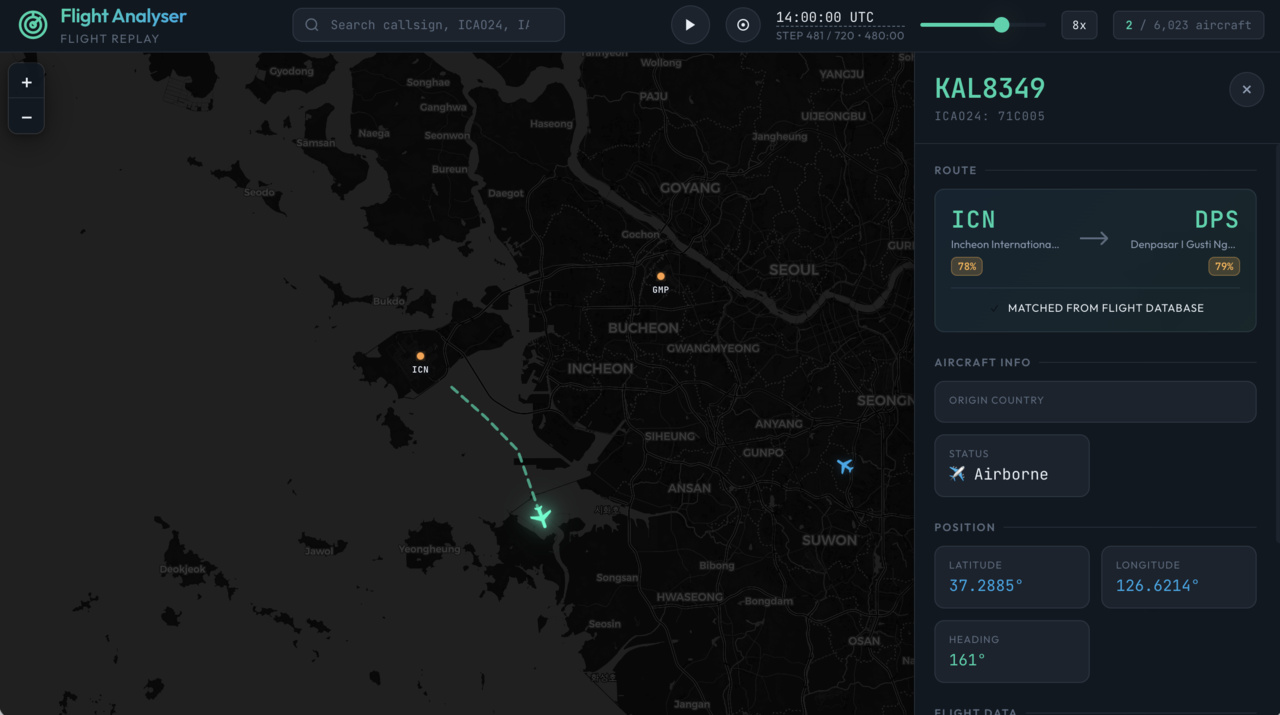}
} \\
\hline
\noalign{\vskip 6pt}
\hline
\cellcolor{orange!15}10 &
{\raggedright\scriptsize\ttfamily \# Count Landings at an Airport Over a Time Window \newline  \newline \#\# GOAL \newline Observe an airport and count how many aircraft land during a specified period. \newline  \newline \#\# STEPS \newline 1. Set the playback time to 16:14 UTC. \newline 2. Navigate to \textbf{Budapest Ferenc Liszt Airport (BUD)}. \newline 3. Observe the airport for 10 minutes of playback time. \newline 4. Count how many new aircraft land at the airport during this period. \newline  \newline \# RESULT FORMAT \newline  \newline ```json \newline \{ \newline  "count": <N> \newline \} \newline ```}
&
\begin{minipage}[t]{\linewidth}
\begin{BVerbatim}[fontsize=\scriptsize,breaklines=true,breakanywhere=true]
{
  "count": 5
}
\end{BVerbatim}
\end{minipage}
\\
\hline
\multicolumn{3}{|c|}{
\includegraphics[width=0.95\linewidth]{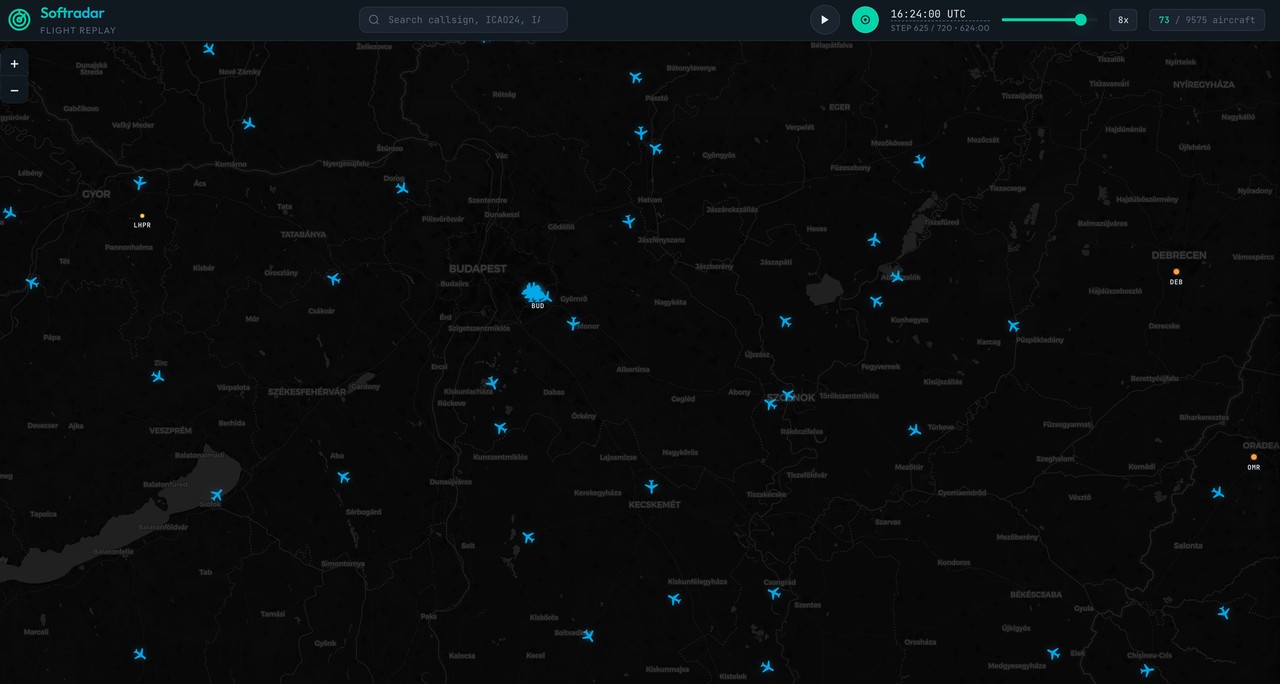}
} \\
\hline
\noalign{\vskip 6pt}
\hline
\cellcolor{orange!15}11 &
{\raggedright\scriptsize\ttfamily \# Count Aircraft Within a Circle at a Specific Time \newline  \newline \#\# GOAL \newline Count how many aircraft are located inside a defined radius around an airport. \newline  \newline \#\# STEPS \newline 1. Set the playback time to 11:41 UTC. \newline 2. Search for \textbf{Thiruvananthapuram International Airport (TRV)} to center the map on it. \newline 3. Draw a 450 km radius circle centered on the airport. \newline 4. Count the number of aircraft located within this circle. \newline  \newline \# RESULT FORMAT \newline  \newline ```json \newline \{ \newline  "count": <N> \newline \} \newline ```}
&
\begin{minipage}[t]{\linewidth}
\begin{BVerbatim}[fontsize=\scriptsize,breaklines=true,breakanywhere=true]
{
  "count": 4
}
\end{BVerbatim}
\end{minipage}
\\
\hline
\multicolumn{3}{|c|}{
\includegraphics[width=0.95\linewidth]{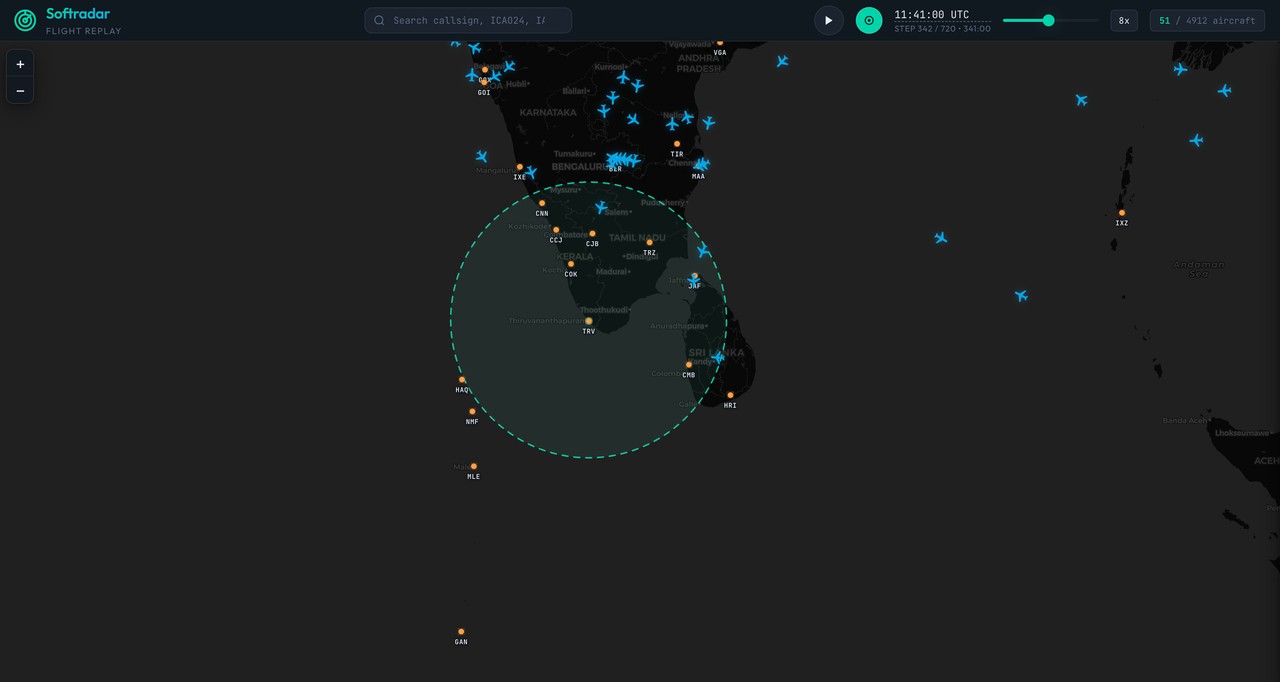}
} \\
\hline
\noalign{\vskip 6pt}
\hline
\cellcolor{orange!15}12 &
{\raggedright\scriptsize\ttfamily \# Identify the First Departure After a Given Time \newline  \newline \#\# GOAL \newline Find the first flight that departs from an airport after a specific time. \newline  \newline \#\# STEPS \newline 1. Set the playback time to 11:30 UTC. \newline 2. Zoom into \textbf{John F. Kennedy International Airport (JFK)} in New York. \newline 3. Observe departures from the airport. \newline 4. Identify the first flight that departs after this time. \newline 5. Report that flight's callsign. \newline  \newline \# RESULT FORMAT \newline  \newline ```json \newline \{ \newline  "callsign": "<callsign>" \newline \} \newline ```}
&
\begin{minipage}[t]{\linewidth}
\begin{BVerbatim}[fontsize=\scriptsize,breaklines=true,breakanywhere=true]
{
  "callsign": "AAL1535"
}
\end{BVerbatim}
\end{minipage}
\\
\hline
\multicolumn{3}{|c|}{
\includegraphics[width=0.95\linewidth]{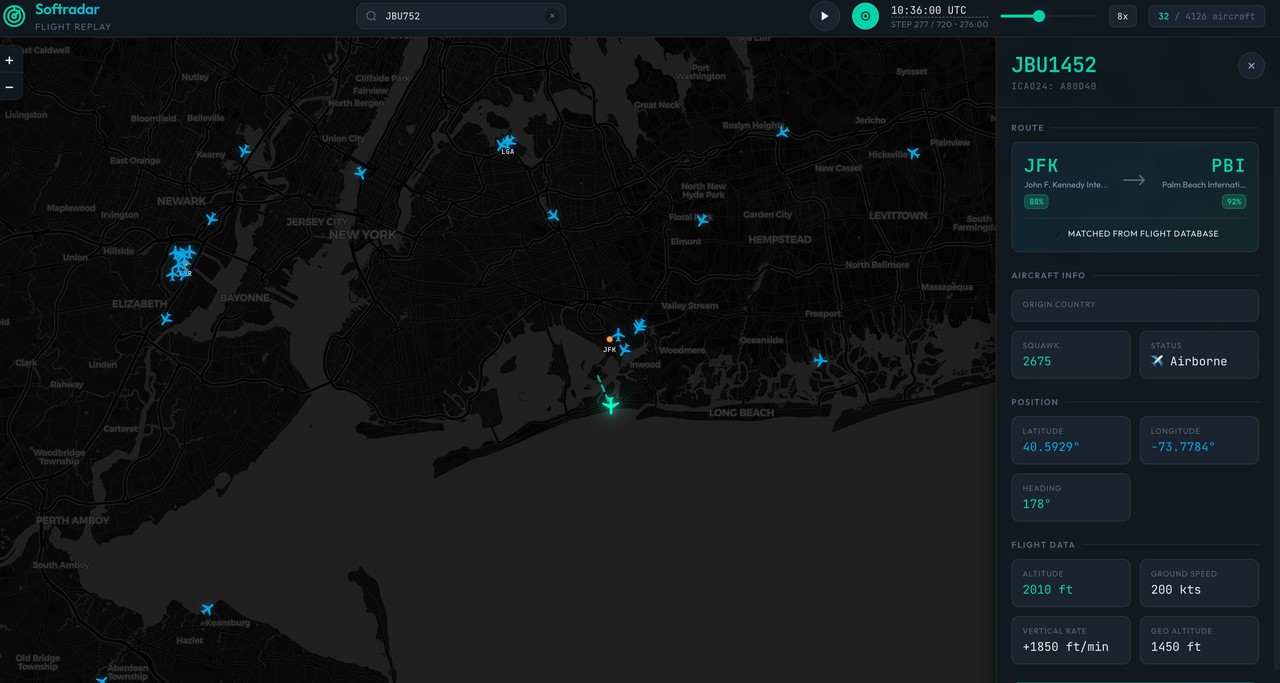}
} \\
\hline
\cellcolor{orange!15}13 &
{\raggedright\scriptsize\ttfamily \# Determine a Border Crossing Time (Canada to US) \newline  \newline \#\# GOAL \newline Track a flight and report the exact time it crosses the Canada–United States border. \newline  \newline \#\# STEPS \newline 1. Set the playback time to 16:00 UTC. \newline 2. Search for flight \textbf{SAS907} using the search bar. \newline 3. Locate the flight on the map and begin tracking it forward in time. \newline 4. Report the UTC time at which it crosses the Canada–United States border. \newline  \newline \# RESULT FORMAT \newline  \newline ```json \newline \{ \newline  "crossing\_time\_utc": "<HH:MM UTC>" \newline \} \newline ```}
&
\begin{minipage}[t]{\linewidth}
\begin{BVerbatim}[fontsize=\scriptsize,breaklines=true,breakanywhere=true]
{
  "crossing_time_utc": "16:35 UTC"
}
\end{BVerbatim}
\end{minipage}
\\
\hline
\multicolumn{3}{|c|}{
\includegraphics[width=0.95\linewidth]{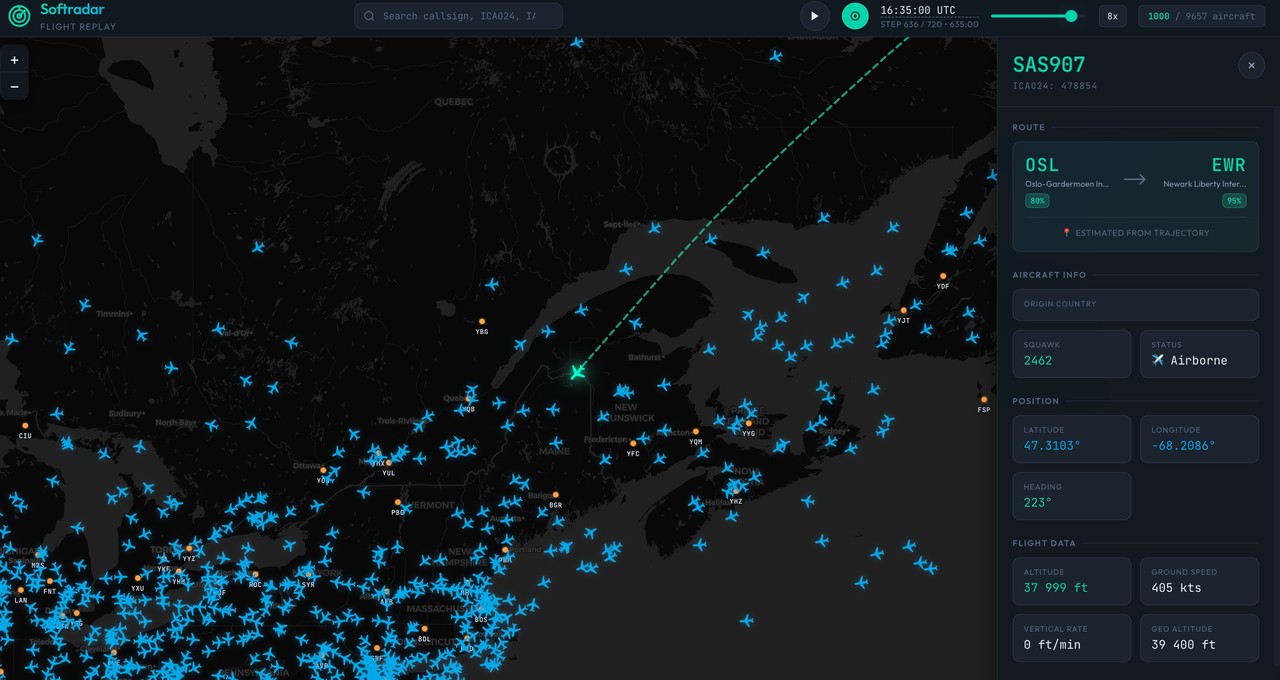}
} \\
\hline
\noalign{\vskip 6pt}
\hline
\cellcolor{orange!15}14 &
{\raggedright\scriptsize\ttfamily \# Identify the First Channel Crossing (UK to France) \newline  \newline \#\# GOAL \newline Find the first flight to cross from England into France over the English Channel after a given time. \newline  \newline \#\# STEPS \newline 1. Set the playback time to 16:08 UTC. \newline 2. Zoom into the narrowest part of the English Channel. \newline 3. Identify all flights that were over England at 16:08 UTC. \newline 4. Observe which of those flights is the first to cross into France. \newline 5. Report that flight's callsign. \newline  \newline \# RESULT FORMAT \newline  \newline ```json \newline \{ \newline  "callsign": "<callsign>" \newline \} \newline ```}
&
\begin{minipage}[t]{\linewidth}
\begin{BVerbatim}[fontsize=\scriptsize,breaklines=true,breakanywhere=true]
{
  "callsign": "TOM7KW"
}
\end{BVerbatim}
\end{minipage}
\\
\hline
\multicolumn{3}{|c|}{
\includegraphics[width=0.95\linewidth]{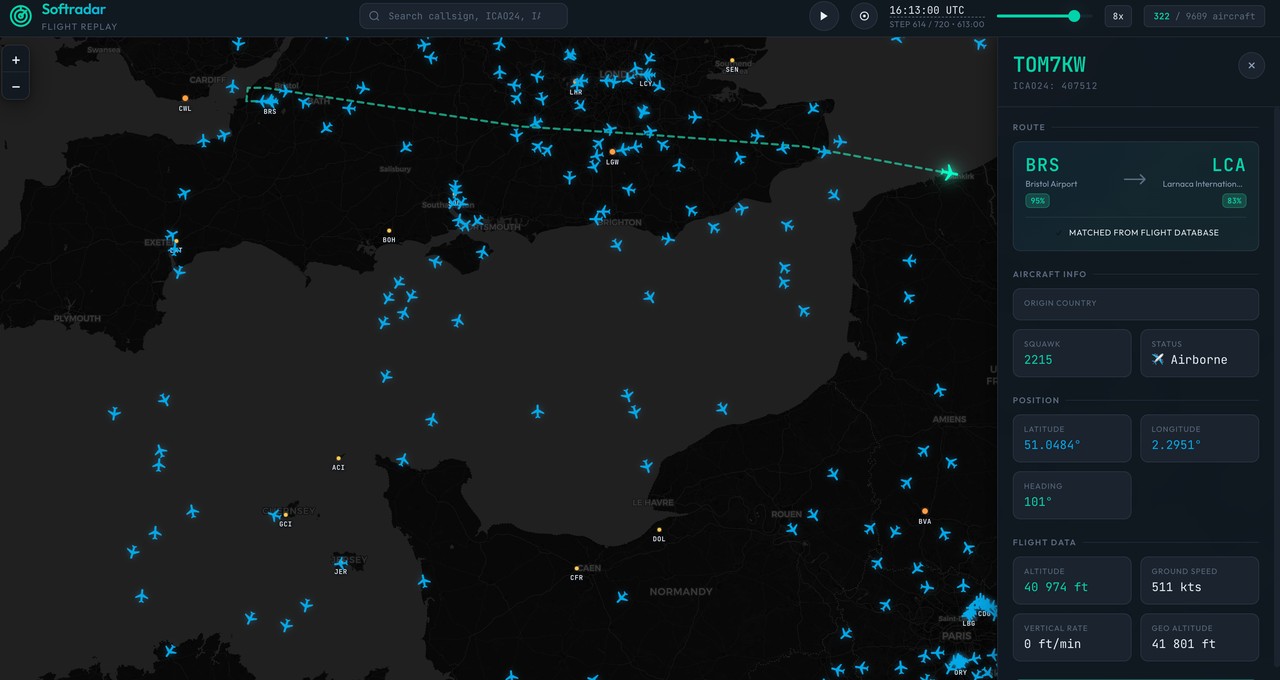}
} \\
\hline
\noalign{\vskip 6pt}
\hline
\cellcolor{orange!15}15 &
{\raggedright\scriptsize\ttfamily \# Find the Last Flight to Leave a Circle \newline  \newline \#\# GOAL \newline Determine which flight is the last to exit a radius around a given airport. \newline  \newline \#\# STEPS \newline 1. Set the playback time to 16:00 UTC. \newline 2. Navigate to \textbf{Karlovy Vary Airport}. \newline 3. Draw a 40 km radius circle around the airport. \newline 4. Identify all flights currently inside the circle. \newline 5. Track them forward in time and determine which flight leaves the circle last. \newline 6. Report that flight's callsign. \newline  \newline \# RESULT FORMAT \newline  \newline ```json \newline \{ \newline  "callsign": "<callsign>" \newline \} \newline ```}
&
\begin{minipage}[t]{\linewidth}
\begin{BVerbatim}[fontsize=\scriptsize,breaklines=true,breakanywhere=true]
{
  "callsign": "TVS5AD"
}
\end{BVerbatim}
\end{minipage}
\\
\hline
\multicolumn{3}{|c|}{
\includegraphics[width=0.95\linewidth]{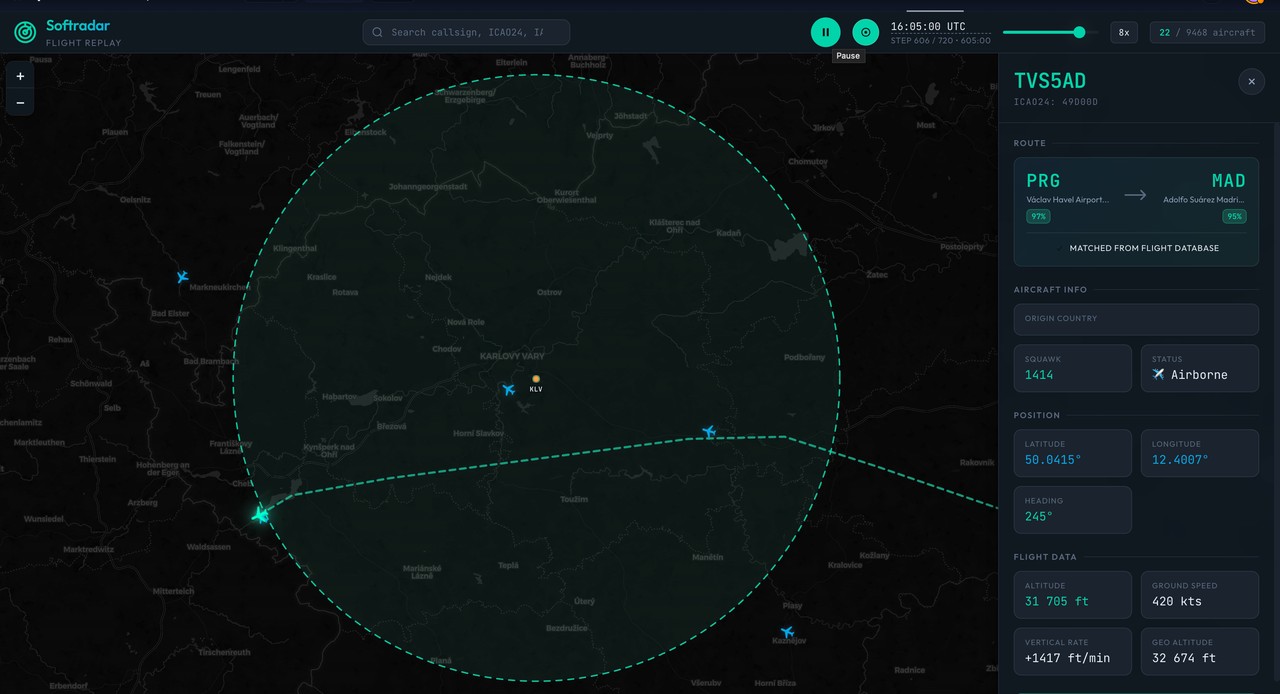}
} \\
\hline
\noalign{\vskip 6pt}
\hline
\cellcolor{orange!15}16 &
{\raggedright\scriptsize\ttfamily \# Count Airborne Aircraft Near an Airport \newline  \newline \#\# GOAL \newline Count how many aircraft within a radius of a given airport have an "Airborne" status. \newline  \newline \#\# STEPS \newline 1. Set the playback time to 16:00 UTC. \newline 2. Search for airport \textbf{ORY} to center the map on it. \newline 3. Zoom in exactly 3 steps. \newline 4. Within a 20 km radius of ORY, open the details of each aircraft in view. \newline 5. Count how many have an "Airborne" status at 16:00 UTC and collect their callsigns. \newline  \newline \# RESULT FORMAT \newline  \newline ```json \newline \{ \newline  "count": <N>, \newline  "callsigns": ["<CS1>", "<CS2>", "..."] \newline \} \newline ```}
&
\begin{minipage}[t]{\linewidth}
\begin{BVerbatim}[fontsize=\scriptsize,breaklines=true,breakanywhere=true]
{
  "count": 5,
  "callsigns": [
    "CTM3813",
    "TVF3828",
    "EJU54MG",
    "TVF3124",
    "FMY8055"
  ]
}
\end{BVerbatim}
\end{minipage}
\\
\hline
\multicolumn{3}{|c|}{
\includegraphics[width=0.95\linewidth]{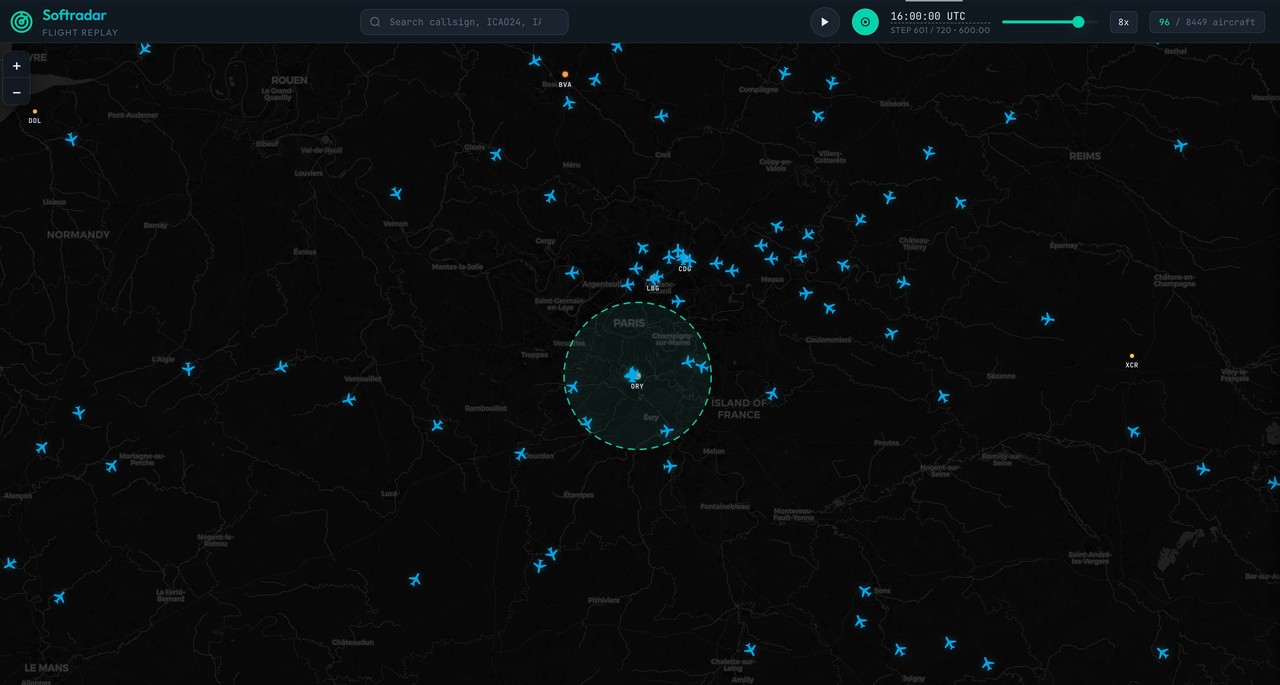}
} \\
\hline
\noalign{\vskip 6pt}
\hline
\cellcolor{orange!15}17 &
{\raggedright\scriptsize\ttfamily \# Check Whether a Flight Path Enters a Circle \newline \newline\#\# GOAL \newline Determine whether a specific flight's path ever enters a defined radius around an airport. \newline \newline\#\# STEPS \newline1. Set the playback time to 11:15 UTC. \newline2. Search for \textbf{Missoula Montana Airport (MSO)} to center the map on it. \newline3. Draw a 500 km radius circle centered on the airport. \newline4. Search for flight \textbf{ASA1217} and identify the flight closest to \textbf{ASA1217} at that moment whose track crosses the 500 km radius circle. \newline5. Report the callsign of that closest flight. \newline \newline\# RESULT FORMAT \newline \newline```json \newline\{ \newline  "callsign": "<callsign>" \newline\} \newline```}
&
\begin{minipage}[t]{\linewidth}
\begin{BVerbatim}[fontsize=\scriptsize,breaklines=true,breakanywhere=true]
{"callsign": N213AL}
\end{BVerbatim}
\end{minipage}
\\
\hline
\multicolumn{3}{|c|}{
\includegraphics[width=0.95\linewidth]{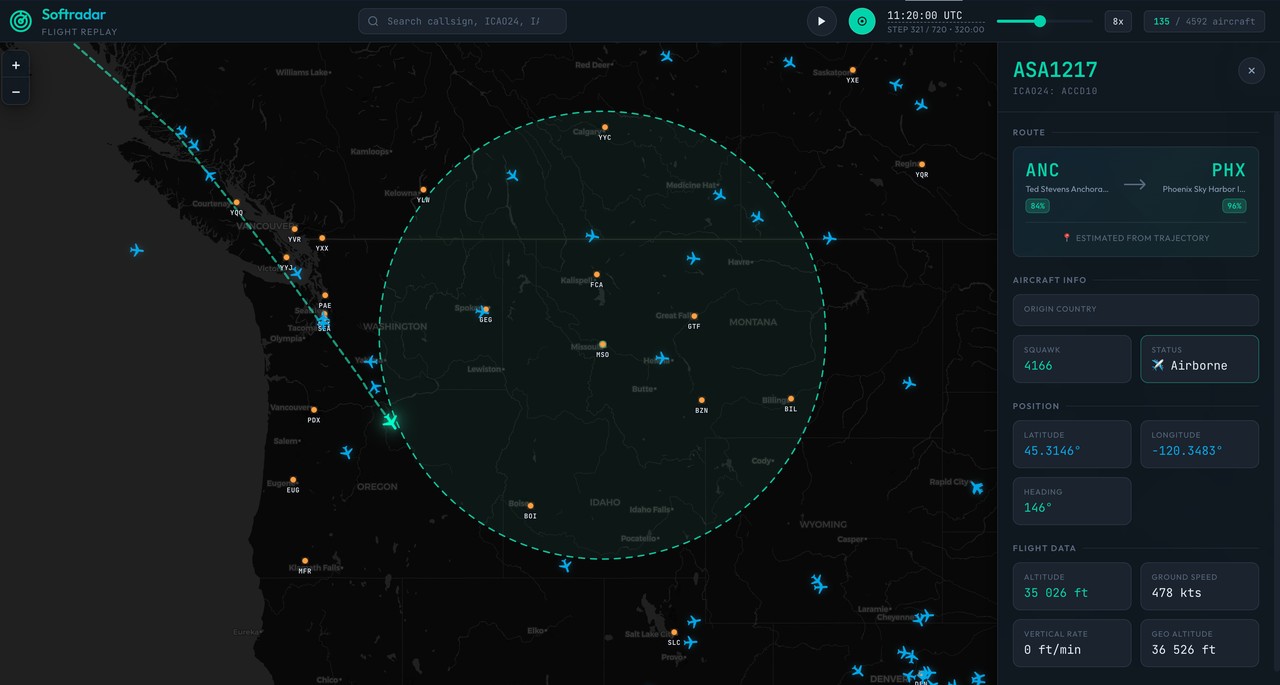}
} \\
\hline
\noalign{\vskip 6pt}
\hline
\cellcolor{orange!15}18 &
{\raggedright\scriptsize\ttfamily \# Detect a Border Crossing (Poland to Germany) \newline  \newline \#\# GOAL \newline Find the first aircraft whose track crosses from Poland into Germany after a given time. \newline  \newline \#\# STEPS \newline 1. Set the playback time to 16:00 UTC. \newline 2. Navigate to the Germany–Poland border region. \newline 3. Observe aircraft tracks crossing the border. \newline 4. Identify the first aircraft that crosses from Poland into Germany (not the reverse direction) and does not re-enter Polish airspace afterward. \newline 5. Confirm the crossing direction by comparing two successive positions — one before and one after the border. \newline 6. Report that aircraft's callsign. \newline  \newline \# RESULT FORMAT \newline  \newline ```json \newline \{ \newline  "callsign": "<callsign>" \newline \} \newline ```}
&
\begin{minipage}[t]{\linewidth}
\begin{BVerbatim}[fontsize=\scriptsize,breaklines=true,breakanywhere=true]
{
  "callsign": "LOT335"
}
\end{BVerbatim}
\end{minipage}
\\
\hline
\multicolumn{3}{|c|}{
\includegraphics[width=0.95\linewidth]{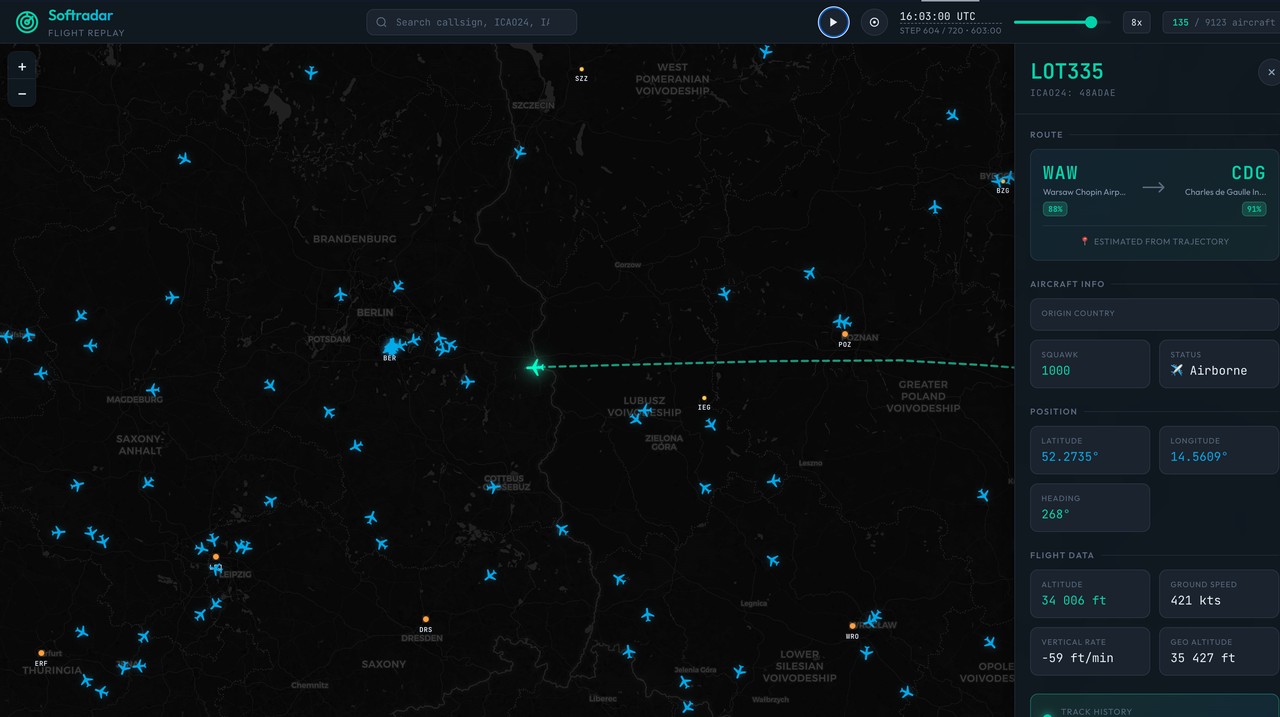}
} \\
\hline
\noalign{\vskip 6pt}
\hline
\cellcolor{red!15}19 &
{\raggedright\scriptsize\ttfamily \# Count Airports Passed Along a Flight Path \newline  \newline \#\# GOAL \newline Track a flight over a period and count how many airports it passes within close proximity. \newline  \newline \#\# STEPS \newline 1. Set the playback time to 17:15 UTC. \newline 2. Search for flight \textbf{SAS907} using the search bar and locate it on the map. \newline 3. Track the flight for the next 30 minutes of playback time. \newline 4. Count how many airports the flight passes within a 30 km radius and list their airport codes. \newline  \newline \# RESULT FORMAT \newline  \newline ```json \newline \{ \newline  "count": <N>, \newline  "airports": ["<airport1>", "<airport2>", "..."] \newline \} \newline ```}
&
\begin{minipage}[t]{\linewidth}
\begin{BVerbatim}[fontsize=\scriptsize,breaklines=true,breakanywhere=true]
{
  "count": 3,
  "airports": [
    "RUT",
    "ALB",
    "SWF"
  ]
}
\end{BVerbatim}
\end{minipage}
\\
\hline
\multicolumn{3}{|c|}{
\includegraphics[width=0.95\linewidth]{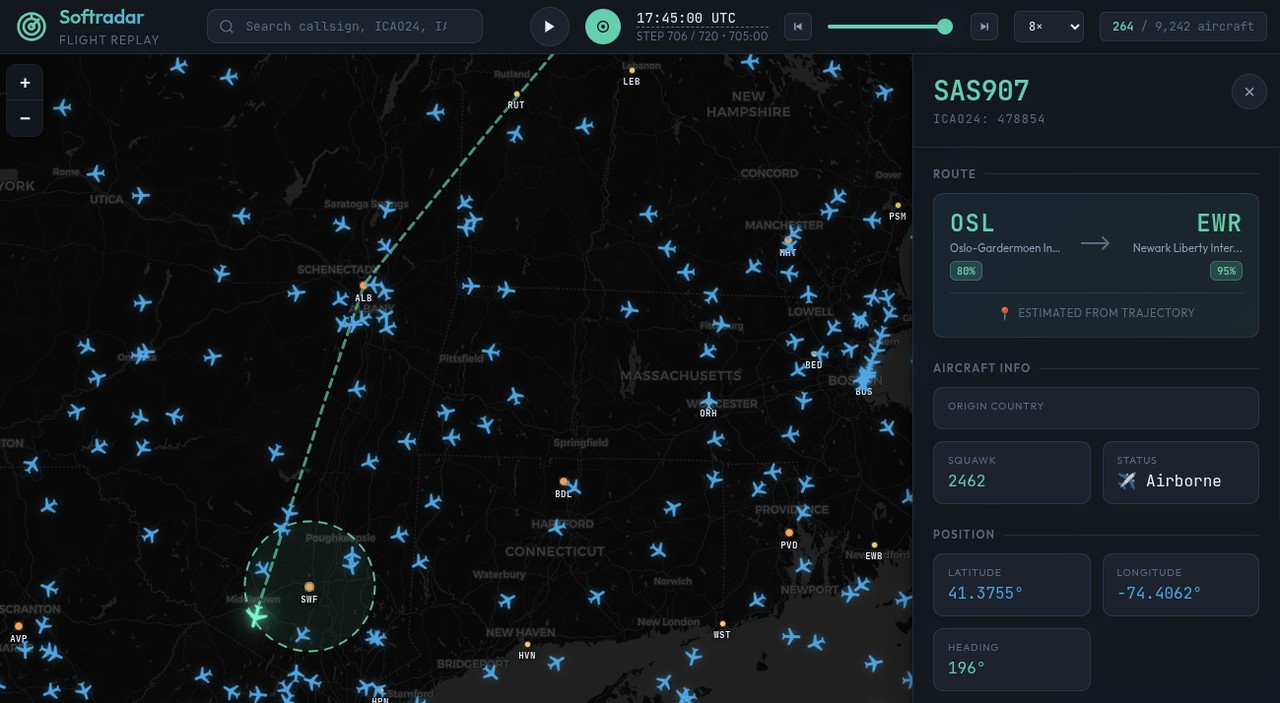}
} \\
\hline
\noalign{\vskip 6pt}
\hline
\cellcolor{red!15}20 &
{\raggedright\scriptsize\ttfamily \# Detect the First Stable East-to-West Ordering of Three Flights \newline \newline\#\# GOAL \newline Determine the first time three aircraft reach a specified west-to-east ordering and keep it for a sustained period. \newline \newline\#\# STEPS \newline1. Set the playback time to 16:20 UTC. \newline2. Search for flights \textbf{AFR24NL}, \textbf{RYR61XN}, and \textbf{BEL0GA} and track them forward in time. \newline3. Find the first UTC minute when the west-to-east order of the flights becomes \textbf{BEL0GA} -> \textbf{AFR24NL} -> \textbf{RYR61XN}. \newline4. Report that UTC time. \newline \newline\# RESULT FORMAT \newline \newline```json \newline\{ \newline  "order\_start\_utc": "<HH:MM UTC>" \newline\} \newline```}
&
\begin{minipage}[t]{\linewidth}
\begin{BVerbatim}[fontsize=\scriptsize,breaklines=true,breakanywhere=true]
{
  "order_start_utc": "16:27 UTC"
}
\end{BVerbatim}
\end{minipage}
\\
\hline
\multicolumn{3}{|c|}{
\includegraphics[width=0.95\linewidth]{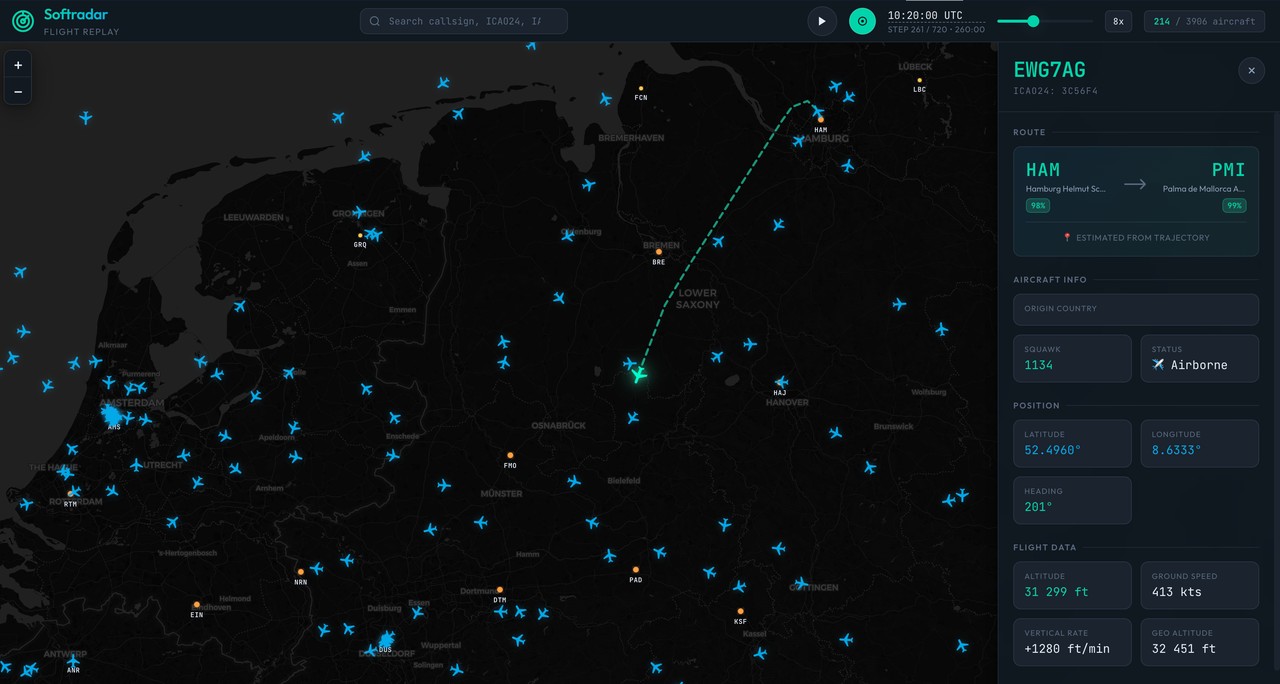}
} \\
\hline
\noalign{\vskip 6pt}
\hline
\cellcolor{red!15}21 &
{\raggedright\scriptsize\ttfamily \# Count Country Borders Crossed from Track History \newline  \newline \#\# GOAL \newline Determine how many country borders a flight has crossed based on its visible track history. \newline  \newline \#\# STEPS \newline 1. Set the playback time to 16:00 UTC. \newline 2. Search for flight \textbf{MSR726} using the search bar. \newline 3. Examine the flight's visible track history (trajectory) as shown at the current moment. \newline 4. Count how many country borders the track crosses and list each country entered. \newline  \newline \# RESULT FORMAT \newline  \newline ```json \newline \{ \newline  "num\_country\_borders": <N>, \newline  "countries": ["<country1>", "<country2>", "..."] \newline \} \newline ```}
&
\begin{minipage}[t]{\linewidth}
\begin{BVerbatim}[fontsize=\scriptsize,breaklines=true,breakanywhere=true]
{
  "num_country_borders": 4,
  "countries": [
    "Belgium",
    "Germany",
    "Austria",
    "Slovenia"
  ]
}
\end{BVerbatim}
\end{minipage}
\\
\hline
\multicolumn{3}{|c|}{
\includegraphics[width=0.95\linewidth]{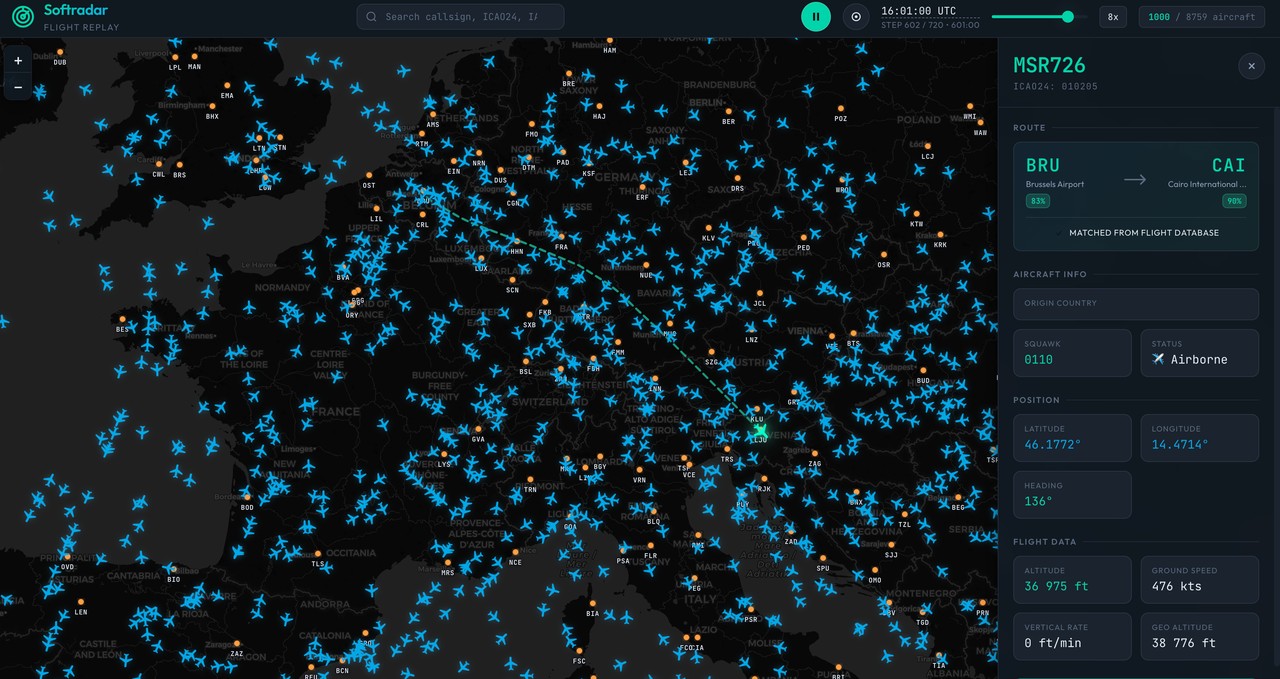}
} \\
\hline
\noalign{\vskip 6pt}
\hline
\cellcolor{red!15}22 &
{\raggedright\scriptsize\ttfamily \# Find the Closest Aircraft and Its Overtake Time \newline  \newline \#\# GOAL \newline Locate the aircraft nearest to a given flight and determine when it overtakes that flight. \newline  \newline \#\# STEPS \newline 1. Set the playback time to 07:50 UTC. \newline 2. Search for flight \textbf{VOE25HV} using the search bar. \newline 3. Identify the aircraft currently closest to VOE25HV. \newline 4. Track both aircraft forward in time. \newline 5. Report the closest aircraft's callsign and the UTC time at which it overtakes (passes ahead of) VOE25HV. \newline  \newline \# RESULT FORMAT \newline  \newline ```json \newline \{ \newline  "closest\_callsign": "<callsign>", \newline  "overtake\_time\_utc": "<HH:MM UTC>" \newline \} \newline ```}
&
\begin{minipage}[t]{\linewidth}
\begin{BVerbatim}[fontsize=\scriptsize,breaklines=true,breakanywhere=true]
{
  "closest_callsign": "N656FN",
  "overtake_time_utc": "07:54 UTC"
}
\end{BVerbatim}
\end{minipage}
\\
\hline
\multicolumn{3}{|c|}{
\includegraphics[width=0.95\linewidth]{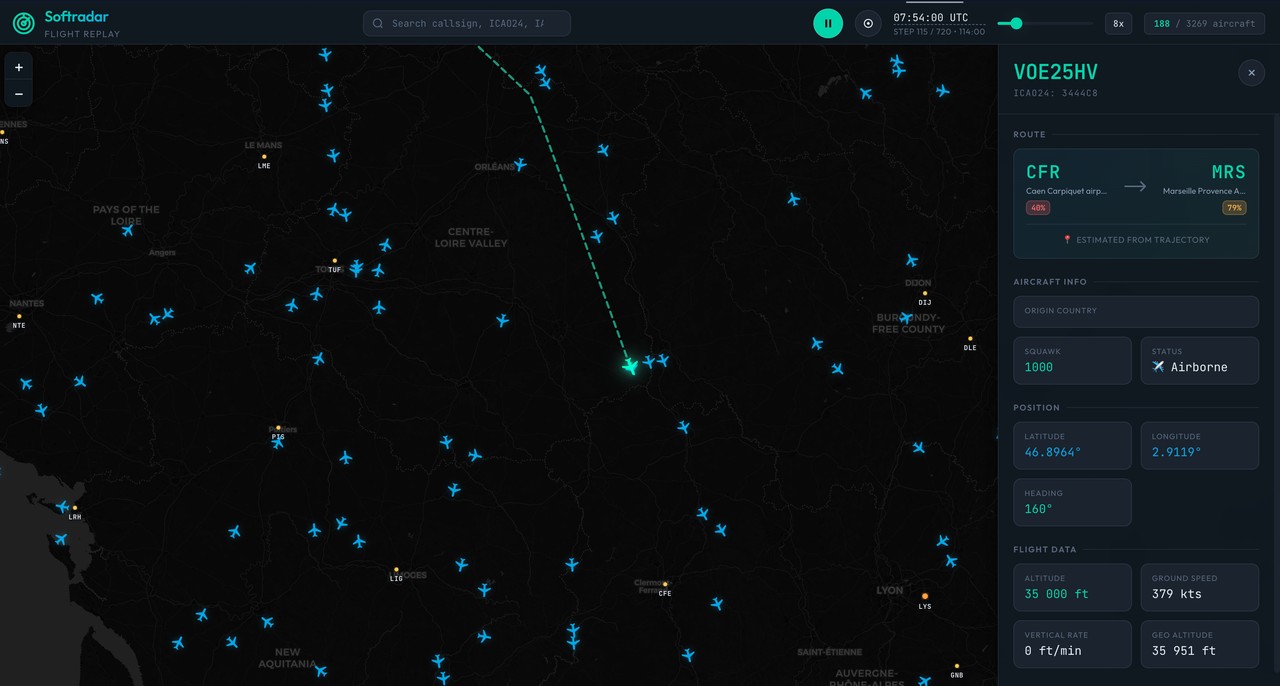}
} \\
\hline
\noalign{\vskip 6pt}
\hline
\cellcolor{red!15}23 &
{\raggedright\scriptsize\ttfamily \# Find the Closest Aircraft Pair Near an Airport \newline  \newline \#\# GOAL \newline Among airborne aircraft near Zurich, find the two that are closest to each other. \newline  \newline \#\# STEPS \newline 1. Set the playback time to 16:00 UTC. \newline 2. Zoom into Zurich, Switzerland. \newline 3. Identify all aircraft within 40 km of the airport at the time. Exclude any aircraft on the ground in the airport area. \newline 4. Determine which two airborne aircraft are closest to each other. \newline 5. Report both callsigns. List the aircraft with the higher geographical altitude first. \newline  \newline \# RESULT FORMAT \newline  \newline ```json \newline \{ \newline  "aircraft\_1\_callsign": "<callsign>", \newline  "aircraft\_2\_callsign": "<callsign>" \newline \} \newline ```}
&
\begin{minipage}[t]{\linewidth}
\begin{BVerbatim}[fontsize=\scriptsize,breaklines=true,breakanywhere=true]
{
  "aircraft_1_callsign": "AFR33SD",
  "aircraft_2_callsign": "SWR60L"
}
\end{BVerbatim}
\end{minipage}
\\
\hline
\multicolumn{3}{|c|}{
\includegraphics[width=0.95\linewidth]{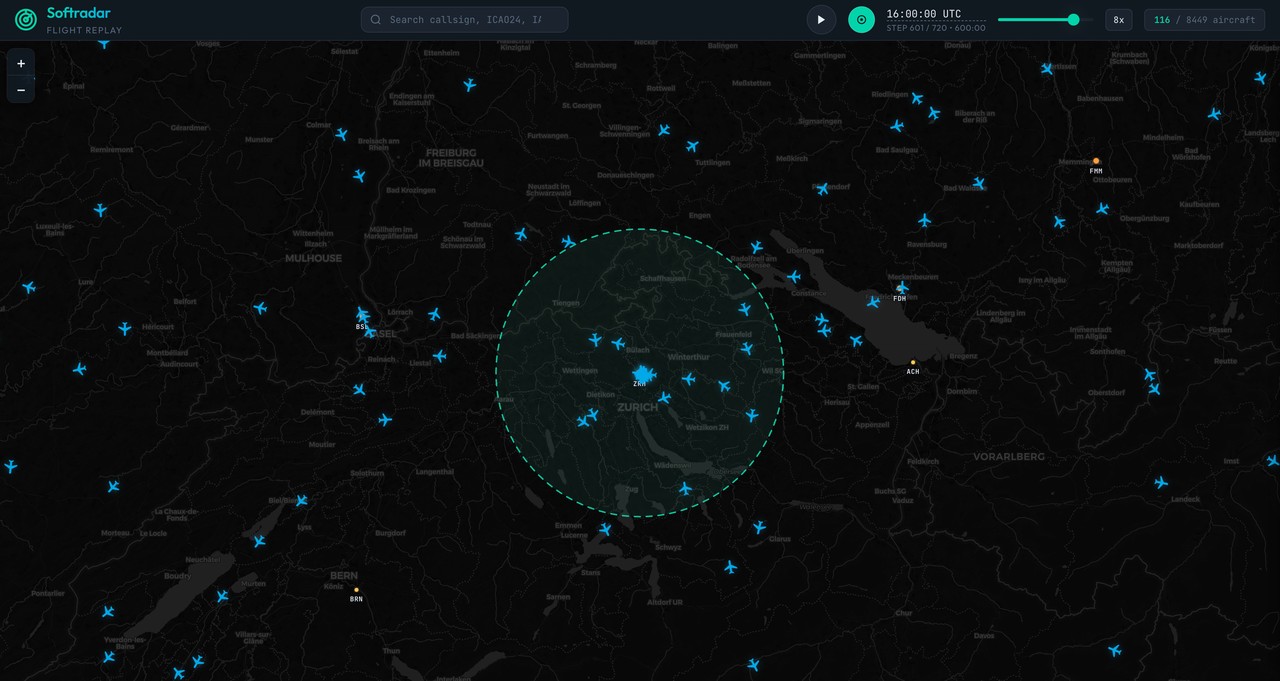}
} \\
\hline
\noalign{\vskip 6pt}
\hline
\cellcolor{red!15}24 &
{\raggedright\scriptsize\ttfamily \# Detect the First Landing and First Departure from an Airport \newline \newline\#\# GOAL \newline Monitor one airport over a short time window and identify the first aircraft that lands and the first aircraft that departs after the start time. \newline \newline\#\# STEPS \newline1. Set the playback time to 10:05 UTC. \newline2. Find Vienna Airport (VIE). \newline3. Observe aircraft activity for the next 10 minutes of playback time. \newline4. Identify: \newline   - the first aircraft that \textbf{departs} from the airport after the start time. \newline   - the aircraft that is \textbf{second} to \textbf{land} at the airport after the start time. \newline5. Report both callsigns and their UTC times. \newline \newline\#\# RESULT FORMAT \newline \newline```json \newline\{ \newline  "second\_landing\_callsign": "<callsign>", \newline  "first\_departure\_callsign": "<callsign>" \newline\} \newline```}
&
\begin{minipage}[t]{\linewidth}
\begin{BVerbatim}[fontsize=\scriptsize,breaklines=true,breakanywhere=true]
{
  "second_landing_callsign": "AUA87",
  "first_departure_callsign": "AUA7F",
}
\end{BVerbatim}
\end{minipage}
\\
\hline
\multicolumn{3}{|c|}{
\includegraphics[width=0.95\linewidth]{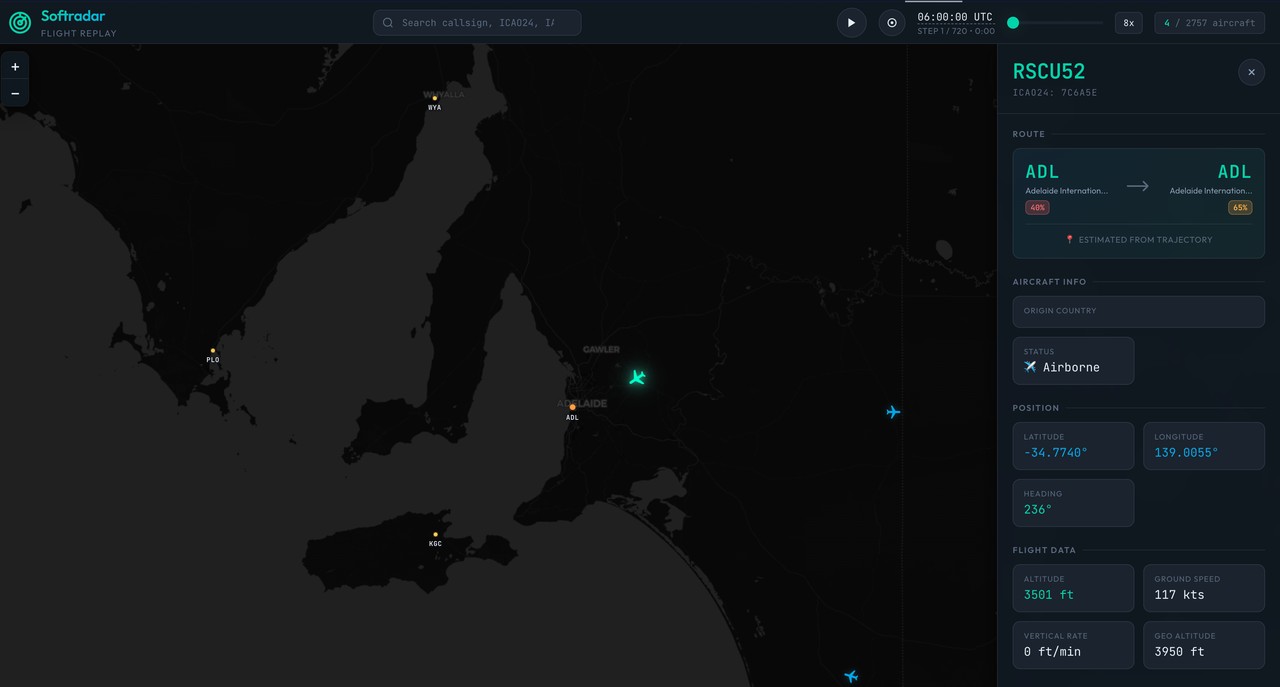}
} \\
\hline
\noalign{\vskip 6pt}
\hline
\cellcolor{red!15}25 &
{\raggedright\scriptsize\ttfamily \# Find the Previous Departure Before a Target Flight \newline \newline\#\# GOAL \newline Monitor departures from \textbf{Manchester Airport} within a defined radius and identify which flight departed immediately \textbf{before} a given target flight. \newline \newline\#\# STEPS \newline1. Set the playback time to 14:22 UTC. \newline2. Search for \textbf{Manchester Airport (MAN)}. \newline3. Draw a \textbf{10 km} radius circle centered on the airport. \newline4. Monitor departures from the airport. \newline5. Locate the departure of flight \textbf{EZY14GQ}. \newline6. Determine which flight \textbf{departed from Manchester Airport immediately before} \textbf{EZY14GQ}. \newline7. Report: \newline   - the previous flight callsign, \newline   - the UTC time when the previous flight left the radius circle. \newline \newline\#\# RESULT FORMAT \newline \newline```json \newline\{ \newline  "previous\_callsign": "<callsign>", \newline  "previous\_left\_time": "<HH:MM UTC>" \newline\} \newline```}
&
\begin{minipage}[t]{\linewidth}
\begin{BVerbatim}[fontsize=\scriptsize,breaklines=true,breakanywhere=true]
{
  "previous_callsign": "RUK6797",
  "previous_left_time": "14:25 UTC",
}
\end{BVerbatim}
\end{minipage}
\\
\hline
\multicolumn{3}{|c|}{
\includegraphics[width=0.95\linewidth]{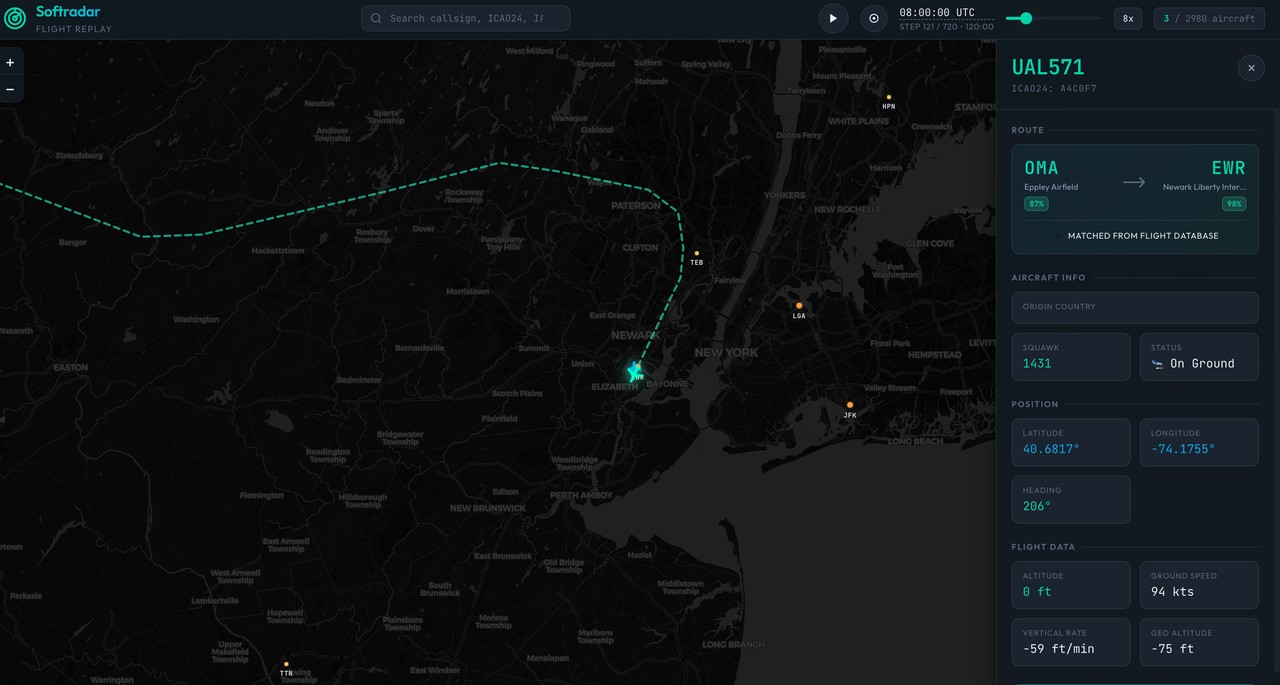}
} \\
\hline
\noalign{\vskip 6pt}
\hline
\cellcolor{red!15}26 &
{\raggedright\scriptsize\ttfamily \# Identify the Last Remaining Flight Inside a Zone \newline \newline\#\# GOAL \newline Among several flights near an airport, determine which one is the last to remain inside a monitored radius. \newline \newline\#\# STEPS \newline1. Set the playback time to 16:21 UTC. \newline2. Search for airport \textbf{XCR} and draw a 50 km radius circle around it. \newline3. Locate flights \textbf{SWR441R}, \textbf{UAE4PV}, \textbf{THY6FE}, and \textbf{S5BBC} in the area. \newline4. Track all four flights forward in time. \newline5. Determine the moment when all of those flights left the circle. \newline6. Report: \newline   - the UTC time when all 4 initial flights left the circle, \newline   - the callsign of the first new flight entering the circle, \newline   - the first UTC time when the circle contains 4 new flights. \newline \newline\# RESULT FORMAT \newline \newline```json \newline\{ \newline  "initial\_left\_utc": "<HH:MM UTC>", \newline  "first\_new\_callsign": "<callsign>", \newline  "first\_time\_with\_four\_new\_utc": "<HH:MM UTC>" \newline\} \newline```}
&
\begin{minipage}[t]{\linewidth}
\begin{BVerbatim}[fontsize=\scriptsize,breaklines=true,breakanywhere=true]
{
  "initial_left_utc": "16:25 UTC",
  "first_new_callsign": "EJU79XU",
  "first_time_with_four_new_utc": 
      "16:28 UTC"
}
\end{BVerbatim}
\end{minipage}
\\
\hline
\multicolumn{3}{|c|}{
\includegraphics[width=0.95\linewidth]{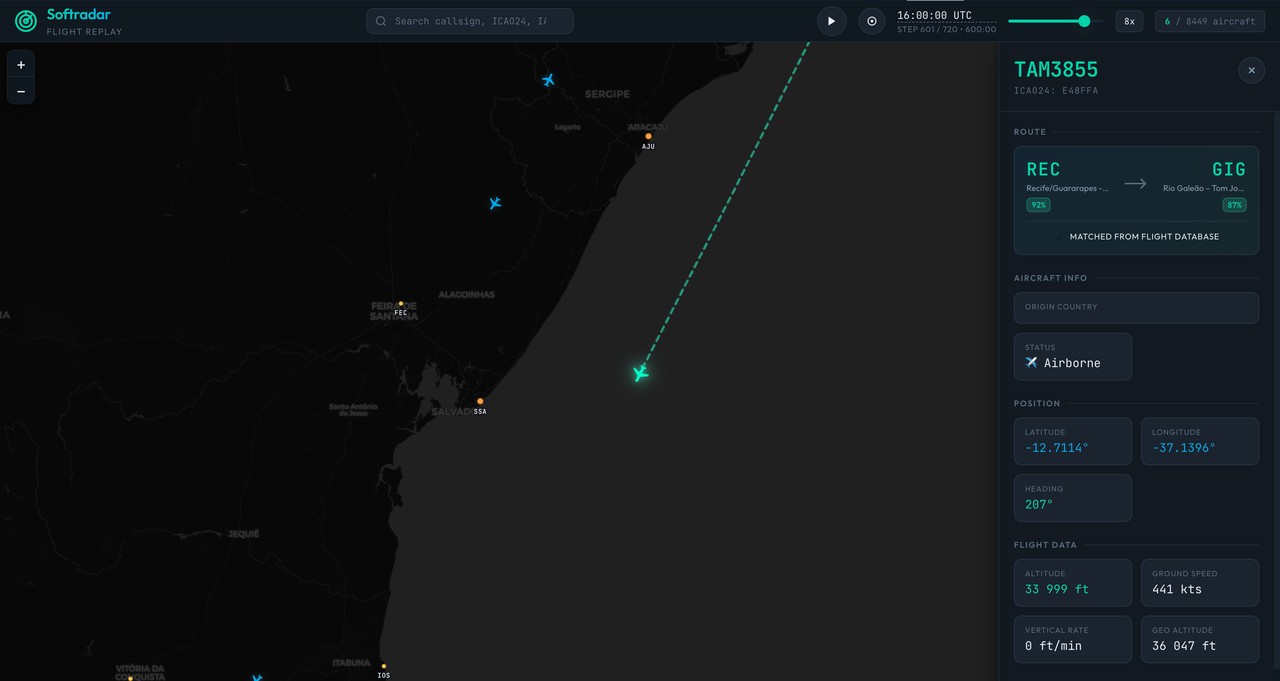}
} \\
\hline
\noalign{\vskip 6pt}
\hline
\cellcolor{red!15}27 &
{\raggedright\scriptsize\ttfamily \# Track a Two-Stage Zone Transition for Two Flights \newline \newline\#\# GOAL \newline Determine a multi-step temporal relationship between two flights around the same airport. \newline \newline\#\# STEPS \newline1. Set the playback time to 11:43 UTC. \newline2. Search for the target airport, \textbf{Ostend-Bruges International Airport (OST)}. \newline3. Identify the two approaching flights that will be the first new flights to enter a 35 km radius zone around the airport. \newline4. Find the first UTC time when \textbf{exactly one} of the two flights enters the 35 km zone around the airport. \newline5. Continue tracking until \textbf{both} flights are outside the circle. \newline6. Report: \newline   - the first time either of the two planes enters the zone, \newline   - which flight is the second to be inside the zone, \newline   - the UTC time when both flights finally exit the zone. \newline \newline\# RESULT FORMAT \newline \newline```json \newline\{ \newline  "any\_enters\_utc": "<HH:MM UTC>", \newline  "last\_inside\_callsign": "<callsign>", \newline  "final\_exit\_time\_utc": "<HH:MM UTC>" \newline\} \newline```}
&
\begin{minipage}[t]{\linewidth}
\begin{BVerbatim}[fontsize=\scriptsize,breaklines=true,breakanywhere=true]
{
  "any_enters_utc": "11:44 UTC",
  "last_inside_callsign": "RYR496K",
  "final_exit_time_utc": "11:50 UTC"
}
\end{BVerbatim}
\end{minipage}
\\
\hline
\multicolumn{3}{|c|}{
\includegraphics[width=0.95\linewidth]{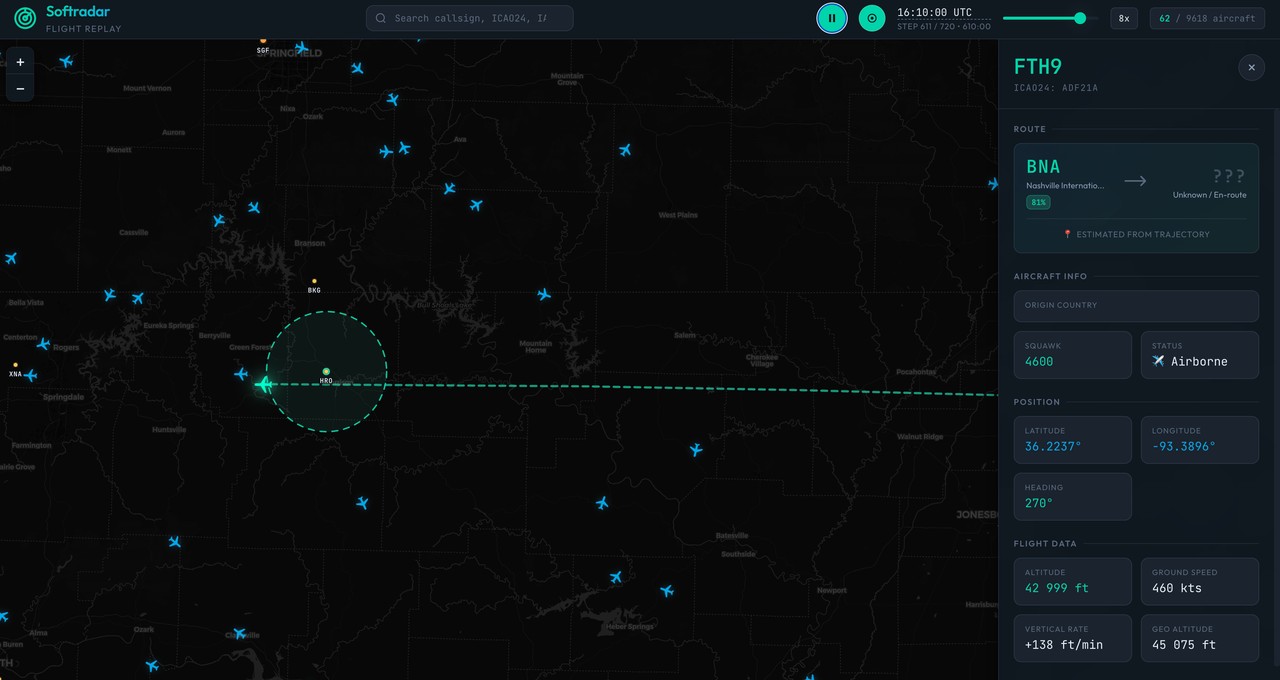}
} \\
\hline
\noalign{\vskip 6pt}
\hline
\end{longtable}
\normalsize

%% file: applications_testcases/testcases_circuit.tex
\scriptsize
\renewcommand{\arraystretch}{1.18}
\setlength{\LTleft}{0pt}
\setlength{\LTright}{0pt}
\setlength{\tabcolsep}{6pt}

\newcommand{\circuitgt}[2]{%
  \begingroup
  \centering
  \vspace{3pt}
  \includegraphics[width=1.0\linewidth]{images/circuit_gt_pngs/tc_circuit_#1.png}\par
  \def\temparg{#2}%
  \ifx\temparg\empty\else
    \vspace{4pt}
    \hrule height 0.4pt\relax
    \vspace{4pt}
    {\raggedright\ttfamily\tiny #2\par}
  \fi
  \endgroup
}


\begin{longtable}{|>{\vspace{0pt}}p{0.03\linewidth}|>{\vspace{0pt}}p{0.50\linewidth}|>{\vspace{0pt}\centering\arraybackslash}p{0.39\linewidth}|}
\hline
\textbf{No.} & \textbf{Task} & \textbf{Ground truth} \\
\hline
\endfirsthead
\hline
\textbf{No.} & \textbf{Task} & \textbf{Ground truth} \\
\hline
\endhead
\hline
\endfoot
\cellcolor{green!15}1 & {\scriptsize\ttfamily \# Build a NOT Gate Using a NAND Gate \newline
 \newline
\#\# GOAL \newline
Implement a logical NOT function using only a universal NAND gate. \newline
 \newline
\#\# STEPS \newline
 \newline
1. Add the following components: \newline
- One input switch: A \newline
- One output switch: Y \newline
- Logical gates as needed \newline
 \newline
2. Wiring: \newline
- Make a NOT gate using only NAND gate. \newline
 \newline
3. Define output: \newline
- The output of the scheme must be the circuit output Y. \newline
 \newline
\# RESULT FORMAT \newline
 \newline
```json \newline
\{"answer": "done"\} \newline
```} &
\circuitgt{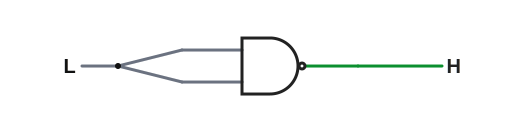}{} \\
\hline
\cellcolor{green!15}2 & {\scriptsize\ttfamily \# Build a 3-Input Odd-Parity Circuit \newline
 \newline
\#\# GOAL \newline
Construct a circuit with three binary inputs (A, B, C) and one output (Y) such that: \newline
- Y = 1 when an odd number of inputs are 1 \newline
- Y = 0 when an even number of inputs are 1 \newline
 \newline
\#\# STEPS \newline
 \newline
1. Add the following components: \newline
- Three input switches: A, B, C \newline
- Two XOR gates \newline
- One output switch: Y \newline
 \newline
2. Wiring: \newline
- Connect A and B to the inputs of the first XOR gate. \newline
- Connect the output of the first XOR gate and C to the inputs of the second XOR gate. \newline
 \newline
3. Define output: \newline
- The output of the second XOR gate must be the circuit output Y. \newline
 \newline
\# RESULT FORMAT \newline
 \newline
```json \newline
\{"answer": "done"\} \newline
```} &
\circuitgt{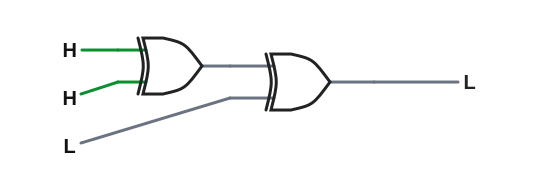}{} \\
\hline
\cellcolor{green!15}3 & {\scriptsize\ttfamily \# Build a NAND Gate From an AND Gate and a NOT Gate \newline
 \newline
\#\# GOAL \newline
Implement the NAND function without using a NAND gate, i.e. Y = NOT (A AND B). \newline
 \newline
\#\# STEPS \newline
 \newline
1. Add the following components: \newline
 - Two input switches: A, B \newline
 - One AND gate \newline
 - One NOT gate \newline
 - One output switch: Y \newline
 \newline
2. Wiring: \newline
 - Connect A and B to the inputs of the AND gate. \newline
 - Connect the output of the AND gate to the input of the NOT gate. \newline
 \newline
3. Define output: \newline
 - The output of the NOT gate must be the circuit output Y. \newline
 \newline
\# RESULT FORMAT \newline
 \newline
```json \newline
\{"answer": "done"\} \newline
```} &
\circuitgt{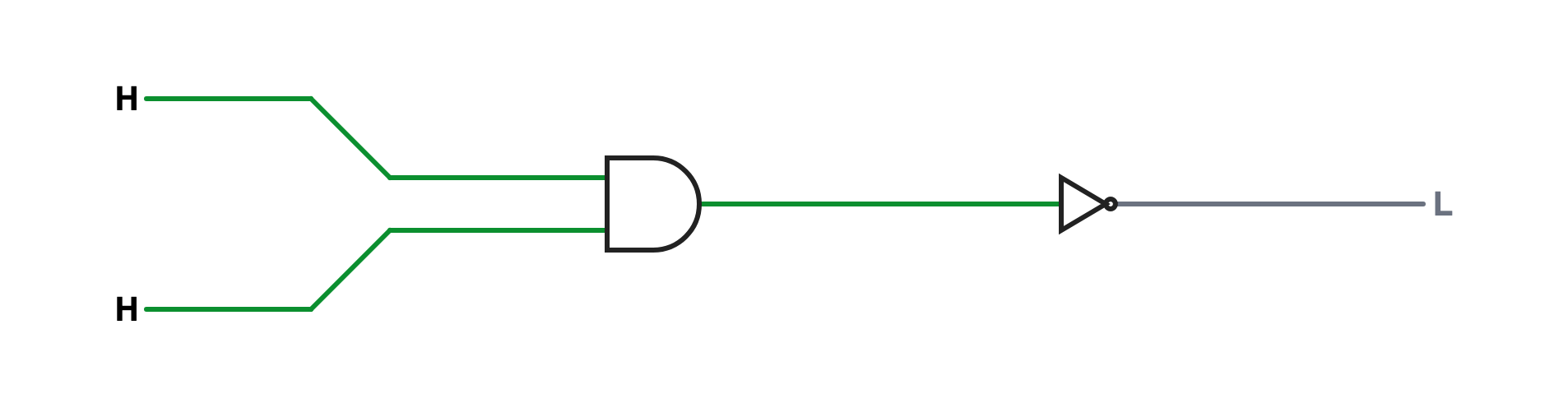}{} \\
\hline
\cellcolor{green!15}4 & {\scriptsize\ttfamily \# Build an AND Gate With an Override Input \newline
 \newline
\#\# GOAL \newline
Construct a circuit with two data inputs (A, B) and one override input (C) such that Y = (A AND B) OR C — the output is high when both A and B are high, and is also forced high whenever C is high. \newline
 \newline
\#\# STEPS \newline
 \newline
1. Add the following components: \newline
 - Three input switches: A, B, C \newline
 - One AND gate \newline
 - One OR gate \newline
 - One output switch: Y \newline
 \newline
2. Wiring: \newline
 - Connect A and B to the inputs of the AND gate. \newline
 - Connect the output of the AND gate and C to the inputs of the OR gate. \newline
 \newline
3. Define output: \newline
 - The output of the OR gate must be the circuit output Y. \newline
 \newline
\# RESULT FORMAT \newline
 \newline
```json \newline
\{"answer": "done"\} \newline
```} &
\circuitgt{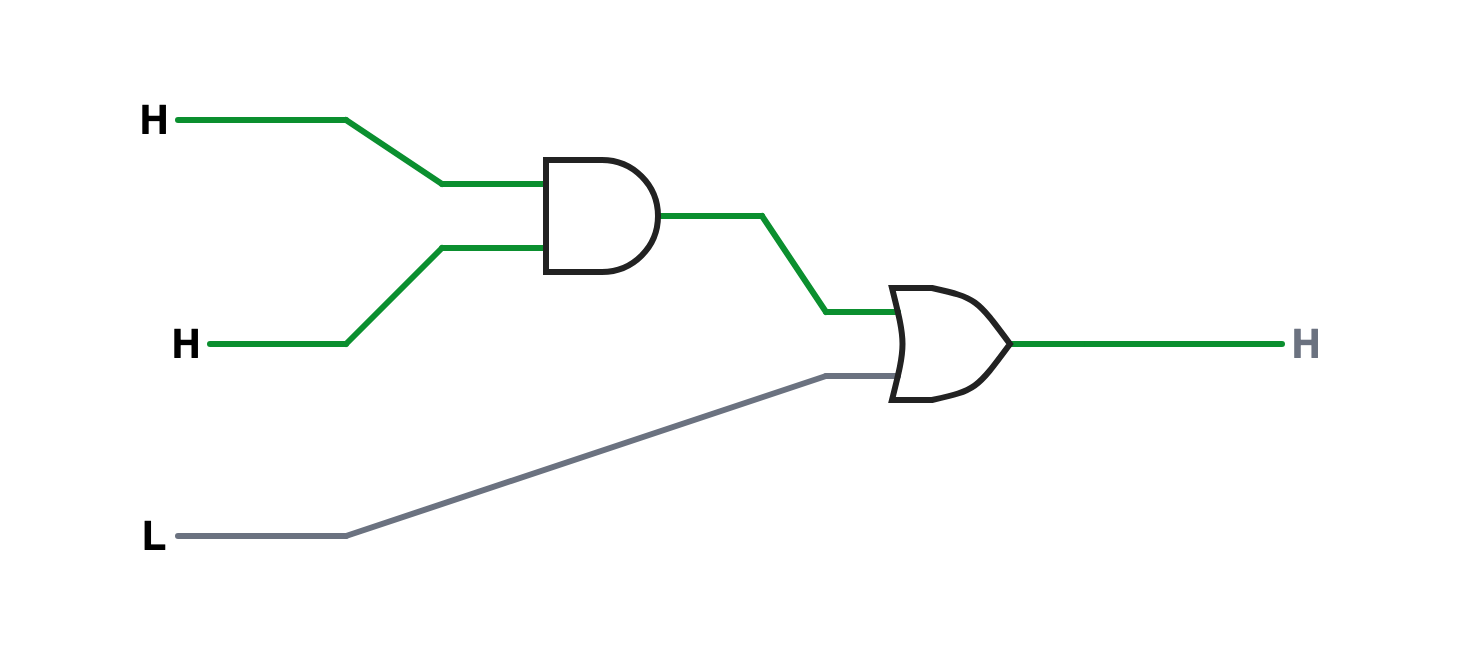}{} \\
\hline
\cellcolor{green!15}5 & {\scriptsize\ttfamily \# Build a 2-Bit Equality Comparator \newline
 \newline
\#\# GOAL \newline
Compare two 2-bit numbers A = (A1, A0) and B = (B1, B0) and drive one output EQ high exactly when the two numbers are equal. \newline
 \newline
\#\# STEPS \newline
 \newline
1. Add the following components: \newline
 - Four input switches: A1, A0, B1, B0 \newline
 - Two XNOR gates \newline
 - One AND gate \newline
 - One output switch: EQ \newline
 \newline
2. Wiring: \newline
 - Connect A1 and B1 to the inputs of the first XNOR gate. \newline
 - Connect A0 and B0 to the inputs of the second XNOR gate. \newline
 - Connect the outputs of both XNOR gates to the inputs of the AND gate. \newline
 \newline
3. Define output: \newline
 - The output of the AND gate must be the circuit output EQ. \newline
 \newline
\# RESULT FORMAT \newline
 \newline
```json \newline
\{"answer": "done"\} \newline
```} &
\circuitgt{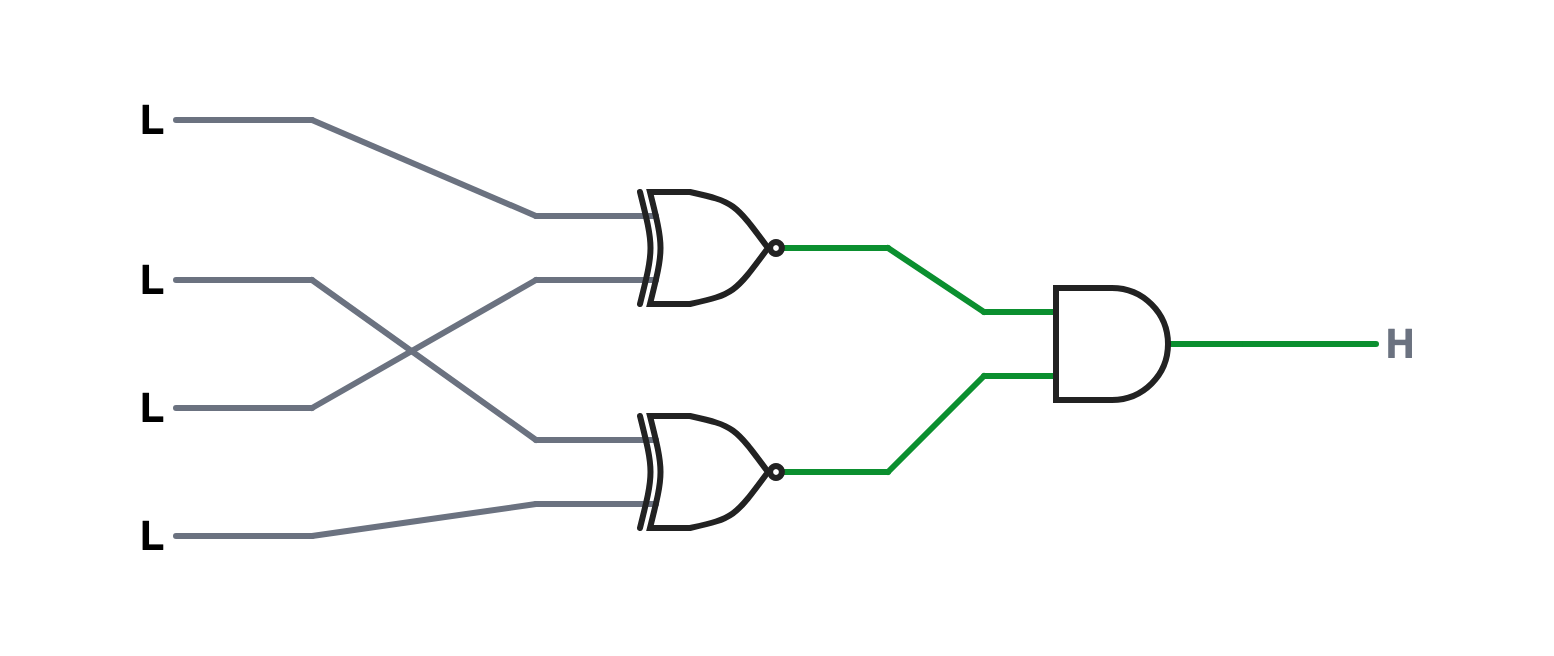}{} \\
\hline
\cellcolor{green!15}6 & {\scriptsize\ttfamily \# Build an OR Gate Using Only NAND Gates \newline
 \newline
\#\# GOAL \newline
Implement the OR function using only universal NAND gates, i.e. Y = (NOT A) NAND (NOT B). \newline
 \newline
\#\# STEPS \newline
 \newline
1. Add the following components: \newline
 - Two input switches: A, B \newline
 - Three NAND gates \newline
 - One output switch: Y \newline
 \newline
2. Wiring: \newline
 - Use the first NAND gate as an inverter for A by connecting A to both of its inputs. \newline
 - Use the second NAND gate as an inverter for B by connecting B to both of its inputs. \newline
 - Connect the outputs of those two NAND gates to the inputs of the third NAND gate. \newline
 \newline
3. Define output: \newline
 - The output of the third NAND gate must be the circuit output Y. \newline
 \newline
\# RESULT FORMAT \newline
 \newline
```json \newline
\{"answer": "done"\} \newline
```} &
\circuitgt{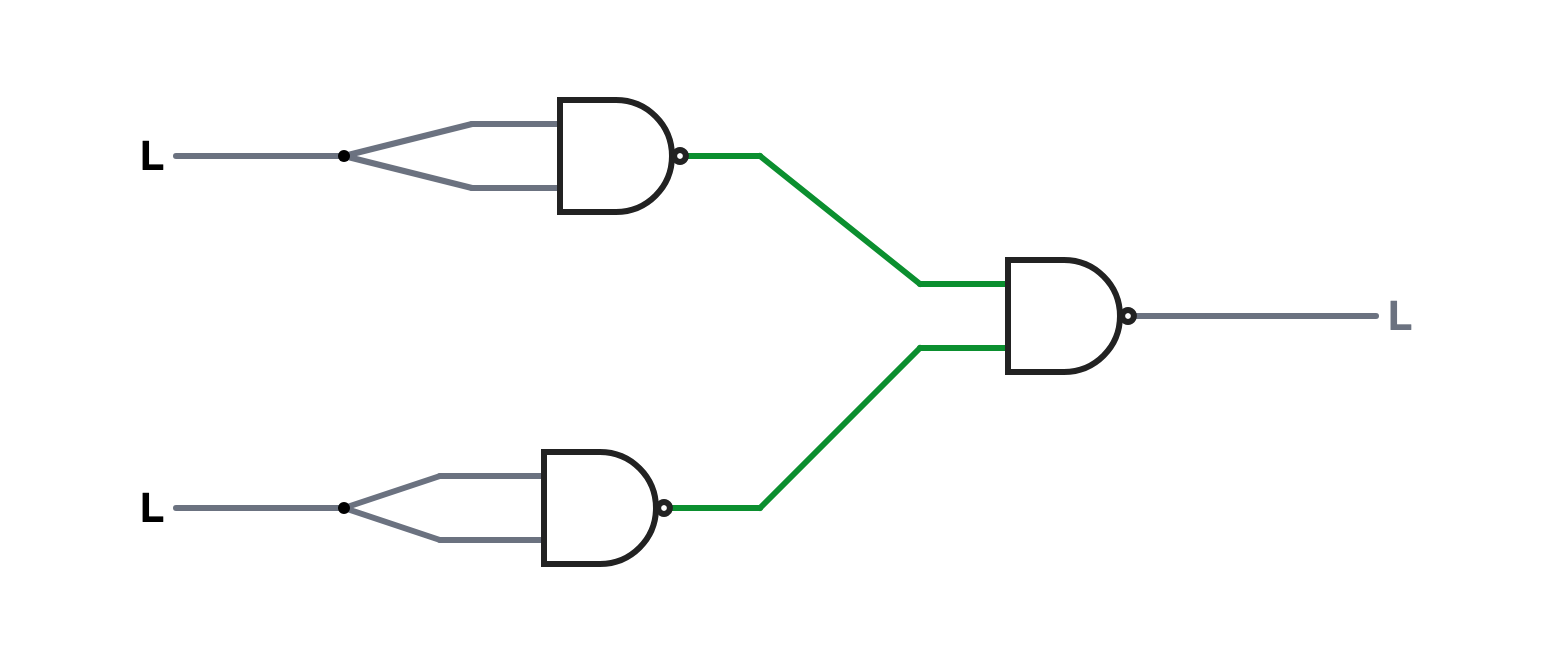}{} \\
\hline
\cellcolor{green!15}7 & {\scriptsize\ttfamily \# Build an OR Gate With an Inhibit Input \newline
 \newline
\#\# GOAL \newline
Construct a circuit with two data inputs (A, B) and one inhibit input (C) such that Y = (A OR B) AND NOT C — the output follows "A or B", but is forced low whenever C is high. \newline
 \newline
\#\# STEPS \newline
 \newline
1. Add the following components: \newline
 - Three input switches: A, B, C \newline
 - One OR gate \newline
 - One NOT gate \newline
 - One AND gate \newline
 - One output switch: Y \newline
 \newline
2. Wiring: \newline
 - Connect A and B to the inputs of the OR gate. \newline
 - Connect C to the input of the NOT gate. \newline
 - Connect the output of the OR gate and the output of the NOT gate to the inputs of the AND gate. \newline
 \newline
3. Define output: \newline
 - The output of the AND gate must be the circuit output Y. \newline
 \newline
\# RESULT FORMAT \newline
 \newline
```json \newline
\{"answer": "done"\} \newline
```} &
\circuitgt{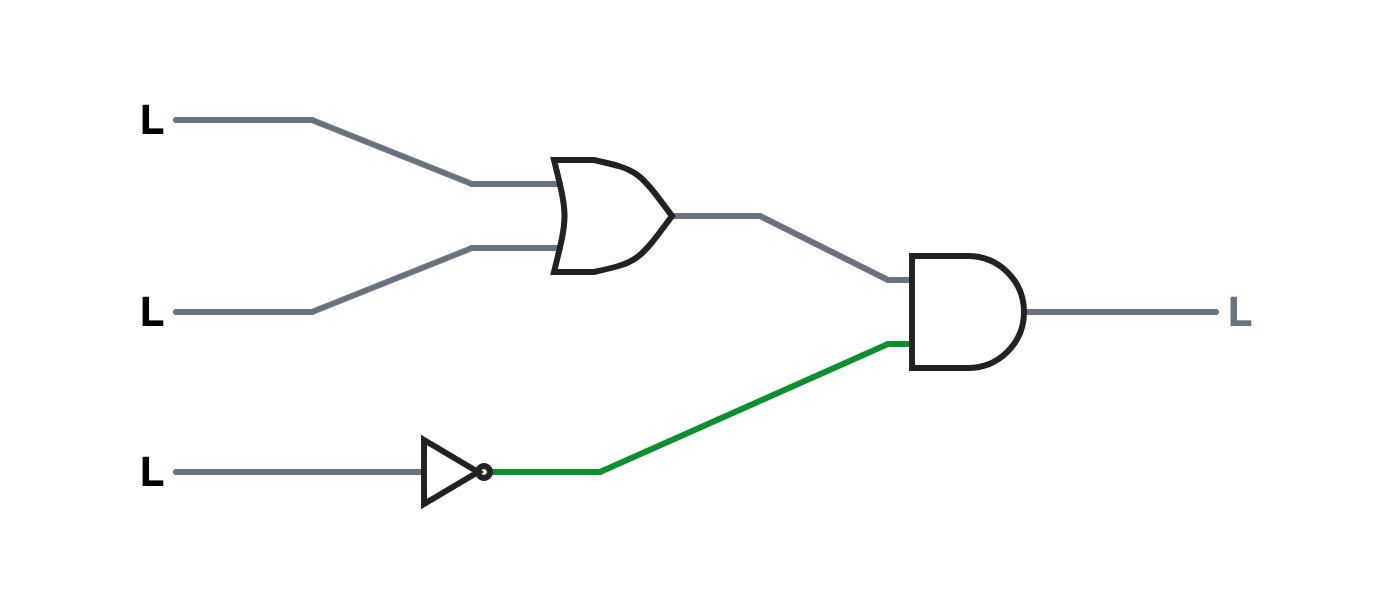}{} \\
\hline
\cellcolor{green!15}8 & {\scriptsize\ttfamily \# Build a Half Adder \newline
 \newline
\#\# GOAL \newline
Construct a 1-bit half adder with two inputs (A, B) and two outputs: \newline
 \newline
SUM = A XOR B \newline
CARRY = A AND B \newline
 \newline
\#\# STEPS \newline
 \newline
1. Add the following components: \newline
 - Two input switches: A, B \newline
 - One XOR gate \newline
 - One AND gate \newline
 - Two output switches: SUM and CARRY \newline
 \newline
2. Wiring: \newline
 - Connect A and B to the inputs of the XOR gate. \newline
 - Connect A and B to the inputs of the AND gate. \newline
 \newline
3. Define outputs: \newline
 - The output of the XOR gate must be the circuit output SUM. \newline
 - The output of the AND gate must be the circuit output CARRY. \newline
 \newline
\# RESULT FORMAT \newline
 \newline
```json \newline
\{"answer": "done"\} \newline
```} &
\circuitgt{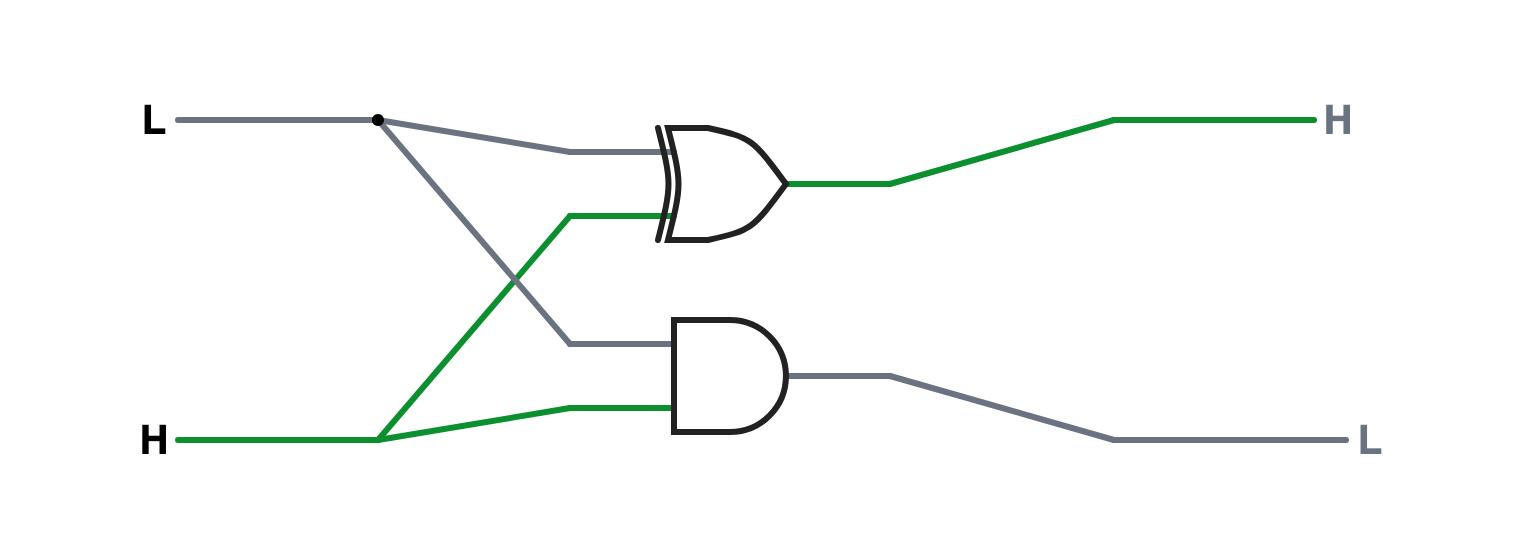}{} \\
\hline
\cellcolor{green!15}9 & {\scriptsize\ttfamily \# Build a Half Subtractor \newline
 \newline
\#\# GOAL \newline
Construct a 1-bit half subtractor with two inputs (A, B) that computes A - B and produces two outputs: \newline
 \newline
DIFF = A XOR B \newline
BORROW = (NOT A) AND B \newline
 \newline
\#\# STEPS \newline
 \newline
1. Add the following components: \newline
 - Two input switches: A, B \newline
 - One XOR gate \newline
 - One NOT gate \newline
 - One AND gate \newline
 - Two output switches: DIFF and BORROW \newline
 \newline
2. Wiring: \newline
 - Connect A and B to the inputs of the XOR gate. \newline
 - Connect A to the input of the NOT gate. \newline
 - Connect the output of the NOT gate and B to the inputs of the AND gate. \newline
 \newline
3. Define outputs: \newline
 - The output of the XOR gate must be the circuit output DIFF. \newline
 - The output of the AND gate must be the circuit output BORROW. \newline
 \newline
\# RESULT FORMAT \newline
 \newline
```json \newline
\{"answer": "done"\} \newline
```} &
\circuitgt{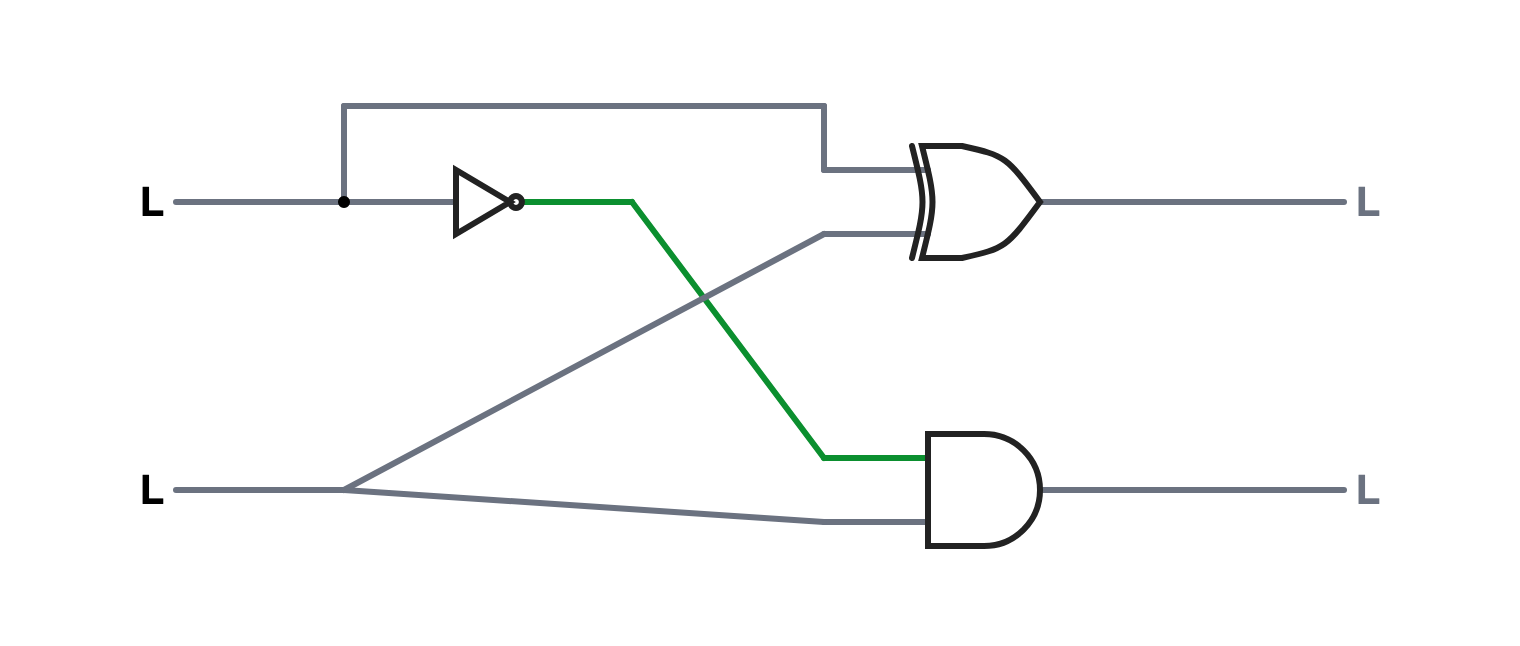}{} \\
\hline
\cellcolor{orange!15}10 & {\scriptsize\ttfamily \# Build a Full-Adder Using Logic Gates \newline
 \newline
\#\# GOAL \newline
Construct a circuit that computes the 1-bit full-adder functions: \newline
 \newline
SUM = A XOR B XOR CIN \newline
COUT = (A AND B) OR (A AND CIN) OR (B AND CIN) \newline
 \newline
\#\# STEPS \newline
 \newline
1. Add the following components: \newline
- Three input switches: A, B, CIN \newline
- Two output switches: SUM and COUT \newline
- Logical gates as needed \newline
 \newline
2. Build the circuit so that: \newline
- SUM is computed as the XOR of A, B, and CIN \newline
- COUT is computed as the carry bit of the full adder \newline
 \newline
3. Define outputs: \newline
- The output of the second XOR gate must be the circuit output SUM. \newline
- The output of the second OR gate must be the circuit output COUT. \newline
 \newline
\# RESULT FORMAT \newline
 \newline
```json \newline
\{"answer": "done"\} \newline
```} &
\circuitgt{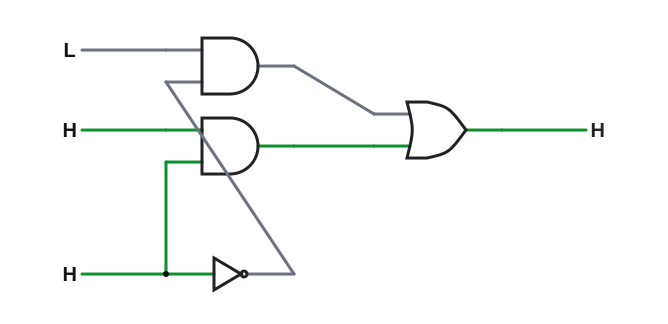}{} \\
\hline
\cellcolor{orange!15}11 & {\scriptsize\ttfamily \# Build a 1-Bit ALU With 2 Operations \newline
 \newline
\#\# GOAL \newline
Construct a simple 1-bit ALU that takes two data inputs A, B and one operation-select bit OP, then produces one output RESULT. \newline
 \newline
The ALU must support two operations OR and AND \newline
 \newline
\#\# STEPS \newline
 \newline
1. Add three input switches: A, B and OP \newline
 \newline
2. Add one output switch: RESULT \newline
 \newline
3. Build both candidate operation wires: \newline
- W\_AND = A AND B \newline
- W\_OR = A OR B \newline
 \newline
4. Use OP to select which operation appears at RESULT: \newline
- When OP = 0, RESULT = W\_AND \newline
- When OP = 1, RESULT = W\_OR \newline
 \newline
5. Verification: \newline
- For OP = 0, the circuit must behave like an AND gate. \newline
- For OP = 1, the circuit must behave like an OR gate. \newline
 \newline
\# RESULT FORMAT \newline
 \newline
```json \newline
\{"answer": "done"\} \newline
```} &
\circuitgt{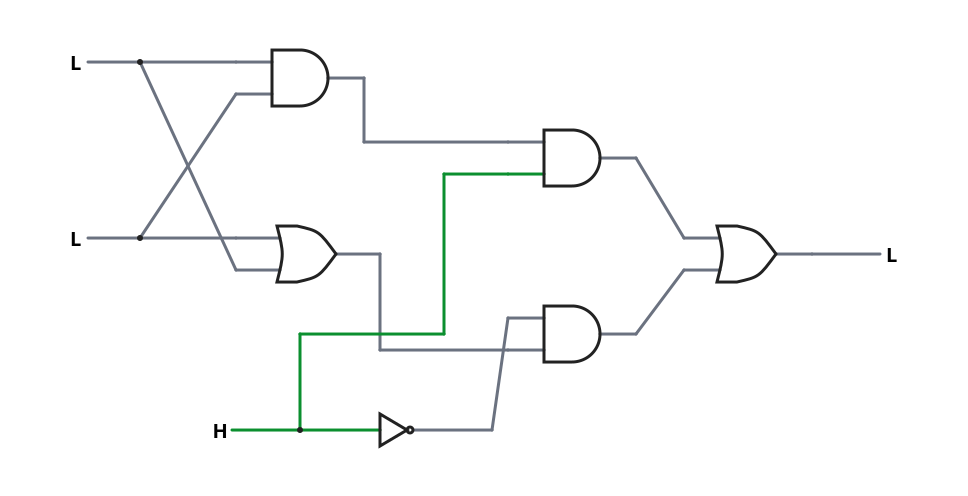}{} \\
\hline
\cellcolor{orange!15}12 & {\scriptsize\ttfamily \# Build a BCD-to-7-Segment Decoder for Segment a \newline
 \newline
\#\# GOAL \newline
Implement a circuit that takes a BCD input D = (D3, D2, D1, D0) and drives output SEG\_A high when segment a of a 7-segment display must be lit. \newline
 \newline
Use the following Boolean expression: \newline
SEG\_A = D3 + D1 + (D2 AND D0) + (NOT D2 AND NOT D0) \newline
 \newline
\#\# STEPS \newline
 \newline
1. Add the following components: \newline
- Four input switches: D3, D2, D1, D0 \newline
- One output switch: SEG\_A \newline
- Logical gates as needed \newline
 \newline
2. Build the circuit: \newline
- Treat D3 as the most significant bit. \newline
- Implement `SEG\_A` exactly as given in the goal equation. \newline
- Use NOT gates where negated signals are required. \newline
- Combine the terms using the necessary AND and OR gates. \newline
 \newline
3. Define the output: \newline
- The single circuit output must be SEG\_A. \newline
 \newline
\# RESULT FORMAT \newline
 \newline
```json \newline
\{"answer": "done"\} \newline
```} &
\circuitgt{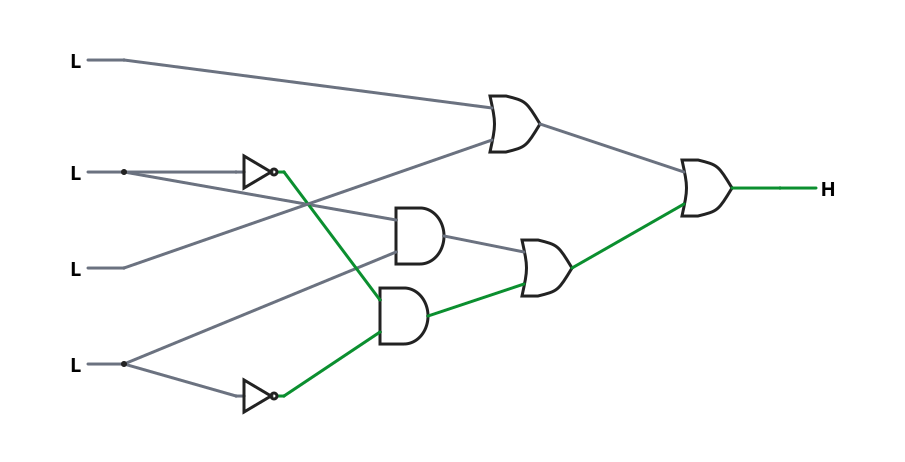}{} \\
\hline
\cellcolor{orange!15}13 & {\scriptsize\ttfamily \# Build a 1-to-4 Demultiplexer \newline
 \newline
\#\# GOAL \newline
 \newline
Implement a circuit with one data input D, two select bits S1 and S0, and four outputs Y0..Y3, such that D is routed to exactly one output chosen by (S1, S0) while the others stay low: \newline
 \newline
Y0 = (NOT S1 AND NOT S0 AND D), Y1 = (NOT S1 AND S0 AND D), Y2 = (S1 AND NOT S0 AND D), Y3 = (S1 AND S0 AND D) \newline
 \newline
\#\# STEPS \newline
 \newline
1. Add the following components: \newline
- One data input switch: D \newline
- Two select input switches: S1, S0 \newline
- Four output switches: Y0, Y1, Y2, Y3 \newline
- Logical gates as needed \newline
 \newline
2. Build the circuit so that: \newline
- For each Yi, build the 3-input AND of D and the select pattern for i. \newline
- Exactly one Yi can be 1 at any time, and only when D = 1. \newline
 \newline
3. Define outputs: \newline
- The four circuit outputs must be Y0, Y1, Y2, Y3. \newline
 \newline
\# RESULT FORMAT \newline
 \newline
```json \newline
\{"answer": "done"\} \newline
```} &
\circuitgt{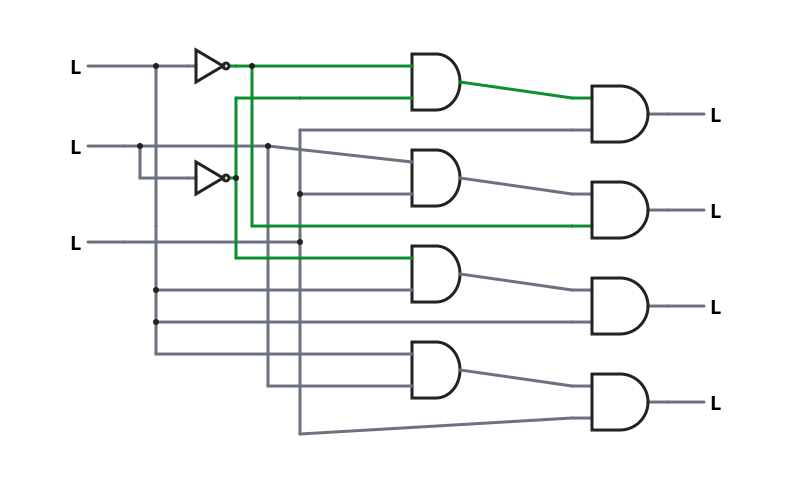}{} \\
\hline
\cellcolor{orange!15}14 & {\scriptsize\ttfamily \# Build a 1-Bit Comparator with Three Outputs \newline
 \newline
\#\# GOAL \newline
Construct a circuit that compares two 1-bit inputs A and B and produces three outputs: \newline
 \newline
A\_GT\_B = 1 when A > B \newline
A\_EQ\_B = 1 when A = B \newline
A\_LT\_B = 1 when A < B \newline
 \newline
\#\# STEPS \newline
 \newline
1. Add the following components: \newline
- Two input switches: A and B \newline
- Two NOT gates \newline
- Two AND gates \newline
- One XNOR gate \newline
- Three output switches: A\_GT\_B, A\_EQ\_B, and A\_LT\_B \newline
 \newline
2. Wiring: \newline
- Connect input B to the first NOT gate. \newline
- Connect input A and the output of the first NOT gate to the inputs of the first AND gate. \newline
- Connect input A to the second NOT gate. \newline
- Connect the output of the second NOT gate and input B to the inputs of the second AND gate. \newline
- Connect inputs A and B to the inputs of the XNOR gate. \newline
 \newline
3. Define outputs: \newline
- The output of the first AND gate must be the circuit output A\_GT\_B. \newline
- The output of the second AND gate must be the circuit output A\_LT\_B. \newline
- The output of the XNOR gate must be the circuit output A\_EQ\_B. \newline
 \newline
\# RESULT FORMAT \newline
 \newline
```json \newline
\{"answer": "done"\} \newline
```} &
\circuitgt{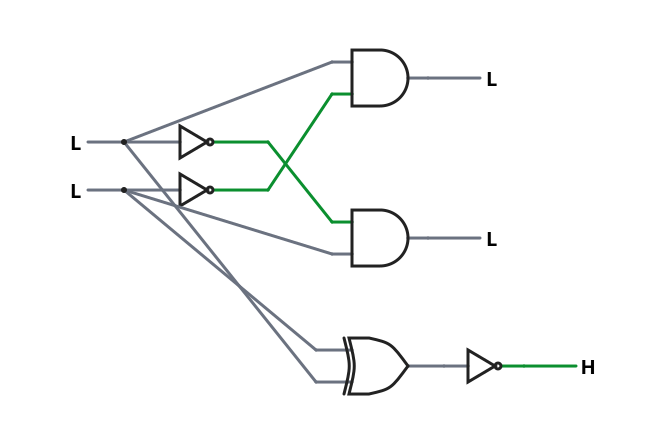}{} \\
\hline
\cellcolor{orange!15}15 & {\scriptsize\ttfamily \# Inverting Summing Amplifier \newline
 \newline
\#\# GOAL \newline
Construct an inverting summing amplifier using an ideal op-amp and measure its output voltage. \newline
 \newline
\#\# STEPS \newline
 \newline
1. Add and configure components: \newline
- Add an Ideal Op-Amp. \newline
- Add a 1-terminal DC Voltage Source (V1) and set it to 1.5 V. \newline
- Add a Resistor (R1) and set its resistance to 2 k ohms. \newline
- Add a 1-terminal DC Voltage Source (V2) and set it to -3.0 V. \newline
- Add a Resistor (R2) and set its resistance to 5 k ohms. \newline
- Add a Feedback Resistor (Rf) and set its resistance to 10 k ohms. \newline
- Add a Ground connection. \newline
 \newline
2. Wiring: \newline
- Connect the non-inverting (+) input of the Op-Amp directly to Ground. \newline
- Connect V1 to one end of R1. \newline
- Connect V2 to one end of R2. \newline
- Connect the free ends of both R1 and R2 to the inverting (-) input of the Op-Amp. \newline
- Connect Rf between the Op-Amp's output terminal and its inverting (-) input terminal. \newline
 \newline
3. Once the circuit reaches a steady state, hover your mouse over the Op-Amp and report the exact Output Voltage (Vout) displayed in the information box at the bottom right corner of the screen. \newline
 \newline
\# RESULT FORMAT \newline
 \newline
```json \newline
\{ \newline
"opamp\_output\_voltage": "value" \newline
\} \newline
```} &
\circuitgt{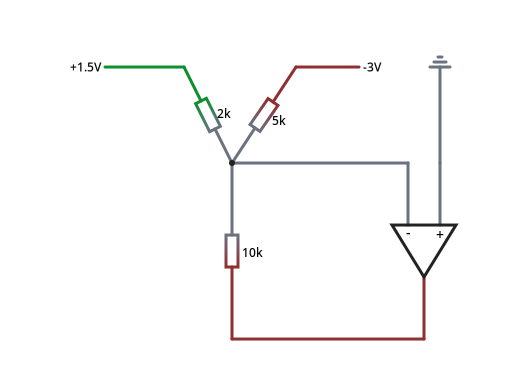}{\{ \newline
"opamp\_output\_voltage": "-1.5 V" \newline
\}} \\
\hline
\cellcolor{orange!15}16 & {\scriptsize\ttfamily \# Wheatstone Bridge Voltage Measurement \newline
 \newline
\#\# GOAL \newline
Construct a Wheatstone bridge circuit and measure the voltage drop across one of the bridge resistors. \newline
 \newline
\#\# STEPS \newline
 \newline
1. Add and configure components: \newline
- Add a 2-terminal DC Voltage Source and set its voltage to 10 V. \newline
- Add a Resistor (R1) and set its resistance to 1 k ohms. \newline
- Add a Resistor (R2) and set its resistance to 2.2 k ohms. \newline
- Add a Resistor (R3) and set its resistance to 3.3 k ohms. \newline
- Add a Resistor (R4) and set its resistance to 4.7 k ohms. \newline
- Add a Ground connection. \newline
 \newline
2. Wiring: \newline
- Ground the negative terminal of the voltage source. \newline
- Connect one end of R1 to the positive terminal of the voltage source. \newline
- Connect one end of R2 to the positive terminal of the voltage source. \newline
- Connect the free end of R1 to one end of R3. \newline
- Connect the free end of R2 to one end of R4. \newline
- Connect the free ends of R3 and R4 together and to Ground. \newline
 \newline
3. Once the circuit reaches a steady state, hover your mouse over Resistor R4 (the 4.7 k ohm resistor) and report the exact Voltage Drop (Vd) displayed in the information box at the bottom right corner of the screen. \newline
 \newline
\# RESULT FORMAT \newline
 \newline
```json \newline
\{ \newline
"voltage\_drop\_R4": "value" \newline
\} \newline
```} &
\circuitgt{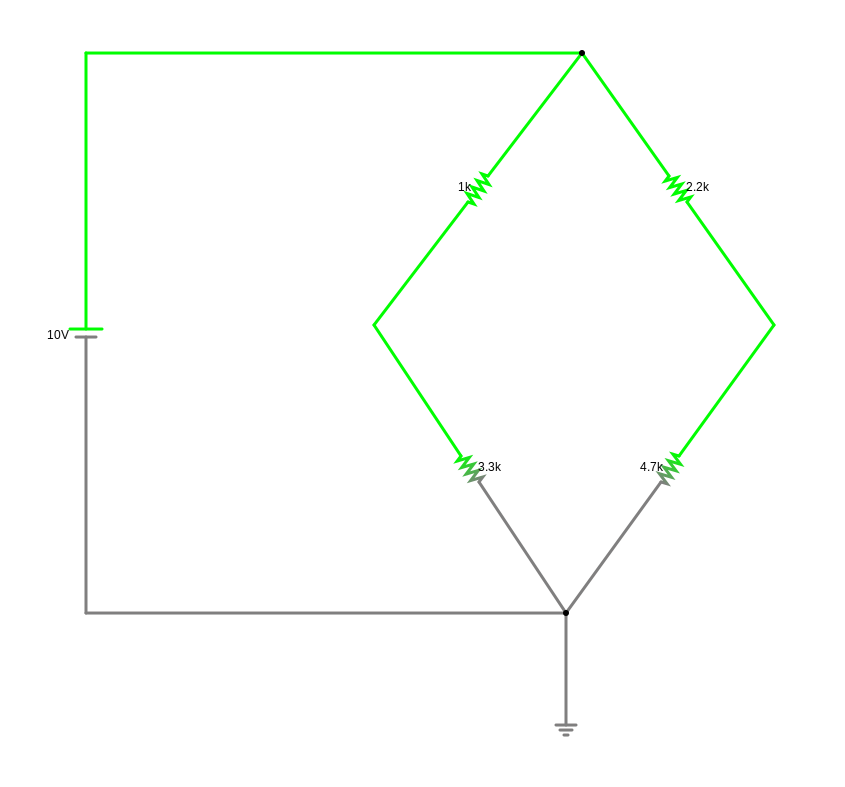}{\{ \newline
"voltage\_drop\_R4": "6.812 V" \newline
\}} \\
\hline
\cellcolor{orange!15}17 & {\scriptsize\ttfamily \# RLC DC Steady-State Trick \newline
 \newline
\#\# GOAL \newline
Construct an RLC circuit driven by DC and measure the capacitor voltage at steady state. \newline
 \newline
\#\# STEPS \newline
 \newline
1. Add and configure components: \newline
- Add a 2-terminal DC Voltage Source and set its voltage to 24 V. \newline
- Add a Resistor (R1) and set its resistance to 1 k ohms. \newline
- Add an Inductor and set its inductance to 100 mH. \newline
- Add a Resistor (R2) and set its resistance to 2.2 k ohms. \newline
- Add a Capacitor and set its capacitance to 10 uF. \newline
- Add a Resistor (R3) and set its resistance to 4.7 k ohms. \newline
- Add a Ground connection. \newline
 \newline
2. Wiring: \newline
- Ground the negative terminal of the voltage source. \newline
- Connect one end of R1 to the positive terminal of the source. Call the other end of R1 "Node A". \newline
- From Node A, create a path to Ground containing the Inductor in series with R2. \newline
- From Node A, create a second parallel path to Ground containing the Capacitor in series with R3. \newline
 \newline
3. Hover your mouse over the Capacitor. Wait for the simulation to charge the capacitor and reach a complete DC steady state (capacitor current displayed in the information box is less than 20 nA) and report the exact Voltage Drop (Vd) displayed in the information box at the bottom right corner of the screen. \newline
 \newline
\# RESULT FORMAT \newline
 \newline
```json \newline
\{ \newline
"capacitor\_voltage\_drop": "value" \newline
\} \newline
```} &
\circuitgt{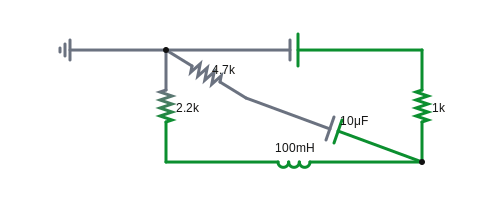}{\{ \newline
"capacitor\_voltage\_drop": "16.5 V" \newline
\}} \\
\hline
\cellcolor{orange!15}18 & {\scriptsize\ttfamily \# 555 Timer Astable: Output Frequency \newline
 \newline
\#\# GOAL \newline
Configure a 555 timer in astable (free-running) mode and measure the exact output frequency and duty cycle on the oscilloscope. \newline
 \newline
\#\# STEPS \newline
 \newline
1. Add and configure components: \newline
- Add a 555 Timer chip. \newline
- Add a Voltage Source 1-terminal (Vcc) and set it to 5 V. \newline
- Add a Resistor (Ra) and set its resistance to 10 k ohms. \newline
- Add a Resistor (Rb) and set its resistance to 22 k ohms. \newline
- Add a Capacitor (Ct) and set its capacitance to 100 nF. \newline
- Add a Ground connection. \newline
 \newline
2. Wiring: \newline
- Connect Vcc to Pin 8 (V+) of the 555. \newline
- Connect Pin 1 (GND) of the 555 to Ground. \newline
- Connect Pin 4 (RESET) of the 555 directly to Vcc (keep reset inactive). \newline
- Connect Ra between Vcc and Pin 7 (DISCH). \newline
- Connect Rb between Pin 7 (DISCH) and Pin 6 (THRESH). \newline
- Connect Pin 6 (THRESH) directly to Pin 2 (TRIG). \newline
- Connect Ct between Pin 6 (THRESH) and Ground. \newline
 \newline
3. Right-click Pin 3 (OUT) of the 555 and select "View in New Scope". In the scope's properties, enable BOTH "Show Frequency" AND "Show Duty Cycle". Let the simulation run for at least 30 output cycles to let the measurements settle, then report the exact Frequency and Duty Cycle values displayed on the scope. \newline
 \newline
\# RESULT FORMAT \newline
 \newline
```json \newline
\{ \newline
"output\_frequency": "value", \newline
"duty\_cycle": "value" \newline
\} \newline
```} &
\circuitgt{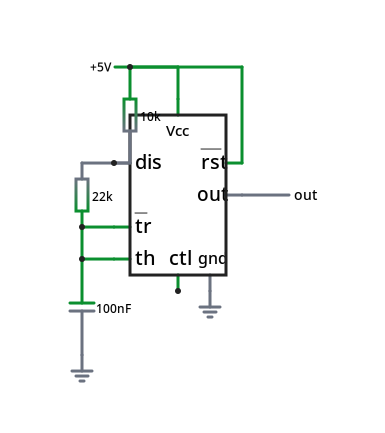}{\{ \newline
"output\_frequency": "260.417 Hz", \newline
"duty\_cycle": "58\%" \newline
\}} \\
\hline
\cellcolor{red!15}19 & {\scriptsize\ttfamily \# Build a 4-bit Monotonic Sequence Detector \newline
 \newline
\#\# GOAL \newline
Construct a combinational circuit with four binary inputs A, B, C, D (read left-to-right as a sequence in that order) and one output Y, such that Y = 1 if and only if the bit sequence A, B, C, D is non-decreasing — i.e., a 1 is never followed by a 0 anywhere in the sequence. \newline
 \newline
\#\# STEPS \newline
 \newline
1. Add the following components: \newline
- Four input switches: A, B, C, D \newline
- One output switch: Y \newline
- AND, OR, NOT, NAND, NOR, XOR, XNOR gates as needed \newline
 \newline
2. Build the circuit so that: \newline
- Y must produce the correct value for all 16 possible input combinations. \newline
 \newline
\# RESULT FORMAT \newline
 \newline
```json \newline
\{"answer": "done"\} \newline
```} &
\circuitgt{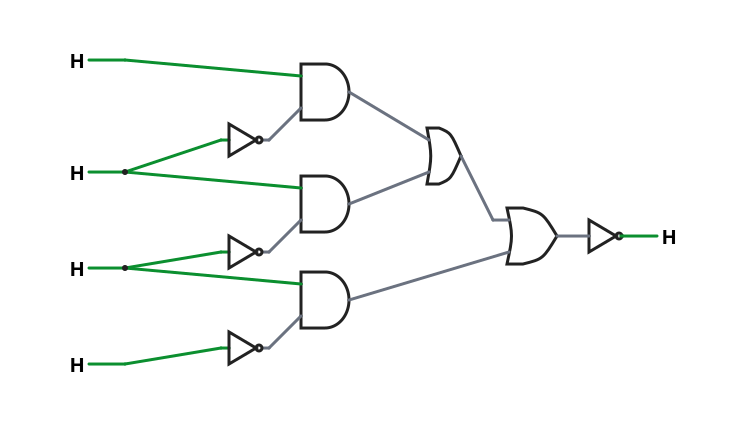}{} \\
\hline
\cellcolor{red!15}20 & {\scriptsize\ttfamily \# Build a 2-bit Binary Squarer \newline
 \newline
\#\# GOAL \newline
Construct a combinational circuit with two binary inputs A1, A0 (A1 the most significant bit) representing the unsigned integer N = 2·A1 + A0, and four binary outputs P3, P2, P1, P0 (P3 the most significant bit) representing the unsigned binary value of N², such that for every one of the 4 input combinations: \newline
 \newline
N² = 8·P3 + 4·P2 + 2·P1 + P0 \newline
 \newline
\#\# STEPS \newline
 \newline
1. Add the following components: \newline
- Two input switches: A1, A0 \newline
- Four output switches: P3, P2, P1, P0 \newline
- AND, OR, NOT, NAND, NOR, XOR, XNOR gates as needed \newline
 \newline
2. The circuit must produce correct values for all 4 input combinations. \newline
 \newline
\# RESULT FORMAT \newline
 \newline
```json \newline
\{"answer": "done"\} \newline
```} &
\circuitgt{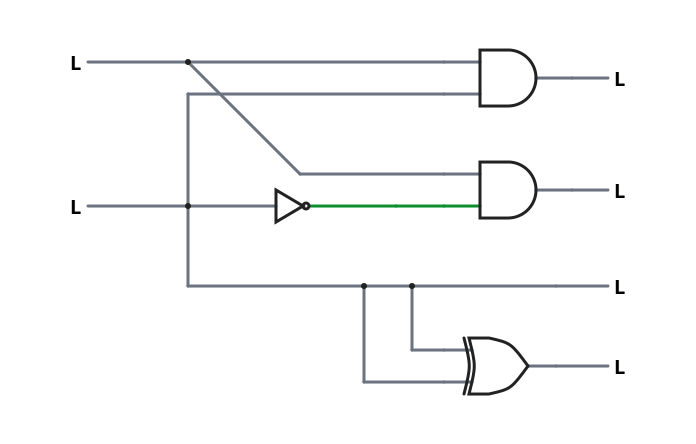}{} \\
\hline
\cellcolor{red!15}21 & {\scriptsize\ttfamily \# Build a 4-to-2 Priority Encoder with Valid Flag \newline
 \newline
\#\# GOAL \newline
Construct a combinational circuit with four request inputs R3, R2, R1, R0 (R3 is the highest priority) and three outputs Y1, Y0, V such that: \newline
 \newline
- V = 1 if at least one request is high; V = 0 otherwise. \newline
- When V = 1, (Y1, Y0) is the binary index of the highest-priority asserted input: \newline
R3 high → (Y1, Y0) = (1, 1) \newline
R3 low, R2 high → (1, 0) \newline
R3 low, R2 low, R1 high → (0, 1) \newline
only R0 high → (0, 0) \newline
- When V = 0, (Y1, Y0) is don't-care. \newline
 \newline
Implement the outputs using these expressions (priority is enforced by masking lower-priority requests with NOT of higher-priority ones): \newline
 \newline
Y1 = R3 OR (NOT R3 AND R2)         = R3 OR R2 \newline
Y0 = R3 OR (NOT R2 AND R1) \newline
V  = R3 OR R2 OR R1 OR R0 \newline
 \newline
Outputs must be correct for all 16 input combinations. \newline
 \newline
\#\# STEPS \newline
 \newline
1. Add the following components: \newline
- Four input switches: R3, R2, R1, R0 \newline
- Three output switches: Y1, Y0, V \newline
- AND, OR, NOT, NAND, NOR, XOR, XNOR gates as needed \newline
 \newline
2. Build Y1, Y0, V exactly as given above. \newline
 \newline
3. Define outputs: \newline
- The three circuit outputs must be Y1, Y0, V. \newline
 \newline
\# RESULT FORMAT \newline
 \newline
```json \newline
\{"answer": "done"\} \newline
```} &
\circuitgt{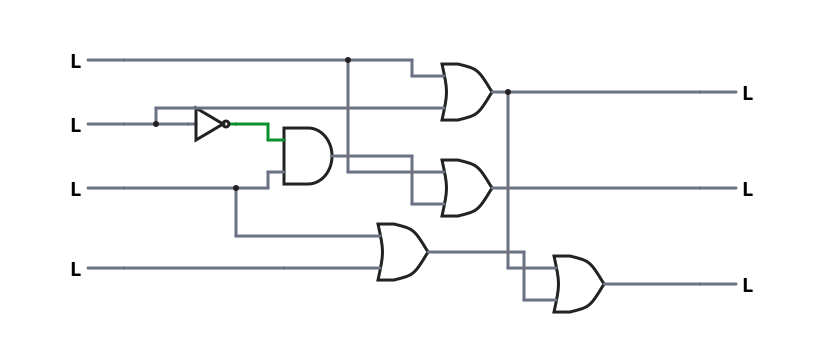}{} \\
\hline
\cellcolor{red!15}22 & {\scriptsize\ttfamily \# Build a 3-Input Majority and Minority Detector \newline
 \newline
\#\# GOAL \newline
Construct a combinational circuit with three binary inputs A, B, C and two outputs MAJ and MIN such that: \newline
 \newline
- MAJ = 1 when at least two of the three inputs are 1 \newline
- MIN = 1 when at least two of the three inputs are 0 \newline
 \newline
Exactly one of MAJ and MIN is high for every input combination, and they are logical complements. \newline
 \newline
Implement the outputs using these expressions, sharing the three pairwise-AND terms across both outputs: \newline
 \newline
T\_AB = A AND B \newline
T\_AC = A AND C \newline
T\_BC = B AND C \newline
MAJ  = T\_AB OR T\_AC OR T\_BC \newline
MIN  = (NOT T\_AB) AND (NOT T\_AC) AND (NOT T\_BC) \newline
 \newline
The three pairwise-AND wires (T\_AB, T\_AC, T\_BC) must each be computed by a single AND gate and fan out to both the MAJ OR-tree and the MIN NOT/AND-tree. \newline
 \newline
\#\# STEPS \newline
 \newline
1. Add the following components: \newline
- Three input switches: A, B, C \newline
- Two output switches: MAJ, MIN \newline
- AND, OR, NOT, NAND, NOR, XOR, XNOR gates as needed \newline
 \newline
2. Build T\_AB, T\_AC, T\_BC exactly as given. Then build MAJ and MIN by reusing those three wires. \newline
 \newline
3. Define outputs: \newline
- The two circuit outputs must be MAJ and MIN. \newline
- Outputs must be correct for all 8 input combinations. \newline
 \newline
\# RESULT FORMAT \newline
 \newline
```json \newline
\{"answer": "done"\} \newline
```} &
\circuitgt{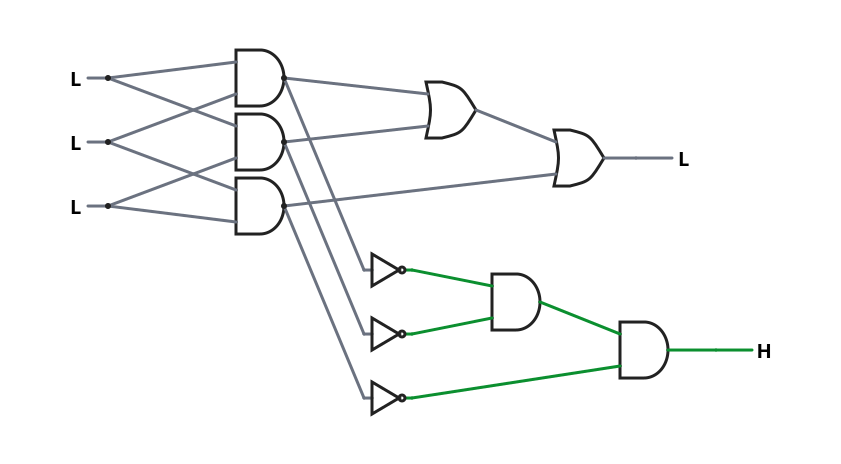}{} \\
\hline
\cellcolor{red!15}23 & {\scriptsize\ttfamily \# BJT Differential Pair with Resistor Loads \newline
 \newline
\#\# GOAL \newline
Build a BJT differential pair with resistor collector loads and an NPN current-mirror tail. Apply a small differential input and measure the peak-to-peak output voltage at the collector of the right-side transistor. \newline
 \newline
\#\# STEPS \newline
 \newline
1. Add and configure components: \newline
- Add three supply nodes: Vcc = +10 V, Vee = -10 V, and Ground. \newline
- Add two default NPN transistors for the input pair: Q1 and Q2. \newline
- Add two collector-load resistors: \newline
- Rc1 = 10 k ohms \newline
- Rc2 = 10 k ohms \newline
- Add two default NPN transistors for the tail current mirror: Q5 and Q6. \newline
- Add a reference resistor Rref and set its resistance to 19 k ohms. \newline
- Add a 1-terminal A/C Voltage Source (Vin) and set: Max Voltage = 1 mV, Frequency = 1000 Hz, DC offset = 0 V, phase = 0 deg. \newline
 \newline
2. Wiring: \newline
- Connect the emitters of Q1 and Q2 together. Call this node TAIL. \newline
- Connect Rc1 from Vcc to the collector of Q1. \newline
- Connect Rc2 from Vcc to the collector of Q2. Call the collector of Q2 OUT. \newline
- Connect the emitters of Q5 and Q6 to Vee. \newline
- Diode-connect Q5 by shorting its collector to its base. \newline
- Connect Rref from Vcc to the collector/base node of Q5. \newline
- Connect the base of Q6 to the base of Q5. \newline
- Connect the collector of Q6 to the TAIL node. \newline
- Connect Vin to the base of Q1. \newline
- Connect the base of Q2 directly to Ground. \newline
 \newline
3. Measurement: \newline
- Run the simulation. \newline
- Confirm that the DC operating point places OUT between the rails, approximately around +5 V DC. \newline
- Scope the OUT node. \newline
- In steady state, read the Peak-to-Peak value shown in the scope overlay. \newline
- Report the measured peak-to-peak AC voltage at OUT. \newline
 \newline
\# RESULT FORMAT \newline
 \newline
```json \newline
\{ \newline
"diff\_pair\_output\_peak\_to\_peak": "value" \newline
\}} &
\circuitgt{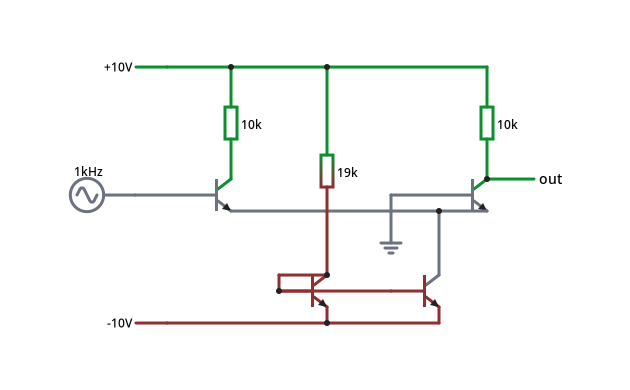}{\{ \newline
"diff\_pair\_output\_peak\_to\_peak": "191.615 mV" \newline
\}} \\
\hline
\cellcolor{red!15}24 & {\scriptsize\ttfamily \# Build a 3-bit Two's-Complement Negator with Overflow Flag \newline
 \newline
\#\# GOAL \newline
Construct a combinational circuit that takes a 3-bit two's-complement signed integer A = (A2, A1, A0), where A2 is the sign bit and the represented value is \newline
 \newline
val(A) = -4·A2 + 2·A1 + A0   (range: -4 .. +3) \newline
 \newline
and produces: \newline
- A 3-bit two's-complement output Y = (Y2, Y1, Y0) such that val(Y) = -val(A) whenever -val(A) is representable in 3-bit two's complement. \newline
- A 1-bit overflow flag OVF that is 1 if and only if A = 100 (i.e. val(A) = -4), because +4 cannot be represented in 3-bit two's complement. When OVF = 1, the value of Y is unconstrained (don't-care). \newline
 \newline
The negation must be computed as Y = (NOT A) + 1 using bitwise inversion followed by a 3-bit ripple-carry incrementer. \newline
 \newline
\#\# STEPS \newline
 \newline
1. Add the following components: \newline
- Three input switches: A2, A1, A0 \newline
- Four output switches: Y2, Y1, Y0, OVF \newline
- AND, OR, NOT, NAND, NOR, XOR, XNOR gates as needed \newline
 \newline
2. Build the circuit so that: \newline
- First stage: compute B0 = NOT A0, B1 = NOT A1, B2 = NOT A2 using NOT gates. \newline
- Second stage: ripple-carry increment B by 1 using three half-adders chained by carry: \newline
- Bit 0: Y0 = B0 XOR 1 = NOT B0 (equivalently, Y0 = A0); carry C1 = B0 AND 1 = B0. \newline
- Bit 1: Y1 = B1 XOR C1; carry C2 = B1 AND C1. \newline
- Bit 2: Y2 = B2 XOR C2. (The final carry-out is discarded — it does not feed OVF.) \newline
- Overflow: OVF = A2 AND (NOT A1) AND (NOT A0). This is the only input pattern for which the negation overflows. \newline
- Y must produce the correct value for all 7 non-overflow input combinations, and OVF must be correct for all 8 input combinations. \newline
 \newline
3. Define outputs: \newline
- The four circuit outputs must be Y2, Y1, Y0, OVF. \newline
 \newline
\# RESULT FORMAT \newline
 \newline
```json \newline
\{"answer": "done"\} \newline
```} &
\circuitgt{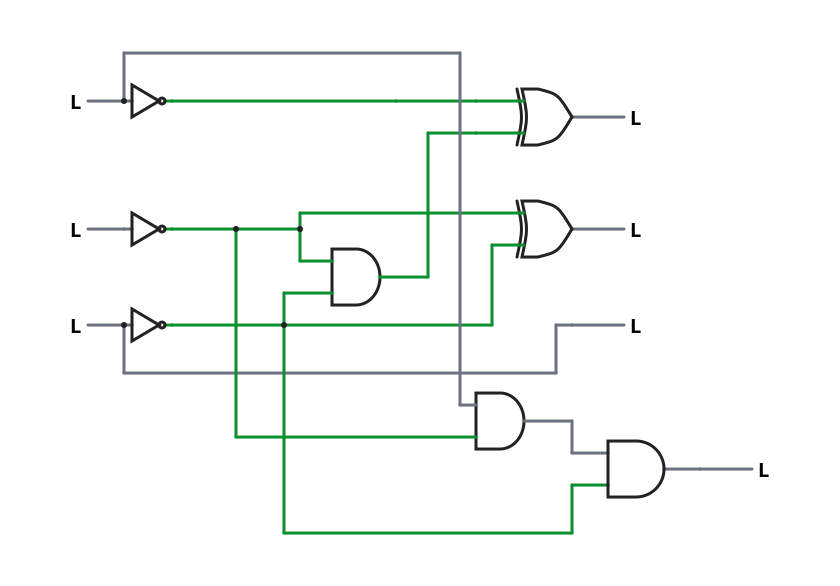}{} \\
\hline
\cellcolor{red!15}25 & {\scriptsize\ttfamily \# 4-Bit Asynchronous Ripple Counter: MSB Output Frequency \newline
 \newline
\#\# GOAL \newline
Build a 4-bit asynchronous (ripple) counter from JK flip-flops driven by a 1 kHz clock, and measure the output frequency at the MSB (Q3). Each flip-flop divides its clock input by 2, so a 4-bit ripple counter divides the master clock by 2\textasciicircum{}4 = 16. \newline
 \newline
\#\# STEPS \newline
 \newline
1. Add and configure components: \newline
- Add a Clock source (Inputs/Outputs menu → Clock, or use a Square-wave Voltage Source). Set: frequency = 1000 Hz, duty cycle = 50\%, high level = 5 V, low level = 0 V. \newline
- Add four JK Flip-Flops (Chips menu → JK Flip-Flop). Label them FF0 (LSB) through FF3 (MSB). \newline
- Add a Logic Input set permanently to high (5 V) — used as a constant logic-1 source for the J and K pins. \newline
- Add a Logic Output for visualising Q3. \newline
- Add Ground. \newline
 \newline
2. Wiring (asynchronous toggle counter): \newline
- Tie BOTH J and K of every flip-flop (FF0–FF3) to the constant-high Logic Input — this configures each one to toggle on every clock edge. \newline
- Connect the Clock source to the CLK input of FF0. \newline
- Connect Q (output) of FF0 → CLK of FF1. \newline
- Connect Q of FF1 → CLK of FF2. \newline
- Connect Q of FF2 → CLK of FF3. \newline
- Connect Q of FF3 → the Logic Output indicator. \newline
- Leave the !Q (inverted) outputs unconnected, and leave any RESET/SET pins unconnected (or tie them to the inactive level if the chip block exposes them). \newline
 \newline
3. Click RUN. \newline
 \newline
4. Right-click the wire at Q3 (the output of FF3) → View in New Scope. In the scope's properties, enable "Show Frequency". Let the simulation run for at least 1 second of simulated time so the counter cycles through all 16 states many times and the frequency overlay locks in. \newline
 \newline
5. Report the frequency value displayed on the scope at Q3. \newline
 \newline
\# RESULT FORMAT \newline
 \newline
The value must be a string containing the numeric frequency in Hz with units, e.g. "62.5 Hz". Do NOT emit the literal placeholder text "value". Do NOT omit units. Round to 1 decimal place if needed. \newline
 \newline
```json \newline
\{ \newline
"msb\_output\_frequency": "<frequency>" \newline
\} \newline
```} &
\circuitgt{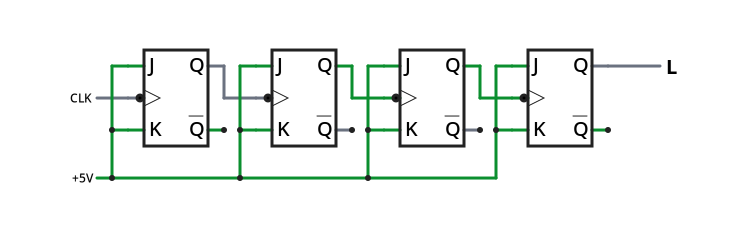}{\{ \newline
"msb\_output\_frequency": "62.5 Hz" \newline
\}} \\
\hline
\cellcolor{red!15}26 & {\scriptsize\ttfamily \# Common-Emitter BJT Amplifier: Mid-Band Peak-to-Peak Output \newline
 \newline
\#\# GOAL \newline
Build a classic common-emitter BJT voltage amplifier with a voltage-divider bias network and a bypassed emitter resistor, and measure the peak-to-peak output voltage at the collector for a small AC input. \newline
 \newline
\#\# STEPS \newline
 \newline
1. Add and configure components: \newline
- Add a Voltage Source 1-terminal (Vcc) and set it to 12 V. \newline
- Add a Transistor (bipolar, NPN). Leave default model parameters. \newline
- Add a Resistor (R1) and set it to 47 k ohms (top bias). \newline
- Add a Resistor (R2) and set it to 10 k ohms (bottom bias). \newline
- Add a Resistor (Rc) and set it to 3.3 k ohms (collector). \newline
- Add a Resistor (Re) and set it to 1 k ohms (emitter). \newline
- Add a Capacitor (Cin) and set it to 1 uF (input coupling). \newline
- Add a Capacitor (Ce) and set it to 100 uF (emitter bypass). \newline
- Add a Capacitor (Cout) and set it to 1 uF (output coupling). \newline
- Add a Resistor (R\_load) and set it to 10 k ohms. \newline
- Add a 1-terminal A/C Voltage Source (Vin) and set: Max Voltage = 10 mV, Frequency = 10 kHz, DC offset = 0, phase = 0. \newline
- Add a Ground connection. \newline
 \newline
2. Wiring: \newline
- Connect Vcc to the top of R1 and to the top of Rc. \newline
- Connect the bottom of R1 and the top of R2 together — call this node "Base\_Bias". Connect the transistor's base to Base\_Bias. \newline
- Connect the bottom of R2 to Ground. \newline
- Connect the bottom of Rc to the transistor's collector. Call this node "Collector". \newline
- Connect the transistor's emitter to the top of Re. Call this node "Emitter". \newline
- Connect the bottom of Re to Ground. \newline
- Connect Ce between Emitter and Ground (parallel with Re). \newline
- Connect Cin between Vin and Base\_Bias. \newline
- Connect Cout between Collector and one end of R\_load. Call the other end of R\_load "Node OUT". \newline
- Connect Node OUT to Ground through R\_load (i.e., R\_load is between Cout and Ground, and Node OUT is the junction with Cout). \newline
 \newline
3. Right-click the wire at Node OUT (the load side of Cout) and select "View in New Scope". In the scope's properties, enable "Show Peak-to-Peak". Let the simulation run for at least 50 input cycles so all coupling capacitors fully charge and the amplifier settles into its operating point, then report the exact Peak-to-Peak value displayed on the scope. \newline
 \newline
\# RESULT FORMAT \newline
 \newline
```json \newline
\{ \newline
"amplifier\_output\_peak\_to\_peak": "value" \newline
\} \newline
```} &
\circuitgt{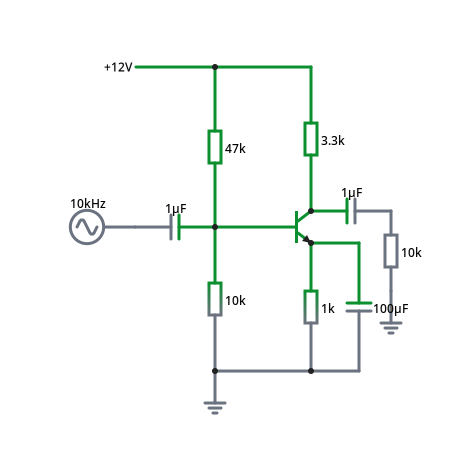}{\{ \newline
"amplifier\_output\_peak\_to\_peak": "2.605 V" \newline
\}} \\
\hline
\cellcolor{red!15}27 & {\scriptsize\ttfamily \# Wien Bridge Oscillator with Diode AGC \newline
 \newline
\#\# GOAL \newline
Build a Wien-bridge oscillator using an op-amp with positive-feedback RC network and a negative-feedback gain network. Use anti-parallel diodes for soft amplitude limiting (automatic gain control), let the loop self-start from a small initial-condition seed, and report both the steady-state oscillation frequency and the peak-to-peak output voltage at the op-amp output. \newline
 \newline
\#\# STEPS \newline
 \newline
1. Add and configure components: \newline
- Add an Op Amp (use the default model, finite-GBW, rails at ±15 V). \newline
- Positive-feedback (Wien) network: \newline
- Resistor Rs = 10 k ohms (series arm). \newline
- Capacitor Cs = 10 nF (series arm). \newline
- Resistor Rp = 10 k ohms (parallel arm). \newline
- Capacitor Cp = 10 nF (parallel arm). \newline
- Negative-feedback (gain-setting) network with diode AGC: \newline
- Resistor Rg = 10 k ohms (from − input to ground). \newline
- Resistor Rf\_a = 12 k ohms (first half of the feedback resistor). \newline
- Resistor Rf\_b = 10 k ohms (second half of the feedback resistor; the diodes shunt this part as the swing grows). \newline
- Two anti-parallel diodes (default Si model), each connected across Rf\_b. \newline
- Add Ground connections. \newline
 \newline
2. Wiring: \newline
- Positive-feedback network (between op-amp OUTPUT and the + input): \newline
- Rs from op-amp OUTPUT to a node "Node\_S". \newline
- Cs from Node\_S to the op-amp + input (call this node "Node\_Plus"). \newline
- Rp from Node\_Plus to Ground. \newline
- Cp from Node\_Plus to Ground (in parallel with Rp). \newline
- Negative-feedback network: \newline
- Rg from the op-amp − input to Ground. \newline
- Rf\_a from the op-amp − input to a midpoint node "Node\_FB\_Mid". \newline
- Rf\_b from Node\_FB\_Mid to op-amp OUTPUT. \newline
- Diode D1 from Node\_FB\_Mid (anode) to op-amp OUTPUT (cathode) — i.e., across Rf\_b. \newline
- Diode D2 from op-amp OUTPUT (anode) to Node\_FB\_Mid (cathode) — i.e., the anti-parallel partner of D1, also across Rf\_b. \newline
 \newline
3. Seed start-up: \newline
- Right-click Cp (the cap in the parallel arm) → Edit → set Initial Voltage = 0.1 V (or higher if needed). CircuitJS1's components are noiseless, so without a seed the oscillator will not spontaneously start. \newline
4. Run the simulation. \newline
5. Right-click the op-amp OUTPUT terminal node and select "View in New Scope". In the scope's properties, enable BOTH "Show Frequency" AND "Show Peak-to-Peak". Let the simulation run long enough for the amplitude to grow into the diode-clamped steady state (\textasciitilde{}50–100 ms simulated). The frequency overlay needs at least 5–10 cycles visible per screen to lock in — if it reads blank, zoom the scope's TIME axis (not the voltage axis) until cycles are clearly resolved, then wait for the readout to stabilise. \newline
 \newline
6. Report the exact Frequency and Peak-to-Peak values displayed on the scope. \newline
 \newline
\# RESULT FORMAT \newline
 \newline
```json \newline
\{ "oscillation\_frequency": "value", \newline
"output\_peak\_to\_peak": "value" \} \newline
```} &
\circuitgt{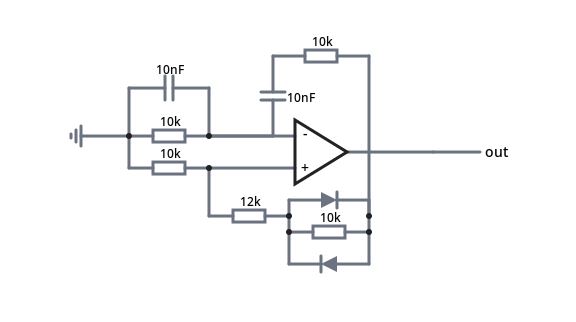}{\{ \newline
"oscillation\_frequency": "1.589 kHz", \newline
"output\_peak\_to\_peak": "1.422 V" \newline
\}} \\
\hline
\end{longtable}
\normalsize